\documentclass{book}
\usepackage{authblk}
\usepackage{roboto}

\usepackage[english]{babel}
\usepackage[letterpaper,top=2cm,bottom=2cm,left=3cm,right=3cm,marginparwidth=1.75cm]{geometry}

\usepackage{amsfonts}
\UseRawInputEncoding

\usepackage{pdflscape}
\usepackage{rotating}
\usepackage{longtable}
\usepackage{array}
\usepackage{float}
\usepackage{amsmath}
\usepackage{graphicx}
\usepackage{inconsolata}
\usepackage[colorlinks=true, allcolors=blue]{hyperref}
\usepackage{enumitem}
\usepackage{tikz}  
\usepackage{listings}
\usepackage{xcolor}
\usepackage[T1]{fontenc}
\usepackage{IEEEtrantools}  
\usepackage{epigraph}  
\usepackage[hang,flushmargin]{footmisc} 
\usetikzlibrary{positioning} 

\definecolor{bgcolor}{rgb}{0.97,0.97,0.97}
\definecolor{codeblue}{rgb}{0.1,0.1,0.8}
\definecolor{codegreen}{rgb}{0,0.4,0}
\definecolor{codegray}{rgb}{0.4,0.4,0.4}
\definecolor{codepurple}{rgb}{0.5,0,0.5}
\definecolor{codered}{rgb}{0.6,0.2,0.2}
\definecolor{lightgray}{rgb}{0.9,0.9,0.9}
\definecolor{darkgray}{rgb}{0.6,0.6,0.6}  

\makeatletter
\renewcommand{\paragraph}{
  \@startsection{paragraph}{4}{\z@}{1ex}{-1em}{\normalfont\normalsize\bfseries\color{gray}}}
\makeatother

\lstdefinestyle{python}{
    language=Python,
    basicstyle=\ttfamily\small\color{black}\usefont{T1}{zi4}{m}{n},  
    keywordstyle=\bfseries\color{codeblue},  
    stringstyle=\color{codegreen},  
    commentstyle=\slshape\color{codegray},  
    showstringspaces=false,
    numbers=left,
    numberstyle=\tiny\color{codegray},  
    stepnumber=1,
    numbersep=8pt,
    frame=single,
    rulecolor=\color{darkgray},  
    breaklines=true,
    backgroundcolor=\color{bgcolor},
    tabsize=4,
    captionpos=b,
    morekeywords={self},  
}

\lstdefinestyle{cmd}{
    language=bash,
    basicstyle=\ttfamily\small\color{black}\usefont{T1}{zi4}{m}{n},  
    keywordstyle=\bfseries\color{blue},
    stringstyle=\color{codegreen},
    commentstyle=\itshape\color{gray},
    showstringspaces=false,
    numbers=none,
    frame=single,
    rulecolor=\color{darkgray},  
    breaklines=true,
    backgroundcolor=\color{bgcolor},
    tabsize=4,
    captionpos=b,
}

\title{Generative Adversarial Networks Bridging Art and Machine Intelligence}
\footnotesize
\author{
    Junhao Song\textsuperscript{*}\textsuperscript{$\dagger$}\thanks{Imperial College London, junhao.song23@imperial.ac.uk}, 
    Yichao Zhang\textsuperscript{*}\thanks{The University of Texas at Dallas, yichao.zhang.us@gmail.com}
    Ziqian Bi\textsuperscript{*}\thanks{Indiana University, bizi@iu.edu}, 
    Tianyang Wang\thanks{Xi'an Jiaotong-Liverpool University, tianyang.wang21@student.xjtlu.edu.cn}, 
    Keyu Chen\thanks{Georgia Institute of Technology, kchen637@gatech.edu}, 
    Ming Li\thanks{Georgia Institute of Technology, mli694@gatech.edu}, 
    Qian Niu\thanks{Kyoto University, niu.qian.f44@kyoto-u.ac.jp}, 
    Junyu Liu\thanks{Kyoto University, liu.junyu.82w@kyoto-u.ac.jp}, 
    Benji Peng\thanks{AppCubic, benji@appcubic.com}, 
    Sen Zhang\thanks{Rutgers University, sen.z@rutgers.edu}, 
    Ming Liu\thanks{Purdue University, liu3183@purdue.edu}, 
    Jiawei Xu\thanks{Purdue University, xu1644@purdue.edu}, 
    Xuanhe Pan\thanks{University of Wisconsin-Madison, xpan73@wisc.edu}, 
    Jinlang Wang\thanks{University of Wisconsin-Madison, jinlang.wang@wisc.edu}, 
    Pohsun Feng\thanks{National Taiwan Normal University, 41075018h@ntnu.edu.tw}, 
    Yizhu Wen\thanks{University of Hawaii, yizhuw@hawaii.edu}, 
    Lawrence K.Q. Yan\thanks{Hong Kong University of Science and Technology, kqyan@connect.ust.hk}, 
    Hong-Ming Tseng\thanks{School of Visual Arts, htseng@sva.edu}, 
    Xinyuan Song\thanks{Emory University, songxinyuan@pku.edu.cn}, 
    Jintao Ren\thanks{Aarhus University, jintaoren@clin.au.dk}, 
    Silin Chen\thanks{Zhejiang University, 
    A1033439225@gmail.com}, 
    Yunze Wang\thanks{University of Edinburgh, 
    Y.Wang-861@sms.ed.ac.uk}, 
    Weiche Hsieh\thanks{National Tsing Hua University, s112033645@m112.nthu.edu.tw}, 
    Bowen Jing\thanks{University of Manchester, bowen.jing@postgrad.manchester.ac.uk}, 
    Junjie Yang\thanks{Pingtan Research Institute of Xiamen University, youngboy@xmu.edu.cn}, 
    Jun Zhou\thanks{The University of Texas at Dallas, jun.zhou.tx@gmail.com},
    Zheyu Yao\thanks{University of Liverpool, pszyao2@liverpool.ac.uk},
    Chia Xin Liang\thanks{JTB Technology Corp., cxldun@gmail.com}
}

\date{}  

\begin{document}

\maketitle

\begingroup
    \renewcommand\thefootnote{}\footnote{
    \textsuperscript{*} Equal contribution. \\
    \textsuperscript{$\dagger$} Corresponding author.
}
\addtocounter{footnote}{0}
\endgroup

\epigraph{"Even today's networks, which we consider quite large from a computational systems point of view, are smaller than the nervous system of even relatively primitive vertebrate animals like frogs."}{\textit{Ian Goodfellow}}

\tableofcontents  

\part{Basic Theories}

\chapter{Fundamentals of Generative Adversarial Networks}
\section{Definition and Background of GANs}
Generative Adversarial Networks (GANs)~\cite{goodfellow2014generative} are one of the most groundbreaking advancements in machine learning, particularly in the field of unsupervised learning. Introduced by Ian Goodfellow~\cite{goodfellow2014generative} and his colleagues in 2014, GANs represent a novel approach to generating data that looks similar to the data the model was trained~\cite{creswell2018generative,metz2016unrolled}.

\begin{figure}[htbp]
    \centering
    \includegraphics[width=0.8\textwidth]{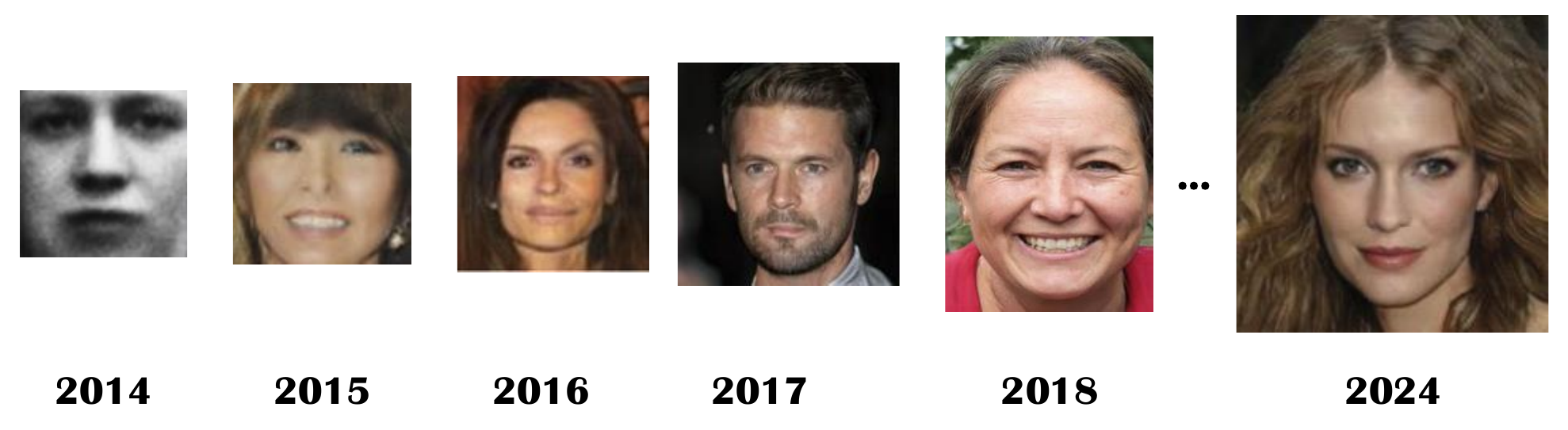}
    \caption{Evolution of GAN performance from 2014 to 2018 and 2024. The results for 2014 to 2018 are based on the demonstration by Goodfellow~\cite{goodfellow2014generative} at the International Conference on Learning Representations (ICLR) 2019 invited talk, showcasing the rapid advancements in GAN quality over the years~\cite{radford2015unsupervised, liu2016coupled, karras2017progressive, karras2019style}. The figure of 2024 from ISFB-GAN~\cite{peng2024isfb}.}
\end{figure}

\subsection{Definition of GAN}
At its core, a GAN consists of two neural networks, referred to as the \textbf{generator} and the \textbf{discriminator}, which are pitted against each other in a zero-sum game~\cite{goodfellow2014generative,wang2017generative}. The generator attempts to create fake data that resembles the real data, while the discriminator tries to distinguish between real and fake data~\cite{aggarwal2021generative, gui2021review}. These two networks are trained simultaneously:

\begin{itemize}
    \item \textbf{Generator:} A neural network that takes random noise as input and attempts to generate data that mimics the real dataset~\cite{goodfellow2014generative}.
    \item \textbf{Discriminator:} Another neural network that evaluates the data and determines whether the input data is from the real dataset or generated by the generator~\cite{goodfellow2014generative,gui2021review}.
\end{itemize}

The goal is to train the generator to the point where the discriminator can no longer reliably distinguish between real and fake data.

\subsection{Historical Development of GANs}
The journey of GANs began with the work of Ian Goodfellow in 2014~\cite{goodfellow2014generative}, but the concepts that led to GANs can be traced back to earlier advancements in deep learning and neural networks~\cite{gui2021review}. Here's a brief overview of the major milestones in the history of GANs:

\begin{itemize}
    \item \textbf{2014:} GAN was introduced by Ian Goodfellow. In the original paper, the authors proposed a novel framework for generative models~\cite{goodfellow2014generative}.
    \item \textbf{2016:} The introduction of techniques like Wasserstein GAN (WGAN)~\cite{arjovsky2017wasserstein} helped to improve the stability of training, which was a significant issue in the early implementations~\cite{gui2021review}.
    \item \textbf{2017:} Progressive Growing of GANs (PGGANs)~\cite{karras2017progressive} was proposed, enabling the generation of high-resolution images.
    \item \textbf{2018:} GANs were used to generate high-quality human faces~\cite{gui2021review,park2019semantic} (e.g., StyleGAN~\cite{karras2019style}).
    \item \textbf{2019:} BigGAN~\cite{brock2018large} introduced large-scale training and incorporated self-attention mechanisms, achieving significant improvements in image quality and diversity. It set a new benchmark for high-resolution image generation.
    \item \textbf{2020:} StyleGAN2~\cite{karras2020analyzing} enhanced its predecessor by improving normalization techniques and architecture design, leading to more realistic images and reducing artifacts in the generated outputs.
    \item \textbf{2021:} GauGAN~\cite{park2019semantic} demonstrated the ability of GANs to transform simple sketches into photorealistic images, showcasing the strength of GANs in semantic image synthesis. It became a powerful tool for creative applications.
    \item \textbf{2022:} StyleGAN3~\cite{karras2021alias} addressed aliasing artifacts present in earlier versions and improved the spatial consistency of generated images under transformations, achieving higher-quality and more stable outputs.
    \item \textbf{2023:} GigaGAN~\cite{kang2023scaling} introduced a scalable architecture capable of generating ultra-high-resolution images with improved quality and significantly faster generation times, advancing the frontier of GAN research.
    \item \textbf{2024:} GAN (SparseGAN~\cite{zhou2020sparse}) focused on reducing computational overhead while maintaining high-quality image generation. By leveraging sparsity in network design, it provided a more efficient approach to GAN training and deployment.
\end{itemize}

\subsection{Comparison Between GAN and Traditional Generative Models}
To understand the significance of GANs, it's important to compare them with more traditional generative models like Variational Autoencoders (VAEs)~\cite{kingma2013auto,kingma2019introduction} and Restricted Boltzmann Machines (RBMs)~\cite{ackley1985learning,nair2010rectified}.

\subsubsection{Traditional Generative Models}
Before the advent of GANs, most generative models relied on certain assumptions or simplifications in modeling the data distribution~\cite{aggarwal2021generative}. For instance:

\begin{itemize}
    \item \textbf{Restricted Boltzmann Machines (RBM):} RBMs are energy-based models that learn a probability distribution~\cite{ramberg1979probability} over input data. They were widely used for dimensionality reduction~\cite{van2009dimensionality} and pretraining of deep networks~\cite{nair2010rectified,erhan2010does}.
    \item \textbf{Variational Autoencoders (VAE):} VAEs aim to learn latent representations of data by optimizing the Evidence Lower Bound (ELBO)~\cite{hoffman2016elbo} and using a combination of an encoder and decoder architecture~\cite{kingma2019introduction}.
\end{itemize}

While these methods were effective in some tasks, they have notable limitations:

\begin{itemize}
    \item VAEs often produce blurry outputs due to their reliance on the Gaussian distribution~\cite{goodman1963statistical} in latent space~\cite{hoff2002latent}.
    \item RBMs have limitations in terms of scalability and convergence~\cite{nair2010rectified}.
\end{itemize}

\subsubsection{Advantages of GANs over Traditional Models}
GANs differ from traditional generative models primarily in their adversarial training approach~\cite{goodfellow2014generative,creswell2018generative,aggarwal2021generative,gui2021review}. Instead of relying on a fixed probability distribution, GANs employ the generator and discriminator in a game-theoretic setup~\cite{goodfellow2014generative,aggarwal2021generative}. The advantages of GANs include:

\begin{itemize}
    \item \textbf{High-quality data generation:} GANs often generate sharper and more realistic outputs than VAEs~\cite{wang2017generative}.
    \item \textbf{Flexibility:} GANs do not require explicit probability distributions for their outputs, making them more flexible in generating various types of data~\cite{kazeminia2020gans}.
    \item \textbf{Adversarial training:} The discriminator provides a continuous feedback loop to the generator, leading to improved performance over time~\cite{aggarwal2021generative}.
\end{itemize}

\section{Understanding GAN with Python: A Simple Example}
In this section, we will implement a simple GAN using PyTorch~\cite{paszke2019pytorch}, a popular deep learning framework. For the purpose of illustration, let's consider a basic problem: generating a distribution that mimics the behavior of a 1D Gaussian distribution~\cite{stevens2020deep}.

First, we need to set up the environment and libraries:

\begin{lstlisting}[style=cmd]
pip install torch torchvision matplotlib numpy
\end{lstlisting}

Next, we implement the generator and discriminator models:

\subsection{Step 1: Import Necessary Libraries}
We begin by importing the required libraries:

\begin{lstlisting}[style=python]
import torch
import torch.nn as nn
import torch.optim as optim
import numpy as np
import matplotlib.pyplot as plt
\end{lstlisting}

\subsection{Step 2: Define the Generator and Discriminator}
The generator network will take a random noise vector~\cite{gui2021review} as input and output a single scalar. The discriminator, on the other hand, will take a scalar and attempt to classify it as either real (coming from a true Gaussian distribution) or fake (generated by the generator).

\begin{lstlisting}[style=python]
# Define the Generator model
class Generator(nn.Module):
    def __init__(self, input_size, hidden_size, output_size):
        super(Generator, self).__init__()
        self.net = nn.Sequential(
            nn.Linear(input_size, hidden_size),
            nn.ReLU(),
            nn.Linear(hidden_size, output_size)
        )

    def forward(self, x):
        return self.net(x)

# Define the Discriminator model
class Discriminator(nn.Module):
    def __init__(self, input_size, hidden_size, output_size):
        super(Discriminator, self).__init__()
        self.net = nn.Sequential(
            nn.Linear(input_size, hidden_size),
            nn.ReLU(),
            nn.Linear(hidden_size, output_size),
            nn.Sigmoid()
        )

    def forward(self, x):
        return self.net(x)
\end{lstlisting}

\subsection{Step 3: Training the GAN}
Now that we have our models defined, we will train the GAN. The generator will learn to generate samples that match a target 1D Gaussian distribution~\cite{goodman1963statistical}, while the discriminator will try to distinguish between real and fake samples~\cite{goodfellow2014generative}.

First, let's initialize the models and the optimizers:

\begin{lstlisting}[style=python]
# Hyperparameters
input_size = 1
hidden_size = 128
output_size = 1
learning_rate = 0.0002

# Create the models
generator = Generator(input_size, hidden_size, output_size)
discriminator = Discriminator(output_size, hidden_size, 1)

# Optimizers
optimizer_g = optim.Adam(generator.parameters(), lr=learning_rate)
optimizer_d = optim.Adam(discriminator.parameters(), lr=learning_rate)

# Loss function
loss_function = nn.BCELoss()
\end{lstlisting}

We'll generate real data from a Gaussian distribution and train the discriminator and generator iteratively:

\begin{lstlisting}[style=python]
# Training loop
num_epochs = 5000
real_data_mean = 4
real_data_stddev = 1.25

for epoch in range(num_epochs):
    # Train Discriminator: maximize log(D(x)) + log(1 - D(G(z)))
    real_data = torch.randn(32, 1) * real_data_stddev + real_data_mean
    fake_data = generator(torch.randn(32, 1)).detach()

    real_labels = torch.ones(32, 1)
    fake_labels = torch.zeros(32, 1)

    d_loss_real = loss_function(discriminator(real_data), real_labels)
    d_loss_fake = loss_function(discriminator(fake_data), fake_labels)
    d_loss = d_loss_real + d_loss_fake

    optimizer_d.zero_grad()
    d_loss.backward()
    optimizer_d.step()

    # Train Generator: minimize log(1 - D(G(z))) or maximize log(D(G(z)))
    noise = torch.randn(32, 1)
    g_loss = loss_function(discriminator(generator(noise)), real_labels)

    optimizer_g.zero_grad()
    g_loss.backward()
    optimizer_g.step()

    if epoch % 1000 == 0:
        print(f'Epoch [{epoch}/{num_epochs}], d_loss: {d_loss.item():.4f}, g_loss: {g_loss.item():.4f}')
\end{lstlisting}
In this code, the generator is trained to improve its ability to fool the discriminator by producing outputs that resemble the real data distribution. The discriminator, in turn, is trained to distinguish between real and fake samples.

\section{Summary}
In this chapter, we explored the basic concepts of GANs, including their definition, historical development, and advantages over traditional generative models~\cite{goodfellow2014generative,gui2021review}. We also implemented a simple GAN using PyTorch to generate a 1D Gaussian distribution, providing a practical example for beginners to understand how GANs work. By progressively refining both the generator and discriminator, the GAN is able to learn to produce realistic data~\cite{kazeminia2020gans}.

\section{GAN's Basic Structure}
Generative Adversarial Networks (GANs) are a class of machine learning frameworks designed to generate data similar to a given dataset. GANs consist of two primary components: the Generator and the Discriminator, both of which are neural networks that compete against each other in a zero-sum game~\cite{goodfellow2014generative,aggarwal2021generative}. This section will explain each component in detail and describe how they interact with each other.

\subsection{Generator}
The Generator is responsible for generating synthetic data that resembles real data from the dataset. Its goal is to learn the distribution of the real data and produce samples that the Discriminator cannot distinguish from real samples.

The Generator starts with random noise, typically drawn from a Gaussian or uniform distribution, and transforms it into data (e.g., an image) using a series of neural network layers~\cite{goodfellow2014generative}. Initially, the Generator's output will not resemble real data, but as it gets trained, it gradually improves.

\textbf{Example of a Generator's forward pass in PyTorch:}

\begin{lstlisting}[style=python]
import torch
import torch.nn as nn

class Generator(nn.Module):
    def __init__(self, input_dim, output_dim):
        super(Generator, self).__init__()
        self.model = nn.Sequential(
            nn.Linear(input_dim, 128),   # Input layer (noise)
            nn.ReLU(),
            nn.Linear(128, 256),         # Hidden layer
            nn.ReLU(),
            nn.Linear(256, output_dim),  # Output layer (generated data)
            nn.Tanh()                    # Activation function for image generation
        )

    def forward(self, x):
        return self.model(x)

# Example usage:
noise = torch.randn((1, 100))  # 100-dimensional random noise
gen = Generator(100, 784)      # 784 = 28x28 pixels (for image generation)
generated_data = gen(noise)
print(generated_data.shape)    # Should output: torch.Size([1, 784])
\end{lstlisting}

In the above example, the Generator network consists of several fully connected (linear) layers with ReLU activations, except for the output layer where we use a Tanh activation function. Tanh is commonly used when generating image data because it restricts the output to values between -1 and 1, matching the normalized pixel values of images.

\subsection{Discriminator}
The Discriminator's role is to distinguish between real data (from the dataset) and the fake data produced by the Generator. It outputs a probability indicating whether it believes a given sample is real or fake~\cite{goodfellow2014generative}. Its objective is to maximize the accuracy of distinguishing between real and fake data.

\textbf{Example of a Discriminator's forward pass in PyTorch:}

\begin{lstlisting}[style=python]
class Discriminator(nn.Module):
    def __init__(self, input_dim):
        super(Discriminator, self).__init__()
        self.model = nn.Sequential(
            nn.Linear(input_dim, 256),   # Input layer (real or fake data)
            nn.LeakyReLU(0.2),
            nn.Linear(256, 128),         # Hidden layer
            nn.LeakyReLU(0.2),
            nn.Linear(128, 1),           # Output layer (real/fake probability)
            nn.Sigmoid()                 # Sigmoid function to get probability
        )

    def forward(self, x):
        return self.model(x)

# Example usage:
disc = Discriminator(784)  # Assuming the input is a 28x28 image flattened to 784 dimensions
real_data = torch.randn((1, 784))  # A real data sample from the dataset
discriminator_output = disc(real_data)
print(discriminator_output)  # Outputs a probability value between 0 and 1
\end{lstlisting}

The Discriminator network is also a fully connected neural network, but its final layer uses a Sigmoid activation function, which outputs a value between 0 and 1, representing the probability that the input is real.

\subsection{The Adversarial Game Between Generator and Discriminator}
GANs are based on a two-player game between the Generator and the Discriminator. The Generator tries to fool the Discriminator by producing data that is as close as possible to real data. Meanwhile, the Discriminator is trained to correctly classify real and fake data~\cite{goodfellow2014generative}.

The training process is adversarial, meaning the Generator improves by learning how to trick the Discriminator, and the Discriminator improves by becoming better at spotting fake data.

\subsubsection{Loss Functions}
The standard loss functions for GANs are:

\textbf{Generator Loss}: The Generator aims to minimize the following loss function~\cite{goodfellow2014generative}:
  \[
  \mathcal{L}_{G} = -\log(D(G(z)))
  \]
  Here, \( G(z) \) represents the fake data generated from noise \( z \), and \( D(G(z)) \) is the Discriminator's estimate of the probability that the fake data is real. The Generator wants to maximize this probability.

\textbf{Discriminator Loss}: The Discriminator aims to maximize the following loss function:
  \[
  \mathcal{L}_{D} = -\left( \log(D(x)) + \log(1 - D(G(z))) \right)
  \]
  Here, \( D(x) \) is the Discriminator's estimate that a real sample \( x \) is real, and \( D(G(z)) \) is its estimate that the fake data is real. The Discriminator tries to correctly classify both real and fake data~\cite{kazeminia2020gans}.

\textbf{Example of the training loop in PyTorch:}

\begin{lstlisting}[style=python]
# Hyperparameters
lr = 0.0002
epochs = 10000

# Instantiate models
gen = Generator(input_dim=100, output_dim=784)
disc = Discriminator(input_dim=784)

# Loss and optimizers
criterion = nn.BCELoss()
optimizer_gen = torch.optim.Adam(gen.parameters(), lr=lr)
optimizer_disc = torch.optim.Adam(disc.parameters(), lr=lr)

for epoch in range(epochs):
    # Train Discriminator
    optimizer_disc.zero_grad()

    # Real data
    real_data = torch.randn((64, 784))  # Batch of real data
    real_labels = torch.ones((64, 1))   # Label = 1 for real data
    output_real = disc(real_data)
    loss_real = criterion(output_real, real_labels)

    # Fake data
    noise = torch.randn((64, 100))      # Batch of noise
    fake_data = gen(noise)              # Generated fake data
    fake_labels = torch.zeros((64, 1))  # Label = 0 for fake data
    output_fake = disc(fake_data.detach())
    loss_fake = criterion(output_fake, fake_labels)

    # Total Discriminator loss and backpropagation
    loss_disc = loss_real + loss_fake
    loss_disc.backward()
    optimizer_disc.step()

    # Train Generator
    optimizer_gen.zero_grad()
    
    # Generate fake data again
    output_fake_for_gen = disc(fake_data)
    loss_gen = criterion(output_fake_for_gen, real_labels)  # We want the generator to fool the discriminator

    # Generator backpropagation
    loss_gen.backward()
    optimizer_gen.step()

    if epoch % 1000 == 0:
        print(f'Epoch [{epoch}/{epochs}], Loss D: {loss_disc.item()}, Loss G: {loss_gen.item()}')
\end{lstlisting}

In this code, both the Generator and Discriminator are trained alternately. First, the Discriminator is trained to distinguish between real and fake data. Then, the Generator is updated to produce better fake data that fools the Discriminator.

\subsection{Visualization of the GAN Structure}
Here is a simple tree-like representation of the GAN structure using \texttt{tikzpicture}~\cite{kottwitz2023latex}:

\begin{center}
\begin{tikzpicture}
  [scale=1, every node/.style={scale=1}, 
  block/.style={rectangle, draw, fill=blue!20, text centered, minimum height=3em},
  arrow/.style={->, thick}]

  \node[block] (noise) {Noise (z)};
  \node[block, right=of noise] (gen) {Generator (G)};
  \node[block, right=of gen] (fake) {Fake Data};
  \node[block, below=of fake] (real) {Real Data};
  \node[block, right=of fake] (disc) {Discriminator (D)};
  \node[block, right=of disc] (output) {Real or Fake};

  \draw[arrow] (noise) -- (gen);
  \draw[arrow] (gen) -- (fake);
  \draw[arrow] (real) -- (disc);
  \draw[arrow] (fake) -- (disc);
  \draw[arrow] (disc) -- (output);
\end{tikzpicture}
\end{center}

\begin{figure}[htbp]
    \centering
    \includegraphics[width=0.8\textwidth]{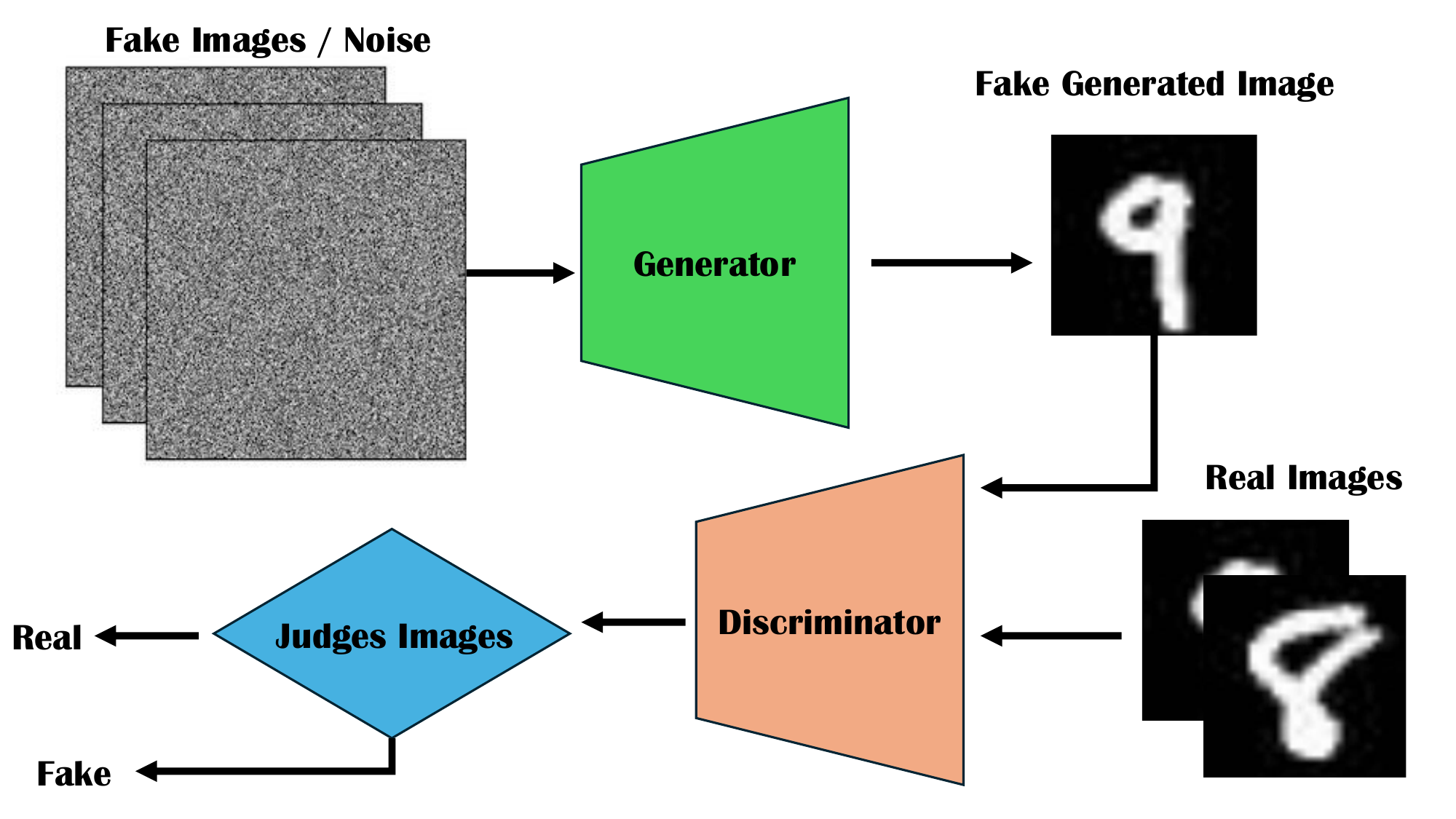}
    \caption{The basic architecture of a Generative Adversarial Network (GAN). The Generator creates fake images from random noise, while the Discriminator evaluates images to determine whether they are real or fake. Both networks are trained adversarially to improve the quality of the generated samples.}
    \label{fig:gan_basic_architecture}
\end{figure}

Generative Adversarial Networks (GANs) are composed of two primary components (as shown Fig \ref{fig:gan_basic_architecture}): a \textbf{Generator} and a \textbf{Discriminator}. These two networks work in opposition to each other to achieve a common goal, generating realistic data.

\begin{itemize}
    \item \textbf{Generator:}
    The Generator takes random noise as input and transforms it into synthetic data that resembles real samples. Its objective is to learn the data distribution and produce outputs that are indistinguishable from the real data. Over the course of training, the Generator improves by learning to ``fool'' the Discriminator.
    
    \item \textbf{Discriminator:}
    The Discriminator acts as a binary classifier, distinguishing between real data samples (from the dataset) and fake data samples (produced by the Generator). Its goal is to maximize its ability to correctly classify inputs as real or fake. 
\end{itemize}

The training process of a GAN is based on a minimax optimization game, where the Generator minimizes the classification performance of the Discriminator, while the Discriminator maximizes its classification accuracy. This dynamic adversarial process leads to the Generator creating more realistic data over time, eventually achieving a balance where the Discriminator can no longer reliably distinguish real from fake samples.

\section{GAN's Objective Function and Optimization}
Generative Adversarial Networks (GANs) are an essential tool for generating synthetic data, and understanding their underlying objective functions is crucial~\cite{kazeminia2020gans}. In this section, we will explore the key objective functions involved in training GANs, with detailed examples and Python code using PyTorch.

\subsection{Binary Cross-Entropy Loss}
In a standard GAN, the generator and discriminator have competing objectives~\cite{goodfellow2014generative}, which can be defined using binary cross-entropy loss~\cite{ho2019real}.

The generator \(G\) tries to generate data that resembles the true data distribution, while the discriminator \(D\) aims to distinguish between real data and fake data generated by \(G\).

The loss for the discriminator can be defined as:
\[
\mathcal{L}_D = - \mathbb{E}_{x \sim p_{data}}[\log D(x)] - \mathbb{E}_{z \sim p_z}[\log (1 - D(G(z)))]
\]
Here, \(x \sim p_{data}\) represents samples from the real data distribution, and \(z \sim p_z\) are random noise vectors input to the generator~\cite{aggarwal2021generative}.

For the generator, the objective is to maximize the discriminator's error in classifying fake samples:
\[
\mathcal{L}_G = - \mathbb{E}_{z \sim p_z}[\log D(G(z))]
\]
This is also referred to as minimizing the negative log-likelihood~\cite{pinheiro1995approximations} of fooling the discriminator.

\begin{lstlisting}[style=python]
import torch
import torch.nn as nn
import torch.optim as optim

# Define the Discriminator model
class Discriminator(nn.Module):
    def __init__(self):
        super(Discriminator, self).__init__()
        self.main = nn.Sequential(
            nn.Linear(784, 512),
            nn.LeakyReLU(0.2),
            nn.Linear(512, 256),
            nn.LeakyReLU(0.2),
            nn.Linear(256, 1),
            nn.Sigmoid()
        )

    def forward(self, x):
        return self.main(x)

# Define the Generator model
class Generator(nn.Module):
    def __init__(self):
        super(Generator, self).__init__()
        self.main = nn.Sequential(
            nn.Linear(100, 256),
            nn.ReLU(),
            nn.Linear(256, 512),
            nn.ReLU(),
            nn.Linear(512, 784),
            nn.Tanh()
        )

    def forward(self, x):
        return self.main(x)

# Instantiate models
D = Discriminator()
G = Generator()

# Binary Cross-Entropy loss and optimizers
criterion = nn.BCELoss()
optimizer_D = optim.Adam(D.parameters(), lr=0.0002)
optimizer_G = optim.Adam(G.parameters(), lr=0.0002)

# Labels for real and fake data
real_label = torch.ones(64, 1)
fake_label = torch.zeros(64, 1)

# Example training step
for epoch in range(epochs):
    # Train the Discriminator
    optimizer_D.zero_grad()
    
    # Real data loss
    real_data = torch.randn(64, 784)  # Random real data for demonstration
    output = D(real_data)
    loss_real = criterion(output, real_label)
    
    # Fake data loss
    noise = torch.randn(64, 100)
    fake_data = G(noise)
    output = D(fake_data.detach())
    loss_fake = criterion(output, fake_label)
    
    # Total discriminator loss and backward
    loss_D = loss_real + loss_fake
    loss_D.backward()
    optimizer_D.step()
    
    # Train the Generator
    optimizer_G.zero_grad()
    output = D(fake_data)
    loss_G = criterion(output, real_label)
    
    loss_G.backward()
    optimizer_G.step()
\end{lstlisting}

\subsection{JS Divergence and KL Divergence}
A fundamental aspect of GANs is their reliance on the Jensen-Shannon (JS) divergence~\cite{gomez2000analysis} to measure the difference between the real data distribution and the distribution generated by \(G\). The JS divergence is a symmetrized and smoothed version of the Kullback-Leibler (KL) divergence~\cite{hershey2007approximating}, which is defined as:
\[
D_{KL}(P || Q) = \sum_{i} P(i) \log \left( \frac{P(i)}{Q(i)} \right)
\]
The KL divergence is non-symmetric and becomes infinite if there are samples that exist in \(P\) but not in \(Q\). This limitation can cause instability in GAN training~\cite{goodfellow2014generative}.

The JS divergence addresses this issue by computing the average between the real and fake distributions:
\[
JS(P || Q) = \frac{1}{2} D_{KL}(P || M) + \frac{1}{2} D_{KL}(Q || M)
\]
where \(M = \frac{1}{2}(P + Q)\) is the mixture distribution. GANs implicitly minimize the JS divergence between the real and generated data distributions during training.

\section{Training and Challenges of GANs}
Training GANs is challenging due to several issues, including mode collapse, vanishing gradients, and instability. Let's explore these problems and discuss how to mitigate them.

\subsection{Mode Collapse}
Mode collapse~\cite{srivastava2017veegan} occurs when the generator produces limited varieties of samples, even though the training data is diverse. In this situation, the generator might map multiple input noise vectors \(z\) to the same output, causing a lack of diversity in the generated data~\cite{thanh2020catastrophic}.

For example, suppose the real data consists of different images of digits (e.g., 0-9), but the generator only produces images of the digit '5'. This is mode collapse.

One common technique to mitigate mode collapse is to use \textbf{Mini-batch Discrimination}, where the discriminator looks at small batches of data instead of individual samples~\cite{thanh2020catastrophic}. This allows the discriminator to detect when the generator is producing similar outputs for different inputs.

\subsection{Vanishing Gradient and Instability}
Another challenge in GAN training is vanishing gradients~\cite{hochreiter1998vanishing}. If the discriminator becomes too strong, the generator may receive extremely small gradient updates, making it difficult to improve~\cite{hanin2018neural}.

This problem can be mitigated using techniques such as:
\begin{itemize}
    \item \textbf{Label Smoothing}: Instead of using binary labels (0 and 1), use slightly smoothed labels like 0.9 for real data and 0.1 for fake data. This helps the discriminator avoid becoming too confident and dominating the training process~\cite{muller2019does}.
    \item \textbf{Batch Normalization}: Adding batch normalization layers to the generator and discriminator helps stabilize the training process by normalizing activations and preventing exploding or vanishing gradients~\cite{santurkar2018does,bjorck2018understanding}.
    \item \textbf{Improved Loss Functions}: Using alternative loss functions such as the \textbf{Wasserstein loss}~\cite{frogner2015learning} can lead to more stable training and better convergence. The \textbf{Wasserstein loss} will be discussed in more detail in a later section.
\end{itemize}

\subsection{Techniques in Adversarial Training}
Training GANs effectively requires several advanced techniques. Below are some important practices:
\begin{center}
\begin{tikzpicture}[level distance=2cm, sibling distance=4cm, edge from parent/.style={draw,-latex}]
  \node[rectangle, draw] {GAN Training Challenges}
    child {node[rectangle, draw] {Mode Collapse}}
    child {node[rectangle, draw] {Vanishing Gradients}}
    child {node[rectangle, draw] {Instability}};
\end{tikzpicture}
\end{center}
\begin{itemize}
    \item \textbf{Feature Matching}: Instead of trying to fool the discriminator, the generator can be trained to match the intermediate features of the real and fake data produced by the discriminator. This can lead to more diverse and realistic samples~\cite{gui2021review}.
    \item \textbf{Progressive Growing}: This technique involves starting with a small model (low resolution) and gradually increasing the model size and resolution as training progresses. It helps GANs learn high-resolution images efficiently~\cite{karras2017progressive}.
    \item \textbf{Noise Injection}: Adding noise to the inputs of the discriminator or generator can help regularize the model, making it less sensitive to small variations in the data, and can improve the generalization of the network~\cite{hoff2002latent}.
\end{itemize}

\chapter{Theoretical Foundations of GANs}
In this chapter, we will explore the fundamental theoretical concepts that form the backbone of Generative Adversarial Networks (GANs). The understanding of GANs requires a firm grasp of probability theory~\cite{he2018probgan}, statistics~\cite{bau2019seeing, ma2021must}, and game theory~\cite{fudenberg1991game}. Additionally, we will examine the concept of Nash Equilibrium~\cite{daskalakis2009complexity, heusel2017gans, farnia2020gans} in the context of GANs, as it plays a critical role in the convergence of the model during training.

\section{Fundamentals of Probability Theory and Statistics}
Probability theory~\cite{he2018probgan} and statistics provide the mathematical foundation for modeling uncertainty in machine learning. In the context of GANs, these concepts help us define the distributions from which data is sampled, as well as how to measure the likelihood of certain outcomes~\cite{farnia2020gans}.

\subsection{Random Variables and Distributions}
In GANs, the generator typically learns to map random noise, sampled from a specific distribution, to a distribution that mimics real data~\cite{goodfellow2014generative}. A random variable is a quantity that can take on different values, each with a specific probability. For example, a simple random variable \( Z \) could represent a Gaussian noise vector:

\[
Z \sim \mathcal{N}(0, 1)
\]

This means that \( Z \) is sampled from a normal distribution with mean $0$ and variance $1$~\cite{goodfellow2014generative, kazeminia2020gans}. In practice, the generator in a GAN takes such noise as input and transforms it into data that approximates the real distribution \( p_{\text{data}} \).

\subsection{Expectation and Variance}
Two important concepts in probability theory~\cite{he2018probgan} are expectation and variance, which help quantify the behavior of random variables:

\begin{itemize}
    \item \textbf{Expectation:} The expected value (or mean) of a random variable represents the average outcome of a large number of samples. For a random variable \( X \), the expectation is denoted as \( \mathbb{E}[X] \).
    \item \textbf{Variance:} The variance measures the spread of a random variable's values around the mean. It is defined as \( \mathbb{V}(X) = \mathbb{E}[(X - \mathbb{E}[X])^2] \).
\end{itemize}

These concepts are useful in GANs when analyzing the output of the generator and the real data (Fig. \ref{fig:gan_basic_architecture}). The generator attempts to produce samples that have similar statistical properties (such as mean and variance) to the real data~\cite{srivastava2017veegan, he2018probgan}.

\subsection{Probability Density Functions (PDF)}
The probability density function (PDF) describes the likelihood of a continuous random variable taking on a specific value~\cite{parzen1962estimation, saatci2017bayesian}. For example, the PDF of a Gaussian distribution is given by:

\[
p(x) = \frac{1}{\sqrt{2 \pi \sigma^2}} \exp\left(-\frac{(x - \mu)^2}{2 \sigma^2}\right)
\]

In GANs, we often want the generated samples to follow a specific probability distribution~\cite{goodfellow2014generative}, such as a Gaussian~\cite{goodman1963statistical} or uniform~\cite{kuipers2012uniform} distribution, and we measure how closely the generated samples approximate the real data distribution using statistical metrics like the Jensen-Shannon divergence~\cite{menendez1997jensen, fuglede2004jensen}.

\section{Game Theory and Optimal Equilibria}
Game theory is a critical component of understanding the adversarial nature of GANs~\cite{fudenberg1991game, farnia2020gans}. GANs can be viewed as a game between two players: the generator and the discriminator~\cite{goodfellow2014generative}. To fully comprehend the dynamics of this interaction, we need to understand key concepts in game theory, particularly the notion of equilibrium.

\subsection{Basic Concepts of Game Theory}
In game theory, players make decisions that influence each other's outcomes. In GANs, the generator and the discriminator can be considered as two players engaged in a zero-sum game~\cite{friedman1998work}, where one player's gain is the other player's loss.

\begin{itemize}
    \item \textbf{Players:} In GANs, the two players are the generator (G) and the discriminator (D)~\cite{goodfellow2014generative, kazeminia2020gans}.
    \item \textbf{Strategies:} The generator's strategy is to create data that can fool the discriminator, while the discriminator's strategy is to correctly classify real versus fake data.
    \item \textbf{Payoffs:} The payoff for the generator is based on how well it can fool the discriminator. The payoff for the discriminator is based on how accurately it can classify the data~\cite{goodfellow2014generative}.
\end{itemize}

The goal of this game is to find an equilibrium where neither player can improve their strategy without the other player's strategy changing~\cite{fudenberg1991game}.

\subsection{Zero-Sum Games}
A GAN can be viewed as a zero-sum game. In such games, the total gain of all players is zero~\cite{friedman1998work}. In other words, the gain of one player is exactly offset by the loss of the other. The objective of the discriminator is to minimize the following loss function:

\[
\min_D \mathbb{E}_{x \sim p_{\text{data}}(x)}[\log D(x)] + \mathbb{E}_{z \sim p_z(z)}[\log (1 - D(G(z)))]
\]

At the same time, the generator tries to maximize the discriminator's loss:

\[
\max_G \mathbb{E}_{z \sim p_z(z)}[\log D(G(z))]
\]

This adversarial process creates a dynamic where both networks are constantly improving in response to each other~\cite{farnia2020gans}.

\section{Nash Equilibrium in GANs}
The concept of \textbf{Nash Equilibrium} is central to understanding the training dynamics of GANs~\cite{farnia2020gans}. A Nash Equilibrium occurs in a game when no player can improve their strategy unilaterally, assuming that the other player's strategy remains fixed. In the context of GANs, Nash Equilibrium represents the ideal state where the generator has learned to generate data that is indistinguishable from the real data, and the discriminator is no longer able to tell the difference between real and fake data~\cite{farnia2020gans}.

\subsection{Formal Definition of Nash Equilibrium}
For a two-player game, a Nash Equilibrium occurs when both players adopt strategies such that neither player can improve their outcome by changing their strategy unilaterally. In GANs, this means that:

\begin{itemize}
    \item The discriminator is optimized to correctly classify real and fake data, given the generator's current strategy~\cite{ho2019real}.
    \item The generator is optimized to produce realistic data, given the discriminator's current strategy.
\end{itemize}

\begin{table}[htbp]
    \centering
    \begin{tabular}{|c|c|c|}
        \hline
        \textbf{Player 1 (Discriminator)} & \textbf{Stick with Strategy} & \textbf{Change Strategy} \\
        \hline
        \textbf{Player 2 (Generator)} &  &  \\
        \textbf{Stick with Strategy} & \textbf{(Nash Equilibrium)} & Player 1 improves \\
        \hline
        \textbf{Change Strategy} & Player 2 improves & Both players adjust \\
        \hline
    \end{tabular}
    \caption{Nash Equilibrium for a Two-Player Game.}
\end{table}

At Nash Equilibrium, the discriminator's performance is no better than random guessing, and the generator has effectively learned to mimic the real data distribution~\cite{farnia2020gans}.

\subsection{Challenges in Reaching Nash Equilibrium in GANs}
While the concept of Nash Equilibrium is theoretically appealing, in practice, reaching equilibrium in GANs can be challenging. Some common issues include:

\begin{itemize}
    \item \textbf{Mode collapse:} The generator might produce a limited variety of samples, focusing on a few modes of the real data distribution, causing the model to collapse to a narrow range of outputs~\cite{he2018probgan}.
    \item \textbf{Non-convergence:} GAN training can be unstable, with the generator and discriminator failing to reach a steady state.
    \item \textbf{Vanishing gradients:} The discriminator may become too strong, making it difficult for the generator to learn effectively due to diminishing gradient updates~\cite{farnia2020gans}.
\end{itemize}

\subsection{Example of Nash Equilibrium in GANs}
Let's consider an example in which the generator and discriminator reach Nash Equilibrium~\cite{kazeminia2020gans}. In a simplified scenario, suppose we are generating a 1D Gaussian distribution with mean \( \mu = 4 \) and standard deviation \( \sigma = 1.25 \).

Initially, the generator might produce random samples that look nothing like the real data. The discriminator will easily classify these as fake~\cite{goodfellow2014generative, farnia2020gans}. Over time, the generator improves by producing samples closer to the real distribution, and the discriminator becomes less certain in its classifications.

Once Nash Equilibrium is reached, the generator produces samples that closely match the real distribution, and the discriminator's accuracy drops to 50\%, which is no better than random guessing~\cite{farnia2020gans}.

\subsection{Training GANs to Approach Nash Equilibrium}
In practical GAN training, we aim to iteratively update the generator and discriminator in a way that moves them toward Nash Equilibrium. This process can be seen in the alternating optimization steps of GAN training.

Here's an illustrative training loop in Python using PyTorch:

\begin{lstlisting}[style=python]
import torch
import torch.nn as nn
import torch.optim as optim

# Define the Generator and Discriminator models (simplified for illustration)
class Generator(nn.Module):
    def __init__(self):
        super(Generator, self).__init__()
        self.net = nn.Sequential(
            nn.Linear(100, 128),
            nn.ReLU(),
            nn.Linear(128, 1)
        )

    def forward(self, x):
        return self.net(x)

class Discriminator(nn.Module):
    def __init__(self):
        super(Discriminator, self).__init__()
        self.net = nn.Sequential(
            nn.Linear(1, 128),
            nn.ReLU(),
            nn.Linear(128, 1),
            nn.Sigmoid()
        )

    def forward(self, x):
        return self.net(x)

# Initialize models
generator = Generator()
discriminator = Discriminator()

# Optimizers
optimizer_g = optim.Adam(generator.parameters(), lr=0.0002)
optimizer_d = optim.Adam(discriminator.parameters(), lr=0.0002)

# Loss function
loss_function = nn.BCELoss()

# Training loop (simplified)
for epoch in range(10000):
    # Generate fake data
    noise = torch.randn(32, 100)
    fake_data = generator(noise)

    # Train Discriminator
    real_data = torch.randn(32, 1) * 1.25 + 4
    real_labels = torch.ones(32, 1)
    fake_labels = torch.zeros(32, 1)

    d_loss_real = loss_function(discriminator(real_data), real_labels)
    d_loss_fake = loss_function(discriminator(fake_data.detach()), fake_labels)
    d_loss = d_loss_real + d_loss_fake

    optimizer_d.zero_grad()
    d_loss.backward()
    optimizer_d.step()

    # Train Generator
    g_loss = loss_function(discriminator(fake_data), real_labels)

    optimizer_g.zero_grad()
    g_loss.backward()
    optimizer_g.step()

    if epoch % 1000 == 0:
        print(f'Epoch [{epoch}/10000], d_loss: {d_loss.item()}, g_loss: {g_loss.item()}')
\end{lstlisting}

In this example, the generator and discriminator are trained in alternating steps, with the goal of approaching Nash Equilibrium. The generator improves its ability to produce realistic samples, while the discriminator becomes increasingly uncertain about classifying them as real or fake~\cite{farnia2020gans}.

\section{Summary}
In this chapter, we explored the theoretical foundations of GANs, focusing on probability theory, statistics, and game theory. Understanding these concepts is crucial for grasping how GANs function. We also delved into the concept of Nash Equilibrium and how it relates to GAN training. Although reaching equilibrium in practice can be difficult, it serves as the theoretical goal of GAN training, where both the generator and discriminator reach a balanced state.

\section{Learning Distributions and Generative Models}
In the field of machine learning, one of the key challenges is learning the underlying distribution of a dataset~\cite{quinonero2022dataset}. Generative models, such as GANs, aim to capture this distribution so that they can generate new data points that are similar to the real data~\cite{goodfellow2014generative}. This section explores the concept of learning distributions, particularly in the context of real and generated data, and how GANs approximate the true data distribution.

\subsection{Real Data Distribution vs Generated Data Distribution}
The goal of any generative model is to learn the \textit{real data distribution}, denoted as \( p_{data}(x) \). This distribution describes the likelihood of observing different data points in a given dataset. For example, if we have a dataset of images of handwritten digits~\cite{lecun1998gradient}, \( p_{data}(x) \) represents the probability of seeing different digit images in the dataset.

On the other hand, a generative model, such as the Generator in a GAN, produces a \textit{generated data distribution}, denoted as \( p_{g}(x) \). Initially, this distribution is random because the Generator has no information about the real data. However, as the Generator trains, it updates its parameters to produce data that increasingly resembles the real data~\cite{goodfellow2014generative}.

\subsubsection{Example: Real and Generated Data}
Consider an example where we are working with a dataset of images of cats and dogs. The \textit{real data distribution} \( p_{data}(x) \) captures how likely we are to see a given image of a cat or a dog. For example, it may be more likely to see an image of a cat lying down than a dog standing up.

Initially, the Generator produces images randomly from a \textit{generated data distribution} \( p_g(x) \). These generated images may look like blobs of noise, as the Generator hasn't learned the features of cats or dogs yet. Over time, through training, the Generator's distribution \( p_g(x) \) will start to resemble the real distribution \( p_{data}(x) \), producing images that look increasingly like real cats or dogs~\cite{goodfellow2014generative, heusel2017gans, he2018probgan, farnia2020gans}.

To formalize this, the Generator in a GAN learns a mapping from a simple distribution (e.g., a Gaussian distribution) to the complex real data distribution. Let's visualize this in a simplified flowchart using \texttt{tikzpicture}:

\begin{center}
\begin{tikzpicture}
  [scale=1, every node/.style={scale=1}, 
  block/.style={rectangle, draw, fill=blue!20, text centered, minimum height=3em},
  arrow/.style={->, thick}]

  \node[block] (noise) {Simple Distribution $p_z(z)$};
  \node[block, right=of noise] (gen) {Generator $G(z)$};
  \node[block, right=of gen] (fake) {Generated Data Distribution $p_g(x)$};
  \node[block, below=of fake] (real) {Real Data Distribution $p_{data}(x)$};

  \draw[arrow] (noise) -- (gen);
  \draw[arrow] (gen) -- (fake);
  \draw[dashed, arrow] (fake) -- (real) node[midway, right] {Approximates};
\end{tikzpicture}
\end{center}

In this diagram, the Generator takes noise sampled from a simple distribution, \( p_z(z) \), and maps it to the generated data distribution, \( p_g(x) \). Over time, the goal of the Generator is for \( p_g(x) \) to approximate the real data distribution \( p_{data}(x) \), so the generated data becomes indistinguishable from real data.

\subsection{GAN's Ability to Approximate Data Distributions}
The key strength of GANs lies in their ability to approximate complex data distributions. GANs achieve this by training two neural networks-the Generator and the Discriminator-in an adversarial process~\cite{he2018probgan, farnia2020gans}. The Generator learns to produce data that mimics the real distribution, while the Discriminator learns to distinguish between real and generated data.

\subsubsection{How GANs Learn to Approximate Distributions}
GANs work through the following iterative process~\cite{goodfellow2014generative, he2018probgan, kazeminia2020gans}:

\begin{enumerate}
    \item The Generator takes random noise as input and generates synthetic data.
    \item The Discriminator is presented with both real data (from the dataset) and the generated data. It predicts whether each data sample is real or fake.
    \item The Generator is trained to produce data that maximizes the Discriminator's error, i.e., it tries to generate data that the Discriminator classifies as real.
    \item The Discriminator is trained to minimize its error, i.e., it tries to accurately classify real and fake data.
\end{enumerate}

This adversarial process drives the Generator to improve its approximation of the real data distribution~\cite{aggarwal2021generative}. Over time, the generated data becomes more similar to the real data, and the generated distribution \( p_g(x) \) approaches the real distribution \( p_{data}(x) \).

\textbf{Example of GAN training in PyTorch:}

Here's a step-by-step example of how GANs learn to approximate the data distribution:

\begin{lstlisting}[style=python]
import torch
import torch.nn as nn

# Generator class
class Generator(nn.Module):
    def __init__(self, input_dim, output_dim):
        super(Generator, self).__init__()
        self.model = nn.Sequential(
            nn.Linear(input_dim, 128),
            nn.ReLU(),
            nn.Linear(128, 256),
            nn.ReLU(),
            nn.Linear(256, output_dim),
            nn.Tanh()  # For generating image data normalized between -1 and 1
        )

    def forward(self, x):
        return self.model(x)

# Discriminator class
class Discriminator(nn.Module):
    def __init__(self, input_dim):
        super(Discriminator, self).__init__()
        self.model = nn.Sequential(
            nn.Linear(input_dim, 256),
            nn.LeakyReLU(0.2),
            nn.Linear(256, 128),
            nn.LeakyReLU(0.2),
            nn.Linear(128, 1),
            nn.Sigmoid()  # Outputs probability
        )

    def forward(self, x):
        return self.model(x)

# GAN Training
def train_gan(generator, discriminator, epochs, batch_size, input_dim, data_dim):
    optimizer_gen = torch.optim.Adam(generator.parameters(), lr=0.0002)
    optimizer_disc = torch.optim.Adam(discriminator.parameters(), lr=0.0002)
    criterion = nn.BCELoss()  # Binary Cross Entropy Loss
    
    for epoch in range(epochs):
        # Training Discriminator
        real_data = torch.randn(batch_size, data_dim)  # Example real data
        real_labels = torch.ones(batch_size, 1)  # Real labels
        
        noise = torch.randn(batch_size, input_dim)  # Random noise for Generator
        fake_data = generator(noise)
        fake_labels = torch.zeros(batch_size, 1)  # Fake labels
        
        # Train on real data
        optimizer_disc.zero_grad()
        output_real = discriminator(real_data)
        loss_real = criterion(output_real, real_labels)
        
        # Train on fake data
        output_fake = discriminator(fake_data.detach())
        loss_fake = criterion(output_fake, fake_labels)
        
        loss_disc = loss_real + loss_fake
        loss_disc.backward()
        optimizer_disc.step()
        
        # Training Generator
        optimizer_gen.zero_grad()
        output_fake_for_gen = discriminator(fake_data)
        loss_gen = criterion(output_fake_for_gen, real_labels)  # Try to fool the discriminator
        
        loss_gen.backward()
        optimizer_gen.step()

        if epoch % 1000 == 0:
            print(f"Epoch [{epoch}/{epochs}], Loss D: {loss_disc.item()}, Loss G: {loss_gen.item()}")
            
# Hyperparameters
input_dim = 100
data_dim = 784  # For 28x28 images (e.g., MNIST)
epochs = 10000
batch_size = 64

# Create instances of Generator and Discriminator
gen = Generator(input_dim, data_dim)
disc = Discriminator(data_dim)

# Train the GAN
train_gan(gen, disc, epochs, batch_size, input_dim, data_dim)
\end{lstlisting}

In this example, the Generator takes random noise and gradually learns to map it to data that approximates the real data distribution. The Discriminator tries to identify whether the data is real or generated, and the Generator is updated to fool the Discriminator over time~\cite{kingma2019introduction}.

\subsubsection{Convergence of GANs}
The ideal outcome of this adversarial training process is that the generated data becomes indistinguishable from real data. Mathematically, this means the generated distribution \( p_g(x) \) converges to the real data distribution \( p_{data}(x) \). At this point, the Discriminator cannot tell the difference between real and fake data, and the Generator has successfully learned the true data distribution.

In practice, however, achieving perfect convergence can be difficult due to factors like unstable training, mode collapse (where the Generator produces only a limited variety of samples), and the sensitivity of GANs to hyperparameters~\cite{farnia2020gans}. Researchers continue to develop techniques to address these challenges and improve the performance of GANs~\cite{heusel2017gans, kazeminia2020gans}.

\subsection{Visualizing Distribution Convergence}
Below is a conceptual diagram of how the generated data distribution \( p_g(x) \) approaches the real data distribution \( p_{data}(x) \) during GAN training:

\begin{center}
\resizebox{\textwidth}{!}{%
\begin{tikzpicture}
  [
    every node/.style={font=\small}, 
    block/.style={
      rectangle, 
      draw, 
      fill=blue!20, 
      text centered, 
      minimum height=3em,
      text width=5cm,  
      align=center  
    },
    arrow/.style={->, thick},
    dashed arrow/.style={->, thick, dashed}
  ]

  \node[block] (p_real) {Real Data Distribution \\ $p_{\text{data}}(x)$};
  \node[block, right=6cm of p_real] (p_fake1) {Initial Generated Distribution \\ $p_g(x)$};
  \node[block, right=6cm of p_fake1] (p_fake2) {Converged Generated Distribution \\ $p_g(x)$};

  \draw[arrow] (p_fake1) -- (p_real) node[midway, above] {Approaches with Training};

  \draw[dashed arrow, bend right=20] (p_fake2) to node[midway, below] {At Convergence} (p_real);
\end{tikzpicture}%
}
\end{center}

Initially, the generated data distribution \( p_g(x) \) is far from the real distribution, but over time, it converges closer to \( p_{data}(x) \), allowing the Generator to produce highly realistic samples.

\section{Mathematical Properties of GANs}
Understanding the mathematical properties of GANs is essential for training effective models and addressing the challenges that arise during the learning process. In this section, we will explore the convergence behavior of GANs and the effects of different loss functions in their training~\cite{goodfellow2014generative, he2018probgan, farnia2020gans}.

\subsection{Convergence of GANs}
GAN convergence refers to the point at which the generator and discriminator reach an equilibrium during training~\cite{goodfellow2014generative}. Ideally, at convergence, the generator has learned to produce data that is indistinguishable from real data, and the discriminator cannot reliably distinguish between the two.

\subsubsection{Minimax Game and Nash Equilibrium}
GANs are formulated as a minimax game between the generator \(G\) and the discriminator \(D\). The objective of the generator is to minimize the probability of the discriminator correctly classifying real and fake data, while the discriminator aims to maximize this probability~\cite{farnia2020gans}. Mathematically, this can be expressed as:
\[
\min_G \max_D \mathbb{E}_{x \sim p_{data}}[\log D(x)] + \mathbb{E}_{z \sim p_z}[\log (1 - D(G(z)))]
\]

At convergence, the generator and discriminator should reach a Nash equilibrium, where neither can improve their performance by unilaterally changing their strategy~\cite{fudenberg1991game, farnia2020gans}. In practical terms, this means the discriminator assigns equal probabilities to real and fake data, i.e., \(D(x) = 0.5\) for both real and generated data.

\subsubsection{Challenges in Achieving Convergence}
Achieving convergence in GANs is notoriously difficult due to the dynamic nature of the minimax game. The generator and discriminator continuously adapt to each other, which can result in instability and oscillations instead of convergence. Some common issues include~\cite{heusel2017gans, farnia2020gans}:
\begin{itemize}
    \item \textbf{Non-stationarity}: As the generator and discriminator improve, the optimization landscape changes dynamically, making it difficult to find a stable point.
    \item \textbf{Mode collapse}: The generator may find a shortcut solution by producing limited types of data, which leads to poor generalization.
    \item \textbf{Vanishing gradients}: If the discriminator becomes too powerful, it provides very small gradient updates to the generator, slowing down the learning process.
\end{itemize}

One method to encourage convergence is to maintain a balance between the generator and discriminator. This can be achieved by carefully tuning the learning rates of both networks, adjusting their architectures, and using techniques like \textit{Wasserstein loss}~\cite{frogner2015learning}, which we'll explore later.

\begin{lstlisting}[style=python]
# Example: Tuning learning rates to improve convergence
import torch
import torch.nn as nn
import torch.optim as optim

# Define simple Discriminator and Generator
class Discriminator(nn.Module):
    def __init__(self):
        super(Discriminator, self).__init__()
        self.main = nn.Sequential(
            nn.Linear(784, 512),
            nn.LeakyReLU(0.2),
            nn.Linear(512, 256),
            nn.LeakyReLU(0.2),
            nn.Linear(256, 1),
            nn.Sigmoid()
        )

    def forward(self, x):
        return self.main(x)

class Generator(nn.Module):
    def __init__(self):
        super(Generator, self).__init__()
        self.main = nn.Sequential(
            nn.Linear(100, 256),
            nn.ReLU(),
            nn.Linear(256, 512),
            nn.ReLU(),
            nn.Linear(512, 784),
            nn.Tanh()
        )

    def forward(self, x):
        return self.main(x)

# Instantiate models
D = Discriminator()
G = Generator()

# Binary Cross-Entropy loss
criterion = nn.BCELoss()

# Optimizers with different learning rates for better balance
optimizer_D = optim.Adam(D.parameters(), lr=0.0004)  # Faster learning for Discriminator
optimizer_G = optim.Adam(G.parameters(), lr=0.0001)  # Slower learning for Generator

# Labels
real_label = torch.ones(64, 1)
fake_label = torch.zeros(64, 1)

# Training example loop
for epoch in range(epochs):
    # Discriminator training
    optimizer_D.zero_grad()
    real_data = torch.randn(64, 784)
    output_real = D(real_data)
    loss_real = criterion(output_real, real_label)
    
    noise = torch.randn(64, 100)
    fake_data = G(noise)
    output_fake = D(fake_data.detach())
    loss_fake = criterion(output_fake, fake_label)
    
    loss_D = loss_real + loss_fake
    loss_D.backward()
    optimizer_D.step()
    
    # Generator training
    optimizer_G.zero_grad()
    output_fake = D(fake_data)
    loss_G = criterion(output_fake, real_label)
    
    loss_G.backward()
    optimizer_G.step()
\end{lstlisting}

\subsection{Effects of Different Loss Functions}
The choice of loss function in GANs significantly affects the model's convergence and the quality of generated data. The original GAN formulation uses the \textbf{binary cross-entropy loss}~\cite{ruby2020binary}, but alternative loss functions can be employed to address issues like mode collapse, vanishing gradients, and unstable training~\cite{ho2019real}.

\subsubsection{Binary Cross-Entropy Loss (Standard GAN Loss)}
In the original GAN formulation, the generator and discriminator are trained using the binary cross-entropy loss~\cite{ho2019real, ruby2020binary}. As shown earlier, the loss for the discriminator is:
\[
\mathcal{L}_D = - \mathbb{E}_{x \sim p_{data}}[\log D(x)] - \mathbb{E}_{z \sim p_z}[\log (1 - D(G(z)))]
\]
The generator's objective is:
\[
\mathcal{L}_G = - \mathbb{E}_{z \sim p_z}[\log D(G(z))]
\]

While effective, this loss can lead to issues like vanishing gradients if the discriminator becomes too strong, as \(D(G(z))\) approaches zero, leading to very small updates for the generator.

\subsubsection{Wasserstein Loss}
The \textbf{Wasserstein loss} (used in WGANs) is designed to address the instability and vanishing gradient problems in standard GANs~\cite{frogner2015learning}. Instead of minimizing the cross-entropy, it minimizes the \textbf{Earth Mover's Distance (EMD)}~\cite{rubner2000earth}, which is more stable and provides better gradients for the generator:
\[
W(P_r, P_g) = \inf_{\gamma \in \Pi(P_r, P_g)} \mathbb{E}_{(x, y) \sim \gamma} [|| x - y ||]
\]
Here, \(P_r\) is the real data distribution and \(P_g\) is the generated data distribution~\cite{ho2019real}. This loss function ensures smoother convergence and has better gradient properties compared to the binary cross-entropy loss~\cite{frogner2015learning}.

In practice, the Wasserstein loss is approximated as:
\[
\mathcal{L}_D = \mathbb{E}_{x \sim p_{data}}[D(x)] - \mathbb{E}_{z \sim p_z}[D(G(z))]
\]
The generator's objective is to minimize:
\[
\mathcal{L}_G = - \mathbb{E}_{z \sim p_z}[D(G(z))]
\]

The key difference here is that \(D(x)\) outputs unbounded real values instead of probabilities (0 to 1), and the training is constrained by applying a \textbf{weight clipping}~\cite{han2019dimension, elsayed2024weight} technique to the discriminator's weights to enforce the Lipschitz constraint~\cite{li2019preventing}.

\begin{lstlisting}[style=python]
# Example: Wasserstein GAN loss implementation with weight clipping
class WGAN_Discriminator(nn.Module):
    def __init__(self):
        super(WGAN_Discriminator, self).__init__()
        self.main = nn.Sequential(
            nn.Linear(784, 512),
            nn.LeakyReLU(0.2),
            nn.Linear(512, 256),
            nn.LeakyReLU(0.2),
            nn.Linear(256, 1)  # No Sigmoid, output is a real number
        )

    def forward(self, x):
        return self.main(x)

# Weight clipping function for Lipschitz constraint
def clip_weights(model, clip_value):
    for param in model.parameters():
        param.data.clamp_(-clip_value, clip_value)

# WGAN loss
for epoch in range(epochs):
    optimizer_D.zero_grad()
    real_data = torch.randn(64, 784)
    loss_D_real = -torch.mean(D(real_data))
    
    noise = torch.randn(64, 100)
    fake_data = G(noise)
    loss_D_fake = torch.mean(D(fake_data.detach()))
    
    loss_D = loss_D_real + loss_D_fake
    loss_D.backward()
    optimizer_D.step()
    
    # Apply weight clipping
    clip_weights(D, 0.01)

    optimizer_G.zero_grad()
    loss_G = -torch.mean(D(fake_data))
    loss_G.backward()
    optimizer_G.step()
\end{lstlisting}

\subsubsection{Hinge Loss}
Another popular loss function for GANs is \textbf{hinge loss}~\cite{gentile1998linear, bartlett2008classification}, which is often used in \textbf{SAGAN} (Self-Attention GAN)~\cite{zhang2019self}. Hinge loss modifies the discriminator loss to be:
\[
\mathcal{L}_D = \mathbb{E}_{x \sim p_{data}}[\max(0, 1 - D(x))] + \mathbb{E}_{z \sim p_z}[\max(0, 1 + D(G(z)))]
\]
The generator's objective is:
\[
\mathcal{L}_G = -\mathbb{E}_{z \sim p_z} [D(G(z))]
\]
Hinge loss encourages the discriminator to push its output values closer to 1 for real data and closer to -1 for generated data, which helps in stabilizing the training process~\cite{bartlett2008classification, zhang2019self}.

Each of these loss functions has unique properties that affect the behavior of GANs during training. Depending on the application and dataset, the appropriate loss function can significantly improve the performance and stability of GAN models~\cite{zhang2019self}.

\begin{figure}[h]
    \centering
    \includegraphics[width=0.8\textwidth]{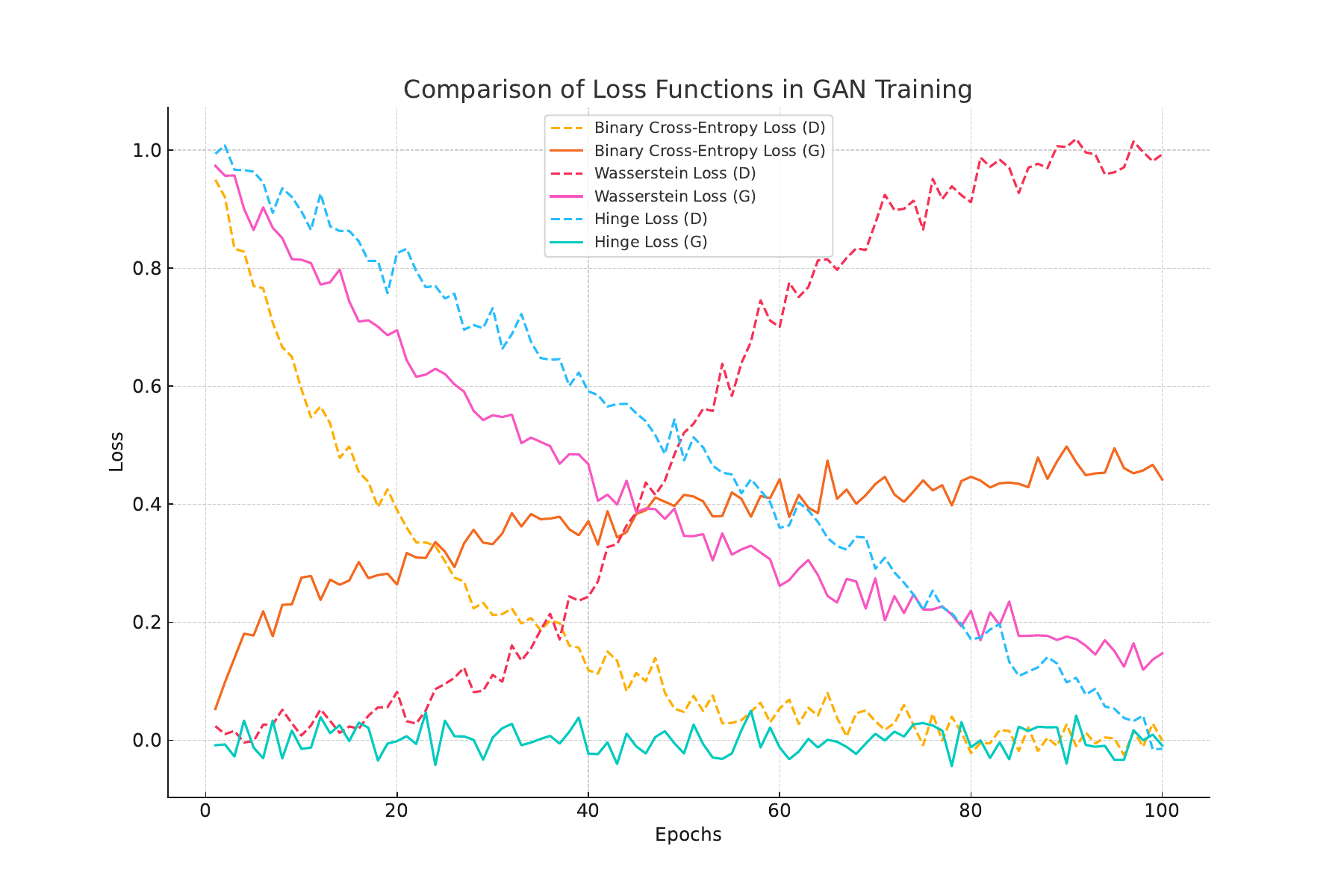}
    \caption{Comparison of Loss Functions in GAN Training.}
\end{figure}

The chart above illustrates the loss trends for three different loss functions (Binary Cross-Entropy, Wasserstein, and Hinge) for both the generator (G) and discriminator (D) during GAN training over 100 epochs. Each line shows how the respective loss changes as training progresses. This visual can help to understand the convergence patterns and stability of different loss functions in GANs.
\part{Classic Variants and Improvements}

\chapter{Classic Variants of GAN}
Generative Adversarial Networks (GANs) have seen numerous improvements and adaptations since their introduction. Among these variations, Conditional GANs (CGANs)~\cite{mirza2014conditional} have gained significant attention for their ability to incorporate additional information during the generation process, allowing more control over the output~\cite{wang2018cgan}. In this chapter, we will explore CGANs in detail, discussing their fundamental concepts and their applications, especially in image generation tasks.

\section{Conditional Generative Adversarial Networks (CGAN)}
Conditional GANs (CGANs) extend the original GAN framework by conditioning the generation process on some external information~\cite{li2020gan}. This allows the generator to not only generate random samples but to generate samples based on specific input conditions. This conditioning can be any type of information, such as class labels or image attributes~\cite{mirza2014conditional, wang2018cgan}.

\subsection{Basic Concept of Conditional GAN}
In a traditional GAN, the generator produces output solely based on random noise. In a CGAN, however, the generator and discriminator both receive an additional input: a condition. This condition can be any type of auxiliary information, such as a class label in a supervised learning problem or some attribute of the data~\cite{wang2018cgan}. The key idea is that this condition is incorporated into both the generator and discriminator to influence the data generation process.

Formally, the CGAN objective can be written as:

\[
\min_G \max_D \mathbb{E}_{x \sim p_{\text{data}}(x)}[\log D(x|y)] + \mathbb{E}_{z \sim p_z(z)}[\log(1 - D(G(z|y)))]
\]

Where:
\begin{itemize}
    \item \( G(z|y) \): The generator that produces data based on random noise \( z \) and a condition \( y \).
    \item \( D(x|y) \): The discriminator that classifies whether a given data sample \( x \) is real or fake, conditioned on \( y \).
    \item \( y \): The conditional input, such as a label or feature.
\end{itemize}

The generator aims to produce samples that not only look real but also match the given condition \( y \), while the discriminator tries to distinguish between real and generated data while also considering the condition~\cite{wang2018cgan, devries2019evaluation}.

\subsection{Illustrative Example of Conditional GAN}
To better understand the concept, let's consider an example where we want to generate images of handwritten digits from the MNIST dataset~\cite{deng2012mnist}, but with the ability to control which digit the generator should produce (i.e., we want to generate a specific digit like 3 or 7)~\cite{thekumparampil2018robustness}.

\begin{figure}[htbp]
    \centering
    \includegraphics[width=0.8\textwidth]{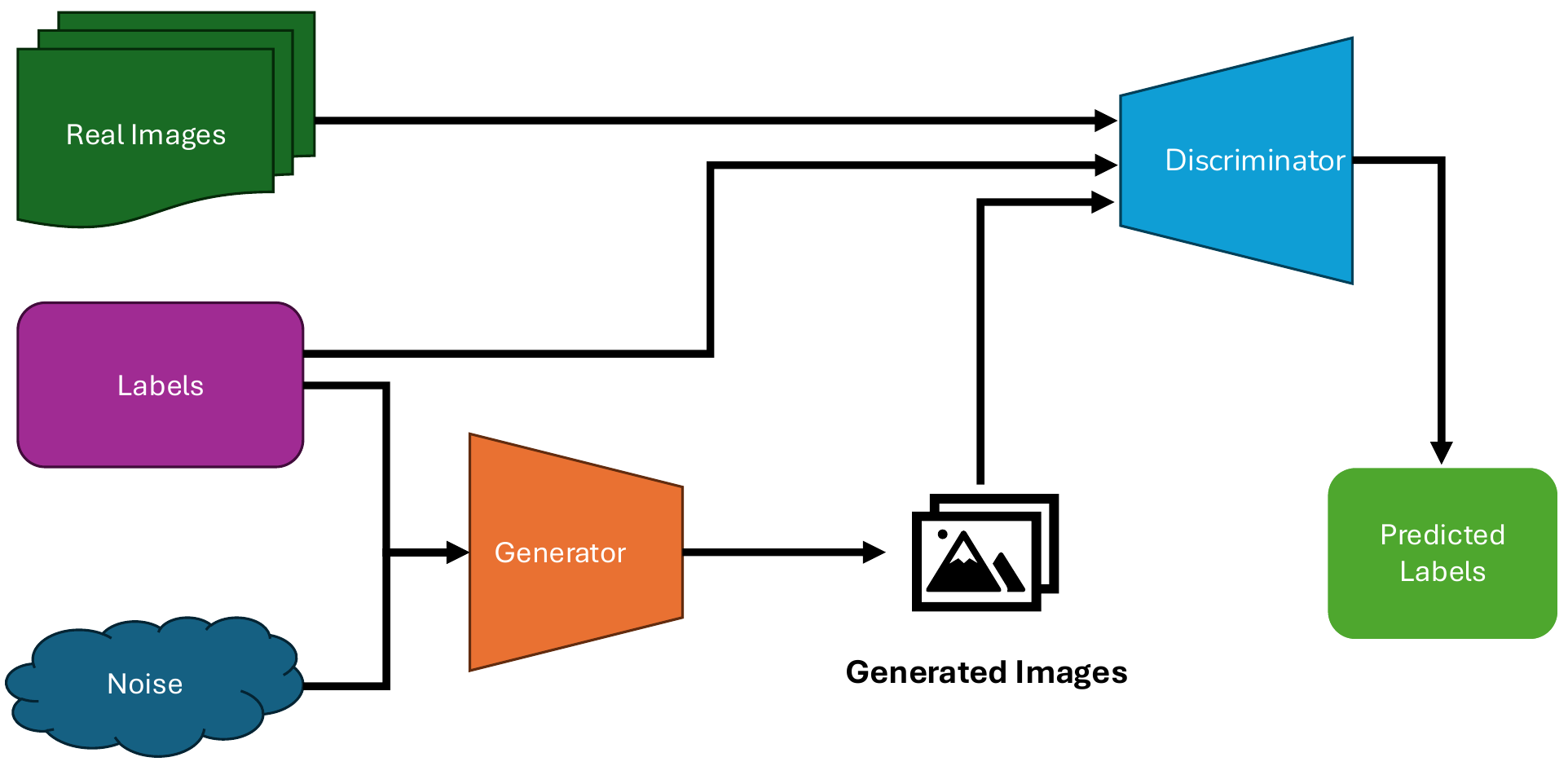}
    \caption{The basic architecture of a Conditional GAN (CGAN).}
\end{figure}

In this case, the condition \( y \) is the label of the digit (0 through 9), and the generator will learn to produce an image of the specified digit based on both the random noise \( z \) and the label \( y \). The discriminator will evaluate not only whether the image looks real, but also whether the generated image corresponds to the specified label.

\subsection{How Conditioning Works in CGAN}
In a CGAN, the condition \( y \) can be concatenated with the input noise vector \( z \) and fed into the generator. Similarly, the discriminator takes both the condition \( y \) and the generated or real data as input. This can be done by either concatenating \( y \) with the input or by using other more sophisticated mechanisms such as embedding layers~\cite{mirza2014conditional}.

Let's break this down in Python using PyTorch.

\subsection{Step-by-Step Example of CGAN}
We will implement a Conditional GAN for generating MNIST digits conditioned on the digit labels. First, we will set up the required libraries:

\begin{lstlisting}[style=cmd]
pip install torch torchvision matplotlib
\end{lstlisting}

Now, let's implement the CGAN architecture using PyTorch.

\subsubsection{Step 1: Import Necessary Libraries}
We will start by importing the necessary libraries and setting up some basic parameters.

\begin{lstlisting}[style=python]
import torch
import torch.nn as nn
import torch.optim as optim
import torchvision
import torchvision.transforms as transforms
import matplotlib.pyplot as plt

# Hyperparameters
latent_dim = 100  # Dimension of the noise vector
num_classes = 10  # Number of digit classes in MNIST
image_size = 28   # Image dimensions (28x28 for MNIST)
batch_size = 64
lr = 0.0002
epochs = 50
\end{lstlisting}

\subsubsection{Step 2: Define the Generator and Discriminator}
We need to modify the generator and discriminator to accept both the noise vector \( z \) and the conditional label \( y \). One common approach is to concatenate \( z \) with a one-hot encoded label vector for \( y \).

\begin{lstlisting}[style=python]
# Generator model
class Generator(nn.Module):
    def __init__(self, latent_dim, num_classes, img_size):
        super(Generator, self).__init__()
        self.label_emb = nn.Embedding(num_classes, num_classes)
        self.model = nn.Sequential(
            nn.Linear(latent_dim + num_classes, 128),
            nn.ReLU(),
            nn.Linear(128, 256),
            nn.BatchNorm1d(256, 0.8),
            nn.ReLU(),
            nn.Linear(256, 512),
            nn.BatchNorm1d(512, 0.8),
            nn.ReLU(),
            nn.Linear(512, img_size * img_size),
            nn.Tanh()
        )
    
    def forward(self, noise, labels):
        # Concatenate noise and label embeddings
        gen_input = torch.cat((noise, self.label_emb(labels)), -1)
        img = self.model(gen_input)
        img = img.view(img.size(0), 1, image_size, image_size)
        return img

# Discriminator model
class Discriminator(nn.Module):
    def __init__(self, num_classes, img_size):
        super(Discriminator, self).__init__()
        self.label_emb = nn.Embedding(num_classes, num_classes)
        self.model = nn.Sequential(
            nn.Linear(num_classes + img_size * img_size, 512),
            nn.LeakyReLU(0.2, inplace=True),
            nn.Linear(512, 256),
            nn.LeakyReLU(0.2, inplace=True),
            nn.Linear(256, 1),
            nn.Sigmoid()
        )

    def forward(self, img, labels):
        # Flatten image and concatenate with label embeddings
        img_flat = img.view(img.size(0), -1)
        d_in = torch.cat((img_flat, self.label_emb(labels)), -1)
        validity = self.model(d_in)
        return validity
\end{lstlisting}

In the generator, we take the noise \( z \) and the label \( y \) as inputs and concatenate them before feeding them into the network. Similarly, in the discriminator, we flatten the image and concatenate it with the label embedding.

\subsubsection{Step 3: Training the CGAN}
Next, we set up the loss function and optimizers, and then train the CGAN.

\begin{lstlisting}[style=python]
# Loss function
adversarial_loss = nn.BCELoss()

# Initialize generator and discriminator
generator = Generator(latent_dim, num_classes, image_size)
discriminator = Discriminator(num_classes, image_size)

# Optimizers
optimizer_g = optim.Adam(generator.parameters(), lr=lr, betas=(0.5, 0.999))
optimizer_d = optim.Adam(discriminator.parameters(), lr=lr, betas=(0.5, 0.999))

# Load MNIST dataset
transform = transforms.Compose([transforms.ToTensor(), transforms.Normalize([0.5], [0.5])])
dataloader = torch.utils.data.DataLoader(
    torchvision.datasets.MNIST('./data', train=True, download=True, transform=transform),
    batch_size=batch_size, shuffle=True
)

# Training loop
for epoch in range(epochs):
    for i, (imgs, labels) in enumerate(dataloader):
        
        # Adversarial ground truths
        valid = torch.ones(imgs.size(0), 1)
        fake = torch.zeros(imgs.size(0), 1)

        # Configure input
        real_imgs = imgs
        labels = labels

        # ---------------------
        #  Train Discriminator
        # ---------------------
        optimizer_d.zero_grad()

        # Sample noise and labels as generator input
        z = torch.randn(imgs.size(0), latent_dim)
        gen_labels = torch.randint(0, num_classes, (imgs.size(0),))

        # Generate a batch of images
        gen_imgs = generator(z, gen_labels)

        # Loss for real images
        real_loss = adversarial_loss(discriminator(real_imgs, labels), valid)
        # Loss for fake images
        fake_loss = adversarial_loss(discriminator(gen_imgs.detach(), gen_labels), fake)
        # Total discriminator loss
        d_loss = (real_loss + fake_loss) / 2

        d_loss.backward()
        optimizer_d.step()

        # -----------------
        #  Train Generator
        # -----------------
        optimizer_g.zero_grad()

        # Loss for generator
        g_loss = adversarial_loss(discriminator(gen_imgs, gen_labels), valid)

        g_loss.backward()
        optimizer_g.step()

        print(f"[Epoch {epoch}/{epochs}] [Batch {i}/{len(dataloader)}] [D loss: {d_loss.item():.4f}] [G loss: {g_loss.item():.4f}]")
\end{lstlisting}

In this training loop, the generator learns to produce MNIST digits conditioned on the class labels, while the discriminator learns to classify whether an image is real or generated, based on both the image and its corresponding label.

\section{Application of CGAN in Image Generation}
Conditional GANs are widely used in various tasks that require controlled data generation~\cite{mirza2014conditional, wang2018cgan, devries2019evaluation}. One of the most common applications is in image generation tasks~\cite{bao2017cvae, liu2019multi}, where CGANs allow users to generate specific types of images based on certain conditions.

\subsection{Example: Handwritten Digit Generation}
As seen in the above implementation, CGAN can be used to generate handwritten digits conditioned on the label of the digit~\cite{liu2019multi}. This means that we can specify which digit (from 0 to 9) we want the generator to create, providing more control over the output compared to a standard GAN.

\subsection{Image-to-Image Translation}
Another popular application of CGANs is in image-to-image translation~\cite{yi2017dualgan}, where the goal is to generate a target image based on an input image. For example:
\begin{itemize}
    \item Generating a colored image from a grayscale image.
    \item Translating a daytime image to a nighttime image.
    \item Converting a sketch to a photorealistic image.
\end{itemize}
In such tasks, the condition \( y \) is often the input image, and the generator learns to translate the input into a desired output based on the condition.

\section{Summary}
In this chapter, we explored the concept of Conditional GANs (CGANs), which extend the original GAN framework by conditioning the generation process on additional information such as labels or attributes. CGANs allow for more control over the generated output and have been successfully applied to various tasks, including digit generation~\cite{liu2019multi} and image-to-image translation~\cite{yi2017dualgan}. Through a detailed PyTorch implementation, we demonstrated how to build and train a CGAN, offering a practical example for beginners to better understand how CGANs function.

\section{Deep Convolutional Generative Adversarial Networks (DCGAN)}
Deep Convolutional Generative Adversarial Networks (DCGAN)~\cite{radford2015unsupervised} are a variant of GANs where convolutional neural networks (CNNs) are used instead of fully connected layers~\cite{luo2021case}, especially in the Generator and Discriminator. This architectural change allows DCGANs to leverage the spatial hierarchical nature of images, making them particularly powerful for image generation tasks. 

\subsection{The Role of Convolutional Networks in GANs}
Convolutional neural networks (CNNs)~\cite{lecun2015deep} are specifically designed to work with grid-like data, such as images. Unlike fully connected layers, where each neuron is connected to all neurons in the previous layer, convolutional layers~\cite{radford2015unsupervised} use filters (also called kernels) to perform localized operations over small patches of the image~\cite{lecun2015deep, o2015cnn}. This process captures spatial dependencies, such as edges or textures, that are essential for image understanding and generation.

In the context of GANs, using CNNs allows the Generator to produce images with better quality and sharper details~\cite{goodfellow2014generative, radford2015unsupervised}. Similarly, the Discriminator can use convolutional layers to more effectively distinguish between real and generated images, recognizing subtle differences in structure and texture.

\textbf{Example of a Convolutional Layer in PyTorch:}

\begin{lstlisting}[style=python]
import torch
import torch.nn as nn

# Example of a simple convolutional layer in PyTorch
conv_layer = nn.Conv2d(in_channels=3, out_channels=64, kernel_size=4, stride=2, padding=1)

# Input image: 3 channels (RGB), size 64x64
input_image = torch.randn(1, 3, 64, 64)

# Apply convolution
output = conv_layer(input_image)
print(output.shape)  # Output will have 64 channels, reduced spatial dimensions
\end{lstlisting}

In this example, the convolutional layer takes a 64x64 RGB image as input and applies 64 filters with a kernel size of 4x4. The stride of 2 reduces the spatial dimensions, while padding ensures the output image size is manageable. This operation helps extract important features such as edges, corners, and textures.

\subsection{DCGAN Architecture and Implementation}
The DCGAN architecture introduces several modifications compared to standard GANs:

\begin{itemize}
    \item \textbf{No fully connected layers}: Both the Generator and Discriminator avoid using fully connected layers in favor of convolutional layers. This helps the networks scale better with image size and capture local patterns effectively~\cite{radford2015unsupervised}.
    \item \textbf{Batch normalization}: Batch normalization is used after most layers to stabilize training by normalizing the activations, which allows for faster convergence.
    \item \textbf{Leaky ReLU}: In the Discriminator, the Leaky ReLU~\cite{dubey2019comparative} activation function is used, which allows a small gradient when the activation is negative, addressing the problem of dying ReLUs.
    \item \textbf{Transposed convolutions}: In the Generator, transposed convolutions~\cite{jansson2012deconvolution} (also known as deconvolutions) are used to upsample noise into an image.
\end{itemize}

\textbf{DCGAN Generator and Discriminator in PyTorch:}

\begin{lstlisting}[style=python]
# DCGAN Generator
class DCGAN_Generator(nn.Module):
    def __init__(self, noise_dim):
        super(DCGAN_Generator, self).__init__()
        self.model = nn.Sequential(
            nn.ConvTranspose2d(noise_dim, 512, 4, 1, 0),  # First layer, fully connected equivalent
            nn.BatchNorm2d(512),
            nn.ReLU(True),
            nn.ConvTranspose2d(512, 256, 4, 2, 1),  # Upsample to 8x8
            nn.BatchNorm2d(256),
            nn.ReLU(True),
            nn.ConvTranspose2d(256, 128, 4, 2, 1),  # Upsample to 16x16
            nn.BatchNorm2d(128),
            nn.ReLU(True),
            nn.ConvTranspose2d(128, 64, 4, 2, 1),  # Upsample to 32x32
            nn.BatchNorm2d(64),
            nn.ReLU(True),
            nn.ConvTranspose2d(64, 3, 4, 2, 1),    # Upsample to 64x64 (RGB)
            nn.Tanh()  # Tanh activation for output images
        )

    def forward(self, x):
        return self.model(x)

# DCGAN Discriminator
class DCGAN_Discriminator(nn.Module):
    def __init__(self):
        super(DCGAN_Discriminator, self).__init__()
        self.model = nn.Sequential(
            nn.Conv2d(3, 64, 4, 2, 1),   # Downsample to 32x32
            nn.LeakyReLU(0.2, inplace=True),
            nn.Conv2d(64, 128, 4, 2, 1), # Downsample to 16x16
            nn.BatchNorm2d(128),
            nn.LeakyReLU(0.2, inplace=True),
            nn.Conv2d(128, 256, 4, 2, 1),# Downsample to 8x8
            nn.BatchNorm2d(256),
            nn.LeakyReLU(0.2, inplace=True),
            nn.Conv2d(256, 512, 4, 2, 1),# Downsample to 4x4
            nn.BatchNorm2d(512),
            nn.LeakyReLU(0.2, inplace=True),
            nn.Conv2d(512, 1, 4, 1, 0),  # Output a single scalar value (real or fake)
            nn.Sigmoid()                 # Sigmoid activation for binary classification
        )

    def forward(self, x):
        return self.model(x)

# Example usage:
noise = torch.randn(1, 100, 1, 1)  # Random noise for Generator
gen = DCGAN_Generator(100)
disc = DCGAN_Discriminator()

generated_image = gen(noise)
disc_output = disc(generated_image)

print(generated_image.shape)  # Should output: torch.Size([1, 3, 64, 64])
print(disc_output.shape)      # Should output: torch.Size([1, 1, 1, 1])
\end{lstlisting}

In this example, the Generator starts with noise of shape \(100 \times 1 \times 1\), which is upsampled through a series of transposed convolutions to a \(64 \times 64\) RGB image. The Discriminator takes this image and progressively downsamples it through convolutions~\cite{lecun2015deep}, outputting a single value indicating whether the image is real or fake.

\section{Information Maximizing Generative Adversarial Networks (InfoGAN)}
InfoGAN~\cite{chen2016infogan} is an extension of GANs that introduces an information-theoretic objective to maximize mutual information between a subset of latent variables and the generated data. This enables InfoGAN to learn interpretable and disentangled representations in an unsupervised manner~\cite{kurutach2018learning}, making it highly useful for understanding the structure of the data without requiring labeled examples.

\subsection{Introducing the Information Maximization Objective}
The key innovation in InfoGAN is the addition of a new objective to maximize the mutual information between the latent code \(c\) and the generated data \(G(z, c)\). In a standard GAN, the latent vector \(z\) is random noise, and the generated data is not necessarily interpretable. However, in InfoGAN, we split \(z\) into two parts:

\begin{itemize}
    \item \textbf{Random noise} \(z\), which is the standard noise vector used by GANs.
    \item \textbf{Latent code} \(c\), which encodes specific information that we want the Generator to learn.
\end{itemize}

Maximizing the mutual information \(I(c; G(z, c))\) encourages the Generator to produce data that reflects the information encoded in \(c\). This gives us control over certain aspects of the generated data, such as the orientation of a digit in an image or its style, while still operating in an unsupervised learning setting~\cite{chen2016infogan}.

The InfoGAN architecture introduces a separate network called the \textbf{Q-network}, which approximates the posterior distribution of the latent code \(c\) given the generated data. This allows InfoGAN to compute and maximize the mutual information efficiently.

\subsubsection{Example: InfoGAN Latent Code Control}
Let's assume we are generating images of handwritten digits using InfoGAN. The latent code \(c\) might encode the following information:

\begin{itemize}
    \item \(c_1\): The digit's thickness.
    \item \(c_2\): The digit's rotation angle.
    \item \(c_3\): The digit's style.
\end{itemize}

By controlling the values of \(c_1\), \(c_2\), and \(c_3\), we can generate images with specific thickness, rotation, or style, even though the model was trained without labeled data.

\textbf{InfoGAN Training in PyTorch:}

\begin{lstlisting}[style=python]
# InfoGAN with additional latent code c
class InfoGAN_Generator(nn.Module):
    def __init__(self, noise_dim, code_dim):
        super(InfoGAN_Generator, self).__init__()
        self.model = nn.Sequential(
            nn.ConvTranspose2d(noise_dim + code_dim, 512, 4, 1, 0),
            nn.BatchNorm2d(512),
            nn.ReLU(True),
            nn.ConvTranspose2d(512, 256, 4, 2, 1),
            nn.BatchNorm2d(256),
            nn.ReLU(True),
            nn.ConvTranspose2d(256, 128, 4, 2, 1),
            nn.BatchNorm2d(128),
            nn.ReLU(True),
            nn.ConvTranspose2d(128, 64, 4, 2, 1),
            nn.BatchNorm2d(64),
            nn.ReLU(True),
            nn.ConvTranspose2d(64, 3, 4, 2, 1),
            nn.Tanh()
        )

    def forward(self, noise, code):
        x = torch.cat([noise, code], dim=1)  # Concatenate noise and code
        return self.model(x)

# Q-network to approximate posterior q(c|x)
class InfoGAN_Q_Network(nn.Module):
    def __init__(self):
        super(InfoGAN_Q_Network, self).__init__()
        self.model = nn.Sequential(
            nn.Conv2d(3, 64, 4, 2, 1),
            nn.LeakyReLU(0.2, inplace=True),
            nn.Conv2d(64, 128, 4, 2, 1),
            nn.BatchNorm2d(128),
            nn.LeakyReLU(0.2, inplace=True),
            nn.Conv2d(128, 256, 4, 2, 1),
            nn.BatchNorm2d(256),
            nn.LeakyReLU(0.2, inplace=True),
            nn.Conv2d(256, 512, 4, 2, 1),
            nn.BatchNorm2d(512),
            nn.LeakyReLU(0.2, inplace=True),
            nn.Flatten(),
            nn.Linear(512 * 4 * 4, 128),
            nn.ReLU(True),
            nn.Linear(128, 10)  # Assume latent code c is 10-dimensional
        )

    def forward(self, x):
        return self.model(x)

# Example usage:
noise = torch.randn(1, 100, 1, 1)  # Random noise
code = torch.randn(1, 10, 1, 1)    # Latent code

gen = InfoGAN_Generator(100, 10)
q_net = InfoGAN_Q_Network()

generated_image = gen(noise, code)
q_output = q_net(generated_image)

print(generated_image.shape)  # Should output: torch.Size([1, 3, 64, 64])
print(q_output.shape)         # Should output: torch.Size([1, 10])
\end{lstlisting}

In this example, the Generator takes both noise and a latent code as input, producing an image that is influenced by the code. The Q-network tries to estimate the latent code from the generated image, allowing the model to learn how the latent code affects the generated data.

\subsection{InfoGAN in Unsupervised Learning}
One of the key advantages of InfoGAN is its ability to learn interpretable features in an unsupervised setting. In many real-world scenarios, labeled data is scarce or expensive to obtain, so having a model that can automatically discover and disentangle important features without supervision is highly valuable.

In InfoGAN, the latent code \(c\) provides a mechanism for this unsupervised learning. By maximizing the mutual information between the latent code and the generated data, InfoGAN encourages the Generator to create data that reflects the structure of the input code. This allows InfoGAN to discover meaningful and disentangled representations of the data, such as variations in object shape, color, or orientation, without needing explicit labels~\cite{mugunthan2021dpd}.

\textbf{Example: Unsupervised Learning of Handwritten Digits}

Consider a dataset of handwritten digits (e.g., MNIST). InfoGAN can learn to control different aspects of the digits, such as:

\begin{itemize}
    \item The digit's thickness (controlled by \(c_1\)).
    \item The rotation angle (controlled by \(c_2\)).
    \item The style or stroke (controlled by \(c_3\)).
\end{itemize}

Even though the model is trained without knowing these specific features, InfoGAN learns to disentangle them naturally~\cite{chen2016infogan}. By manipulating the latent code during generation, we can generate digits with specific characteristics, gaining insight into the structure of the data in an unsupervised manner.

\section{Laplacian Pyramid GAN (LAPGAN)}
Understanding LAPGAN~\cite{denton2015deep} is crucial for generating high-resolution images with fine details. In this section, we will delve into the hierarchical generation process~\cite{jin2020hierarchical} of LAPGAN and explore its applications in image detail generation.

\subsection{Hierarchical Generation Process}

The Laplacian Pyramid Generative Adversarial Network (LAPGAN) is a GAN architecture that generates images in a coarse-to-fine fashion using a pyramid of generators and discriminators. Instead of generating a high-resolution image in one pass, LAPGAN breaks down the image generation process into multiple stages, each responsible for generating images at different resolutions~\cite{jin2020hierarchical}.

\subsubsection{Laplacian Pyramid Concept}

The Laplacian Pyramid is a technique used in image processing to represent images at multiple scales or resolutions~\cite{zhang2018photographic}. It involves decomposing an image into a set of band-pass filtered images (Laplacian images) and a low-resolution residual image.

To construct a Laplacian Pyramid, we perform the following steps:

\begin{enumerate}
    \item \textbf{Gaussian Pyramid Construction}: Create a series of images where each subsequent image is a downsampled (usually by a factor of 2) version of the previous one using a Gaussian filter.
    \item \textbf{Laplacian Images Computation}: Subtract the upsampled version of each lower-resolution image from the current resolution image to obtain the Laplacian images.
\end{enumerate}

By reconstructing the original image from the Laplacian Pyramid, we can add back the details at each level, starting from the lowest resolution.

\subsubsection{LAPGAN Architecture}

In LAPGAN, the image generation process is divided into multiple levels corresponding to different resolutions. Each level consists of a generator and a discriminator:

\begin{itemize}
    \item \textbf{Generator at Level \( i \)}: Generates a high-resolution image \( x_i \) conditioned on the upsampled image \( x_{i-1}^{\uparrow} \) from the previous level and a random noise vector \( z_i \).
    \item \textbf{Discriminator at Level \( i \)}: Evaluates the authenticity of the generated image \( x_i \) against the real images at the same resolution.
\end{itemize}

The overall generation process can be summarized as:

\[
x_0 = G_0(z_0) \\
x_i = G_i(x_{i-1}^{\uparrow}, z_i), \quad \text{for } i = 1, 2, ..., N
\]

Where:
\begin{itemize}
    \item \( x_0 \) is the initial low-resolution image generated from noise.
    \item \( x_{i-1}^{\uparrow} \) is the upsampled image from the previous level.
    \item \( G_i \) is the generator at level \( i \).
    \item \( z_i \) is the noise vector injected at level \( i \).
\end{itemize}

This hierarchical approach allows the model to focus on adding details progressively, making it easier to generate high-resolution images with fine details~\cite{denton2015deep}.

\subsubsection{Implementation Example}

Let's implement a simplified version of LAPGAN using PyTorch. We'll use a three-level pyramid to generate images of size \( 64 \times 64 \).

\begin{lstlisting}[style=python]
import torch
import torch.nn as nn
import torch.optim as optim
import torch.nn.functional as F

# Define the Generator for level 0 (16x16)
class GeneratorLevel0(nn.Module):
    def __init__(self):
        super(GeneratorLevel0, self).__init__()
        self.main = nn.Sequential(
            nn.Linear(100, 128),
            nn.ReLU(),
            nn.Linear(128, 16*16*3),
            nn.Tanh()
        )
    
    def forward(self, z):
        output = self.main(z)
        output = output.view(-1, 3, 16, 16)
        return output

# Define the Generator for higher levels (32x32 and 64x64)
class GeneratorLevelN(nn.Module):
    def __init__(self, input_channels):
        super(GeneratorLevelN, self).__init__()
        self.main = nn.Sequential(
            nn.Conv2d(input_channels, 64, kernel_size=3, padding=1),
            nn.ReLU(),
            nn.Conv2d(64, 3, kernel_size=3, padding=1),
            nn.Tanh()
        )
    
    def forward(self, x, z):
        z = z.view(-1, 1, x.size(2), x.size(3))
        input = torch.cat([x, z], dim=1)
        output = self.main(input)
        return output

# Define the Discriminator for each level
class Discriminator(nn.Module):
    def __init__(self, image_size):
        super(Discriminator, self).__init__()
        self.main = nn.Sequential(
            nn.Conv2d(3, 64, kernel_size=4, stride=2),
            nn.LeakyReLU(0.2),
            nn.Conv2d(64, 128, kernel_size=4, stride=2),
            nn.LeakyReLU(0.2),
            nn.Flatten(),
            nn.Linear(128 * (image_size//4) * (image_size//4), 1),
            nn.Sigmoid()
        )
    
    def forward(self, x):
        return self.main(x)
\end{lstlisting}

In this example, we have:

\begin{itemize}
    \item \textbf{GeneratorLevel0}: Generates a \( 16 \times 16 \) image from a noise vector \( z \).
    \item \textbf{GeneratorLevelN}: Takes the upsampled image from the previous level, concatenated with a noise map, and outputs a higher-resolution image.
    \item \textbf{Discriminator}: Evaluates images at each resolution.
\end{itemize}

Next, we need to define the training process for each level.

\begin{lstlisting}[style=python]
# Instantiate generators and discriminators
G0 = GeneratorLevel0()
D0 = Discriminator(16)
G1 = GeneratorLevelN(4)  # 3 channels from upsampled image + 1 channel noise
D1 = Discriminator(32)
G2 = GeneratorLevelN(4)
D2 = Discriminator(64)

# Optimizers
optimizer_G0 = optim.Adam(G0.parameters(), lr=0.0002)
optimizer_D0 = optim.Adam(D0.parameters(), lr=0.0002)
optimizer_G1 = optim.Adam(G1.parameters(), lr=0.0002)
optimizer_D1 = optim.Adam(D1.parameters(), lr=0.0002)
optimizer_G2 = optim.Adam(G2.parameters(), lr=0.0002)
optimizer_D2 = optim.Adam(D2.parameters(), lr=0.0002)

# Loss function
criterion = nn.BCELoss()

# Training loop for each level
for epoch in range(num_epochs):
    ############################
    # Level 0 Training (16x16)
    ############################
    # Generate noise and fake images
    z0 = torch.randn(batch_size, 100)
    fake_images0 = G0(z0)
    
    # Get real images at 16x16 resolution
    real_images0 = get_real_images(16)
    
    # Train Discriminator D0
    optimizer_D0.zero_grad()
    # Real images
    outputs_real = D0(real_images0)
    labels_real = torch.ones(batch_size, 1)
    loss_D_real = criterion(outputs_real, labels_real)
    # Fake images
    outputs_fake = D0(fake_images0.detach())
    labels_fake = torch.zeros(batch_size, 1)
    loss_D_fake = criterion(outputs_fake, labels_fake)
    # Backprop and optimize
    loss_D0 = loss_D_real + loss_D_fake
    loss_D0.backward()
    optimizer_D0.step()
    
    # Train Generator G0
    optimizer_G0.zero_grad()
    outputs = D0(fake_images0)
    loss_G0 = criterion(outputs, labels_real)
    loss_G0.backward()
    optimizer_G0.step()
    
    ############################
    # Level 1 Training (32x32)
    ############################
    # Upsample images from Level 0
    upsampled_images0 = F.interpolate(fake_images0.detach(), scale_factor=2)
    # Generate noise map
    z1 = torch.randn(batch_size, 1, 32, 32)
    # Generate fake images at Level 1
    fake_images1 = G1(upsampled_images0, z1)
    
    # Get real images at 32x32 resolution
    real_images1 = get_real_images(32)
    
    # Train Discriminator D1
    optimizer_D1.zero_grad()
    # Real images
    outputs_real = D1(real_images1)
    labels_real = torch.ones(batch_size, 1)
    loss_D_real = criterion(outputs_real, labels_real)
    # Fake images
    outputs_fake = D1(fake_images1.detach())
    labels_fake = torch.zeros(batch_size, 1)
    loss_D_fake = criterion(outputs_fake, labels_fake)
    # Backprop and optimize
    loss_D1 = loss_D_real + loss_D_fake
    loss_D1.backward()
    optimizer_D1.step()
    
    # Train Generator G1
    optimizer_G1.zero_grad()
    outputs = D1(fake_images1)
    loss_G1 = criterion(outputs, labels_real)
    loss_G1.backward()
    optimizer_G1.step()
    
    ############################
    # Level 2 Training (64x64)
    ############################
    # Upsample images from Level 1
    upsampled_images1 = F.interpolate(fake_images1.detach(), scale_factor=2)
    # Generate noise map
    z2 = torch.randn(batch_size, 1, 64, 64)
    # Generate fake images at Level 2
    fake_images2 = G2(upsampled_images1, z2)
    
    # Get real images at 64x64 resolution
    real_images2 = get_real_images(64)
    
    # Train Discriminator D2
    optimizer_D2.zero_grad()
    # Real images
    outputs_real = D2(real_images2)
    labels_real = torch.ones(batch_size, 1)
    loss_D_real = criterion(outputs_real, labels_real)
    # Fake images
    outputs_fake = D2(fake_images2.detach())
    labels_fake = torch.zeros(batch_size, 1)
    loss_D_fake = criterion(outputs_fake, labels_fake)
    # Backprop and optimize
    loss_D2 = loss_D_real + loss_D_fake
    loss_D2.backward()
    optimizer_D2.step()
    
    # Train Generator G2
    optimizer_G2.zero_grad()
    outputs = D2(fake_images2)
    loss_G2 = criterion(outputs, labels_real)
    loss_G2.backward()
    optimizer_G2.step()
\end{lstlisting}

In this code:

\begin{itemize}
    \item We define separate generators and discriminators for each level.
    \item At each level, the generator takes the upsampled image from the previous level and a noise map to generate finer details.
    \item The discriminator at each level evaluates the generated images against real images at the same resolution.
    \item We use the Binary Cross-Entropy loss (\texttt{nn.BCELoss})~\cite{ruby2020binary} for training.
\end{itemize}

\subsection{Applications of LAPGAN in Image Detail Generation}

LAPGAN is particularly useful in generating high-resolution images with fine details, which is challenging for standard GAN architectures. By breaking down the generation process into hierarchical levels, LAPGAN can:

\begin{itemize}
    \item \textbf{Capture Global Structure}: The initial low-resolution generator focuses on generating the overall structure of the image~\cite{cao2015grarep}.
    \item \textbf{Add Fine Details}: Subsequent generators add details at increasingly finer scales, refining the image progressively.
    \item \textbf{Improve Training Stability}: Training smaller generators and discriminators at each level can be more stable and easier than training a single large network.
\end{itemize}

\subsubsection{Example: High-Resolution Face Generation}

Suppose we want to generate high-resolution images of faces at \( 256 \times 256 \) pixels. Using LAPGAN, we can divide the generation process into multiple levels:

\begin{enumerate}
    \item \textbf{Level 0}: Generate a coarse \( 64 \times 64 \) face image capturing the overall facial structure.
    \item \textbf{Level 1}: Refine to \( 128 \times 128 \) resolution, adding details like eyes, nose, and mouth shapes.
    \item \textbf{Level 2}: Finalize at \( 256 \times 256 \) resolution, adding skin textures, hair details, and other fine features.
\end{enumerate}

At each level, the generator focuses on adding the appropriate level of detail, conditioned on the upsampled image from the previous level~\cite{denton2015deep}.

\subsubsection{Benefits in Image Super-Resolution}

LAPGAN can also be applied to image super-resolution~\cite{lai2018fast} tasks, where the goal is to reconstruct high-resolution images from low-resolution inputs. By leveraging the hierarchical structure, LAPGAN can progressively upscale images while adding realistic details.

\subsubsection{Comparison with Other Methods}

Compared to traditional GANs, LAPGAN offers several advantages:

\begin{itemize}
    \item \textbf{Efficiency}: Training smaller networks at each level reduces computational requirements.
    \item \textbf{Quality}: Produces higher-quality images with better detail preservation.
    \item \textbf{Scalability}: Can be extended to generate very high-resolution images by adding more levels~\cite{lai2018fast}.
\end{itemize}

However, LAPGAN also has some limitations:

\begin{itemize}
    \item \textbf{Complexity}: The architecture is more complex due to multiple generators and discriminators~\cite{denton2015deep}.
    \item \textbf{Training Time}: Sequential training of multiple levels can increase the overall training time.
\end{itemize}

\subsection{Visualization of LAPGAN Architecture}

To better understand the structure of LAPGAN, consider the following diagram illustrating the hierarchical generation process:

\begin{center}
\begin{tikzpicture}[node distance=2cm, auto]
    \node [draw, rectangle] (G0) {Generator Level 0};
    \node [draw, rectangle, right of=G0, node distance=4cm] (D0) {Discriminator Level 0};
    \node [draw, rectangle, below of=G0] (Upsample0) {Upsample};
    \node [draw, rectangle, below of=Upsample0] (G1) {Generator Level 1};
    \node [draw, rectangle, right of=G1, node distance=4cm] (D1) {Discriminator Level 1};
    \node [draw, rectangle, below of=G1] (Upsample1) {Upsample};
    \node [draw, rectangle, below of=Upsample1] (G2) {Generator Level 2};
    \node [draw, rectangle, right of=G2, node distance=4cm] (D2) {Discriminator Level 2};
    
    \draw [->] (G0) -- (D0);
    \draw [->] (G0) -- (Upsample0);
    \draw [->] (Upsample0) -- (G1);
    \draw [->] (G1) -- (D1);
    \draw [->] (G1) -- (Upsample1);
    \draw [->] (Upsample1) -- (G2);
    \draw [->] (G2) -- (D2);
\end{tikzpicture}
\end{center}

This diagram illustrates how each generator builds upon the output of the previous level, progressively refining the image.

\subsection{Conclusion}

LAPGAN introduces a novel approach to image generation by leveraging the concept of Laplacian Pyramids~\cite{denton2015deep}. By generating images hierarchically, it effectively captures both global structures and fine details, leading to high-quality high-resolution images~\cite{lai2018fast}. For beginners, implementing LAPGAN provides valuable insights into advanced GAN architectures and techniques for improving image generation.

\chapter{Improved Training Methods and Optimization Strategies}
Training GANs~\cite{mescheder2018training} can be notoriously difficult due to issues such as instability~\cite{becker2022instability}, mode collapse~\cite{durall2020combating}, and vanishing gradients~\cite{ding2022take}. Over time, researchers have proposed several improvements to address these challenges~\cite{kossale2022mode}. In this chapter, we will explore some of the most important improvements, including Wasserstein GAN (WGAN)~\cite{adler2018banach}, WGAN with Gradient Penalty (WGAN-GP)~\cite{gulrajani2017improved}, and Least Squares GAN (LSGAN)~\cite{mao2017least}. These methods introduce modifications to the original GAN training objective, making the training process more stable and improving the quality of generated samples~\cite{kossale2022mode}.

\section{Wasserstein GAN (WGAN)}
Wasserstein GAN (WGAN) is one of the most widely recognized improvements over the traditional GAN architecture. It addresses the problem of instability and mode collapse in GAN training by modifying the loss function to be based on the Wasserstein distance (also known as Earth Mover's Distance)~\cite{adler2018banach}, which provides a better metric for comparing the real and generated distributions.

\subsection{WGAN's Objective and Wasserstein Distance}
The main issue with the original GAN training is that the Jensen-Shannon (JS) divergence~\cite{fuglede2004jensen}, which is implicitly minimized during training, can lead to vanishing gradients, especially when the discriminator becomes too good at distinguishing real from fake data. This can cause the generator to stop learning~\cite{adler2018banach}.

WGAN replaces the JS divergence with the Wasserstein distance~\cite{adler2018banach}, which measures the distance between two probability distributions in a more meaningful way, particularly when the distributions have little or no overlap. The Wasserstein distance is defined as:

\[
W(p_r, p_g) = \inf_{\gamma \in \Pi(p_r, p_g)} \mathbb{E}_{(x,y) \sim \gamma}[\|x - y\|]
\]

Where:
\begin{itemize}
    \item \( p_r \) is the real data distribution.
    \item \( p_g \) is the generated data distribution.
    \item \( \Pi(p_r, p_g) \) is the set of all joint distributions whose marginals are \( p_r \) and \( p_g \).
\end{itemize}

Intuitively, Wasserstein distance measures the cost of transforming one distribution into another. Unlike the JS divergence, it provides useful gradient information even when the two distributions do not overlap significantly, resulting in more stable GAN training~\cite{adler2018banach}.

\subsubsection{WGAN Objective Function}
To optimize the Wasserstein distance in WGAN, the discriminator (or critic, as it's called in WGAN) is trained to approximate the Wasserstein distance between the real and generated distributions. The WGAN objective is:

\[
\min_G \max_D \mathbb{E}_{x \sim p_{\text{data}}} [D(x)] - \mathbb{E}_{z \sim p_z(z)} [D(G(z))]
\]

The key differences from traditional GAN are:
\begin{itemize}
    \item The critic outputs real-valued scores (not probabilities) for real and generated data.
    \item The objective is to maximize the difference between the critic's scores on real and fake data.
\end{itemize}

\subsubsection{Weight Clipping in WGAN}
One of the constraints in WGAN is that the critic must be a 1-Lipschitz function, meaning its gradients must be bounded. To enforce this, WGAN introduces weight clipping, where the weights of the critic are constrained to lie within a certain range after each update. This ensures the critic satisfies the Lipschitz condition, although it can lead to training difficulties.

\subsubsection{WGAN Example Implementation}
Here is a basic implementation of WGAN using PyTorch, demonstrating the use of Wasserstein loss and weight clipping.

\begin{lstlisting}[style=python]
import torch
import torch.nn as nn
import torch.optim as optim

# Generator Model
class Generator(nn.Module):
    def __init__(self, latent_dim):
        super(Generator, self).__init__()
        self.model = nn.Sequential(
            nn.Linear(latent_dim, 128),
            nn.ReLU(),
            nn.Linear(128, 256),
            nn.ReLU(),
            nn.Linear(256, 512),
            nn.ReLU(),
            nn.Linear(512, 784),
            nn.Tanh()
        )
    
    def forward(self, z):
        img = self.model(z)
        return img.view(img.size(0), 1, 28, 28)

# Critic Model (Discriminator in WGAN is called Critic)
class Critic(nn.Module):
    def __init__(self):
        super(Critic, self).__init__()
        self.model = nn.Sequential(
            nn.Linear(784, 512),
            nn.LeakyReLU(0.2),
            nn.Linear(512, 256),
            nn.LeakyReLU(0.2),
            nn.Linear(256, 1)
        )
    
    def forward(self, img):
        img_flat = img.view(img.size(0), -1)
        return self.model(img_flat)

# Hyperparameters
latent_dim = 100
lr = 0.00005
batch_size = 64
epochs = 50

# Initialize models
generator = Generator(latent_dim)
critic = Critic()

# Optimizers
optimizer_g = optim.RMSprop(generator.parameters(), lr=lr)
optimizer_c = optim.RMSprop(critic.parameters(), lr=lr)

# Training loop
for epoch in range(epochs):
    for i, (imgs, _) in enumerate(dataloader):
        # Train Critic
        real_imgs = imgs
        z = torch.randn(imgs.size(0), latent_dim)
        fake_imgs = generator(z)
        
        # Critic loss
        real_loss = torch.mean(critic(real_imgs))
        fake_loss = torch.mean(critic(fake_imgs.detach()))
        c_loss = -(real_loss - fake_loss)
        
        optimizer_c.zero_grad()
        c_loss.backward()
        optimizer_c.step()

        # Weight clipping
        for p in critic.parameters():
            p.data.clamp_(-0.01, 0.01)

        # Train Generator every few critic updates
        if i % 5 == 0:
            fake_imgs = generator(z)
            g_loss = -torch.mean(critic(fake_imgs))

            optimizer_g.zero_grad()
            g_loss.backward()
            optimizer_g.step()

    print(f"[Epoch {epoch}/{epochs}] [Critic Loss: {c_loss.item():.4f}] [Generator Loss: {g_loss.item():.4f}]")
\end{lstlisting}

This example demonstrates a basic WGAN setup where weight clipping ensures the Lipschitz constraint, and the critic is trained more frequently than the generator to ensure that the Wasserstein distance is well approximated.

\section{WGAN-GP: WGAN with Gradient Penalty}
Although WGAN improves the stability of GAN training, weight clipping introduces its own challenges, such as vanishing and exploding gradients~\cite{gulrajani2017improved}. To address this, WGAN-GP (WGAN with Gradient Penalty) was introduced, which replaces weight clipping with a gradient penalty to enforce the Lipschitz constraint more effectively.

\subsection{The Gradient Penalty Term}
Instead of clipping the weights of the critic, WGAN-GP adds a penalty to the loss function~\cite{gulrajani2017improved} to ensure that the gradients of the critic with respect to its input have a norm of at most 1. The gradient penalty term is defined as:

\[
\lambda \cdot \mathbb{E}_{\hat{x} \sim \mathbb{P}_{\hat{x}}} \left[ \left( \|\nabla_{\hat{x}} D(\hat{x}) \|_2 - 1 \right)^2 \right]
\]

Where \( \hat{x} \) is sampled uniformly along the straight line between a real data point and a generated data point. The penalty encourages the gradients of the critic to have a norm close to 1, ensuring that the critic is a 1-Lipschitz function~\cite{anil2019sorting} without the need for weight clipping.

\subsection{WGAN-GP Implementation Example}
Here is a basic implementation of WGAN-GP in PyTorch:

\begin{lstlisting}[style=python]
# Gradient Penalty Function
def gradient_penalty(critic, real_imgs, fake_imgs):
    alpha = torch.rand(real_imgs.size(0), 1, 1, 1).expand_as(real_imgs)
    interpolates = (alpha * real_imgs + (1 - alpha) * fake_imgs).requires_grad_(True)
    d_interpolates = critic(interpolates)
    fake = torch.ones(real_imgs.size(0), 1)
    gradients = torch.autograd.grad(
        outputs=d_interpolates, inputs=interpolates,
        grad_outputs=fake, create_graph=True, retain_graph=True, only_inputs=True
    )[0]
    gradients = gradients.view(gradients.size(0), -1)
    gradient_penalty = ((gradients.norm(2, dim=1) - 1) ** 2).mean()
    return gradient_penalty

# WGAN-GP Training Loop
lambda_gp = 10  # Gradient penalty coefficient
for epoch in range(epochs):
    for i, (imgs, _) in enumerate(dataloader):
        # Train Critic
        real_imgs = imgs
        z = torch.randn(imgs.size(0), latent_dim)
        fake_imgs = generator(z)
        
        real_loss = torch.mean(critic(real_imgs))
        fake_loss = torch.mean(critic(fake_imgs.detach()))
        gp = gradient_penalty(critic, real_imgs, fake_imgs)
        c_loss = -(real_loss - fake_loss) + lambda_gp * gp

        optimizer_c.zero_grad()
        c_loss.backward()
        optimizer_c.step()

        # Train Generator every few critic updates
        if i % 5 == 0:
            fake_imgs = generator(z)
            g_loss = -torch.mean(critic(fake_imgs))

            optimizer_g.zero_grad()
            g_loss.backward()
            optimizer_g.step()

    print(f"[Epoch {epoch}/{epochs}] [Critic Loss: {c_loss.item():.4f}] [Generator Loss: {g_loss.item():.4f}]")
\end{lstlisting}

In this implementation, the gradient penalty is applied to the critic's loss, ensuring the Lipschitz constraint~\cite{li2019preventing} without the need for weight clipping.

\section{LSGAN: Least Squares Generative Adversarial Networks}
Least Squares GAN (LSGAN)~\cite{mao2017least} is another variant of GANs aimed at addressing the problem of vanishing gradients during training. Instead of using binary cross-entropy as the loss function, LSGAN uses a least-squares loss, which provides smoother gradients and leads to more stable training~\cite{lee2022least}.

\subsection{LSGAN Objective}
In LSGAN, the discriminator is trained to minimize the following least-squares loss for real and generated data:

\[
\min_D \frac{1}{2} \mathbb{E}_{x \sim p_{\text{data}}} [(D(x) - 1)^2] + \frac{1}{2} \mathbb{E}_{z \sim p_z(z)} [D(G(z))^2]
\]

The generator is trained to minimize:

\[
\min_G \frac{1}{2} \mathbb{E}_{z \sim p_z(z)} [(D(G(z)) - 1)^2]
\]

This loss function encourages the discriminator to output values close to 1 for real data and close to 0 for fake data~\cite{lee2022least}. Similarly, the generator is encouraged to produce data that leads the discriminator to output values close to 1.

\subsection{LSGAN Implementation Example}
Here is an example of implementing LSGAN using PyTorch:

\begin{lstlisting}[style=python]
# LSGAN Loss Functions
def lsgan_discriminator_loss(real_preds, fake_preds):
    real_loss = 0.5 * torch.mean((real_preds - 1) ** 2)
    fake_loss = 0.5 * torch.mean(fake_preds ** 2)
    return real_loss + fake_loss

def lsgan_generator_loss(fake_preds):
    return 0.5 * torch.mean((fake_preds - 1) ** 2)

# LSGAN Training Loop
for epoch in range(epochs):
    for i, (imgs, _) in enumerate(dataloader):
        # Train Discriminator
        real_imgs = imgs
        z = torch.randn(imgs.size(0), latent_dim)
        fake_imgs = generator(z)
        
        real_preds = discriminator(real_imgs)
        fake_preds = discriminator(fake_imgs.detach())
        d_loss = lsgan_discriminator_loss(real_preds, fake_preds)

        optimizer_d.zero_grad()
        d_loss.backward()
        optimizer_d.step()

        # Train Generator
        fake_preds = discriminator(fake_imgs)
        g_loss = lsgan_generator_loss(fake_preds)

        optimizer_g.zero_grad()
        g_loss.backward()
        optimizer_g.step()

    print(f"[Epoch {epoch}/{epochs}] [D loss: {d_loss.item():.4f}] [G loss: {g_loss.item():.4f}]")
\end{lstlisting}

This implementation uses least-squares loss for both the discriminator and the generator, leading to more stable training and better gradient flow compared to binary cross-entropy loss.

\section{Summary}
In this chapter, we explored several important GAN variants that aim to improve the stability and performance of GAN training. We covered Wasserstein GAN (WGAN) and its improved version WGAN-GP, which introduces a gradient penalty to enforce the Lipschitz constraint without weight clipping. We also discussed Least Squares GAN (LSGAN), which uses a least-squares loss to provide smoother gradients and more stable training. Each of these methods represents a significant step forward in making GANs easier to train and more reliable in generating high-quality data~\cite{goodfellow2014generative}.

\section{SNGAN: Spectral Normalization GAN}
Spectral Normalization GAN (SNGAN)~\cite{miyato2018spectral} is an extension of GANs that introduces spectral normalization as a method to stabilize GAN training~\cite{lin2021spectral}. Spectral normalization ensures that the weight matrices of the Discriminator have controlled Lipschitz continuity, preventing gradients from exploding or vanishing, which is a common issue in GAN training~\cite{miyato2018spectral}. This technique helps to improve the stability and performance of the model, particularly when training deep architectures.

\subsection{The Role of Spectral Normalization}
Spectral normalization is a technique that stabilizes the training of GANs by normalizing the spectral norm (the largest singular value) of each layer's weight matrix in the Discriminator~\cite{farnia2018generalizable}. By controlling the spectral norm, we can ensure that the Discriminator remains within a specific Lipschitz constant, preventing drastic changes in output when small changes are made to the input.

The spectral norm of a matrix \( W \) is the largest singular value of \( W \), and it is computed as:

\[
\sigma(W) = \max \left\{ \sqrt{\lambda} : \lambda \text{ is an eigenvalue of } W^T W \right\}
\]

By normalizing the weight matrix \( W \) by its spectral norm, we ensure that the function represented by the Discriminator is Lipschitz continuous, meaning that small changes in the input will not cause disproportionately large changes in the output~\cite{bjorck2021towards}.

\textbf{Why is this important?}  
In GAN training, the Discriminator plays a critical role in determining the gradients that the Generator uses to improve. If the Discriminator's gradients are too large, the Generator can receive overly aggressive updates, leading to instability or mode collapse. Spectral normalization helps mitigate this issue by ensuring that the Discriminator's gradients remain well-behaved~\cite{miyato2018spectral}.

\textbf{Example of Spectral Normalization in PyTorch:}

\begin{lstlisting}[style=python]
import torch
import torch.nn as nn
import torch.nn.utils.spectral_norm as spectral_norm

# Discriminator with Spectral Normalization applied to its layers
class SNGAN_Discriminator(nn.Module):
    def __init__(self):
        super(SNGAN_Discriminator, self).__init__()
        self.model = nn.Sequential(
            spectral_norm(nn.Conv2d(3, 64, 4, 2, 1)),   # Apply spectral normalization
            nn.LeakyReLU(0.2, inplace=True),
            spectral_norm(nn.Conv2d(64, 128, 4, 2, 1)),
            nn.LeakyReLU(0.2, inplace=True),
            spectral_norm(nn.Conv2d(128, 256, 4, 2, 1)),
            nn.LeakyReLU(0.2, inplace=True),
            spectral_norm(nn.Conv2d(256, 512, 4, 2, 1)),
            nn.LeakyReLU(0.2, inplace=True),
            nn.Conv2d(512, 1, 4, 1, 0),  # Output a single scalar (real or fake)
            nn.Sigmoid()                 # Sigmoid activation for binary classification
        )

    def forward(self, x):
        return self.model(x)

# Example usage:
input_image = torch.randn(1, 3, 64, 64)  # Random 64x64 RGB image
disc = SNGAN_Discriminator()
output = disc(input_image)
print(output.shape)  # Should output: torch.Size([1, 1, 1, 1])
\end{lstlisting}

In this example, spectral normalization is applied to each convolutional layer of the Discriminator using PyTorch's built-in \texttt{spectral\_norm} function. This ensures that the gradients remain controlled during the training process, leading to more stable and consistent updates~\cite{bjorck2021towards}.

\subsection{Theoretical Analysis of Stabilizing GAN Training}
Spectral normalization enforces a Lipschitz constraint on the Discriminator, which has been shown to stabilize the GAN training process. The stability comes from preventing the Discriminator from becoming too strong, which can lead to vanishing gradients for the Generator. When the Discriminator's gradient becomes too large, the Generator struggles to make meaningful updates, often leading to training failure~\cite{miyato2018spectral, bjorck2021towards}.

The key idea behind this constraint is to prevent the Discriminator from becoming too ``sharp'' in its classification between real and fake data. If the Discriminator's decision boundary is too aggressive, the Generator cannot follow the gradient smoothly, leading to instability or even divergence~\cite{lin2021spectral}. Spectral normalization mitigates this by ensuring that the Discriminator's response to changes in the input is smooth and controlled.

This technique works particularly well with deep architectures, where the risk of gradient explosion or vanishing is higher due to the depth of the network~\cite{miyato2018spectral, lin2021spectral}. By normalizing the weight matrices, we effectively regularize the Discriminator, making the entire GAN framework more robust to training issues.

\textbf{Visualizing the Effect of Spectral Normalization on GAN Training:}

\begin{center}
\begin{tikzpicture}
  [scale=1, every node/.style={scale=1}, 
  block/.style={rectangle, draw, fill=blue!20, text centered, minimum height=3em},
  arrow/.style={->, thick}]

  \node[block] (gen) {Generator};
  \node[block, right=of gen] (disc) {Discriminator};
  \node[block, right=of disc] (spec_norm) {Spectral Normalization};
  \node[block, right=of spec_norm] (stable) {Stable Gradients};

  \draw[arrow] (gen) -- (disc);
  \draw[arrow] (disc) -- (spec_norm);
  \draw[arrow] (spec_norm) -- (stable);
\end{tikzpicture}
\end{center}

In the diagram above, the Generator feeds data into the Discriminator, and spectral normalization ensures that the Discriminator produces stable gradients, which in turn helps stabilize the overall training process.

\section{Unrolled GAN}
Unrolled GAN~\cite{metz2016unrolled} is a variant of GANs that addresses one of the key challenges in GAN training: mode collapse. Mode collapse occurs when the Generator produces a limited variety of outputs, failing to capture the full diversity of the real data distribution~\cite{wu2021modeling}. The unrolled GAN mitigates this issue by unrolling the optimization of the Discriminator for several steps, allowing the Generator to anticipate the Discriminator's updates and adjust accordingly~\cite{wang2022unrolled}.

\subsection{Countermeasures to Mode Collapse}
Mode collapse is a common issue in GANs where the Generator finds a way to fool the Discriminator by producing only a small subset of the real data distribution~\cite{chen2024unsupervised}. For example, in a GAN trained to generate images of handwritten digits, mode collapse might lead the Generator to only produce images of the digit "1", ignoring other digits like "2" or "3". This happens because the Generator finds a way to fool the Discriminator, but only for a narrow range of outputs.

The unrolled GAN introduces a novel solution to this problem by allowing the Generator to "look ahead" at the Discriminator's future updates during training~\cite{metz2016unrolled}. Instead of optimizing the Discriminator for just one step (as in traditional GANs), the Discriminator is unrolled for several steps. This unrolling process helps the Generator anticipate how the Discriminator will change in response to its updates, leading to more diverse and robust generations~\cite{wu2021modeling}.

\textbf{Unrolled GAN Training Process:}

\begin{enumerate}
    \item During each training step, instead of updating the Discriminator after a single forward-backward pass, we simulate multiple updates (i.e., "unroll" the Discriminator's optimization) without actually applying them.
    \item The Generator uses these unrolled updates to predict how the Discriminator will respond to its changes, allowing it to produce more diverse samples~\cite{wang2022unrolled}.
    \item After the unrolling step, we revert the Discriminator to its original state and proceed with the actual update, avoiding computational overhead while still gaining the benefits of unrolling.
\end{enumerate}

\textbf{Example of Unrolled GAN in PyTorch:}

\begin{lstlisting}[style=python]
import torch
import torch.nn as nn
import copy

# Function to unroll the Discriminator for k steps
def unroll_discriminator(discriminator, real_data, fake_data, criterion, k_steps, optimizer_disc):
    disc_copy = copy.deepcopy(discriminator)  # Create a copy of the Discriminator
    for _ in range(k_steps):
        optimizer_disc.zero_grad()
        real_output = disc_copy(real_data)
        fake_output = disc_copy(fake_data)
        loss_real = criterion(real_output, torch.ones_like(real_output))
        loss_fake = criterion(fake_output, torch.zeros_like(fake_output))
        loss = loss_real + loss_fake
        loss.backward()
        optimizer_disc.step()
    return disc_copy  # Return the unrolled Discriminator

# Example usage
disc = SNGAN_Discriminator()  # Spectral Normalized Discriminator
gen = DCGAN_Generator(100)    # DCGAN Generator
optimizer_disc = torch.optim.Adam(disc.parameters(), lr=0.0002)
criterion = nn.BCELoss()

real_data = torch.randn(64, 3, 64, 64)  # Batch of real images
noise = torch.randn(64, 100, 1, 1)     # Random noise for Generator
fake_data = gen(noise)                  # Fake images generated

# Unroll the Discriminator for 5 steps
disc_unrolled = unroll_discriminator(disc, real_data, fake_data, criterion, k_steps=5, optimizer_disc=optimizer_disc)

# After unrolling, update the Generator
optimizer_gen = torch.optim.Adam(gen.parameters(), lr=0.0002)
optimizer_gen.zero_grad()
fake_output = disc_unrolled(fake_data)
loss_gen = criterion(fake_output, torch.ones_like(fake_output))
loss_gen.backward()
optimizer_gen.step()
\end{lstlisting}

In this code, the Discriminator is unrolled for 5 steps before the Generator is updated~\cite{metz2016unrolled}. This unrolling process allows the Generator to see how the Discriminator would evolve and adapt accordingly, helping to mitigate mode collapse~\cite{wang2022unrolled}.

\subsection{Theoretical Insights into Unrolled GAN}
The unrolling technique allows the Generator to better account for the dynamics of the Discriminator. By simulating how the Discriminator will change in the future, the Generator can make more informed updates that lead to a more diverse set of generated outputs.

Unrolling introduces a form of anticipatory learning, where the Generator does not just react to the current state of the Discriminator but also considers its future state. This forward-looking approach helps prevent mode collapse because the Generator can no longer "latch onto" a single mode to fool the Discriminator~\cite{wang2022unrolled}. Instead, it must produce a more diverse range of outputs to continue fooling the Discriminator as it evolves over multiple steps.

\textbf{Visualizing the Unrolled GAN Process:}

\begin{center}
\begin{tikzpicture}
  [scale=1, every node/.style={scale=1}, 
  block/.style={rectangle, draw, fill=blue!20, text centered, minimum height=3em},
  arrow/.style={->, thick}]

  \node[block] (gen) {Generator};
  \node[block, right=of gen] (disc) {Discriminator};
  \node[block, right=of disc] (unroll) {Unroll Discriminator};
  \node[block, right=of unroll] (update_gen) {Update Generator};

  \draw[arrow] (gen) -- (disc);
  \draw[arrow] (disc) -- (unroll);
  \draw[arrow] (unroll) -- (update_gen);
\end{tikzpicture}
\end{center}

In this process, the Generator uses the unrolled Discriminator to make better decisions, leading to more robust training and more diverse generations.

\section{PacGAN: Pack Discriminating GAN}
PacGAN~\cite{lin2018pacgan}, or Pack Discriminating GAN, is an extension of the standard GAN framework aimed at addressing one of the common problems in GAN training: \textbf{mode collapse}. Mode collapse occurs when the generator produces limited diversity in its outputs, meaning different input noise vectors might generate highly similar or identical outputs~\cite{dou2023machine}.

In this section, we will explore how PacGAN tackles this issue, along with its implications for improving GAN training and generating diverse samples.

\subsection{A New Approach to Handling Mode Collapse}

The main innovation in PacGAN is its ability to mitigate mode collapse~\cite{zhang2018convergence} by modifying the discriminator's input. Instead of evaluating individual real or fake samples one at a time, PacGAN passes a \textbf{pack of samples} to the discriminator~\cite{lin2018pacgan}. This allows the discriminator to evaluate whether a set of generated samples has sufficient diversity, rather than just focusing on whether a single sample looks real or fake.

\subsubsection{PacGAN Architecture}
In PacGAN, the discriminator does not take a single image as input but rather a pack of \(k\) images. For instance, if \(k=2\), the discriminator receives two images at once and determines whether they are both real, both fake, or a mixture~\cite{lin2018pacgan}.

Let \(x_i\) represent a real sample and \(G(z_i)\) represent a generated sample. In a standard GAN, the discriminator's objective is to distinguish between individual real and fake samples:
\[
D(x_i) \quad \text{vs} \quad D(G(z_i))
\]
In PacGAN, the discriminator takes a pack of \(k\) images and decides whether the pack contains all real samples or all fake samples:
\[
D([x_1, x_2, \ldots, x_k]) \quad \text{vs} \quad D([G(z_1), G(z_2), \ldots, G(z_k)])
\]

By evaluating multiple samples simultaneously, the discriminator becomes more sensitive to the lack of diversity in the generator's outputs~\cite{lin2018pacgan}. If the generator produces similar images for different noise inputs, the discriminator will recognize the similarity and penalize the generator, forcing it to generate more diverse outputs.

\subsubsection{Implementation of PacGAN in PyTorch}
Below is an example of how to implement PacGAN in PyTorch, using a pack size of 2:

\begin{lstlisting}[style=python]
import torch
import torch.nn as nn
import torch.optim as optim

# Discriminator model for PacGAN (input pack of 2 images)
class PacDiscriminator(nn.Module):
    def __init__(self, pack_size):
        super(PacDiscriminator, self).__init__()
        self.pack_size = pack_size
        self.main = nn.Sequential(
            nn.Linear(784 * pack_size, 512),
            nn.LeakyReLU(0.2),
            nn.Linear(512, 256),
            nn.LeakyReLU(0.2),
            nn.Linear(256, 1),
            nn.Sigmoid()
        )
    
    def forward(self, x):
        # Flatten the pack of images into a single vector for the discriminator
        x = x.view(x.size(0), -1)
        return self.main(x)

# Generator model (same as standard GAN)
class Generator(nn.Module):
    def __init__(self):
        super(Generator, self).__init__()
        self.main = nn.Sequential(
            nn.Linear(100, 256),
            nn.ReLU(),
            nn.Linear(256, 512),
            nn.ReLU(),
            nn.Linear(512, 784),
            nn.Tanh()
        )
    
    def forward(self, x):
        return self.main(x).view(-1, 1, 28, 28)

# Instantiate models and optimizers
pack_size = 2
D = PacDiscriminator(pack_size=pack_size)
G = Generator()

optimizer_D = optim.Adam(D.parameters(), lr=0.0002)
optimizer_G = optim.Adam(G.parameters(), lr=0.0002)
criterion = nn.BCELoss()

# Training loop
for epoch in range(num_epochs):
    # Generate fake images
    noise = torch.randn(batch_size, 100)
    fake_images = G(noise)

    # Create packs of real and fake images
    real_images = get_real_images(batch_size // pack_size, 28*28)
    real_packs = real_images.view(-1, pack_size, 28*28)
    fake_packs = fake_images.view(-1, pack_size, 28*28)

    # Train Discriminator
    optimizer_D.zero_grad()
    # Real packs
    output_real = D(real_packs)
    loss_real = criterion(output_real, torch.ones(real_packs.size(0), 1))
    # Fake packs
    output_fake = D(fake_packs.detach())
    loss_fake = criterion(output_fake, torch.zeros(fake_packs.size(0), 1))
    # Backprop
    loss_D = loss_real + loss_fake
    loss_D.backward()
    optimizer_D.step()

    # Train Generator
    optimizer_G.zero_grad()
    output_fake = D(fake_packs)
    loss_G = criterion(output_fake, torch.ones(fake_packs.size(0), 1))
    loss_G.backward()
    optimizer_G.step()
\end{lstlisting}

In this implementation, the discriminator evaluates packs of 2 images at a time. The rest of the training loop is similar to a standard GAN, but with the discriminator focusing on packs instead of individual samples.

\subsection{Advantages of PacGAN}
PacGAN introduces several advantages compared to traditional GANs:

\begin{itemize}
    \item \textbf{Better Diversity}: By forcing the discriminator to evaluate multiple samples, PacGAN encourages the generator to produce a wider variety of outputs, reducing mode collapse.
    \item \textbf{Improved Sample Quality}: The generator is penalized if it fails to produce distinct samples for different noise vectors, leading to higher-quality images~\cite{lin2018pacgan}.
    \item \textbf{Ease of Implementation}: The PacGAN architecture builds on standard GAN frameworks, requiring only minimal changes to the discriminator's input and output processing.
\end{itemize}

\section{Regularization Techniques in GANs}
Regularization techniques in GANs are crucial for stabilizing training and ensuring that the generator and discriminator learn effectively. In this section, we will explore several important regularization techniques, including \textbf{gradient penalty}~\cite{gao2020data}, \textbf{experience replay}~\cite{wu2018memory}, \textbf{noise injection}~\cite{feng2021understanding}, and \textbf{gradient clipping}~\cite{zhang2019gradient}.

\subsection{Gradient Penalty}
The gradient penalty is a regularization technique used to enforce the Lipschitz continuity of the discriminator~\cite{gao2020data}. This is especially important in Wasserstein GANs (WGANs), where the discriminator (or critic) must satisfy the 1-Lipschitz constraint to ensure the Wasserstein distance is properly estimated~\cite{arjovsky2017wasserstein}.

\subsubsection{WGAN-GP: Gradient Penalty in WGANs}
Instead of using weight clipping (which can lead to optimization issues), WGAN-GP introduces a gradient penalty term. The gradient penalty encourages the gradient norm of the discriminator to stay close to 1 for all inputs, helping to maintain the Lipschitz constraint~\cite{arjovsky2017wasserstein}.

The gradient penalty is defined as:
\[
\mathcal{L}_{GP} = \lambda \mathbb{E}_{\hat{x}} \left[ \left( || \nabla_{\hat{x}} D(\hat{x}) ||_2 - 1 \right)^2 \right]
\]
where \(\hat{x}\) is a random interpolation between real and fake samples, and \(\lambda\) is a hyperparameter that controls the strength of the penalty.

\subsubsection{PyTorch Implementation of WGAN-GP}
Below is an example of how to implement the gradient penalty in WGAN-GP using PyTorch:

\begin{lstlisting}[style=python]
# Function to compute the gradient penalty
def gradient_penalty(D, real_samples, fake_samples):
    batch_size = real_samples.size(0)
    epsilon = torch.rand(batch_size, 1, 1, 1).to(real_samples.device)
    interpolated = epsilon * real_samples + (1 - epsilon) * fake_samples
    interpolated.requires_grad_(True)
    
    d_interpolated = D(interpolated)
    gradients = torch.autograd.grad(
        outputs=d_interpolated,
        inputs=interpolated,
        grad_outputs=torch.ones_like(d_interpolated),
        create_graph=True,
        retain_graph=True,
        only_inputs=True
    )[0]
    
    gradients = gradients.view(batch_size, -1)
    gradient_norm = gradients.norm(2, dim=1)
    penalty = ((gradient_norm - 1) ** 2).mean()
    return penalty

# WGAN-GP training loop
lambda_gp = 10  # Gradient penalty weight
for epoch in range(num_epochs):
    # Train Discriminator
    optimizer_D.zero_grad()
    
    # Real and fake data
    real_data = get_real_images(batch_size, 28*28)
    noise = torch.randn(batch_size, 100)
    fake_data = G(noise)
    
    # Discriminator outputs
    real_output = D(real_data)
    fake_output = D(fake_data.detach())
    
    # Gradient penalty
    gp = gradient_penalty(D, real_data, fake_data)
    
    # Losses and backprop
    loss_D = torch.mean(fake_output) - torch.mean(real_output) + lambda_gp * gp
    loss_D.backward()
    optimizer_D.step()

    # Train Generator
    optimizer_G.zero_grad()
    fake_output = D(fake_data)
    loss_G = -torch.mean(fake_output)
    loss_G.backward()
    optimizer_G.step()
\end{lstlisting}

\subsection{Experience Replay and Noise Injection}

Another regularization technique in GANs is \textbf{experience replay}, which borrows concepts from reinforcement learning. The idea is to store past generated samples and occasionally reintroduce them into the training process to prevent the discriminator from forgetting about earlier parts of the data distribution.

\subsubsection{Noise Injection for Smoother Training}
\textbf{Noise injection} is a technique where small amounts of noise are added to the inputs of the discriminator or generator during training. This can help smooth out training, making the models less sensitive to small changes in the input data~\cite{feng2021understanding}.

For example, Gaussian noise can be added to the input data:
\begin{lstlisting}[style=python]
# Adding noise to the discriminator input
noise_level = 0.05

# Apply noise to real and fake images
real_images_with_noise = real_images + noise_level * torch.randn_like(real_images)
fake_images_with_noise = fake_images + noise_level * torch.randn_like(fake_images)

# Train the discriminator with noisy images
output_real = D(real_images_with_noise)
output_fake = D(fake_images_with_noise)
\end{lstlisting}

This method encourages the generator to produce images that are robust to small perturbations, which can improve generalization and reduce overfitting.

\subsection{Gradient Clipping Techniques}
\textbf{Gradient clipping} is a simple but effective technique to stabilize GAN training. During backpropagation, gradients can sometimes explode or vanish, leading to instability in the training process. Gradient clipping ensures that the gradient norms do not exceed a specified threshold, preventing large updates that could destabilize the training~\cite{zhang2019gradient}.

\begin{lstlisting}[style=python]
# Gradient clipping example
for p in D.parameters():
    p.grad.data.clamp_(-0.01, 0.01)  # Clip gradients between -0.01 and 0.01
\end{lstlisting}

This technique is especially useful in WGANs but can be applied to other GAN architectures as well.

\section{Conclusion}

In this chapter, we explored advanced GAN techniques like PacGAN for addressing mode collapse, gradient penalty for stabilizing training in WGANs, and regularization strategies such as noise injection~\cite{feng2021understanding} and gradient clipping~\cite{zhang2019gradient}. Each of these techniques helps improve the robustness and performance of GANs, enabling them to generate more diverse and high-quality outputs. Understanding and applying these techniques is key to mastering GAN training.

\chapter{Architectural Improvements in Generators and Discriminators}
As GANs have evolved, researchers have continually proposed architectural improvements to the generator and discriminator to enhance performance, particularly for high-resolution image generation~\cite{wang2018cgan, goodfellow2014generative}. One of the most influential approaches is Progressive Growing of GANs (ProGAN)~\cite{karras2017progressive}, which introduces a unique training strategy that significantly improves the quality of generated high-resolution images. In this chapter, we will explore the core ideas behind progressive training and how ProGAN achieves high-quality results, especially in generating large-scale images.

\section{Progressive Growing of GANs (ProGAN)}
Progressive Growing of GANs, introduced by Karras et al. in 2017~\cite{karras2017progressive}, is a method specifically designed to stabilize GAN training for high-resolution image generation. The idea is to gradually increase the complexity of the task by starting with a low-resolution image and progressively adding layers to both the generator and discriminator as training progresses. This gradual increase allows the network to learn the basic structure of the images at a low resolution before handling finer details, which significantly improves both training stability and the quality of the generated images~\cite{song2021gansim}.

\subsection{Core Idea of Progressive Training}
The core idea behind ProGAN is to train the GAN in phases, starting with small images (e.g., 4x4 pixels) and gradually increasing the resolution (e.g., 8x8, 16x16, 32x32, etc.) by adding layers to both the generator and discriminator.

\subsubsection{Progressive Layer Addition}
The training begins with a small resolution (e.g., 4x4 pixels). Once the network stabilizes at this resolution, new layers are added to both the generator and the discriminator, doubling the resolution (e.g., 8x8). This process continues until the desired resolution is reached (e.g., 1024x1024 for very high-resolution images). 

In each phase, the network learns to generate increasingly complex image features, starting with basic shapes and structures and progressing to finer details such as textures~\cite{karras2017progressive}. This progressive approach allows the model to focus on learning the essential structures of the image first, which leads to higher-quality results at larger resolutions~\cite{zhang2019progressive}.

\subsubsection{Fade-in Transition}
To avoid abrupt changes when new layers are added, ProGAN introduces a ``fade-in'' mechanism~\cite{zheng2022exploratory} during the transition between resolutions. Initially, when a new layer is added, its influence is weighted by a factor that gradually increases over time. This smooth transition helps maintain stability during training and prevents the network from being overwhelmed by the sudden increase in complexity~\cite{karras2017progressive}.

For example, if the network is transitioning from 8x8 to 16x16 resolution, the output from the new layers that handle 16x16 resolution is blended with the output from the previous layers (which handle 8x8 resolution) during the early stages of the transition. Over time, the contribution from the new layers increases until they fully take over.

\subsection{Improving the Quality of High-Resolution Image Generation}
The key benefit of ProGAN is its ability to generate high-quality, high-resolution images that maintain coherent global structure and fine details. This is achieved through a combination of architectural improvements and training strategies.

\subsubsection{Handling Large-Scale Data}
Generating large-scale images with traditional GAN architectures often leads to problems such as mode collapse, poor diversity, and instability~\cite{pathak2018efficient, kang2023scaling}. ProGAN overcomes these issues by training in stages, ensuring that the generator and discriminator learn to handle complexity progressively. By the time the model reaches high resolutions, it has already learned the core features of the data at lower resolutions, making it more stable and capable of producing diverse, realistic images~\cite{kang2023scaling}.

\subsubsection{Fine Details and Texture Learning}
As the resolution increases during training, ProGAN layers are able to focus on finer details, such as textures~\cite{ma1996texture} and edges, while preserving the overall structure of the image~\cite{dundar2023fine}. For example, in generating human faces, ProGAN can first learn the basic layout of facial features (eyes, nose, mouth) at low resolution and then gradually add details like skin texture, hair strands, and lighting effects as the resolution increases~\cite{dundar2023fine}.

\subsubsection{Training Stability}
Training GANs is often unstable, particularly when dealing with high-resolution images. The progressive training strategy of ProGAN helps alleviate this by simplifying the task in the early stages~\cite{dundar2023fine}. As the generator and discriminator are initially trained on small images, they can learn stable representations before tackling the more challenging task of generating high-resolution images. This reduces the likelihood of training collapse and leads to more consistent results~\cite{pathak2018efficient}.

\subsection{Step-by-Step Example of ProGAN using PyTorch}
To help you understand how ProGAN works in practice, let's walk through a simplified example using PyTorch. In this example, we will create a basic ProGAN-like architecture that starts with a small image resolution and progressively grows to handle larger resolutions.

\subsubsection{Step 1: Importing Necessary Libraries}
We first need to import the required libraries for our implementation:

\begin{lstlisting}[style=python]
import torch
import torch.nn as nn
import torch.optim as optim
import torch.nn.functional as F
from torch.autograd import Variable
import numpy as np
import matplotlib.pyplot as plt
\end{lstlisting}

\subsubsection{Step 2: Define the Generator and Discriminator}
In ProGAN, both the generator and discriminator architectures are designed to grow progressively as new layers are added during training. For simplicity, we will define basic versions of these models and then progressively add layers to them.

\begin{lstlisting}[style=python]
# Basic block used in both the generator and discriminator
class ConvBlock(nn.Module):
    def __init__(self, in_channels, out_channels, kernel_size=3, padding=1):
        super(ConvBlock, self).__init__()
        self.conv = nn.Conv2d(in_channels, out_channels, kernel_size, padding=padding)
        self.bn = nn.BatchNorm2d(out_channels)
        self.activation = nn.LeakyReLU(0.2)
    
    def forward(self, x):
        x = self.conv(x)
        x = self.bn(x)
        return self.activation(x)

# Generator model
class Generator(nn.Module):
    def __init__(self, latent_dim):
        super(Generator, self).__init__()
        self.initial_layer = nn.Sequential(
            nn.Linear(latent_dim, 128 * 4 * 4),
            nn.ReLU()
        )
        self.conv_blocks = nn.ModuleList([
            ConvBlock(128, 128),
            ConvBlock(128, 64),
            ConvBlock(64, 32)
        ])
        self.to_rgb = nn.Conv2d(32, 3, kernel_size=1, stride=1, padding=0)
    
    def forward(self, z):
        # Start with a 4x4 image
        x = self.initial_layer(z).view(-1, 128, 4, 4)
        for block in self.conv_blocks:
            x = F.interpolate(x, scale_factor=2)  # Upsample image progressively
            x = block(x)
        return torch.tanh(self.to_rgb(x))

# Discriminator model
class Discriminator(nn.Module):
    def __init__(self):
        super(Discriminator, self).__init__()
        self.conv_blocks = nn.ModuleList([
            ConvBlock(3, 32),
            ConvBlock(32, 64),
            ConvBlock(64, 128)
        ])
        self.fc = nn.Sequential(
            nn.Linear(128 * 4 * 4, 1),
            nn.Sigmoid()
        )
    
    def forward(self, x):
        for block in self.conv_blocks:
            x = block(x)
            x = F.avg_pool2d(x, kernel_size=2)  # Downsample image progressively
        x = x.view(x.size(0), -1)
        return self.fc(x)
\end{lstlisting}

Here, the generator starts by generating a small 4x4 image, which is progressively upsampled as it passes through the convolutional blocks. Similarly, the discriminator starts with a high-resolution image and progressively downsamples it before making a final classification.

\subsubsection{Step 3: Training Loop with Progressive Layer Addition}
Next, we define the training loop, where we progressively add layers to both the generator and discriminator as training progresses. We'll use a simplified version of ProGAN's fade-in mechanism to gradually introduce new layers~\cite{karras2017progressive}.

\begin{lstlisting}[style=python]
# Initialize models
latent_dim = 100
generator = Generator(latent_dim)
discriminator = Discriminator()

# Optimizers
optimizer_g = optim.Adam(generator.parameters(), lr=0.0002, betas=(0.5, 0.999))
optimizer_d = optim.Adam(discriminator.parameters(), lr=0.0002, betas=(0.5, 0.999))

# Training loop
epochs = 10
for epoch in range(epochs):
    for i, (real_imgs, _) in enumerate(dataloader):
        
        # Train Discriminator
        z = torch.randn(real_imgs.size(0), latent_dim)
        fake_imgs = generator(z)
        
        real_validity = discriminator(real_imgs)
        fake_validity = discriminator(fake_imgs.detach())
        
        d_loss_real = F.binary_cross_entropy(real_validity, torch.ones_like(real_validity))
        d_loss_fake = F.binary_cross_entropy(fake_validity, torch.zeros_like(fake_validity))
        d_loss = (d_loss_real + d_loss_fake) / 2

        optimizer_d.zero_grad()
        d_loss.backward()
        optimizer_d.step()

        # Train Generator
        fake_validity = discriminator(fake_imgs)
        g_loss = F.binary_cross_entropy(fake_validity, torch.ones_like(fake_validity))

        optimizer_g.zero_grad()
        g_loss.backward()
        optimizer_g.step()

    print(f"Epoch [{epoch}/{epochs}], D Loss: {d_loss.item()}, G Loss: {g_loss.item()}")
\end{lstlisting}

This basic training loop follows the usual GAN framework but with the additional concept of progressively increasing the resolution of generated images as new layers are introduced. In practice, this approach helps to stabilize training and allows the network to generate high-quality, high-resolution images over time.

\section{Summary}
In this chapter, we explored Progressive Growing of GANs (ProGAN), an important architectural innovation that significantly improves the stability and quality of GAN training, particularly for high-resolution image generation~\cite{karras2017progressive}. ProGAN's core idea is to train the GAN in phases, progressively increasing the resolution and complexity of the images. By starting with low-resolution images and gradually adding layers, ProGAN achieves more stable training and produces high-quality results. The use of fade-in transitions during layer addition further ensures smooth training progress. We also walked through an example implementation of ProGAN using PyTorch, providing a clear, step-by-step guide for beginners.

\section{BigGAN: Large-Scale Generative Adversarial Networks}
BigGAN~\cite{donahue2019large} is a powerful extension of the traditional GAN architecture, designed specifically to generate high-quality, large-scale images. Unlike traditional GANs that may struggle with larger, more complex datasets, BigGAN introduces several architectural and training improvements that allow it to generate realistic, high-resolution images on large-scale datasets such as ImageNet~\cite{deng2009imagenet}. This section will explain how BigGAN achieves this, and detail the techniques used to train it effectively`\cite{zhou2024comprehensive}.

\subsection{Generating High-Quality Large-Scale Images}
Generating high-quality, large-scale images is challenging because of the high complexity and variability present in real-world datasets. Traditional GAN architectures often produce blurry or low-quality images when scaled up to higher resolutions (e.g., \(256 \times 256\) or higher) or larger datasets (e.g., ImageNet). BigGAN addresses this problem by introducing several key innovations:

\textbf{1. Class-Conditional Batch Normalization:}  
BigGAN leverages class-conditional batch normalization (CCBN)~\cite{wang2020attentive} to condition both the Generator and Discriminator on class labels. In CCBN, the scale and shift parameters of the batch normalization layers are conditioned on the class label, allowing the Generator to produce images specific to a certain class, while still benefiting from the regularization properties of batch normalization~\cite{donahue2019large}.

\textbf{2. Larger Batch Sizes:}  
One of the fundamental challenges in GAN training is maintaining stability, especially as image resolution increases. BigGAN addresses this by utilizing larger batch sizes during training, which helps to reduce gradient noise and stabilize training. Larger batch sizes allow for more consistent updates to both the Generator and Discriminator, leading to higher-quality images~\cite{donahue2019large}.

\textbf{3. Orthogonal Regularization:}  
To prevent the Discriminator from becoming overly powerful and causing training instability, BigGAN applies orthogonal regularization to the weight matrices of the Generator. This prevents the weights from becoming too correlated, thus encouraging diversity in the generated images~\cite{wang2015deep}.

\textbf{4. Truncated Sampling:}  
BigGAN also introduces truncated sampling~\cite{brock2018large}, a method for controlling the diversity-quality tradeoff. Instead of sampling noise \(z\) from a normal distribution, BigGAN samples from a truncated normal distribution, which restricts the noise to a certain range. By limiting the noise input, the Generator is forced to focus on generating high-quality images that are more consistent with the target distribution~\cite{donahue2019large}. The truncation parameter can be adjusted to balance diversity and quality.

\textbf{Example: BigGAN Generator with Class-Conditional Batch Normalization in PyTorch}

\begin{lstlisting}[style=python]
import torch
import torch.nn as nn

# Class-Conditional BatchNorm2d
class ConditionalBatchNorm2d(nn.Module):
    def __init__(self, num_features, num_classes):
        super(ConditionalBatchNorm2d, self).__init__()
        self.bn = nn.BatchNorm2d(num_features, affine=False)
        self.embed = nn.Embedding(num_classes, num_features * 2)
        self.embed.weight.data[:, :num_features].normal_(1, 0.02)  # Scale
        self.embed.weight.data[:, num_features:].zero_()           # Shift

    def forward(self, x, y):
        out = self.bn(x)
        gamma, beta = self.embed(y).chunk(2, 1)
        gamma = gamma.view(-1, out.size(1), 1, 1)
        beta = beta.view(-1, out.size(1), 1, 1)
        return gamma * out + beta

# BigGAN Generator block
class BigGAN_GeneratorBlock(nn.Module):
    def __init__(self, in_channels, out_channels, num_classes):
        super(BigGAN_GeneratorBlock, self).__init__()
        self.conv1 = nn.ConvTranspose2d(in_channels, out_channels, 4, 2, 1)
        self.bn1 = ConditionalBatchNorm2d(out_channels, num_classes)
        self.relu = nn.ReLU(True)
        self.conv2 = nn.ConvTranspose2d(out_channels, out_channels, 3, 1, 1)
        self.bn2 = ConditionalBatchNorm2d(out_channels, num_classes)
    
    def forward(self, x, y):
        x = self.conv1(x)
        x = self.bn1(x, y)
        x = self.relu(x)
        x = self.conv2(x)
        x = self.bn2(x, y)
        return self.relu(x)

# Example usage
noise = torch.randn(16, 128, 1, 1)  # Noise vector
labels = torch.randint(0, 1000, (16,))  # Random class labels for 1000 classes
gen_block = BigGAN_GeneratorBlock(128, 256, 1000)
output = gen_block(noise, labels)
print(output.shape)  # Output should be torch.Size([16, 256, 4, 4])
\end{lstlisting}

In this example, the Generator block uses class-conditional batch normalization to condition the generation process on class labels. This is crucial for BigGAN's ability to generate diverse, high-quality images across many different categories.

\subsection{Training Techniques for Large-Scale Datasets}
Training a GAN on large-scale datasets like ImageNet presents several challenges, including the need for stable training, efficient resource utilization, and ensuring diversity in the generated images~\cite{donahue2019large}. BigGAN introduces several techniques to handle these challenges effectively:

\textbf{1. Gradient Accumulation for Large Batch Sizes:}  
BigGAN uses very large batch sizes (up to 2048 samples) during training, which can be resource-intensive. When the hardware cannot support such large batches in memory, gradient accumulation is used. Gradient accumulation involves accumulating gradients over multiple smaller batches and then updating the model parameters as if a larger batch was used. This allows the model to simulate training with a large batch size without the need for extensive hardware resources~\cite{brock2018large}.

\textbf{2. Self-Attention Mechanism:}  
To improve the generation of fine details and capture long-range dependencies in the images, BigGAN incorporates a self-attention mechanism~\cite{vaswani2017attention}. This helps the model focus on different parts of the image, allowing it to generate globally coherent images. The self-attention module is especially useful in generating high-resolution images where capturing global structure is critical.

\textbf{3. Spectral Normalization:}  
BigGAN also applies spectral normalization to both the Generator and the Discriminator. Spectral normalization helps to stabilize training by controlling the Lipschitz constant of the networks. This technique, already discussed in the context of SNGAN, is crucial for ensuring that the gradients do not explode or vanish, making it possible to train on large, complex datasets~\cite{donahue2019large}.

\textbf{4. Adaptive Learning Rates:}  
Since different layers of the Generator and Discriminator can have different magnitudes of gradient updates, BigGAN uses adaptive learning rates for different layers~\cite{ma2019adaptive}. This helps to balance the training dynamics and ensure that no single layer dominates the learning process.

\textbf{Example: Adding Self-Attention to BigGAN in PyTorch}

\begin{lstlisting}[style=python]
class SelfAttention(nn.Module):
    def __init__(self, in_channels):
        super(SelfAttention, self).__init__()
        self.query = nn.Conv2d(in_channels, in_channels // 8, 1)
        self.key = nn.Conv2d(in_channels, in_channels // 8, 1)
        self.value = nn.Conv2d(in_channels, in_channels, 1)
        self.gamma = nn.Parameter(torch.zeros(1))
    
    def forward(self, x):
        batch_size, C, width, height = x.size()
        query = self.query(x).view(batch_size, -1, width * height)  # B x C/8 x N
        key = self.key(x).view(batch_size, -1, width * height)      # B x C/8 x N
        value = self.value(x).view(batch_size, -1, width * height)  # B x C x N
        
        attention = torch.bmm(query.permute(0, 2, 1), key)          # B x N x N
        attention = torch.softmax(attention, dim=-1)
        
        out = torch.bmm(value, attention.permute(0, 2, 1))          # B x C x N
        out = out.view(batch_size, C, width, height)
        
        return self.gamma * out + x

# Example of using Self-Attention in BigGAN
class BigGAN_GeneratorWithAttention(nn.Module):
    def __init__(self, noise_dim, num_classes):
        super(BigGAN_GeneratorWithAttention, self).__init__()
        self.block1 = BigGAN_GeneratorBlock(noise_dim, 256, num_classes)
        self.attention1 = SelfAttention(256)
        self.block2 = BigGAN_GeneratorBlock(256, 128, num_classes)
    
    def forward(self, x, y):
        x = self.block1(x, y)
        x = self.attention1(x)  # Apply attention mechanism
        x = self.block2(x, y)
        return x

# Example usage
noise = torch.randn(16, 128, 1, 1)
labels = torch.randint(0, 1000, (16,))
gen = BigGAN_GeneratorWithAttention(128, 1000)
output = gen(noise, labels)
print(output.shape)  # Should output torch.Size([16, 128, 8, 8])
\end{lstlisting}

In this example, self-attention is integrated into the BigGAN architecture to capture global dependencies in the generated images, improving the quality of fine details and structure.

\subsubsection{Techniques for Efficient Training on Large Datasets}
To handle the complexity and size of large-scale datasets like ImageNet, BigGAN implements several advanced training techniques~\cite{adadi2021survey}:

\begin{itemize}
    \item \textbf{Multi-GPU Training:} To accommodate large models and batch sizes, BigGAN is often trained on multiple GPUs, distributing the workload and reducing training time.
    \item \textbf{Data Augmentation:} To improve generalization and prevent overfitting, BigGAN employs extensive data augmentation techniques such as random cropping, flipping, and color jittering.
    \item \textbf{Truncated Sampling:} Truncated sampling is used to improve the visual quality of generated images by controlling the range of noise inputs.
\end{itemize}

\textbf{Visualizing BigGAN's Training Process:}

\begin{center}
\begin{tikzpicture}
  [scale=1, every node/.style={scale=1}, 
  block/.style={rectangle, draw, fill=blue!20, text centered, minimum height=3em},
  arrow/.style={->, thick}]

  \node[block] (noise) {Noise $z$};
  \node[block, right=of noise] (gen) {BigGAN Generator};
  \node[block, right=of gen] (gen_out) {Generated Images};
  \node[block, below=of gen_out] (disc) {BigGAN Discriminator};
  \node[block, below=of disc] (real_data) {Real Data};
  
  \draw[arrow] (noise) -- (gen);
  \draw[arrow] (gen) -- (gen_out);
  \draw[arrow] (real_data) -- (disc);
  \draw[arrow] (gen_out) -- (disc);
\end{tikzpicture}
\end{center}

In this diagram, noise is passed through the BigGAN Generator to produce high-quality, large-scale images, which are then passed through the Discriminator for real vs fake classification. BigGAN applies techniques like self-attention and class-conditional batch normalization to ensure high-quality outputs.

\section{StyleGAN and StyleGAN2}
StyleGAN and its successor StyleGAN2~\cite{karras2020analyzing} represent a significant advancement in the field of GANs, particularly in terms of controllable image generation and high-quality results~\cite{khan2022transformers}. These models introduce innovative techniques such as style-based architecture and multi-resolution synthesis, which allow for fine-grained control over the features of generated images~\cite{bond2021deep}. In this section, we will explore the key concepts of StyleGAN and StyleGAN2, focusing on style control, multi-resolution generation, style mixing, feature interpolation, and their applications in image editing.

\begin{figure}[htbp]
    \centering
    \includegraphics[width=\textwidth]{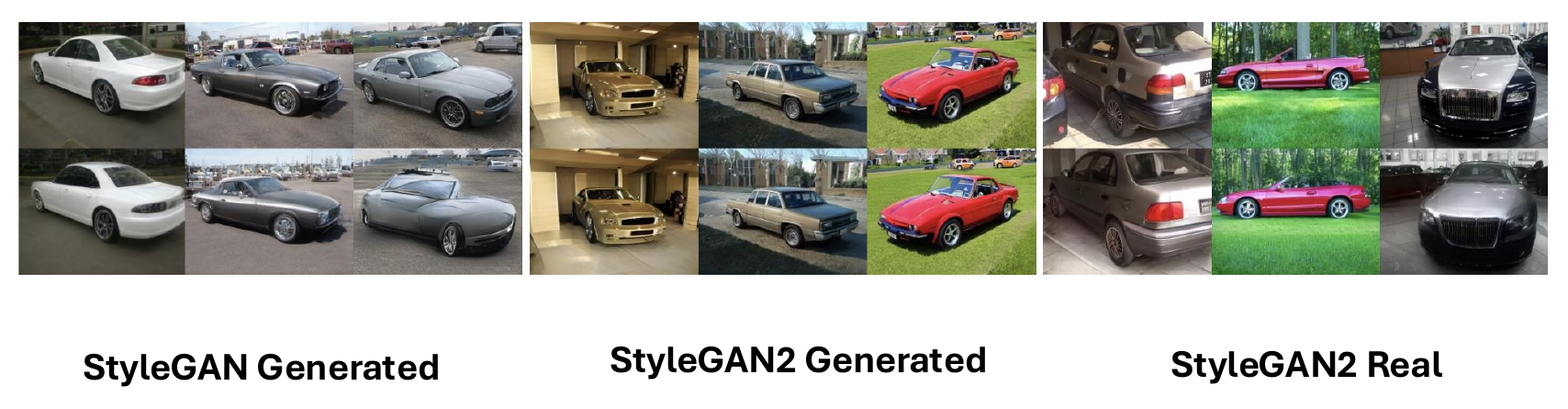}
    \caption{Example images and their projected and re-synthesized counterparts. For each configuration, top row shows the target images
    and bottom row shows the synthesis of the corresponding projected latent vector and noise inputs. With the baseline StyleGAN, projection
    often finds a reasonably close match for generated images, but especially the backgrounds differ from the originals. Image from Karras et al.\cite{karras2020analyzing} in 2020 StyleGAN2 paper.}
\end{figure}

\subsection{Style Control and Multi-Resolution Generation}

One of the main innovations in StyleGAN is its \textbf{style-based generator architecture}~\cite{karras2020analyzing}, which differs from traditional GANs. Instead of feeding the input noise directly into the generator, StyleGAN uses an intermediate latent space that allows for more structured control over the generated images. This architecture enables the separation of high-level and low-level features, leading to better control over the image generation process.

\subsubsection{Latent Space in StyleGAN}
In traditional GANs, a noise vector \( z \) sampled from a distribution (such as a normal distribution) is directly input to the generator~\cite{abdal2019image2stylegan}. However, in StyleGAN, the input noise \( z \) is first mapped to an intermediate latent space \( w \) using a learned function called a \textbf{mapping network}~\cite{karras2020analyzing}:
\[
w = M(z)
\]
Here, \( M \) is a multi-layer perceptron (MLP) that transforms the latent vector into a different space that is better suited for controlling image features. This allows for disentangling the features in a more intuitive way.

\subsubsection{AdaIN (Adaptive Instance Normalization)}
StyleGAN uses \textbf{Adaptive Instance Normalization} (AdaIN)~\cite{huang2017arbitrary} to control the style at different layers of the generator. AdaIN works by modulating the feature maps of the generator based on the style vector \( w \) for each resolution~\cite{kim2020transfer}. Specifically, AdaIN modifies the mean and variance of the feature maps at each layer using the style vector:
\[
\text{AdaIN}(x, y) = y_s \left( \frac{x - \mu(x)}{\sigma(x)} \right) + y_b
\]
Where:
\begin{itemize}
    \item \( x \) is the feature map.
    \item \( y_s \) and \( y_b \) are the style-specific scaling and bias values derived from \( w \).
    \item \( \mu(x) \) and \( \sigma(x) \) are the mean and standard deviation of the feature map.
\end{itemize}
By applying AdaIN at different layers, StyleGAN allows control over different levels of details in the generated image.

\subsubsection{Multi-Resolution Synthesis}
Another major innovation of StyleGAN is its ability to generate images at multiple resolutions, with fine control over different levels of detail. This is achieved by applying the style vector \( w \) at various stages of the generator, which corresponds to different image resolutions (e.g., low-level features like pose and shape at coarse resolutions, and high-level details like texture and color at finer resolutions)~\cite{jing2020dynamic}.

The generator starts by producing a low-resolution image, which is progressively upsampled to higher resolutions, with each stage adding more details. The style vector \( w \) controls the features generated at each resolution, providing fine-grained control over both global structure and local details.

\begin{lstlisting}[style=python]
# Example of AdaIN in PyTorch
import torch
import torch.nn as nn

class AdaIN(nn.Module):
    def __init__(self):
        super(AdaIN, self).__init__()

    def forward(self, content_features, style_features):
        content_mean, content_std = self._get_mean_std(content_features)
        style_mean, style_std = self._get_mean_std(style_features)

        normalized_content = (content_features - content_mean) / content_std
        return style_std * normalized_content + style_mean

    def _get_mean_std(self, features, eps=1e-5):
        size = features.size()
        mean = features.view(size[0], size[1], -1).mean(2).view(size[0], size[1], 1, 1)
        std = features.view(size[0], size[1], -1).std(2).view(size[0], size[1], 1, 1)
        return mean, std
\end{lstlisting}

In this code, AdaIN normalizes the content features (feature maps) using the statistics (mean and standard deviation) derived from the style features~\cite{karras2020analyzing}. This operation modulates the content features according to the desired style, as controlled by the latent vector \( w \).

\subsection{Style Mixing and Feature Interpolation}

\subsubsection{Style Mixing}

\textbf{Style mixing}~\cite{zhang2022styleswin} is another important concept introduced in StyleGAN, which allows for the combination of styles from multiple latent vectors to generate hybrid images~\cite{karras2020analyzing}. This technique helps the model learn more diverse representations and prevents overfitting to specific styles.

In style mixing, two different latent vectors \( w_1 \) and \( w_2 \) are used at different stages of the generator. For example, \( w_1 \) might be applied to the low-resolution layers, controlling global attributes like pose, while \( w_2 \) is applied to the high-resolution layers, controlling finer details like texture. This leads to images that inherit attributes from both latent vectors:
\[
\text{Generator}(w_1, w_2) = \text{AdaIN}(x, w_1) \quad \text{for low-res layers}, \quad \text{AdaIN}(x, w_2) \quad \text{for high-res layers}
\]

\begin{lstlisting}[style=python]
# Example of style mixing in PyTorch
def style_mixing(generator, w1, w2, mixing_point):
    """
    Apply w1 for layers up to mixing_point, and w2 for the rest.
    """
    for i, layer in enumerate(generator.layers):
        if i < mixing_point:
            style = w1
        else:
            style = w2
        x = layer.apply_style(x, style)
    return x
\end{lstlisting}

\subsubsection{Feature Interpolation}
Another powerful feature of StyleGAN is \textbf{feature interpolation}~\cite{upchurch2017deep}, which allows for smooth transitions between two different styles. This is done by interpolating between two latent vectors \( w_1 \) and \( w_2 \) and generating images that smoothly blend the characteristics of both.

The interpolation can be performed linearly between the two latent vectors:
\[
w_{interp} = \alpha w_1 + (1 - \alpha) w_2
\]
Where \( \alpha \in [0, 1] \) controls the blending ratio between the two styles.

\begin{lstlisting}[style=python]
# Example of feature interpolation in PyTorch
def interpolate_styles(generator, w1, w2, alpha):
    w_interp = alpha * w1 + (1 - alpha) * w2
    return generator(w_interp)
\end{lstlisting}

This allows for continuous transformations between different styles, providing rich possibilities for generating new images by blending features such as age, gender, or lighting conditions~\cite{karras2020analyzing}.

\subsection{Applications of StyleGAN in Image Editing}

StyleGAN has found significant applications in the field of \textbf{image editing}, where its ability to control specific attributes of an image makes it an incredibly powerful tool~\cite{bermano2022state}. Some of the key applications include face editing, attribute manipulation, and generating new artistic styles.

\subsubsection{Face Editing}
One of the most popular applications of StyleGAN is in generating and editing realistic human faces. By manipulating the latent vector \( w \), users can control attributes such as age, gender, facial expressions, hairstyle, and more~\cite{alaluf2022hyperstyle}.

For example, to change the age of a face, we can modify the latent vector in the direction corresponding to "age." This allows for intuitive editing of facial features.

\subsubsection{Attribute Manipulation}
In addition to face editing, StyleGAN can also be used to manipulate other attributes in generated images~\cite{patashnik2021styleclip}. For instance, StyleGAN can be used to adjust lighting conditions, change the background of a scene, or even mix different artistic styles (e.g., turning a realistic photo into a painting)~\cite{liu2023gan}.

\begin{lstlisting}[style=python]
# Example of attribute manipulation in PyTorch
def manipulate_attribute(generator, w, attribute_vector, intensity):
    """
    Manipulate a specific attribute by moving the latent vector w in the
    direction of the attribute_vector.
    """
    modified_w = w + intensity * attribute_vector
    return generator(modified_w)
\end{lstlisting}

In this example, the latent vector \( w \) is adjusted by adding the attribute vector (e.g., "smile" or "age") scaled by the desired intensity. This results in a modified image with the corresponding attribute altered.

\subsubsection{Artistic Style Transfer}
StyleGAN can also be employed in \textbf{artistic style transfer}~\cite{kotovenko2019content}, where features from one image (such as texture or color) are transferred onto another image. This can be done by using the style mixing technique, combining the structural features from one image with the artistic features from another.

\begin{center}
\begin{tikzpicture}[level distance=2cm, sibling distance=4cm, edge from parent/.style={draw,-latex}]
  \node[rectangle, draw] {StyleGAN Applications}
    child {node[rectangle, draw] {Face Editing}}
    child {node[rectangle, draw] {Attribute Manipulation}}
    child {node[rectangle, draw] {Artistic Style Transfer}};
\end{tikzpicture}
\end{center}

\subsubsection{Example: Editing Hair Style}
Suppose we want to change the hairstyle of a generated face. By manipulating the latent vector in the direction corresponding to "hair style," we can generate new images with varying hairstyles while preserving other facial features.

Here's how we can edit the hairstyle of a generated face using StyleGAN:

\begin{lstlisting}[style=python]
# Load pre-trained generator and latent vectors
generator = load_pretrained_stylegan()
latent_vector = sample_latent_vector()

# Hair style attribute direction
hair_style_vector = get_hair_style_direction()

# Modify latent vector to change hair style
modified_latent_vector = latent_vector + 0.5 * hair_style_vector
generated_image = generator(modified_latent_vector)
\end{lstlisting}

In this example, we add the hair style vector to the original latent vector, resulting in a generated face with a different hairstyle.

\section{Conclusion}

StyleGAN and StyleGAN2 represent a major leap forward in controllable image generation. Through techniques such as style-based generation, multi-resolution synthesis, style mixing, and feature interpolation, StyleGAN allows for fine-grained control over the characteristics of generated images~\cite{karras2020analyzing}. These capabilities have found broad applications in areas such as face editing, attribute manipulation, and artistic style transfer, making StyleGAN one of the most powerful and flexible GAN architectures available today.

\chapter{Task-Specific Variants of GANs}
GANs have been adapted to solve a wide range of specific tasks, particularly in image translation and synthesis. One of the most exciting applications of GANs is their ability to transform images from one domain to another~\cite{sharma2024generative}. This process is known as image-to-image translation~\cite{isola2017image}, and it has led to the development of several GAN variants, including Pix2Pix~\cite{qu2019enhanced} and CycleGAN~\cite{chu2017cyclegan}. In this chapter, we will explore these two GAN architectures, focusing on how they handle supervised and unsupervised image translation tasks, respectively.

\section{Image Translation and Synthesis}
Image translation is the process of converting an image from one domain (e.g., grayscale images) to another domain (e.g., color images)~\cite{pang2021image}. GANs are highly effective in this area due to their ability to model complex image distributions and generate realistic outputs. Two popular GAN architectures used for image translation are Pix2Pix and CycleGAN.

\subsection{Pix2Pix: Supervised Image Translation}
Pix2Pix is a GAN variant designed for supervised image-to-image translation~\cite{mustafa2020transformation}. In supervised learning, the model is trained on pairs of images where each input image from one domain (e.g., a sketch) has a corresponding target image in the other domain (e.g., a photorealistic version of the sketch)~\cite{qu2019enhanced}. Pix2Pix uses this paired data to learn a mapping from the input domain to the output domain.

\subsubsection{Core Concept of Pix2Pix}
The main goal of Pix2Pix is to generate an image in the target domain that corresponds to a given input image in the source domain~\cite{guo2020zero}. To achieve this, Pix2Pix uses a conditional GAN (CGAN) framework, where both the generator and discriminator are conditioned on the input image. This is different from a standard GAN, where the generator produces images purely based on random noise.

The objective function of Pix2Pix consists of two parts:
\begin{itemize}
    \item \textbf{Adversarial Loss:} Encourages the generator to produce images that are indistinguishable from real images in the target domain.
    \item \textbf{L1 Loss:} Ensures that the generated image is close to the ground truth image in terms of pixel-wise similarity.
\end{itemize}

The total objective function is:

\[
\mathcal{L}_{\text{Pix2Pix}} = \mathcal{L}_{\text{GAN}} + \lambda \mathcal{L}_{L1}
\]

Where:
\begin{itemize}
    \item \( \mathcal{L}_{\text{GAN}} \) is the adversarial loss.
    \item \( \mathcal{L}_{L1} \) is the pixel-wise L1 loss.
    \item \( \lambda \) is a hyperparameter that balances the two losses.
\end{itemize}

\subsubsection{Pix2Pix Example: Image Translation from Edges to Photos}
A common use case for Pix2Pix~\cite{isola2017image} is translating edge maps (outlines of objects) into photorealistic images. For instance, given an edge map of a building, the generator learns to produce a detailed image of the building.

\begin{figure}[htbp]
    \centering
    \includegraphics[width=\textwidth]{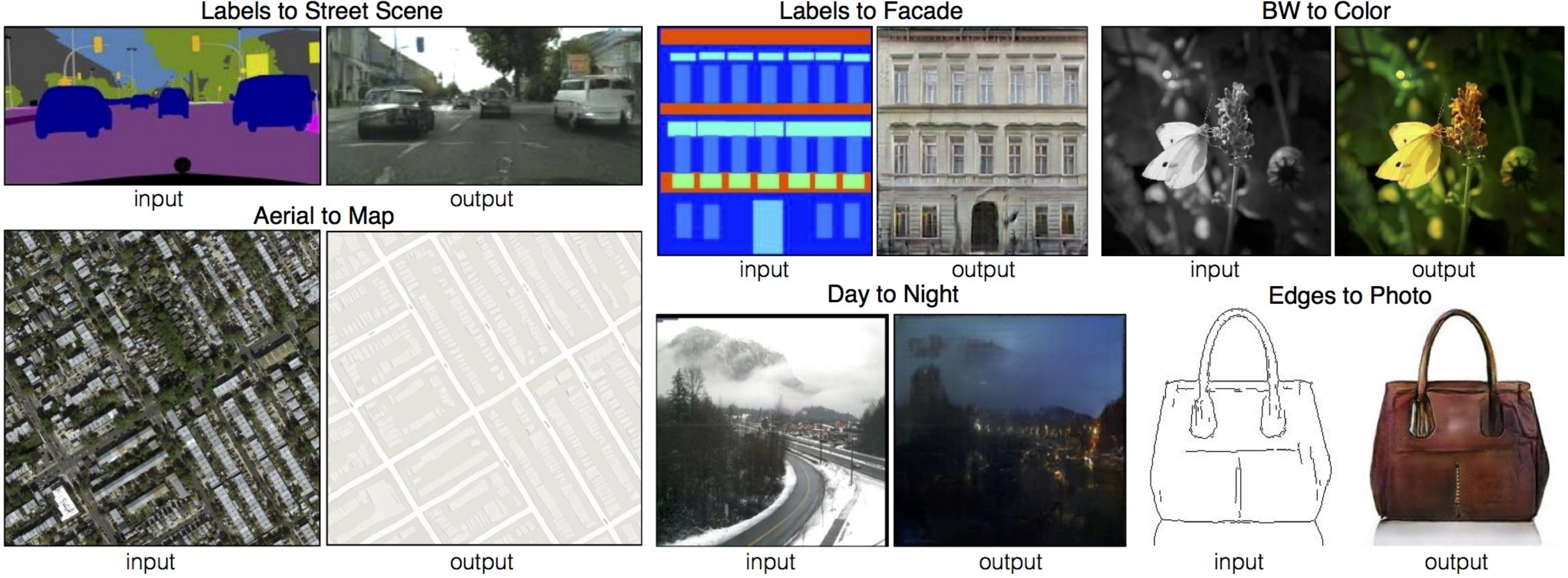}
    \caption{Example images from Pix2Pix official website~\cite{isola2017image}.}
\end{figure}

\subsubsection{Pix2Pix Architecture}
Pix2Pix uses a U-Net~\cite{ronneberger2015u} architecture for the generator and a PatchGAN for the discriminator. The U-Net architecture is particularly well-suited for image translation tasks because it uses skip connections~\cite{peng2023u} that allow low-level information from the input image to directly influence the output image, preserving fine details~\cite{isola2017image}.

\textbf{U-Net Generator:}
\begin{itemize}
    \item The generator is an encoder-decoder architecture with skip connections.
    \item The input image is progressively downsampled to capture the high-level features~\cite{ronneberger2015u}, and then it is upsampled to generate the output image.
    \item Skip connections are used to pass information from corresponding layers in the encoder to the decoder, preserving spatial information and fine details~\cite{peng2023u}.
\end{itemize}

\textbf{PatchGAN Discriminator:}
\begin{itemize}
    \item Instead of classifying the entire image as real or fake, PatchGAN~\cite{isola2017image} classifies individual patches of the image.
    \item This encourages the discriminator to focus on local image features, improving the realism of the generated image at a finer scale.
\end{itemize}

\subsubsection{Pix2Pix Implementation in PyTorch}
Here is a simplified implementation of Pix2Pix in PyTorch, focusing on translating edge maps to photorealistic images.

\begin{lstlisting}[style=python]
import torch
import torch.nn as nn
import torch.optim as optim
import torch.nn.functional as F

# U-Net Generator
class UNetGenerator(nn.Module):
    def __init__(self, in_channels, out_channels):
        super(UNetGenerator, self).__init__()
        # Define the encoder
        self.encoder = nn.ModuleList([
            nn.Conv2d(in_channels, 64, kernel_size=4, stride=2, padding=1), 
            nn.Conv2d(64, 128, kernel_size=4, stride=2, padding=1), 
            nn.Conv2d(128, 256, kernel_size=4, stride=2, padding=1), 
            nn.Conv2d(256, 512, kernel_size=4, stride=2, padding=1)
        ])
        # Define the decoder
        self.decoder = nn.ModuleList([
            nn.ConvTranspose2d(512, 256, kernel_size=4, stride=2, padding=1),
            nn.ConvTranspose2d(256, 128, kernel_size=4, stride=2, padding=1),
            nn.ConvTranspose2d(128, 64, kernel_size=4, stride=2, padding=1),
            nn.ConvTranspose2d(64, out_channels, kernel_size=4, stride=2, padding=1)
        ])
    
    def forward(self, x):
        skip_connections = []
        for layer in self.encoder:
            x = F.leaky_relu(layer(x), 0.2)
            skip_connections.append(x)
        
        for idx, layer in enumerate(self.decoder):
            if idx != 0:
                x = torch.cat((x, skip_connections[-idx]), 1)  # Skip connections
            x = F.relu(layer(x))
        
        return torch.tanh(x)

# PatchGAN Discriminator
class PatchGANDiscriminator(nn.Module):
    def __init__(self, in_channels):
        super(PatchGANDiscriminator, self).__init__()
        self.model = nn.Sequential(
            nn.Conv2d(in_channels * 2, 64, kernel_size=4, stride=2, padding=1),
            nn.LeakyReLU(0.2),
            nn.Conv2d(64, 128, kernel_size=4, stride=2, padding=1),
            nn.LeakyReLU(0.2),
            nn.Conv2d(128, 256, kernel_size=4, stride=2, padding=1),
            nn.LeakyReLU(0.2),
            nn.Conv2d(256, 1, kernel_size=4, stride=1, padding=1)
        )
    
    def forward(self, img_A, img_B):
        x = torch.cat((img_A, img_B), 1)  # Concatenate input and target images
        return self.model(x)

# Initialize models
gen = UNetGenerator(in_channels=3, out_channels=3)
disc = PatchGANDiscriminator(in_channels=3)

# Losses and optimizers
adversarial_loss = nn.MSELoss()  # For GAN loss
l1_loss = nn.L1Loss()            # For L1 loss
optimizer_g = optim.Adam(gen.parameters(), lr=0.0002)
optimizer_d = optim.Adam(disc.parameters(), lr=0.0002)

# Training loop (simplified)
for epoch in range(epochs):
    for i, (real_A, real_B) in enumerate(dataloader):
        # Train Discriminator
        fake_B = gen(real_A)
        real_pred = disc(real_A, real_B)
        fake_pred = disc(real_A, fake_B.detach())
        
        real_loss = adversarial_loss(real_pred, torch.ones_like(real_pred))
        fake_loss = adversarial_loss(fake_pred, torch.zeros_like(fake_pred))
        d_loss = (real_loss + fake_loss) / 2
        
        optimizer_d.zero_grad()
        d_loss.backward()
        optimizer_d.step()
        
        # Train Generator
        fake_pred = disc(real_A, fake_B)
        g_adv_loss = adversarial_loss(fake_pred, torch.ones_like(fake_pred))
        g_l1_loss = l1_loss(fake_B, real_B)
        g_loss = g_adv_loss + lambda_l1 * g_l1_loss
        
        optimizer_g.zero_grad()
        g_loss.backward()
        optimizer_g.step()

        print(f"[Epoch {epoch}/{epochs}] [Batch {i}/{len(dataloader)}] [D loss: {d_loss.item()}] [G loss: {g_loss.item()}]")
\end{lstlisting}

In this implementation, the generator uses a U-Net architecture to generate an image from an input image, and the discriminator uses PatchGAN to classify whether the generated image is real or fake.

\subsection{CycleGAN: Unsupervised Image Translation}
While Pix2Pix requires paired training data, CycleGAN~\cite{chu2017cyclegan} allows for unsupervised image translation, meaning that it can translate between two domains without paired examples~\cite{zhu2017unpaired}. For instance, you could use CycleGAN to translate between photos of horses and zebras without having corresponding pairs of horse and zebra images.

\subsubsection{Core Concept of CycleGAN}
The key idea behind CycleGAN is to learn a mapping between two domains \( A \) and \( B \) without requiring paired data. To achieve this, CycleGAN introduces the concept of cycle consistency. This means that if we translate an image from domain \( A \) to domain \( B \), we should be able to translate it back to domain \( A \) and recover the original image.

CycleGAN uses two generators:
\begin{itemize}
    \item \( G: A \to B \) — Translates images from domain \( A \) to domain \( B \).
    \item \( F: B \to A \) — Translates images from domain \( B \) to domain \( A \).
\end{itemize}

And two discriminators:
\begin{itemize}
    \item \( D_B \) — Classifies whether an image in domain \( B \) is real or generated.
    \item \( D_A \) — Classifies whether an image in domain \( A \) is real or generated.
\end{itemize}

The total CycleGAN objective includes:
\begin{itemize}
    \item \textbf{Adversarial Loss:} Encourages each generator to generate images that resemble the target domain.
    \item \textbf{Cycle Consistency Loss:} Ensures that translating an image to the other domain and back results in the original image.
\end{itemize}

\subsubsection{CycleGAN Objective Function}
The full objective function is:

\[
\mathcal{L}_{\text{CycleGAN}} = \mathcal{L}_{\text{GAN}}(G, D_B, A, B) + \mathcal{L}_{\text{GAN}}(F, D_A, B, A) + \lambda \mathcal{L}_{\text{cycle}}(G, F)
\]

Where:
\begin{itemize}
    \item \( \mathcal{L}_{\text{GAN}} \) is the adversarial loss for each generator-discriminator pair.
    \item \( \mathcal{L}_{\text{cycle}} \) is the cycle consistency loss.
    \item \( \lambda \) controls the importance of the cycle consistency loss.
\end{itemize}

\subsubsection{CycleGAN Example: Horse to Zebra Translation}
A popular application of CycleGAN is translating between images of horses and zebras. Given a set of horse images and a set of zebra images, CycleGAN learns to translate horses into zebras and vice versa without needing paired examples~\cite{zhu2017unpaired} of the same horse in both domains.

\subsubsection{CycleGAN Implementation in PyTorch}
Here's a simplified implementation of CycleGAN using PyTorch:

\begin{lstlisting}[style=python]
# CycleGAN Generator
class ResidualBlock(nn.Module):
    def __init__(self, in_features):
        super(ResidualBlock, self).__init__()
        self.block = nn.Sequential(
            nn.Conv2d(in_features, in_features, kernel_size=3, stride=1, padding=1),
            nn.InstanceNorm2d(in_features),
            nn.ReLU(inplace=True),
            nn.Conv2d(in_features, in_features, kernel_size=3, stride=1, padding=1),
            nn.InstanceNorm2d(in_features)
        )
    
    def forward(self, x):
        return x + self.block(x)

class CycleGANGenerator(nn.Module):
    def __init__(self, input_channels, output_channels):
        super(CycleGANGenerator, self).__init__()
        # Define the generator architecture
        self.model = nn.Sequential(
            nn.Conv2d(input_channels, 64, kernel_size=7, stride=1, padding=3),
            nn.InstanceNorm2d(64),
            nn.ReLU(inplace=True),
            # Downsampling
            nn.Conv2d(64, 128, kernel_size=3, stride=2, padding=1),
            nn.InstanceNorm2d(128),
            nn.ReLU(inplace=True),
            nn.Conv2d(128, 256, kernel_size=3, stride=2, padding=1),
            nn.InstanceNorm2d(256),
            nn.ReLU(inplace=True),
            # Residual blocks
            *[ResidualBlock(256) for _ in range(6)],
            # Upsampling
            nn.ConvTranspose2d(256, 128, kernel_size=3, stride=2, padding=1, output_padding=1),
            nn.InstanceNorm2d(128),
            nn.ReLU(inplace=True),
            nn.ConvTranspose2d(128, 64, kernel_size=3, stride=2, padding=1, output_padding=1),
            nn.InstanceNorm2d(64),
            nn.ReLU(inplace=True),
            nn.Conv2d(64, output_channels, kernel_size=7, stride=1, padding=3),
            nn.Tanh()
        )
    
    def forward(self, x):
        return self.model(x)

# CycleGAN Training Loop (simplified)
for epoch in range(epochs):
    for i, (real_A, real_B) in enumerate(dataloader):
        # Translate between domains
        fake_B = G_A2B(real_A)
        fake_A = G_B2A(real_B)

        # Cycle consistency
        rec_A = G_B2A(fake_B)
        rec_B = G_A2B(fake_A)

        # Adversarial loss for generators
        loss_G_A2B = adversarial_loss(D_B(fake_B), torch.ones_like(fake_B))
        loss_G_B2A = adversarial_loss(D_A(fake_A), torch.ones_like(fake_A))

        # Cycle consistency loss
        cycle_loss_A = cycle_loss(rec_A, real_A)
        cycle_loss_B = cycle_loss(rec_B, real_B)
        total_cycle_loss = cycle_loss_A + cycle_loss_B

        # Total generator loss
        g_loss = loss_G_A2B + loss_G_B2A + lambda_cycle * total_cycle_loss

        optimizer_g.zero_grad()
        g_loss.backward()
        optimizer_g.step()

        # Train discriminators
        real_loss_A = adversarial_loss(D_A(real_A), torch.ones_like(real_A))
        fake_loss_A = adversarial_loss(D_A(fake_A.detach()), torch.zeros_like(fake_A))
        d_A_loss = (real_loss_A + fake_loss_A) / 2

        real_loss_B = adversarial_loss(D_B(real_B), torch.ones_like(real_B))
        fake_loss_B = adversarial_loss(D_B(fake_B.detach()), torch.zeros_like(fake_B))
        d_B_loss = (real_loss_B + fake_loss_B) / 2

        optimizer_d_A.zero_grad()
        d_A_loss.backward()
        optimizer_d_A.step()

        optimizer_d_B.zero_grad()
        d_B_loss.backward()
        optimizer_d_B.step()

        print(f"[Epoch {epoch}/{epochs}] [D A loss: {d_A_loss.item()}] [D B loss: {d_B_loss.item()}] [G loss: {g_loss.item()}]")
\end{lstlisting}
In this CycleGAN implementation, two generators ($G_A2B$ and $G_B2A$) and two discriminators ($D_A$ and $D_B$) are trained to translate between two domains without the need for paired examples.

\section{Summary}
In this chapter, we explored two powerful GAN-based architectures for image translation: Pix2Pix and CycleGAN. Pix2Pix is a supervised approach that requires paired training data, while CycleGAN handles unsupervised image translation, making it suitable for tasks where paired examples are not available. Both architectures have been widely applied in various tasks, such as translating edge maps to photorealistic images, and style transfers like horse-to-zebra transformations. Through detailed explanations and code implementations using PyTorch, we have demonstrated how these models function, offering a comprehensive guide for beginners to apply these GAN variants in their projects.

\section{Super-Resolution Generative Adversarial Networks (SRGAN)}
Super-Resolution Generative Adversarial Networks (SRGAN)~\cite{ledig2016photo} are specialized GANs designed to generate high-resolution images from low-resolution inputs. SRGANs are particularly useful for image super-resolution tasks, where the objective is to increase the resolution of an image while maintaining or enhancing the image quality. This section will explore the techniques behind SRGAN and how it achieves high-quality image super-resolution.

\begin{figure}[htbp]
    \centering
    \includegraphics[width=0.8\textwidth]{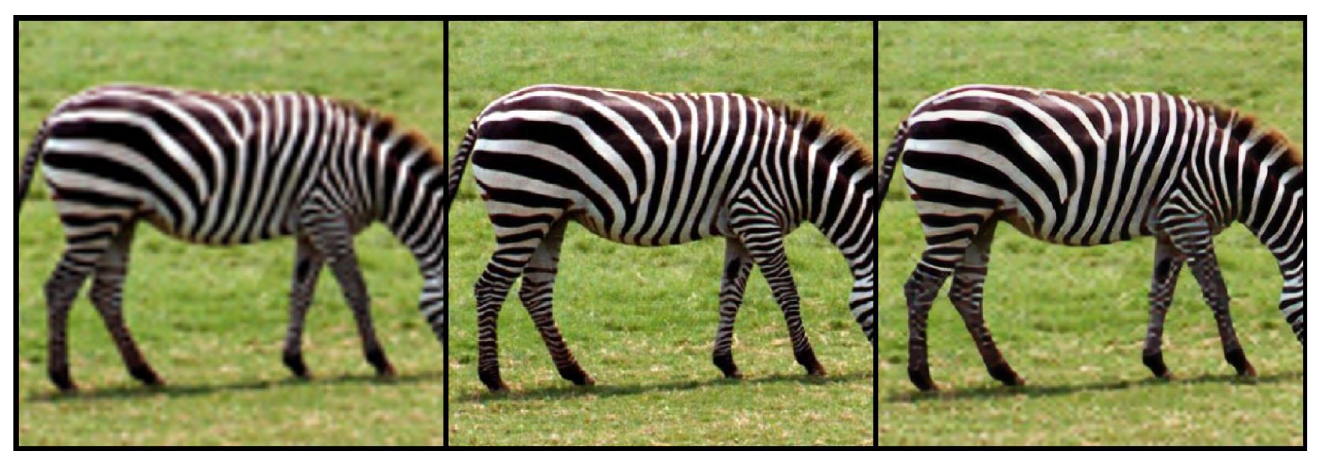}
    \caption{Example resolution from SRGAN.}
\end{figure}

\subsection{Techniques for Super-Resolution Image Generation}
Super-resolution is the process of reconstructing a high-resolution image from a low-resolution counterpart~\cite{ledig2016photo}. This is a challenging task because increasing the resolution of an image involves predicting and generating details that were not present in the original low-resolution image~\cite{xiong2020improved}. SRGAN solves this problem by using both a Generator and a Discriminator to produce high-quality images with realistic textures.

\textbf{1. Perceptual Loss Function:}  
One of the main innovations in SRGAN is the use of a perceptual loss function. Instead of simply minimizing pixel-wise differences between the generated and real images, SRGAN uses a combination of pixel-wise loss and perceptual loss, which compares the high-level features of images extracted from a pre-trained network (such as VGG~\cite{simonyan2014very}). This allows SRGAN to focus on generating images with more realistic textures rather than just matching individual pixel values.

\textbf{2. Residual Blocks in the Generator:}  
The Generator in SRGAN employs residual blocks~\cite{he2016deep}, which help in generating high-resolution details by adding shortcut connections that bypass some layers. These residual blocks improve the learning ability of the Generator by allowing information to flow directly through the network, reducing the vanishing gradient problem in deep networks~\cite{vaswani2017attention, he2016deep}.

\textbf{3. Discriminator with PatchGAN Architecture:}  
The Discriminator in SRGAN is designed to classify whether small patches of the image are real or generated. This PatchGAN~\cite{isola2017image} architecture allows the model to focus on local texture details, making it more effective at distinguishing between realistic and fake images.

\textbf{Example: SRGAN Generator in PyTorch}

\begin{lstlisting}[style=python]
import torch
import torch.nn as nn

# Residual Block used in the SRGAN Generator
class ResidualBlock(nn.Module):
    def __init__(self, channels):
        super(ResidualBlock, self).__init__()
        self.block = nn.Sequential(
            nn.Conv2d(channels, channels, kernel_size=3, stride=1, padding=1),
            nn.BatchNorm2d(channels),
            nn.PReLU(),  # Parametric ReLU activation
            nn.Conv2d(channels, channels, kernel_size=3, stride=1, padding=1),
            nn.BatchNorm2d(channels)
        )

    def forward(self, x):
        return x + self.block(x)  # Add input to output (residual connection)

# SRGAN Generator
class SRGAN_Generator(nn.Module):
    def __init__(self, num_residual_blocks=16):
        super(SRGAN_Generator, self).__init__()
        self.initial = nn.Sequential(
            nn.Conv2d(3, 64, kernel_size=9, stride=1, padding=4),
            nn.PReLU()
        )

        # Residual blocks
        self.residuals = nn.Sequential(
            *[ResidualBlock(64) for _ in range(num_residual_blocks)]
        )

        self.upsample = nn.Sequential(
            nn.Conv2d(64, 256, kernel_size=3, stride=1, padding=1),
            nn.PixelShuffle(2),  # Upscale by factor of 2
            nn.PReLU(),
            nn.Conv2d(64, 256, kernel_size=3, stride=1, padding=1),
            nn.PixelShuffle(2),  # Upscale by factor of 2 again
            nn.PReLU()
        )

        self.final = nn.Conv2d(64, 3, kernel_size=9, stride=1, padding=4)

    def forward(self, x):
        initial = self.initial(x)
        res = self.residuals(initial)
        upsampled = self.upsample(res)
        return self.final(upsampled)

# Example usage:
low_res_image = torch.randn(1, 3, 64, 64)  # Example low-resolution image
srgan_generator = SRGAN_Generator()
high_res_image = srgan_generator(low_res_image)
print(high_res_image.shape)  # Output should be high-resolution, e.g., torch.Size([1, 3, 256, 256])
\end{lstlisting}

In this example, the SRGAN Generator is implemented using residual blocks and PixelShuffle for upscaling. The network takes a low-resolution image as input and generates a higher-resolution version of the same image.

\subsection{Training SRGAN with Perceptual Loss}
The training process of SRGAN is based on the combination of two loss functions~\cite{wang2018esrgan}:

\begin{itemize}
    \item \textbf{Pixel-wise loss}: Measures the difference between the generated high-resolution image and the ground truth using pixel values (e.g., Mean Squared Error).
    \item \textbf{Perceptual loss}: Compares high-level features of the generated and ground truth images extracted from a pre-trained network (such as VGG), encouraging the Generator to produce perceptually realistic images.
\end{itemize}

The Discriminator is trained to classify whether an image is real or generated, while the Generator aims to fool the Discriminator by producing realistic high-resolution images.

\section{3D Generative Adversarial Networks (3DGAN)}
3DGANs~\cite{wu2016learning} are a class of GANs designed to generate three-dimensional objects from 2D images or noise. Unlike traditional GANs that generate 2D images, 3DGANs focus on generating 3D models~\cite{smith2017improved}, which can be represented as voxel grids, point clouds, or meshes. This section explores the techniques used to generate 3D objects and the transition from 2D to 3D in GAN architectures~\cite{chan2022efficient}.

\subsection{Generating 3D Models from 2D Images}
The goal of 3DGAN is to generate realistic 3D models based on 2D input images. For instance, given a 2D image of a car, the model should be able to generate a full 3D representation of the car~\cite{wu2016learning}. This is particularly useful in applications such as computer graphics, augmented reality, and 3D printing.

\begin{figure}[htbp]
    \centering
    \includegraphics[width=\textwidth]{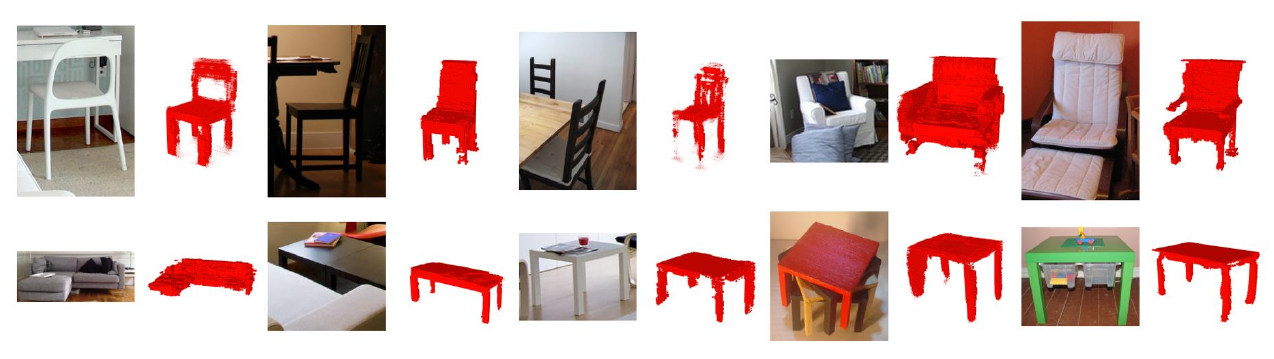}
    \caption{Qualitative results of single image 3D reconstruction from 3DGAN paper~\cite{wu2016learning}.}
\end{figure}

\textbf{1. Voxel Representation:}  
In 3DGAN, one common way to represent 3D objects is by using voxel grids. A voxel is a 3D equivalent of a pixel in 2D images. A voxel grid is a 3D array where each voxel can be filled (indicating the presence of an object) or empty (indicating empty space). The Generator in 3DGAN produces a voxel grid that represents the 3D structure of the object~\cite{liu2020neural}.

\textbf{2. 3D Convolutional Networks:}  
To generate 3D objects, the Generator in 3DGAN uses 3D convolutional layers instead of 2D convolutions. 3D convolutions allow the model to capture spatial dependencies in all three dimensions (height, width, and depth), making it possible to generate consistent 3D structures~\cite{wu2016learning}.

\textbf{Example: 3DGAN Generator Using Voxels in PyTorch}

\begin{lstlisting}[style=python]
import torch
import torch.nn as nn

# 3DGAN Generator
class GAN3D_Generator(nn.Module):
    def __init__(self):
        super(GAN3D_Generator, self).__init__()
        self.model = nn.Sequential(
            nn.ConvTranspose3d(512, 256, kernel_size=4, stride=1, padding=0),
            nn.BatchNorm3d(256),
            nn.ReLU(True),
            nn.ConvTranspose3d(256, 128, kernel_size=4, stride=2, padding=1),
            nn.BatchNorm3d(128),
            nn.ReLU(True),
            nn.ConvTranspose3d(128, 64, kernel_size=4, stride=2, padding=1),
            nn.BatchNorm3d(64),
            nn.ReLU(True),
            nn.ConvTranspose3d(64, 1, kernel_size=4, stride=2, padding=1),
            nn.Sigmoid()  # Output a voxel grid
        )

    def forward(self, x):
        return self.model(x)

# Example usage:
noise = torch.randn(1, 512, 1, 1, 1)  # Random noise vector
generator_3d = GAN3D_Generator()
voxel_grid = generator_3d(noise)
print(voxel_grid.shape)  # Output should be a voxel grid, e.g., torch.Size([1, 1, 32, 32, 32])
\end{lstlisting}

In this example, the 3DGAN Generator uses 3D transposed convolutions to generate a voxel grid representing a 3D object. The noise input is transformed into a structured 3D shape by upsampling through multiple layers.

\subsubsection{Techniques for Generating 3D Objects}
Generating 3D objects involves several challenges that differ from 2D image generation:

\begin{itemize}
    \item \textbf{3D Convolutional Networks:} 3D convolutions allow the model to learn spatial features in three dimensions, making it possible to generate consistent 3D structures from noise or 2D images~\cite{liu2020neural}.
    \item \textbf{Conditional GAN for 3D Reconstruction:} Conditional GANs~\cite{wang2018cgan} can be used to generate 3D objects based on input 2D images. By conditioning on 2D views of an object, the model can predict the full 3D structure.
    \item \textbf{Loss Functions for 3D Shape:} Instead of pixel-wise losses, 3DGANs often use specialized loss functions that take into account the structure of the 3D object, such as intersection-over-union (IoU)~\cite{nowozin2014optimal} or volumetric loss~\cite{xie2018tempogan}.
\end{itemize}

\textbf{Visualizing the Process of 3DGAN:}

\begin{center}
\begin{tikzpicture}
  [scale=1, every node/.style={scale=1}, 
  block/.style={rectangle, draw, fill=blue!20, text centered, minimum height=3em},
  arrow/.style={->, thick}]

  \node[block] (noise) {Noise $z$};
  \node[block, right=of noise] (gen) {3DGAN Generator};
  \node[block, right=of gen] (voxels) {Voxel Grid};
  \node[block, below=of voxels] (disc) {3DGAN Discriminator};
  \node[block, below=of disc] (real_voxel) {Real 3D Model};

  \draw[arrow] (noise) -- (gen);
  \draw[arrow] (gen) -- (voxels);
  \draw[arrow] (real_voxel) -- (disc);
  \draw[arrow] (voxels) -- (disc);
\end{tikzpicture}
\end{center}

In this diagram, the noise vector is transformed into a voxel grid by the 3DGAN Generator, which is then evaluated by the Discriminator to classify whether it is real or generated.

In summary, both SRGAN and 3DGAN tackle complex image and object generation tasks, with SRGAN focusing on high-resolution 2D images and 3DGAN generating 3D models from either noise or 2D images~\cite{wu2016learning}. Each of these models uses specialized techniques to handle the challenges of generating high-quality and complex outputs in their respective domains.

\section{Text-to-Image Generation with GANs}
Text-to-image generation~\cite{reed2016generative} is an exciting application of GANs, where the goal is to generate images that match a given text description. This task is more challenging than standard image generation, as the model must not only generate high-quality images but also ensure that the images align with the semantic meaning of the input text. In this section, we will explore two popular models for text-to-image generation: StackGAN~\cite{zhang2017stackgan} and AttnGAN~\cite{xu2017attngan}.

\subsection{StackGAN: Staged Image Generation}

\textbf{StackGAN} (Stacked Generative Adversarial Networks) is a two-stage architecture designed to generate high-resolution images from text descriptions. The idea behind StackGAN is to divide the image generation process into two stages~\cite{zhang2017stackgan}: a rough low-resolution image is generated in the first stage, and the second stage refines this image to add finer details. This staged approach helps in generating more realistic images that are well-aligned with the input text.

\subsubsection{Stage-I: Coarse Image Generation}

In the first stage, the model takes a text description and a noise vector as input and generates a low-resolution image, typically \(64 \times 64\). This image captures the basic structure of the object described by the text but may lack finer details~\cite{lu2024coarse}.

\[
\text{Stage-I Generator: } G_1(z, t) \rightarrow I_1
\]
Where:
\begin{itemize}
    \item \( z \) is a noise vector.
    \item \( t \) is the text embedding of the input description.
    \item \( I_1 \) is the generated low-resolution image.
\end{itemize}

The generator learns to produce an image that matches the basic structure and layout of the text, while the discriminator evaluates whether the generated image matches the real data distribution for the given description.

\subsubsection{Stage-II: Fine Image Refinement}

The second stage of StackGAN takes the low-resolution image generated by the first stage and refines it to produce a high-resolution image (e.g., \(256 \times 256\)). The text description is used again in this stage to ensure that the refined image remains consistent with the input description. The generator focuses on adding finer details such as texture, color, and small features~\cite{zhang2017stackgan}.

\[
\text{Stage-II Generator: } G_2(I_1, t) \rightarrow I_2
\]
Where:
\begin{itemize}
    \item \( I_1 \) is the low-resolution image from Stage-I.
    \item \( t \) is the text embedding.
    \item \( I_2 \) is the final high-resolution image.
\end{itemize}

\subsubsection{StackGAN Example in PyTorch}

Below is an implementation of the two-stage StackGAN in PyTorch:

\begin{lstlisting}[style=python]
import torch
import torch.nn as nn
import torch.optim as optim

# Stage-I Generator
class StageIGenerator(nn.Module):
    def __init__(self):
        super(StageIGenerator, self).__init__()
        self.fc = nn.Sequential(
            nn.Linear(100 + 1024, 128 * 16 * 16),
            nn.ReLU()
        )
        self.upsample = nn.Sequential(
            nn.ConvTranspose2d(128, 64, 4, 2, 1),
            nn.BatchNorm2d(64),
            nn.ReLU(),
            nn.ConvTranspose2d(64, 3, 4, 2, 1),
            nn.Tanh()
        )
    
    def forward(self, noise, text_embedding):
        x = torch.cat((noise, text_embedding), dim=1)
        x = self.fc(x)
        x = x.view(-1, 128, 16, 16)
        return self.upsample(x)

# Stage-II Generator
class StageIIGenerator(nn.Module):
    def __init__(self):
        super(StageIIGenerator, self).__init__()
        self.fc = nn.Sequential(
            nn.Conv2d(3 + 1024, 128, 3, 1, 1),
            nn.BatchNorm2d(128),
            nn.ReLU(),
            nn.Conv2d(128, 64, 3, 1, 1),
            nn.BatchNorm2d(64),
            nn.ReLU(),
            nn.Conv2d(64, 3, 3, 1, 1),
            nn.Tanh()
        )
    
    def forward(self, low_res_image, text_embedding):
        text_embedding = text_embedding.view(-1, 1024, 1, 1)
        text_embedding = text_embedding.repeat(1, 1, low_res_image.size(2), low_res_image.size(3))
        x = torch.cat((low_res_image, text_embedding), dim=1)
        return self.fc(x)

# Text embedding, noise, and training example
noise = torch.randn(batch_size, 100)
text_embedding = torch.randn(batch_size, 1024)

# Stage-I and Stage-II generators
G1 = StageIGenerator()
G2 = StageIIGenerator()

# Generate low-resolution and high-resolution images
low_res_image = G1(noise, text_embedding)
high_res_image = G2(low_res_image, text_embedding)
\end{lstlisting}

In this code, Stage-I generates a \(64 \times 64\) image, and Stage-II refines it into a high-resolution image of \(256 \times 256\), both based on the input text embedding.

\subsection{AttnGAN: Introducing Attention Mechanism in Image Generation}

\textbf{AttnGAN} (Attention Generative Adversarial Networks) further improves text-to-image generation by introducing an \textbf{attention mechanism}~\cite{vaswani2017attention} that allows the model to focus on specific parts of the text when generating different regions of the image. This makes AttnGAN particularly effective at generating complex images where different parts of the text description correspond to different regions of the image~\cite{xu2017attngan}.

\subsubsection{Attention Mechanism}

The key idea in AttnGAN is to use an attention mechanism that computes an alignment between the words in the text description and the sub-regions of the generated image. This attention mechanism ensures that the generated image accurately reflects all aspects of the input description by selectively focusing on different parts of the text at different stages of the image generation process~\cite{xu2017attngan}.

The attention mechanism is defined as:
\[
\alpha_{i,j} = \frac{\exp(s(h_i, e_j))}{\sum_{k} \exp(s(h_i, e_k))}
\]
Where:
\begin{itemize}
    \item \( h_i \) represents the feature of the image at location \(i\).
    \item \( e_j \) represents the word embedding of the \(j\)-th word in the text description.
    \item \( s(h_i, e_j) \) is a similarity function (often cosine similarity) that measures how relevant the word \(e_j\) is to the image feature at location \(h_i\).
\end{itemize}

This attention mechanism ensures that important words in the description receive more focus during the image generation process.

\subsubsection{AttnGAN Example in PyTorch}

Here's a simplified version of how the attention mechanism is incorporated into AttnGAN using PyTorch:

\begin{lstlisting}[style=python]
class AttentionLayer(nn.Module):
    def __init__(self):
        super(AttentionLayer, self).__init__()
        self.fc_img = nn.Linear(128, 128)  # Image features
        self.fc_txt = nn.Linear(256, 128)  # Text embeddings

    def forward(self, img_features, text_embeddings):
        img_features_proj = self.fc_img(img_features)
        text_embeddings_proj = self.fc_txt(text_embeddings)

        # Compute attention scores
        attention = torch.bmm(img_features_proj, text_embeddings_proj.permute(0, 2, 1))
        attention = torch.softmax(attention, dim=-1)

        # Weighted sum of text embeddings based on attention
        attended_text = torch.bmm(attention, text_embeddings_proj)
        return attended_text
\end{lstlisting}

In this code, we project the image features and text embeddings into the same dimensional space, compute attention scores using matrix multiplication, and apply softmax to obtain attention weights. These weights are used to produce a weighted sum of the text embeddings, focusing on the most relevant parts of the text.

\subsection{Applications of Text-to-Image GANs}

Text-to-image GANs like StackGAN and AttnGAN have several practical applications~\cite{adadi2021survey}:
\begin{itemize}
    \item \textbf{Art and Design}: Artists and designers can use text-to-image GANs to quickly generate concept art or prototypes based on textual descriptions.
    \item \textbf{Content Creation}: These models can be used to automatically generate images for books, advertisements, and websites based on text input.
    \item \textbf{Data Augmentation}: Text-to-image models can be used to generate synthetic data for training other machine learning models, especially when labeled image data is scarce.
\end{itemize}

\section{Temporal Generative Adversarial Networks}

Temporal data generation is a challenging task that involves generating sequences of data that evolve over time~\cite{saito2017temporal}. Examples include video generation, motion synthesis, and time-series forecasting. In this section, we will discuss two key models for temporal data generation: TGAN~\cite{saito2017temporal} and MoCoGAN~\cite{tulyakov2017mocogan}.

\subsection{TGAN: Temporal Data Generation}

\textbf{TGAN} (Temporal Generative Adversarial Network) is designed for generating sequences of data, such as time-series or video frames. The goal is to capture both the temporal dependencies between frames and the spatial structure of each frame~\cite{saito2017temporal}.

\subsubsection{TGAN Architecture}

TGAN extends traditional GANs by introducing a recurrent component to model the temporal dependencies. The generator and discriminator both incorporate LSTM (Long Short-Term Memory) or GRU (Gated Recurrent Unit) layers to process the sequence of frames.

\begin{lstlisting}[style=python]
# TGAN Generator with LSTM for temporal dependencies
class TGANGenerator(nn.Module):
    def __init__(self):
        super(TGANGenerator, self).__init__()
        self.lstm = nn.LSTM(input_size=100, hidden_size=256, batch_first=True)
        self.fc = nn.Sequential(
            nn.Linear(256, 128),
            nn.ReLU(),
            nn.Linear(128, 64 * 64 * 3),
            nn.Tanh()
        )

    def forward(self, noise):
        # Generate temporal sequence
        lstm_out, _ = self.lstm(noise)
        lstm_out = lstm_out.contiguous().view(-1, 256)
        images = self.fc(lstm_out)
        return images.view(-1, 3, 64, 64)
\end{lstlisting}

In this example, noise is passed through an LSTM layer to model temporal relationships, and then fully connected layers generate the individual frames of the sequence.

\subsection{MoCoGAN: Motion and Content Disentanglement}

\textbf{MoCoGAN} (Motion and Content Generative Adversarial Network) is a GAN-based model for video generation that disentangles motion from content~\cite{tulyakov2017mocogan}. In video generation, the challenge is to separate the static content of the scene (e.g., the background or object identity) from the dynamic aspects (e.g., motion or camera movement)~\cite{li2024survey}.

\subsubsection{Motion and Content Disentanglement}

MoCoGAN separates the latent space into two parts:
\begin{itemize}
    \item \textbf{Content Latent Code \( z_c \)}: Encodes the static content of the video, such as the identity of an object or the background.
    \item \textbf{Motion Latent Code \( z_m \)}: Encodes the temporal dynamics, such as motion or changes between frames.
\end{itemize}

The generator uses both the content code and motion code to generate a sequence of frames~\cite{tulyakov2017mocogan}. The motion code changes over time, but the content code remains fixed for the entire sequence~\cite{li2024survey}.

\begin{lstlisting}[style=python]
# MoCoGAN generator
class MoCoGANGenerator(nn.Module):
    def __init__(self):
        super(MoCoGANGenerator, self).__init__()
        self.fc_content = nn.Linear(100, 128)  # Content code
        self.lstm_motion = nn.LSTM(input_size=50, hidden_size=128, batch_first=True)  # Motion code
        self.fc_frame = nn.Sequential(
            nn.Linear(256, 128),
            nn.ReLU(),
            nn.Linear(128, 64 * 64 * 3),
            nn.Tanh()
        )

    def forward(self, content_code, motion_code):
        content_features = self.fc_content(content_code)
        motion_features, _ = self.lstm_motion(motion_code)
        combined_features = torch.cat([content_features, motion_features], dim=2)
        frames = self.fc_frame(combined_features)
        return frames.view(-1, 3, 64, 64)
\end{lstlisting}

In MoCoGAN, the content code remains fixed for the entire sequence, while the motion code evolves over time, allowing the generator to create coherent videos that maintain content consistency while introducing motion dynamics.

\subsubsection{Applications of MoCoGAN}

MoCoGAN has applications in video generation and animation, where it is important to maintain the identity of objects or characters while allowing for natural motion. Some use cases include:
\begin{itemize}
    \item \textbf{Video Synthesis}: Generating realistic video sequences based on content and motion descriptions.
    \item \textbf{Animation}: Creating animated characters that retain their identity while performing different actions.
\end{itemize}

\section{Conclusion}

Text-to-image and temporal GANs open up new possibilities in areas such as image synthesis, video generation, and time-series modeling. Models like StackGAN~\cite{zhang2017stackgan} and AttnGAN~\cite{xu2017attngan} leverage techniques such as staged generation and attention mechanisms to improve text-to-image alignment, while temporal GANs like TGAN~\cite{xie2018tempogan} and MoCoGAN~\cite{tulyakov2017mocogan} focus on generating realistic sequences by disentangling motion and content. These advanced models demonstrate the versatility and potential of GANs in a wide range of applications~\cite{li2024survey}.

\chapter{Other Variants of Generative Adversarial Networks}
Generative Adversarial Networks (GANs) have inspired many variations, each designed to address specific challenges or extend the capabilities of the original GAN framework. In this chapter, we will explore several advanced GAN variants: Energy-Based GANs (EBGANs)~\cite{zhao2016energy}, Adversarial Autoencoders (AAEs)~\cite{makhzani2015adversarial}, Bidirectional GANs (BiGANs), and Autoencoder GANs (AEGANs)~\cite{donahue2016adversarial}. These models offer unique approaches to improving the stability, interpretability, and functionality of GANs. We will provide detailed explanations of each variant, along with examples and practical use cases.

\section{Energy-Based Generative Adversarial Networks (EBGAN)}
EBGAN is a variant of GAN that takes an energy-based approach to the discriminator. Instead of having the discriminator output a probability, EBGAN models the discriminator as an energy function, which assigns a scalar value (energy) to the input~\cite{zhao2016energy}. The generator is trained to produce samples that have low energy, while the discriminator is trained to assign higher energy to fake samples and lower energy to real samples~\cite{zhang2019self}.

\subsection{Core Concept of EBGAN}
In traditional GANs, the discriminator outputs the probability of whether the input is real or generated. EBGAN changes this by treating the discriminator as an energy function. The energy function is minimized for real samples and maximized for fake samples. The generator's goal is to create samples that the discriminator assigns low energy to, thereby making them indistinguishable from real samples.

The key difference between EBGAN and traditional GANs is how the discriminator works. In EBGAN, the discriminator is treated as an autoencoder that reconstructs the input image. The energy of a sample is defined as the reconstruction error, which is minimized for real samples and maximized for fake ones~\cite{zhao2016energy}.

\subsection{EBGAN Objective Function}
The EBGAN loss can be written as:

\[
\mathcal{L}_{\text{EBGAN}} = \mathbb{E}_{x \sim p_{\text{data}}} [E(x)] - \mathbb{E}_{z \sim p_z(z)} [E(G(z))]
\]

Where:
\begin{itemize}
    \item \( E(x) \) is the energy assigned to real samples by the discriminator (autoencoder reconstruction loss).
    \item \( G(z) \) is the generator, which tries to produce low-energy samples.
\end{itemize}

The discriminator is trained to increase the energy (reconstruction error) for fake samples while decreasing it for real samples.

\subsection{EBGAN Architecture}
In EBGAN, the discriminator is implemented as an autoencoder. The generator produces samples, which are passed through the autoencoder (discriminator). The autoencoder tries to reconstruct the input, and the energy is defined as the reconstruction loss.

\textbf{Autoencoder Discriminator:}
\begin{itemize}
    \item The input image is encoded into a low-dimensional representation.
    \item The encoded representation is then decoded back into the original image.
    \item The reconstruction error serves as the energy of the input.
\end{itemize}

\subsection{EBGAN Implementation in PyTorch}
Here's a basic implementation of EBGAN using PyTorch:

\begin{lstlisting}[style=python]
import torch
import torch.nn as nn
import torch.optim as optim

# Autoencoder-based Discriminator (Energy Function)
class AutoencoderDiscriminator(nn.Module):
    def __init__(self):
        super(AutoencoderDiscriminator, self).__init__()
        self.encoder = nn.Sequential(
            nn.Conv2d(3, 64, kernel_size=4, stride=2, padding=1),
            nn.ReLU(),
            nn.Conv2d(64, 128, kernel_size=4, stride=2, padding=1),
            nn.ReLU(),
            nn.Flatten(),
            nn.Linear(128 * 8 * 8, 1024),
            nn.ReLU()
        )
        self.decoder = nn.Sequential(
            nn.Linear(1024, 128 * 8 * 8),
            nn.ReLU(),
            nn.Unflatten(1, (128, 8, 8)),
            nn.ConvTranspose2d(128, 64, kernel_size=4, stride=2, padding=1),
            nn.ReLU(),
            nn.ConvTranspose2d(64, 3, kernel_size=4, stride=2, padding=1),
            nn.Sigmoid()
        )
    
    def forward(self, x):
        encoded = self.encoder(x)
        reconstructed = self.decoder(encoded)
        return reconstructed

# Generator
class Generator(nn.Module):
    def __init__(self, latent_dim):
        super(Generator, self).__init__()
        self.model = nn.Sequential(
            nn.Linear(latent_dim, 128 * 8 * 8),
            nn.ReLU(),
            nn.Unflatten(1, (128, 8, 8)),
            nn.ConvTranspose2d(128, 64, kernel_size=4, stride=2, padding=1),
            nn.ReLU(),
            nn.ConvTranspose2d(64, 3, kernel_size=4, stride=2, padding=1),
            nn.Tanh()
        )
    
    def forward(self, z):
        return self.model(z)

# Loss function (Reconstruction Loss for Discriminator)
def reconstruction_loss(real, reconstructed):
    return nn.functional.mse_loss(reconstructed, real)

# Training loop
latent_dim = 100
generator = Generator(latent_dim)
discriminator = AutoencoderDiscriminator()

optimizer_g = optim.Adam(generator.parameters(), lr=0.0002)
optimizer_d = optim.Adam(discriminator.parameters(), lr=0.0002)

for epoch in range(epochs):
    for i, (real_imgs, _) in enumerate(dataloader):
        z = torch.randn(real_imgs.size(0), latent_dim)
        fake_imgs = generator(z)

        # Train Discriminator
        real_reconstructed = discriminator(real_imgs)
        fake_reconstructed = discriminator(fake_imgs.detach())
        real_energy = reconstruction_loss(real_imgs, real_reconstructed)
        fake_energy = reconstruction_loss(fake_imgs.detach(), fake_reconstructed)
        d_loss = real_energy - fake_energy

        optimizer_d.zero_grad()
        d_loss.backward()
        optimizer_d.step()

        # Train Generator
        fake_reconstructed = discriminator(fake_imgs)
        g_loss = reconstruction_loss(fake_imgs, fake_reconstructed)

        optimizer_g.zero_grad()
        g_loss.backward()
        optimizer_g.step()

    print(f"[Epoch {epoch}/{epochs}] [D loss: {d_loss.item()}] [G loss: {g_loss.item()}]")
\end{lstlisting}

In this implementation, the discriminator is an autoencoder, and the energy is the reconstruction loss. The generator tries to produce images that minimize the reconstruction error, making them indistinguishable from real images.

\section{Adversarial Autoencoders (AAE)}
Adversarial Autoencoders (AAEs) combine autoencoders with GANs to impose a specific prior on the latent space. This makes it possible to generate samples from a structured latent space, similar to Variational Autoencoders (VAEs)~\cite{doersch2016tutorial}, but using adversarial training instead of maximum likelihood~\cite{makhzani2015adversarial}.

\subsection{Core Concept of AAE}
In an Adversarial Autoencoder, the encoder maps the input data into a latent space, and the decoder reconstructs the input from the latent representation. The key difference from a traditional autoencoder is that the latent space is regularized using a GAN. The discriminator ensures that the encoded latent vectors follow a desired distribution (e.g., a Gaussian or uniform distribution)~\cite{li2024survey}.

The adversarial training forces the encoder to map the input data to a latent space that matches the prior distribution, while the decoder reconstructs the data from the latent space.

\subsubsection{AAE Objective Function}
The AAE objective function consists of two parts:
\begin{itemize}
    \item \textbf{Reconstruction Loss:} Encourages the decoder to accurately reconstruct the input from the latent code.
    \item \textbf{Adversarial Loss:} Forces the latent space to match a predefined prior distribution.
\end{itemize}

The total loss is:

\[
\mathcal{L}_{\text{AAE}} = \mathcal{L}_{\text{reconstruction}} + \mathcal{L}_{\text{adversarial}}
\]

Where:
\begin{itemize}
    \item \( \mathcal{L}_{\text{reconstruction}} \) is the pixel-wise reconstruction loss.
    \item \( \mathcal{L}_{\text{adversarial}} \) is the adversarial loss on the latent space.
\end{itemize}

\subsection{AAE Architecture}
The architecture of AAE is similar to a traditional autoencoder, with the addition of a discriminator to enforce the latent space distribution~\cite{donahue2016adversarial}.

\textbf{Encoder:}
\begin{itemize}
    \item Maps the input image to a latent vector.
\end{itemize}

\textbf{Decoder:}
\begin{itemize}
    \item Reconstructs the image from the latent vector.
\end{itemize}

\textbf{Discriminator:}
\begin{itemize}
    \item Tries to distinguish between the latent vectors generated by the encoder and samples from the prior distribution.
\end{itemize}

\subsection{AAE Implementation in PyTorch}
Here's a simplified implementation of Adversarial Autoencoders using PyTorch:

\begin{lstlisting}[style=python]
# Encoder
class AAEEncoder(nn.Module):
    def __init__(self, latent_dim):
        super(AAEEncoder, self).__init__()
        self.model = nn.Sequential(
            nn.Linear(28*28, 512),
            nn.ReLU(),
            nn.Linear(512, latent_dim)
        )
    
    def forward(self, x):
        return self.model(x.view(x.size(0), -1))

# Decoder
class AAEDecoder(nn.Module):
    def __init__(self, latent_dim):
        super(AAEDecoder, self).__init__()
        self.model = nn.Sequential(
            nn.Linear(latent_dim, 512),
            nn.ReLU(),
            nn.Linear(512, 28*28),
            nn.Sigmoid()
        )
    
    def forward(self, z):
        return self.model(z).view(z.size(0), 1, 28, 28)

# Discriminator for Latent Space
class AAEDiscriminator(nn.Module):
    def __init__(self, latent_dim):
        super(AAEDiscriminator, self).__init__()
        self.model = nn.Sequential(
            nn.Linear(latent_dim, 512),
            nn.ReLU(),
            nn.Linear(512, 1),
            nn.Sigmoid()
        )
    
    def forward(self, z):
        return self.model(z)

# Initialize models
latent_dim = 10
encoder = AAEEncoder(latent_dim)
decoder = AAEDecoder(latent_dim)
discriminator = AAEDiscriminator(latent_dim)

# Losses and optimizers
reconstruction_loss = nn.BCELoss()
adversarial_loss = nn.BCELoss()
optimizer_g = optim.Adam(list(encoder.parameters()) + list(decoder.parameters()), lr=0.0002)
optimizer_d = optim.Adam(discriminator.parameters(), lr=0.0002)

# Training loop (simplified)
for epoch in range(epochs):
    for i, (imgs, _) in enumerate(dataloader):
        # Encode images
        z = encoder(imgs)
        real_z = torch.randn(imgs.size(0), latent_dim)

        # Train Discriminator
        real_pred = discriminator(real_z)
        fake_pred = discriminator(z.detach())
        d_loss_real = adversarial_loss(real_pred, torch.ones_like(real_pred))
        d_loss_fake = adversarial_loss(fake_pred, torch.zeros_like(fake_pred))
        d_loss = (d_loss_real + d_loss_fake) / 2

        optimizer_d.zero_grad()
        d_loss.backward()
        optimizer_d.step()

        # Train Generator (Encoder and Decoder)
        fake_pred = discriminator(z)
        g_loss_adv = adversarial_loss(fake_pred, torch.ones_like(fake_pred))
        g_loss_recon = reconstruction_loss(decoder(z), imgs)
        g_loss = g_loss_recon + g_loss_adv

        optimizer_g.zero_grad()
        g_loss.backward()
        optimizer_g.step()

    print(f"[Epoch {epoch}/{epochs}] [D loss: {d_loss.item()}] [G loss: {g_loss.item()}]")
\end{lstlisting}

In this implementation, the encoder and decoder form an autoencoder, while the discriminator regularizes the latent space by ensuring it follows a predefined distribution.

\section{Bidirectional GAN (BiGAN)}
Bidirectional GAN (BiGAN) extends the standard GAN by learning an inverse mapping from the data space back to the latent space. This enables BiGAN to both generate data from latent vectors and infer the latent vector corresponding to a given data sample, making it possible to perform tasks such as representation learning and data compression~\cite{donahue2016adversarial}.

\subsection{Core Concept of BiGAN}
In traditional GANs, only the generator maps from the latent space to the data space. BiGAN introduces an encoder network, which maps from the data space to the latent space. The encoder and generator are trained jointly in an adversarial framework, with the discriminator distinguishing between pairs of real data and real latent vectors, and pairs of generated data and fake latent vectors.

\subsection{BiGAN Objective Function}
The BiGAN loss function is:

\[
\mathcal{L}_{\text{BiGAN}} = \mathcal{L}_{\text{GAN}} + \lambda \mathcal{L}_{\text{encoder}}
\]

Where \( \mathcal{L}_{\text{encoder}} \) ensures that the encoder accurately maps data samples to their corresponding latent vectors.

\section{Autoencoder GAN (AEGAN)}
Autoencoder GAN (AEGAN) combines autoencoders and GANs to improve the quality of generated samples and ensure that the learned representations are useful for downstream tasks. AEGAN uses an autoencoder structure to generate data, and the discriminator ensures that the generated samples are indistinguishable from real data~\cite{donahue2016adversarial}.

\section{Summary}
In this chapter, we explored several advanced GAN variants, each offering unique approaches to improving GAN performance or extending their capabilities. Energy-Based GAN (EBGAN) treats the discriminator as an energy function, while Adversarial Autoencoders (AAE) impose a prior on the latent space using adversarial training~\cite{xu2019cross}. Bidirectional GAN (BiGAN) introduces an encoder to learn mappings from data to the latent space, and Autoencoder GAN (AEGAN) combines autoencoders and GANs to generate high-quality samples with useful latent representations. Each of these variants expands the potential applications of GANs and provides new tools for tasks such as image generation, representation learning, and data synthesis.

\part{Applications of GANs}

\chapter{Image Generation and Editing}
Generative Adversarial Networks (GANs) have gained widespread recognition for their ability to generate and edit images~\cite{goodfellow2014generative, wang2015deep, li2020gan, wang2022unrolled}. The applications of GANs in this domain range from creating high-resolution images to transforming images based on various styles or attributes. In this chapter, we will explore the techniques and methods used for image generation and editing, focusing on high-resolution image generation and artistic style transfer. Each section will provide a detailed, beginner-friendly explanation, along with examples and code snippets to guide readers through the concepts.

\section{Image Generation}
Image generation is one of the most popular applications of GANs. GANs are capable of producing highly realistic images from random noise, especially when trained on large datasets. With the advancement of GAN architectures, such as Progressive GANs (ProGAN)~\cite{he2018probgan} and StyleGAN~\cite{karras2019style}, high-resolution image generation has become a reality. In this section, we will discuss the challenges of generating high-resolution images and demonstrate how GANs can be used to overcome these challenges.

\subsection{High-Resolution Image Generation}
Generating high-resolution images using GANs poses several challenges. As the resolution increases, the complexity of the generated images also increases, making it difficult for the generator to capture fine details and for the discriminator to distinguish between real and fake images. Moreover, training GANs for high-resolution images is computationally expensive and requires stable training techniques~\cite{wang2018esrgan}.

\subsubsection{Challenges of High-Resolution Image Generation}
The main challenges of generating high-resolution images include:
\begin{itemize}
    \item \textbf{Mode Collapse:} The generator might focus on generating a limited variety of images, leading to poor diversity in the generated samples.
    \item \textbf{Training Instability:} As the resolution increases, GAN training can become unstable, with the generator and discriminator oscillating rather than converging.
    \item \textbf{Memory and Computational Requirements:} High-resolution images require more memory and computational resources, making it difficult to train models on standard hardware.
\end{itemize}

To address these challenges, advanced techniques such as progressive growing and multi-scale training have been introduced. One of the most notable architectures for high-resolution image generation is \textbf{Progressive GAN (ProGAN)}~\cite{he2018probgan}.

\subsubsection{Progressive Growing of GANs (ProGAN)}
ProGAN, introduced by Karras et al., is an architecture designed specifically to handle high-resolution image generation. The key idea behind ProGAN is to train the GAN progressively, starting from a low-resolution image and gradually increasing the resolution by adding new layers to both the generator and discriminator.

\textbf{Key Features of ProGAN:}
\begin{itemize}
    \item \textbf{Progressive Layer Addition:} The model starts by generating low-resolution images (e.g., 4x4 pixels) and progressively adds layers to generate higher-resolution images (e.g., 1024x1024 pixels).
    \item \textbf{Fade-in Transition:} When new layers are added, their contribution is gradually increased using a fade-in transition. This smooth transition prevents the model from destabilizing as the resolution increases~\cite{he2018probgan}.
\end{itemize}

\textbf{Example of ProGAN in PyTorch:}
Here is a simplified implementation of Progressive GAN using PyTorch. The generator starts by generating low-resolution images and gradually increases the resolution by adding layers.

\begin{lstlisting}[style=python]
import torch
import torch.nn as nn
import torch.optim as optim
import torch.nn.functional as F

# Simple ProGAN-like Generator
class ProGANGenerator(nn.Module):
    def __init__(self, latent_dim, start_res):
        super(ProGANGenerator, self).__init__()
        self.start_res = start_res  # Starting resolution (e.g., 4x4)
        self.latent_dim = latent_dim
        self.model = nn.ModuleList([self.initial_block(latent_dim, start_res)])

    def initial_block(self, latent_dim, res):
        return nn.Sequential(
            nn.Linear(latent_dim, 128 * res * res),
            nn.ReLU(),
            nn.Unflatten(1, (128, res, res))
        )

    def add_layer(self, in_channels, out_channels):
        block = nn.Sequential(
            nn.Conv2d(in_channels, out_channels, kernel_size=3, padding=1),
            nn.ReLU(),
            nn.Upsample(scale_factor=2)
        )
        self.model.append(block)

    def forward(self, z):
        x = self.model[0](z)
        for block in self.model[1:]:
            x = block(x)
        return torch.tanh(x)

# Initialize generator
latent_dim = 100
start_res = 4
generator = ProGANGenerator(latent_dim, start_res)

# Example of adding layers to increase resolution progressively
generator.add_layer(128, 64)  # 8x8 resolution
generator.add_layer(64, 32)   # 16x16 resolution
generator.add_layer(32, 3)    # Final layer to output 3-channel image
\end{lstlisting}

In this code, the generator starts by generating a low-resolution 4x4 image and progressively adds layers to increase the resolution. Each layer doubles the resolution, allowing the generator to handle higher complexity step by step.

\subsubsection{Training Strategies for High-Resolution Image Generation}
To train GANs for high-resolution image generation, several strategies are commonly employed:
\begin{itemize}
    \item \textbf{Multi-Scale Training:} The generator is trained to produce images at multiple resolutions, starting from low resolution and progressively increasing it. This allows the generator to capture global structure before focusing on finer details~\cite{wang2018cgan}.
    \item \textbf{Batch Normalization and Instance Normalization:} These normalization techniques help stabilize GAN training by ensuring that the generator and discriminator operate on well-behaved data distributions.
    \item \textbf{Noise Injection:} Adding noise at various stages of the generator can help the model generalize better and avoid overfitting to the training data.
\end{itemize}

\subsection{Artistic Style Transfer}
Artistic style transfer refers to the process of transforming the style of one image (e.g., a photograph) into the artistic style of another image (e.g., a painting). GANs have proven to be highly effective for this task, allowing for the seamless transfer of artistic styles between images~\cite{wang2020attentive}. 

\subsubsection{What is Style Transfer?}
Style transfer aims to separate the content and style of an image~\cite{jing2019neural}. The content refers to the objects and structure in the image, while the style refers to the texture, colors, and artistic features. The goal of style transfer is to apply the style of one image to the content of another image.

\textbf{Example:} 
\begin{itemize}
    \item Content Image: A photograph of a landscape.
    \item Style Image: A painting by Van Gogh.
    \item Result: A photograph of the landscape in the style of Van Gogh's painting.
\end{itemize}

\subsubsection{CycleGAN for Unsupervised Style Transfer}
CycleGAN~\cite{chu2017cyclegan} is one of the most popular GAN architectures for unsupervised image translation, including artistic style transfer. CycleGAN does not require paired images from two domains. Instead, it learns to map images from one domain (e.g., photographs) to another domain (e.g., paintings) without needing paired examples.

CycleGAN consists of two generators and two discriminators:
\begin{itemize}
    \item \( G: A \to B \) - Translates images from domain \( A \) (e.g., photographs) to domain \( B \) (e.g., paintings).
    \item \( F: B \to A \) - Translates images from domain \( B \) to domain \( A \).
    \item \( D_A \) - Discriminator for domain \( A \), ensuring that translated images from \( F \) are realistic.
    \item \( D_B \) - Discriminator for domain \( B \), ensuring that translated images from \( G \) are realistic.
\end{itemize}

\subsubsection{Cycle Consistency Loss in Style Transfer}
To ensure that the style transfer does not lose important content information, CycleGAN introduces the concept of cycle consistency. This means that if we translate an image from domain \( A \) to domain \( B \) and then back to domain \( A \), the result should closely resemble the original image~\cite{chu2017cyclegan}.

\[
\mathcal{L}_{\text{cycle}}(G, F) = \mathbb{E}_{x \sim p_{\text{data}}(x)}[\| F(G(x)) - x \|_1] + \mathbb{E}_{y \sim p_{\text{data}}(y)}[\| G(F(y)) - y \|_1]
\]

\subsubsection{CycleGAN Implementation for Style Transfer}
Here's a basic CycleGAN implementation using PyTorch for unsupervised style transfer between photographs and paintings.

\begin{lstlisting}[style=python]
class ResnetBlock(nn.Module):
    def __init__(self, dim):
        super(ResnetBlock, self).__init__()
        self.conv_block = nn.Sequential(
            nn.Conv2d(dim, dim, kernel_size=3, padding=1),
            nn.InstanceNorm2d(dim),
            nn.ReLU(True),
            nn.Conv2d(dim, dim, kernel_size=3, padding=1),
            nn.InstanceNorm2d(dim)
        )

    def forward(self, x):
        return x + self.conv_block(x)

class CycleGANGenerator(nn.Module):
    def __init__(self, in_channels, out_channels, num_resnet_blocks=6):
        super(CycleGANGenerator, self).__init__()
        model = [
            nn.Conv2d(in_channels, 64, kernel_size=7, padding=3),
            nn.InstanceNorm2d(64),
            nn.ReLU(True),
            nn.Conv2d(64, 128, kernel_size=3, stride=2, padding=1),
            nn.InstanceNorm2d(128),
            nn.ReLU(True),
            nn.Conv2d(128, 256, kernel_size=3, stride=2, padding=1),
            nn.InstanceNorm2d(256),
            nn.ReLU(True)
        ]
        for _ in range(num_resnet_blocks):
            model += [ResnetBlock(256)]
        model += [
            nn.ConvTranspose2d(256, 128, kernel_size=3, stride=2, padding=1, output_padding=1),
            nn.InstanceNorm2d(128),
            nn.ReLU(True),
            nn.ConvTranspose2d(128, 64, kernel_size=3, stride=2, padding=1, output_padding=1),
            nn.InstanceNorm2d(64),
            nn.ReLU(True),
            nn.Conv2d(64, out_channels, kernel_size=7, padding=3),
            nn.Tanh()
        ]
        self.model = nn.Sequential(*model)

    def forward(self, x):
        return self.model(x)

# Initialize models
G_A2B = CycleGANGenerator(in_channels=3, out_channels=3)
G_B2A = CycleGANGenerator(in_channels=3, out_channels=3)

# Example of using the generator to apply style transfer
content_image = torch.randn(1, 3, 256, 256)  # Example content image
style_transferred_image = G_A2B(content_image)  # Style transferred image
\end{lstlisting}

In this implementation, the generator consists of convolutional layers and residual blocks, which are effective for learning artistic style mappings between domains.

\section{Summary}
In this chapter, we explored two important applications of GANs: high-resolution image generation and artistic style transfer. GANs such as ProGAN have been specifically designed to handle the challenges of high-resolution image generation by progressively increasing the resolution during training. We also covered CycleGAN, a powerful architecture for unsupervised image translation, which has been successfully applied to tasks like artistic style transfer. Through detailed explanations and code examples, we provided a comprehensive guide for beginners to understand how GANs can be used for various image generation and editing tasks.

\section{Image Editing}
Generative Adversarial Networks (GANs) have been widely applied in the field of image editing, where they enable the manipulation and generation of high-quality, realistic images. Image editing tasks include face generation and editing, image inpainting (repairing damaged images), and denoising (removing noise from images). This section explores how GANs are used in these image editing tasks, detailing the underlying techniques and providing examples.

\subsection{Face Generation and Editing}
Face generation and editing are popular applications of GANs, where the goal is to generate new facial images or edit existing ones in a controlled way. GANs, particularly architectures like StyleGAN~\cite{karras2019style}, have shown incredible results in generating highly realistic faces, allowing users to manipulate various facial attributes, such as age, hair color, expression, and more~\cite{patashnik2021styleclip}.

\textbf{1. Latent Space Interpolation:}  
In GANs, particularly StyleGAN, images are generated by sampling from a latent space, which encodes different attributes of the image. By manipulating vectors in this latent space, we can generate new faces or modify specific attributes of existing faces. For example, moving in a certain direction in the latent space might change the age of a person, while moving in another direction might change their hairstyle.

\textbf{2. Attribute Editing:}  
GANs can be used to edit specific attributes of an image by conditioning the generation process on certain attributes. This can be done by training the Generator to learn how to map latent vectors and specific attributes (e.g., age, gender) to facial images. By modifying these attribute values, we can control how the generated image changes~\cite{abdal2019image2stylegan}.

\textbf{Example: Face Generation with Latent Space Manipulation in PyTorch}

\begin{lstlisting}[style=python]
import torch
import torch.nn as nn

# Simple GAN Generator for face generation
class FaceGenerator(nn.Module):
    def __init__(self, input_dim, output_dim):
        super(FaceGenerator, self).__init__()
        self.model = nn.Sequential(
            nn.Linear(input_dim, 256),
            nn.ReLU(),
            nn.Linear(256, 512),
            nn.ReLU(),
            nn.Linear(512, 1024),
            nn.ReLU(),
            nn.Linear(1024, output_dim),
            nn.Tanh()  # Output scaled between -1 and 1 (for image pixel values)
        )

    def forward(self, x):
        return self.model(x)

# Example usage:
latent_vector = torch.randn(1, 100)  # Latent vector representing face attributes
face_generator = FaceGenerator(100, 3 * 64 * 64)  # Assuming output is 64x64 RGB image
generated_face = face_generator(latent_vector)
print(generated_face.shape)  # Output should be torch.Size([1, 12288]), which is 64x64x3
\end{lstlisting}

In this simple example, the Generator takes a latent vector as input and produces an image of a face. By modifying the latent vector, we can change different facial attributes like expression or hairstyle.

\subsection{Image Inpainting and Denoising}
Image inpainting (image completion) and denoising are important tasks in the field of image restoration. Inpainting refers to the process of filling in missing or damaged parts of an image, while denoising refers to removing noise from a corrupted image. GANs are highly effective at these tasks because they can learn to generate realistic details that match the surrounding context.

\textbf{1. Image Inpainting:}  
Inpainting with GANs involves generating missing pixels in an image by conditioning the generation process on the known surrounding pixels. The Generator learns to fill in the missing areas with realistic content that matches the rest of the image. GANs are particularly useful here because they can generate semantically consistent content, ensuring that the inpainted region fits naturally with the surrounding pixels~\cite{pang2021image}.

\textbf{2. Image Denoising:}  
GANs can also be applied to image denoising by learning to map noisy images to clean, denoised versions. The Generator is trained to remove noise while preserving important image details. The Discriminator ensures that the generated image looks as realistic as possible by distinguishing between real clean images and denoised images.

\textbf{Example: Image Inpainting with GANs in PyTorch}

\begin{lstlisting}[style=python]
class InpaintingGenerator(nn.Module):
    def __init__(self):
        super(InpaintingGenerator, self).__init__()
        self.model = nn.Sequential(
            nn.Conv2d(3, 64, kernel_size=4, stride=2, padding=1),
            nn.ReLU(),
            nn.Conv2d(64, 128, kernel_size=4, stride=2, padding=1),
            nn.ReLU(),
            nn.ConvTranspose2d(128, 64, kernel_size=4, stride=2, padding=1),
            nn.ReLU(),
            nn.ConvTranspose2d(64, 3, kernel_size=4, stride=2, padding=1),
            nn.Tanh()  # Output scaled between -1 and 1
        )

    def forward(self, x):
        return self.model(x)

# Example usage:
damaged_image = torch.randn(1, 3, 64, 64)  # Simulated damaged image
inpainting_generator = InpaintingGenerator()
reconstructed_image = inpainting_generator(damaged_image)
print(reconstructed_image.shape)  # Output should be torch.Size([1, 3, 64, 64])
\end{lstlisting}

In this example, the Generator takes a partially damaged image as input and outputs the inpainted image. The inpainted areas are generated to match the known areas of the image, producing a seamless and realistic result.

\section{Image Translation and Style Transfer}
Image translation and style transfer are tasks in which one image is transformed into another while maintaining some of its key properties. GANs, particularly models like CycleGAN~\cite{chu2017cyclegan}, are well-suited for these tasks as they can learn complex mappings between different image domains, such as transforming a photograph into a painting or converting images between different styles~\cite{karras2019style}.

\subsection{Supervised and Unsupervised Image Translation}
Image translation refers to the process of converting an image from one domain to another. For instance, translating an image of a horse into an image of a zebra, or turning a day-time scene into a night-time scene. This can be done in both supervised and unsupervised ways:

\textbf{1. Supervised Image Translation:}  
In supervised image translation, we have paired examples of images from the source and target domains. For instance, we may have pairs of day-time and night-time images of the same scene. The GAN is trained to map the source image to the target image by learning from these paired examples~\cite{abdal2019image2stylegan}.

\textbf{2. Unsupervised Image Translation:}  
In many cases, paired examples are not available. Unsupervised image translation methods like CycleGAN allow the model to learn mappings between domains without paired examples. CycleGAN uses a cycle consistency loss to ensure that when an image is translated from one domain to another and back, it returns to the original image.

\textbf{Example: CycleGAN for Unsupervised Image Translation in PyTorch}

\begin{lstlisting}[style=python]
class CycleGAN_Generator(nn.Module):
    def __init__(self, in_channels, out_channels):
        super(CycleGAN_Generator, self).__init__()
        self.model = nn.Sequential(
            nn.Conv2d(in_channels, 64, kernel_size=4, stride=2, padding=1),
            nn.ReLU(),
            nn.Conv2d(64, 128, kernel_size=4, stride=2, padding=1),
            nn.ReLU(),
            nn.ConvTranspose2d(128, 64, kernel_size=4, stride=2, padding=1),
            nn.ReLU(),
            nn.ConvTranspose2d(64, out_channels, kernel_size=4, stride=2, padding=1),
            nn.Tanh()
        )

    def forward(self, x):
        return self.model(x)

# Example usage:
image_A = torch.randn(1, 3, 64, 64)  # Image from domain A (e.g., horses)
cyclegan_generator = CycleGAN_Generator(3, 3)
image_B = cyclegan_generator(image_A)  # Translate to domain B (e.g., zebras)
print(image_B.shape)  # Output should be torch.Size([1, 3, 64, 64])
\end{lstlisting}

In this example, a simple CycleGAN Generator is used to translate an image from one domain to another. The same Generator can be used to translate the image back to the original domain using the cycle consistency loss.

\subsection{Cross-Domain Style Transfer}
Style transfer refers to the task of transferring the style of one image onto the content of another. For example, transforming a photograph into a painting by a famous artist. GANs can be used for cross-domain style transfer, where the Generator is trained to map the content of one image into the style of another domain, such as mapping real-world images into artistic styles or transferring the textures of one object to another~\cite{chu2017cyclegan}.

\textbf{1. Neural Style Transfer with GANs:}  
GANs are powerful for performing neural style transfer because they can generate high-quality, stylized images while preserving the underlying content of the original image. By training the Generator to apply the style of a target domain, we can create visually appealing results where the content remains unchanged but the style is dramatically altered.

\textbf{2. Multi-Domain Style Transfer:}  
In multi-domain style transfer, the Generator is trained to transfer images across multiple styles (e.g., turning a photograph into different painting styles). This is done by conditioning the Generator on the target style, allowing it to generate images in a variety of different styles from a single model.

\textbf{Example: Style Transfer with GANs in PyTorch}

\begin{lstlisting}[style=python]
class StyleTransferGenerator(nn.Module):
    def __init__(self, in_channels, out_channels, num_styles):
        super(StyleTransferGenerator, self).__init__()
        self.model = nn.Sequential(
            nn.Conv2d(in_channels, 64, kernel_size=4, stride=2, padding=1),
            nn.ReLU(),
            nn.Conv2d(64, 128, kernel_size=4, stride=2, padding=1),
            nn.ReLU(),
            nn.ConvTranspose2d(128, 64, kernel_size=4, stride=2, padding=1),
            nn.ReLU(),
            nn.ConvTranspose2d(64, out_channels, kernel_size=4, stride=2, padding=1),
            nn.Tanh()
        )
        self.style_embedding = nn.Embedding(num_styles, 128)  # Embedding for different styles

    def forward(self, x, style_idx):
        style = self.style_embedding(style_idx).view(x.size(0), -1, 1, 1)
        x = self.model(x)
        return x * style  # Apply style modulation

# Example usage:
image = torch.randn(1, 3, 64, 64)  # Content image
style_idx = torch.tensor([2])  # Target style index
style_transfer_generator = StyleTransferGenerator(3, 3, num_styles=5)
styled_image = style_transfer_generator(image, style_idx)
print(styled_image.shape)  # Output should be torch.Size([1, 3, 64, 64])
\end{lstlisting}

In this example, a style transfer Generator is implemented, where the style is modulated by an embedding for different styles. The image content remains the same, but the style can be changed by selecting different style indices.

\textbf{Visualizing Cross-Domain Style Transfer~\cite{xu2019cross}:}

\begin{center}
\begin{tikzpicture}
  [scale=1, every node/.style={scale=1}, 
  block/.style={rectangle, draw, fill=blue!20, text centered, minimum height=3em},
  arrow/.style={->, thick}]

  \node[block] (content) {Content Image};
  \node[block, right=of content] (gen) {Style Transfer Generator};
  \node[block, right=of gen] (output) {Stylized Image};
  \node[block, below=of gen] (style) {Style Embedding};

  \draw[arrow] (content) -- (gen);
  \draw[arrow] (gen) -- (output);
  \draw[arrow] (style) -- (gen);
\end{tikzpicture}
\end{center}

In this diagram, the content image is passed through the Style Transfer Generator, which is conditioned on the target style embedding. The output is a stylized version of the content image, reflecting the selected style.

\chapter{Video Generation and Processing}
GANs are not only used for generating and editing images but also have significant applications in video generation and processing~\cite{adadi2021survey}. Video data has an additional temporal dimension, making it more complex than static images. GAN-based models have been extended to handle this temporal aspect, allowing them to generate realistic videos, predict future frames, perform frame interpolation, and even transfer styles between videos~\cite{xu2024videogigagan}. In this chapter, we will cover the core ideas behind GAN-based video generation and the challenges that come with it, providing detailed explanations, examples, and code implementations for beginners~\cite{tulyakov2017mocogan}.

\section{GAN-Based Video Generation}
GANs for video generation extend the principles of image-based GANs to handle both the spatial and temporal dimensions of video~\cite{chen2017coherent}. Instead of generating a single image, the generator now learns to produce a sequence of frames that form a coherent video. The discriminator evaluates not just the individual frames, but the temporal consistency between them.

\subsection{Key Concepts in Video Generation with GANs}
In video generation, it is essential to ensure both the quality of individual frames and the temporal coherence between consecutive frames~\cite{li2024survey}. Several techniques and models have been developed to achieve this, such as:
\begin{itemize}
    \item \textbf{Spatial Consistency:} Each frame in the generated video must maintain high visual quality and be consistent with the overall scene.
    \item \textbf{Temporal Coherence:} The frames must flow naturally from one to the next, ensuring smooth motion and avoiding abrupt changes or artifacts.
    \item \textbf{Recurrent Generators:} Many video GANs use recurrent neural networks (RNNs) or 3D convolutions to model temporal dependencies between frames.
\end{itemize}

\subsubsection{VGAN: Video GAN}
One of the earliest approaches to GAN-based video generation is the Video GAN (VGAN)~\cite{aldausari2022video}. VGAN extends the standard GAN architecture to generate sequences of images, ensuring temporal coherence through the use of 3D convolutions~\cite{brooks2022generating}.

\textbf{VGAN Architecture:}
\begin{itemize}
    \item \textit{3D Convolutional Generator:} The generator takes a noise vector as input and generates a sequence of frames using 3D convolutional layers. This allows the model to capture both spatial and temporal features.
    \item \textit{3D Convolutional Discriminator:} The discriminator evaluates the generated video as a whole, considering both the spatial and temporal dimensions to determine if the video is real or fake.
\end{itemize}

\subsubsection{VGAN Implementation in PyTorch}
Here is a simplified implementation of VGAN in PyTorch:

\begin{lstlisting}[style=python]
import torch
import torch.nn as nn

# 3D Convolutional Generator
class VGANGenerator(nn.Module):
    def __init__(self, latent_dim):
        super(VGANGenerator, self).__init__()
        self.model = nn.Sequential(
            nn.ConvTranspose3d(latent_dim, 512, kernel_size=(4, 4, 4), stride=1, padding=0),
            nn.BatchNorm3d(512),
            nn.ReLU(),
            nn.ConvTranspose3d(512, 256, kernel_size=(4, 4, 4), stride=2, padding=1),
            nn.BatchNorm3d(256),
            nn.ReLU(),
            nn.ConvTranspose3d(256, 128, kernel_size=(4, 4, 4), stride=2, padding=1),
            nn.BatchNorm3d(128),
            nn.ReLU(),
            nn.ConvTranspose3d(128, 3, kernel_size=(4, 4, 4), stride=2, padding=1),
            nn.Tanh()
        )

    def forward(self, z):
        z = z.view(z.size(0), z.size(1), 1, 1, 1)  # Expand latent vector to 5D (batch, channels, depth, height, width)
        return self.model(z)

# 3D Convolutional Discriminator
class VGANDiscriminator(nn.Module):
    def __init__(self):
        super(VGANDiscriminator, self).__init__()
        self.model = nn.Sequential(
            nn.Conv3d(3, 128, kernel_size=(4, 4, 4), stride=2, padding=1),
            nn.LeakyReLU(0.2),
            nn.Conv3d(128, 256, kernel_size=(4, 4, 4), stride=2, padding=1),
            nn.BatchNorm3d(256),
            nn.LeakyReLU(0.2),
            nn.Conv3d(256, 512, kernel_size=(4, 4, 4), stride=2, padding=1),
            nn.BatchNorm3d(512),
            nn.LeakyReLU(0.2),
            nn.Conv3d(512, 1, kernel_size=(4, 4, 4), stride=1, padding=0)
        )

    def forward(self, video):
        return self.model(video).view(-1, 1)

# Initialize models
latent_dim = 100
generator = VGANGenerator(latent_dim)
discriminator = VGANDiscriminator()

# Example input to generate video
z = torch.randn(8, latent_dim)  # Batch of latent vectors
generated_video = generator(z)   # Generate a batch of videos
\end{lstlisting}

In this implementation, the generator takes a latent vector and outputs a sequence of frames using 3D convolutions. The discriminator evaluates the entire video to determine if it is real or fake, ensuring both spatial and temporal coherence.

\section{Video Prediction and Frame Interpolation}
Video prediction involves forecasting future frames of a video based on past frames, while frame interpolation aims to generate intermediate frames between existing ones. These tasks are challenging because they require a model to understand the motion dynamics and predict smooth transitions between frames~\cite{aldausari2022video}.

\subsection{GANs for Video Prediction}
GANs are particularly well-suited for video prediction tasks because they can model the complex dynamics of motion in videos. A common approach is to use a conditional GAN (cGAN), where the generator takes the past frames as input and predicts the future frames.

\subsubsection{Example: Conditional GAN for Video Prediction}
In conditional GANs for video prediction, the generator is conditioned on the past frames, and the discriminator evaluates the predicted future frames along with the past frames.

\begin{lstlisting}[style=python]
# Conditional Generator for Video Prediction
class VideoPredictionGenerator(nn.Module):
    def __init__(self, in_channels, out_channels):
        super(VideoPredictionGenerator, self).__init__()
        self.model = nn.Sequential(
            nn.Conv2d(in_channels, 128, kernel_size=3, padding=1),
            nn.ReLU(),
            nn.Conv2d(128, 256, kernel_size=3, padding=1),
            nn.ReLU(),
            nn.ConvTranspose2d(256, out_channels, kernel_size=4, stride=2, padding=1),
            nn.Tanh()
        )

    def forward(self, x):
        return self.model(x)

# Initialize generator
in_channels = 3 * 5  # Example: 5 past frames with 3 channels each
out_channels = 3 * 1  # Predicting 1 future frame with 3 channels
generator = VideoPredictionGenerator(in_channels, out_channels)

# Example input: 5 past frames concatenated along the channel dimension
past_frames = torch.randn(8, in_channels, 64, 64)  # Batch of past frames
predicted_frame = generator(past_frames)  # Generate the future frame
\end{lstlisting}

In this example, the generator predicts the next frame in a video sequence based on past frames. The model can be trained with a discriminator that ensures the predicted frames are realistic and consistent with the previous frames.

\section{Video Style Transfer}
Video style transfer refers to applying the artistic style of one video (or image) to another video. The challenge here is not only to transfer the style to individual frames but also to maintain temporal consistency between the frames~\cite{aldausari2022video}.

\subsection{Maintaining Temporal Consistency in Video Generation}
One of the key challenges in video generation is ensuring temporal consistency. Temporal consistency refers to the smoothness of transitions between frames, which is critical for creating realistic videos~\cite{li2024survey}. If each frame is generated independently, the result may suffer from flickering or abrupt changes between frames.

\textit{Techniques to Ensure Temporal Consistency:}
\begin{itemize}
    \item \textit{Recurrent Neural Networks (RNNs):} Using RNNs or Long Short-Term Memory (LSTM) networks helps the generator remember information from previous frames, enabling smoother transitions.
    \item \textit{Optical Flow Constraints:} Enforcing optical flow consistency between frames ensures that motion is continuous and realistic.
    \item \textit{Temporal Loss Functions:} Adding a temporal loss that penalizes large differences between consecutive frames helps enforce consistency.
\end{itemize}

\subsubsection{Example: Temporal Loss for Video Style Transfer}
In this example, we apply a temporal loss to ensure smooth transitions between frames during video style transfer.

\begin{lstlisting}[style=python]
# Temporal Loss Function
def temporal_loss(current_frame, previous_frame):
    return nn.functional.mse_loss(current_frame, previous_frame)

# Example usage in training loop
for t in range(1, num_frames):
    current_frame = generated_video[:, :, t, :, :]  # t-th frame
    previous_frame = generated_video[:, :, t-1, :, :]  # (t-1)-th frame
    loss_temporal = temporal_loss(current_frame, previous_frame)
    total_loss = loss_adversarial + lambda_temporal * loss_temporal
    total_loss.backward()
\end{lstlisting}

This temporal loss ensures that consecutive frames in the generated video are smooth and coherent, avoiding artifacts such as flickering.

\section{Challenges and Solutions in Video Generation}
Generating videos with GANs presents several unique challenges that go beyond those encountered in image generation~\cite{chen2017coherent}. Some of the key challenges include~\cite{li2024survey}:

\subsection{Handling High Dimensionality}
Video data is inherently high-dimensional, as it consists of multiple frames over time. This increases the memory and computational requirements for training GANs on video data. One solution is to reduce the resolution of the input frames or use efficient 3D convolutions~\cite{tulyakov2017mocogan}.

\subsection{Ensuring Temporal Coherence}
Temporal coherence is crucial for generating realistic videos. As mentioned earlier, incorporating recurrent layers, optical flow constraints, or temporal loss functions can help maintain smooth transitions between frames.

\subsection{Avoiding Mode Collapse}
Just like in image generation, mode collapse can be an issue in video GANs. In video generation, mode collapse may result in repetitive or static video sequences. Techniques such as feature matching loss and multi-scale discrimination can be used to mitigate this.

\subsection{Training Stability}
Training GANs on video data can be unstable, especially when handling long video sequences. Progressive training strategies, such as starting with short sequences and gradually increasing the sequence length, can help stabilize training~\cite{chen2017coherent}.

\section{Summary}
In this chapter, we explored various aspects of GAN-based video generation and processing. We covered the fundamentals of video GANs, including architectures like VGAN that use 3D convolutions to model the temporal dynamics of videos. We also discussed video prediction, frame interpolation, and video style transfer, emphasizing the importance of temporal consistency in these tasks. Finally, we outlined some of the major challenges in video generation and offered potential solutions to address these issues~\cite{li2024survey}. Through practical examples and code implementations in PyTorch, we provided a clear and comprehensive guide for beginners interested in applying GANs to video generation and processing tasks.


\chapter{Applications in Text, Speech, and Other Domains}

Generative Adversarial Networks (GANs) have demonstrated tremendous versatility across various domains~\cite{yang2022wavegan}, including text generation~\cite{de2021survey}, speech synthesis, medical imaging, and even the creation of virtual worlds~\cite{li2024survey}. While GANs were originally designed for generating realistic images, their potential has been extended to other forms of data, leading to groundbreaking advancements in fields like natural language processing (NLP)~\cite{hirschberg2015advances}, audio engineering~\cite{zwicker1991audio}, and healthcare~\cite{murmu2024reliable}. In this chapter, we will explore how GANs can be applied to these diverse domains, providing step-by-step explanations and practical examples using PyTorch. We will start with text generation, explaining how GANs can be adapted to produce coherent and contextually accurate sequences of words.

\section{Text Generation}

Text generation is one of the most challenging tasks in machine learning, primarily because it involves creating sequences of coherent and grammatically correct text~\cite{lin2019commongen}. Traditional language models have been employed to generate text, but they often suffer from issues like lack of diversity and repetitive phrases. GANs offer a new way of addressing these challenges by employing a generator-discriminator framework that can learn to produce more diverse and natural language outputs. In this section, we will explore two prominent models for text generation: SeqGAN~\cite{yu2017seqgan} and TextGAN~\cite{zhang2017adversarial}.

\subsection{SeqGAN: Sequence Generative Adversarial Networks}

SeqGAN is a pioneering approach that adapts the GAN framework for the generation of discrete sequences, such as text~\cite{yu2017seqgan}. Unlike images, where pixel values are continuous, text is composed of discrete tokens (words or characters), which poses a unique challenge for traditional GANs. SeqGAN addresses this issue by using reinforcement learning (RL)~\cite{kaelbling1996reinforcement} techniques to allow the generator to learn from the feedback provided by the discriminator, even when the data is not continuous.

\textbf{1. Overview of SeqGAN}

SeqGAN is designed to handle the problem of generating sequences by framing it as a reinforcement learning problem. The generator is treated as an agent that generates sequences, and the reward signal is provided by the discriminator, which acts as a critic~\cite{yu2017seqgan}. The key components of SeqGAN are:
\begin{itemize}
    \item \textbf{Generator ($G$):} The generator in SeqGAN produces sequences of tokens (e.g., words or characters). It is trained to generate sequences that are indistinguishable from real data.
    \item \textbf{Discriminator ($D$):} The discriminator evaluates the sequences produced by the generator, providing a probability score that indicates how likely a sequence is to be real or fake. This score is used to train the generator.
    \item \textbf{Reinforcement Learning (RL):} Since text data is discrete, standard backpropagation cannot be applied directly. SeqGAN uses a policy gradient method from RL to enable the generator to learn from the rewards given by the discriminator.
\end{itemize}

\textbf{2. Architecture of SeqGAN}

The architecture of SeqGAN can be illustrated as follows:
\begin{figure}[htbp]
    \centering
    \includegraphics[width=\textwidth]{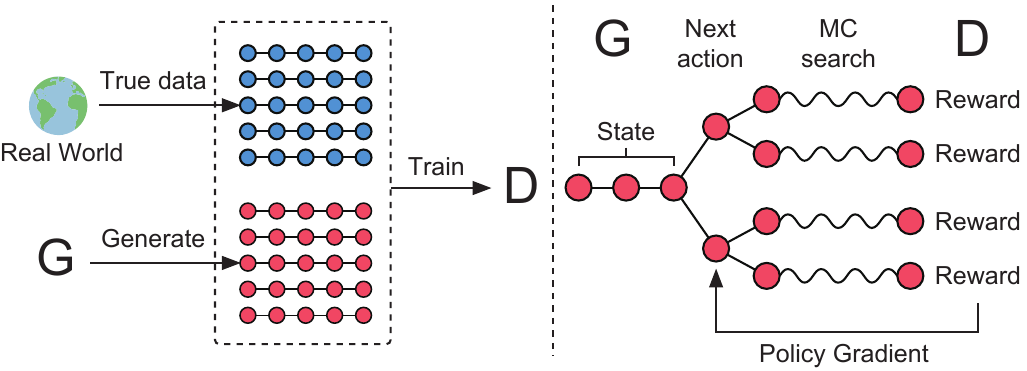}
    \caption{The illustration of SeqGAN. Left: $D$ is trained over the real data and the generated data by $G$. Right: $G$ is trained by policy gradient where the final reward signal is provided by $D$ and is passed back to the intermediate action value via Monte Carlo search~\cite{browne2012survey}. The figure from SeqGAN~\cite{yu2017seqgan}.}
\end{figure}

\textbf{3. Implementation in PyTorch}

Let's look at a simplified implementation of SeqGAN using PyTorch. In this example, we will define the generator and discriminator, and then train the model using the reinforcement learning approach.

\begin{lstlisting}[style=python]
import torch
import torch.nn as nn
import torch.optim as optim
import torch.nn.functional as F

# Define the Generator
class Generator(nn.Module):
    def __init__(self, vocab_size, embed_size, hidden_size):
        super(Generator, self).__init__()
        self.embedding = nn.Embedding(vocab_size, embed_size)
        self.lstm = nn.LSTM(embed_size, hidden_size, batch_first=True)
        self.fc = nn.Linear(hidden_size, vocab_size)
    
    def forward(self, x):
        embedded = self.embedding(x)
        output, _ = self.lstm(embedded)
        output = self.fc(output)
        return F.softmax(output, dim=-1)

# Define the Discriminator
class Discriminator(nn.Module):
    def __init__(self, vocab_size, embed_size, hidden_size):
        super(Discriminator, self).__init__()
        self.embedding = nn.Embedding(vocab_size, embed_size)
        self.lstm = nn.LSTM(embed_size, hidden_size, batch_first=True)
        self.fc = nn.Linear(hidden_size, 1)
    
    def forward(self, x):
        embedded = self.embedding(x)
        output, _ = self.lstm(embedded)
        output = self.fc(output[:, -1, :])  # Use the last hidden state
        return torch.sigmoid(output)

# Hyperparameters
vocab_size = 5000
embed_size = 128
hidden_size = 256

# Initialize Generator and Discriminator
generator = Generator(vocab_size, embed_size, hidden_size)
discriminator = Discriminator(vocab_size, embed_size, hidden_size)

# Optimizers
g_optimizer = optim.Adam(generator.parameters(), lr=0.001)
d_optimizer = optim.Adam(discriminator.parameters(), lr=0.001)

# Example of training loop (simplified)
for epoch in range(100):
    # Generate fake sequences
    fake_data = generator(torch.randint(0, vocab_size, (32, 10)))
    # Train discriminator
    real_data = torch.randint(0, vocab_size, (32, 10))  # Placeholder for real data
    real_labels = torch.ones(32, 1)
    fake_labels = torch.zeros(32, 1)
    
    d_optimizer.zero_grad()
    real_output = discriminator(real_data)
    fake_output = discriminator(fake_data.detach())
    d_loss = F.binary_cross_entropy(real_output, real_labels) + F.binary_cross_entropy(fake_output, fake_labels)
    d_loss.backward()
    d_optimizer.step()
    
    # Train generator using policy gradient (simplified)
    g_optimizer.zero_grad()
    fake_output = discriminator(fake_data)
    g_loss = -torch.mean(torch.log(fake_output))  # Reward is log(D(G(z)))
    g_loss.backward()
    g_optimizer.step()
\end{lstlisting}

In the above code, we defined a simple SeqGAN architecture where the generator and discriminator work together to improve the quality of generated text. This example provides a basic idea of how reinforcement learning can be integrated into the GAN framework for text generation~\cite{yu2017seqgan}. A real-world implementation would involve more complex structures and optimizations.

\section{Speech Generation}

The application of Generative Adversarial Networks (GANs) in the field of speech synthesis has led to significant advancements in generating high-quality, realistic audio~\cite{yang2024integrated}. Unlike image generation, where GANs deal with visual data, speech generation involves producing continuous audio waveforms or spectrograms, which requires different techniques and considerations. In this section, we will explore two notable models for speech generation: WaveGAN and MelGAN. We will provide detailed explanations, along with examples in PyTorch, to help beginners understand how these models work and how to implement them.

\subsection{WaveGAN: Generating Raw Audio Waveforms}

WaveGAN~\cite{donahue2018adversarial} is one of the earliest models designed to generate raw audio waveforms using the GAN framework. Traditional speech synthesis systems convert text to audio by generating spectrograms and then converting those spectrograms to waveforms. However, WaveGAN bypasses this intermediate step by directly generating audio samples, producing continuous waveforms that can be played as audio.

\textbf{1. Overview of WaveGAN}

WaveGAN is designed to produce audio waveforms that are coherent and realistic. The core idea is to treat the generation of waveforms as a one-dimensional sequence generation problem, where the GAN's generator directly outputs samples of the audio waveform. This approach~\cite{donahue2018adversarial} is advantageous because it simplifies the process and can handle tasks like generating speech, music, or other audio effects.

Key components of WaveGAN:
\begin{itemize}
    \item \textbf{Generator ($G$):} The generator produces a sequence of audio samples that form a continuous waveform. It learns to generate realistic audio by mimicking the patterns present in real audio data.
    \item \textbf{Discriminator ($D$):} The discriminator assesses the generated waveforms and distinguishes between real (human-produced) audio and fake (machine-generated) audio.
    \item \textbf{1D Convolutional Layers:} Unlike image-based GANs, WaveGAN uses 1D convolutional layers to process the temporal nature of audio signals.
\end{itemize}

\textbf{2. Methods of WaveGAN}

\begin{figure}[htbp]
    \centering
    \includegraphics[width=\textwidth]{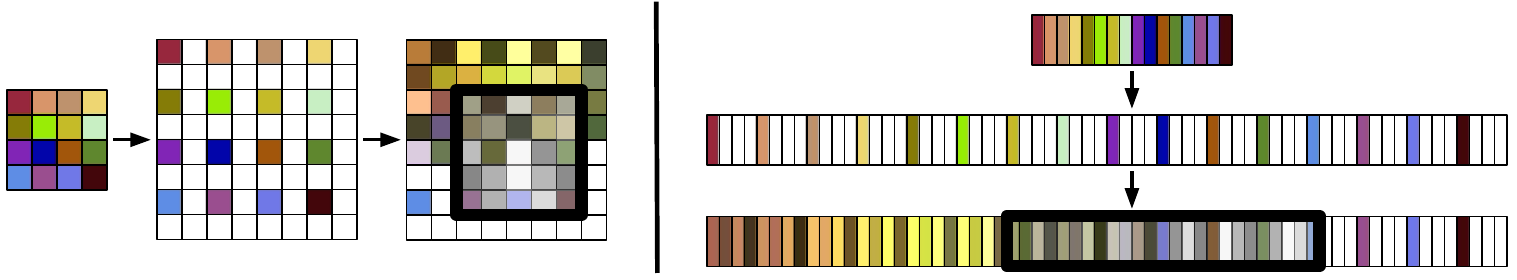}
    \caption{DCGAN uses small (5x5), twodimensional filters while WaveGAN uses longer (length-25), one-dimensional filters and a larger upsampling factor. Both strategies have the same number of parameters and numerical operations. The figure from WaveGAN~\cite{donahue2018adversarial}}
    \label{fig:wavegan0}
\end{figure}

As shown in Figure~\ref{fig:wavegan0}, depiction of the transposed convolution operation for the first layers of the DCGAN~\cite{radford2015unsupervised} (left, and we mentioned before) and WaveGAN~\cite{donahue2018adversarial} (right) generators.

\begin{figure}[htbp]
    \centering
    \includegraphics[width=0.4\textwidth]{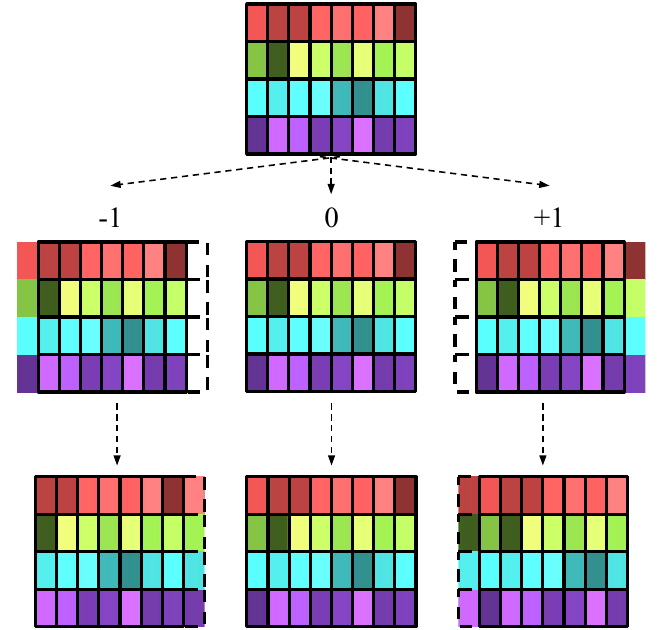}
    \caption{At each layer of the WaveGAN discriminator, the phase shuffle operation perturbs the phase of each feature map. The figure from WaveGAN~\cite{donahue2018adversarial}}
    \label{fig:wavegan1}
\end{figure}

To prevent the discriminator from learning such a solution, we propose the phase shuffle operation with hyperparameter $n$. Phase shuffle randomly perturbs the phase of each layer’s activations by $−n$ to $n$ samples before input to the next layer (Figure~\ref{fig:wavegan1}).

\textbf{3. Implementation in PyTorch}

Below is a simplified example of how to implement WaveGAN using PyTorch. We will define the generator and discriminator, and demonstrate a basic training loop.

\begin{lstlisting}[style=python]
import torch
import torch.nn as nn
import torch.optim as optim
import torch.nn.functional as F

# Define the Generator
class WaveGenerator(nn.Module):
    def __init__(self, latent_dim, output_size):
        super(WaveGenerator, self).__init__()
        self.fc = nn.Linear(latent_dim, 256)
        self.deconv1 = nn.ConvTranspose1d(256, 128, 25, stride=4)
        self.deconv2 = nn.ConvTranspose1d(128, 64, 25, stride=4)
        self.deconv3 = nn.ConvTranspose1d(64, 1, 25, stride=4)
    
    def forward(self, x):
        x = F.relu(self.fc(x).unsqueeze(-1))
        x = F.relu(self.deconv1(x))
        x = F.relu(self.deconv2(x))
        return torch.tanh(self.deconv3(x))

# Define the Discriminator
class WaveDiscriminator(nn.Module):
    def __init__(self, input_size):
        super(WaveDiscriminator, self).__init__()
        self.conv1 = nn.Conv1d(1, 64, 25, stride=4)
        self.conv2 = nn.Conv1d(64, 128, 25, stride=4)
        self.conv3 = nn.Conv1d(128, 256, 25, stride=4)
        self.fc = nn.Linear(256, 1)
    
    def forward(self, x):
        x = F.relu(self.conv1(x))
        x = F.relu(self.conv2(x))
        x = F.relu(self.conv3(x))
        return torch.sigmoid(self.fc(x.view(x.size(0), -1)))

# Initialize models, optimizers, and training loop (simplified)
latent_dim = 100
output_size = 16000  # Example: 1-second audio at 16kHz
generator = WaveGenerator(latent_dim, output_size)
discriminator = WaveDiscriminator(output_size)

g_optimizer = optim.Adam(generator.parameters(), lr=0.0002)
d_optimizer = optim.Adam(discriminator.parameters(), lr=0.0002)

for epoch in range(100):
    # Generate fake audio
    z = torch.randn(32, latent_dim)
    fake_audio = generator(z)
    real_audio = torch.randn(32, 1, output_size)  # Placeholder for real audio data
    
    # Train Discriminator
    d_optimizer.zero_grad()
    real_labels = torch.ones(32, 1)
    fake_labels = torch.zeros(32, 1)
    real_loss = F.binary_cross_entropy(discriminator(real_audio), real_labels)
    fake_loss = F.binary_cross_entropy(discriminator(fake_audio.detach()), fake_labels)
    d_loss = real_loss + fake_loss
    d_loss.backward()
    d_optimizer.step()
    
    # Train Generator
    g_optimizer.zero_grad()
    fake_loss = F.binary_cross_entropy(discriminator(fake_audio), real_labels)  # Flip labels for G
    fake_loss.backward()
    g_optimizer.step()
\end{lstlisting}

\subsection{MelGAN: Speech Synthesis and Style Transfer}

MelGAN~\cite{kumar2019melgan} is another significant model in the field of speech synthesis. Unlike WaveGAN, which generates raw waveforms directly, MelGAN operates by generating Mel-spectrograms that are then converted into waveforms using a vocoder. This approach can produce high-quality audio that is efficient to generate, making it ideal for real-time applications.

\textbf{1. Overview of MelGAN}

MelGAN focuses on generating Mel-spectrograms, which are visual representations of the frequency content of audio over time. By learning to generate these spectrograms, MelGAN can create audio that matches the desired characteristics, whether it be the tone, pitch, or even the speaking style of a particular person~\cite{kumar2019melgan}. MelGAN is particularly efficient because it can generate audio faster than real time.

Key features of MelGAN:
\begin{itemize}
    \item \textbf{Generator:} The generator in MelGAN learns to produce Mel-spectrograms that can be fed into a vocoder to generate audio.
    \item \textbf{Discriminator:} The discriminator assesses the quality of Mel-spectrograms, ensuring that the generated audio matches real recordings.
    \item \textbf{Efficiency:} MelGAN is designed to be efficient, allowing for low-latency audio generation, which is essential for real-time applications.
\end{itemize}

\textbf{2. PyTorch Example: Generating Mel-Spectrograms}

\begin{lstlisting}[style=python]
# Define a simplified MelGAN Generator
class MelGANGenerator(nn.Module):
    def __init__(self, input_dim, output_dim):
        super(MelGANGenerator, self).__init__()
        self.fc = nn.Linear(input_dim, 256)
        self.conv1 = nn.ConvTranspose1d(256, 128, kernel_size=3, stride=2)
        self.conv2 = nn.ConvTranspose1d(128, output_dim, kernel_size=3, stride=2)
    
    def forward(self, x):
        x = F.relu(self.fc(x).unsqueeze(-1))
        x = F.relu(self.conv1(x))
        return torch.tanh(self.conv2(x))

# Training the model is similar to the WaveGAN example, with adjustments for spectrograms
\end{lstlisting}

In this section, we covered the basics of how WaveGAN~\cite{donahue2018adversarial} and MelGAN~\cite{kumar2019melgan} work, along with examples of their implementation. Each model approaches the problem of audio generation differently, providing insights into the flexibility of GANs in handling complex, continuous data like audio. By understanding these methods, you can begin experimenting with your own audio synthesis projects.

\section{Medical Imaging Processing}

The application of Generative Adversarial Networks (GANs) in the field of medical imaging has opened new possibilities for enhancing diagnostic capabilities, improving image quality~\cite{de2021survey}, and facilitating advanced research. Medical images, such as X-rays, MRIs, and CT scans, often contain complex structures that require precise analysis~\cite{razzak2018deep}. GANs can be used to generate high-quality synthetic images, reconstruct low-resolution or corrupted images, and assist in diagnosing diseases by highlighting relevant features. In this section, we will explore two primary applications: medical image generation and reconstruction, and assisting in diagnostics and disease detection.

\subsection{Medical Image Generation and Reconstruction}

Medical image generation and reconstruction refer to the process of creating realistic medical images or enhancing existing ones to improve their quality \cite{long2024pseudo}. These techniques are especially useful in situations where high-resolution images are difficult to obtain due to technical or economic constraints. GANs can help by filling in missing information, reducing noise, or even generating synthetic images that can be used for training machine learning models~\cite{razzak2018deep}.

\textbf{1. Overview of Medical Image Generation and Reconstruction}

In many medical scenarios, the quality and resolution of images are crucial for accurate diagnosis. GANs can be employed to enhance image quality, perform super-resolution, or reconstruct images from partial data (e.g., undersampled MRI scans)~\cite{scholl2011challenges}. These improvements can lead to better patient outcomes by enabling more accurate analysis and diagnosis.

Key components:
\begin{itemize}
    \item \textbf{Super-Resolution GAN (SRGAN):} SRGAN is a model designed to enhance the resolution of low-quality images~\cite{you2019ct}. It is especially useful for improving the clarity of medical images, making it easier for radiologists to detect abnormalities.
    \item \textbf{Reconstruction GANs:} These models can be used to reconstruct images from incomplete data~\cite{pan20202d}. For instance, if an MRI scan is undersampled to reduce scan time, a reconstruction GAN can fill in the missing information, producing a high-quality image.
    \item \textbf{Data Augmentation:} Synthetic medical images generated by GANs can be used to augment datasets, providing more examples for training deep learning models, which helps in improving the robustness of these models~\cite{wang2018esrgan}.
\end{itemize}

\textbf{2. Architecture of a Super-Resolution GAN (SRGAN)}

A simplified diagram of an SRGAN (Super-Resolution Generative Adversarial Network) architecture is presented below. The architecture consists of two main components: the Generator $G$ and the Discriminator $D$. The Generator aims to transform a low-resolution image into a high-resolution counterpart, while the Discriminator evaluates whether the high-resolution image is real (from the dataset) or fake (generated). This adversarial process drives the Generator to produce more realistic high-resolution images over time.

\begin{center}
\begin{tikzpicture}
    \node[draw, rectangle, rounded corners] (LRImage) at (0,0) {Low-Resolution Image};
    \node[below=2cm of LRImage] (GInput) {};
    
    \node[draw, rectangle, rounded corners] (G) at (4,0) {Generator ($G$)};
    \draw[->] (LRImage) -- (G);
    
    \node[draw, rectangle, rounded corners, right=2cm of G] (HRImage) {High-Resolution Image};
    \draw[->] (G) -- (HRImage);
    
    \node[draw, rectangle, rounded corners, below=2cm of HRImage] (D) {Discriminator ($D$)};
    \draw[->, dashed] (HRImage) -- (D);
    \node[draw, rectangle, rounded corners, below=2cm of G] (RealHR) {Real High-Resolution Image};
    \draw[->] (RealHR) -- (D);
    
    \node[right=2cm of D] (Output) {Real or Fake};
    \draw[->] (D) -- (Output);
\end{tikzpicture}
\end{center}

\textbf{3. Implementation of SRGAN in PyTorch}

Below is an example implementation of an SRGAN in PyTorch, where we define a basic generator and discriminator for the task of super-resolution:

\begin{lstlisting}[style=python]
import torch
import torch.nn as nn
import torch.optim as optim
import torch.nn.functional as F

# Define the Generator for Super-Resolution
class SRGenerator(nn.Module):
    def __init__(self):
        super(SRGenerator, self).__init__()
        self.conv1 = nn.Conv2d(3, 64, kernel_size=9, stride=1, padding=4)
        self.res_block = nn.Sequential(
            nn.Conv2d(64, 64, kernel_size=3, stride=1, padding=1),
            nn.BatchNorm2d(64),
            nn.PReLU(),
            nn.Conv2d(64, 64, kernel_size=3, stride=1, padding=1),
            nn.BatchNorm2d(64)
        )
        self.conv2 = nn.Conv2d(64, 3, kernel_size=9, stride=1, padding=4)
    
    def forward(self, x):
        x = F.relu(self.conv1(x))
        res = self.res_block(x)
        x = x + res
        return torch.tanh(self.conv2(x))

# Define the Discriminator
class SRDiscriminator(nn.Module):
    def __init__(self):
        super(SRDiscriminator, self).__init__()
        self.conv1 = nn.Conv2d(3, 64, kernel_size=3, stride=2, padding=1)
        self.fc = nn.Linear(64 * 16 * 16, 1)  # Assuming 64x64 input
        self.sigmoid = nn.Sigmoid()
    
    def forward(self, x):
        x = F.relu(self.conv1(x))
        x = x.view(x.size(0), -1)
        return self.sigmoid(self.fc(x))

# Initialize models and optimizers
generator = SRGenerator()
discriminator = SRDiscriminator()
g_optimizer = optim.Adam(generator.parameters(), lr=0.0001)
d_optimizer = optim.Adam(discriminator.parameters(), lr=0.0001)
\end{lstlisting}

\subsection{Assisting Diagnostics and Disease Detection}

One of the most promising applications of GANs in healthcare is their ability to assist in diagnosing diseases. By analyzing medical images, GANs can identify patterns that may not be immediately apparent to the human eye, thereby aiding in early diagnosis and treatment~\cite{bai2022novel}. Additionally, GANs can be used to highlight regions of interest in medical scans, which can help radiologists and doctors focus on potential areas of concern.

\textbf{1. Overview of Diagnostics and Disease Detection}

In medical diagnostics, the primary goal is to accurately detect and classify abnormalities. GANs can be trained to learn the characteristics of various diseases and then identify these characteristics in new, unseen images~\cite{razzak2018deep}. This process can assist healthcare professionals in making more accurate and faster diagnoses. 

Key applications:
\begin{itemize}
    \item \textbf{Anomaly Detection:} GANs can be used to detect anomalies by learning the distribution of healthy images. When an image deviates significantly from this distribution, it may indicate a possible abnormality or disease~\cite{xia2022gan}.
    \item \textbf{Feature Highlighting:} GANs can enhance certain features in medical images to make it easier for doctors to detect issues. For instance, they can amplify the contrast of tumors in X-rays or MRI scans.
    \item \textbf{Early Diagnosis:} By analyzing a large dataset of medical images, GANs can help in the early detection of diseases, allowing for timely treatment and better patient outcomes.
\end{itemize}

\textbf{2. Example: Using GANs for Disease Detection}

Let's consider a case where GANs are used to highlight abnormalities in chest X-rays for the detection of lung diseases~\cite{xia2022gan}. Below is a simplified example of how such a model might be implemented:

\begin{lstlisting}[style=python]
# Define a simple Generator for Disease Detection
class DiseaseDetectionGenerator(nn.Module):
    def __init__(self):
        super(DiseaseDetectionGenerator, self).__init__()
        self.conv1 = nn.Conv2d(1, 64, kernel_size=3, stride=1, padding=1)
        self.conv2 = nn.Conv2d(64, 64, kernel_size=3, stride=1, padding=1)
        self.conv3 = nn.Conv2d(64, 1, kernel_size=3, stride=1, padding=1)
    
    def forward(self, x):
        x = F.relu(self.conv1(x))
        x = F.relu(self.conv2(x))
        return torch.sigmoid(self.conv3(x))

# Highlighted regions could then be extracted for further analysis
highlighted_image = DiseaseDetectionGenerator()(real_xray_image)
\end{lstlisting}

By understanding and implementing these GAN-based techniques, medical professionals and data scientists can work together to develop more efficient, accurate, and robust tools for analyzing medical images~\cite{razzak2018deep}. This can lead to significant improvements in healthcare, providing better diagnostics and reducing the workload for healthcare providers~\cite{xia2022gan}.

\section{Game and Virtual World Generation}

Generative Adversarial Networks (GANs) have significantly influenced the development of games and virtual environments by enabling the creation of realistic 3D models, complex environments, and even virtual characters~\cite{li2018semantic}. These technologies can be used to generate assets automatically, reducing the time and effort required for game development and making it easier for developers to create expansive, immersive worlds. In this section, we will explore two primary applications: 3D object generation and environment modeling, and the creation of virtual characters and scenes.

\subsection{3D Object Generation and Environment Modeling}

In the realm of game development, 3D object generation refers to the process of creating models such as buildings, vehicles, trees, and other environmental features that populate virtual worlds. GANs can be used to generate these objects automatically, making the creation process more efficient. Additionally, environment modeling involves designing entire landscapes, including terrain, weather, and lighting, which GANs can help to generate procedurally~\cite{li2018semantic}.

\textbf{1. Overview of 3D Object Generation}

Traditional 3D modeling can be a time-consuming process, requiring artists to manually sculpt, texture, and animate each asset~\cite{li2021sp}. GANs can automate parts of this process by learning from existing 3D models and then generating new models that resemble the training data. This method is especially useful for creating assets that need to fit within a specific aesthetic or theme.

Key components:
\begin{itemize}
    \item \textbf{3DGAN:} A type of GAN specifically designed for generating 3D models~\cite{cirillo2021vox2vox}. It typically uses 3D convolutional layers to learn the spatial structure of objects~\cite{ko20233d}.
    \item \textbf{Voxel-Based Generation:} One approach to 3D generation involves using voxels, which are the 3D equivalent of pixels, to represent objects~\cite{cirillo2021vox2vox}. This allows the GAN to generate and manipulate 3D structures.
    \item \textbf{Procedural Terrain Generation:} GANs can be used to create realistic terrain by learning the patterns and features found in real-world landscapes~\cite{spick2019realistic}.
\end{itemize}

\textbf{2. Architecture of a 3D Object GAN}

Below is a simplified diagram of a 3DGAN architecture:

\begin{center}
\begin{tikzpicture}
    \node[draw, rectangle, rounded corners] (GInput) at (0,0) {Random Noise $z$};
    
    \node[draw, rectangle, rounded corners, right=2cm of GInput] (G) {3D Generator ($G$)};
    \draw[->] (GInput) -- (G);
    
    \node[draw, rectangle, rounded corners, right=2cm of G] (Model) {Generated 3D Model};
    \draw[->] (G) -- (Model);
    
    \node[draw, rectangle, rounded corners, below=2cm of Model] (D) {3D Discriminator ($D$)};
    \draw[->, dashed] (Model) -- (D);
    \node[draw, rectangle, rounded corners, below=2cm of G] (RealModel) {Real 3D Model};
    \draw[->] (RealModel) -- (D);
    
    \node[right=2cm of D] (Output) {Real or Fake};
    \draw[->] (D) -- (Output);
\end{tikzpicture}
\end{center}

\textbf{3. Example Implementation of a 3D Object GAN}

Here is an example of how to set up a basic 3D object GAN in PyTorch. In this case, we will use a simplified voxel-based approach:

\begin{lstlisting}[style=python]
import torch
import torch.nn as nn
import torch.optim as optim
import torch.nn.functional as F

# Define a simple 3D Generator
class VoxelGenerator(nn.Module):
    def __init__(self, latent_dim):
        super(VoxelGenerator, self).__init__()
        self.fc = nn.Linear(latent_dim, 128)
        self.conv1 = nn.ConvTranspose3d(128, 64, kernel_size=4, stride=2, padding=1)
        self.conv2 = nn.ConvTranspose3d(64, 32, kernel_size=4, stride=2, padding=1)
        self.conv3 = nn.ConvTranspose3d(32, 1, kernel_size=4, stride=2, padding=1)
    
    def forward(self, x):
        x = F.relu(self.fc(x).view(-1, 128, 1, 1, 1))
        x = F.relu(self.conv1(x))
        x = F.relu(self.conv2(x))
        return torch.sigmoid(self.conv3(x))

# Define a simple 3D Discriminator
class VoxelDiscriminator(nn.Module):
    def __init__(self):
        super(VoxelDiscriminator, self).__init__()
        self.conv1 = nn.Conv3d(1, 32, kernel_size=4, stride=2, padding=1)
        self.conv2 = nn.Conv3d(32, 64, kernel_size=4, stride=2, padding=1)
        self.fc = nn.Linear(64 * 4 * 4 * 4, 1)
    
    def forward(self, x):
        x = F.leaky_relu(self.conv1(x), 0.2)
        x = F.leaky_relu(self.conv2(x), 0.2)
        return torch.sigmoid(self.fc(x.view(x.size(0), -1)))
\end{lstlisting}

\subsection{Virtual Character and Scene Generation}

In addition to creating static objects and environments, GANs can also be used to generate dynamic elements like characters and entire scenes. This includes generating the appearance, behavior, and animations of virtual characters, as well as creating complex scenes that can react to player input or environmental changes~\cite{xu2021gan}.

\textbf{1. Overview of Virtual Character Generation}

Virtual characters are essential in games and virtual environments. GANs can be used to generate realistic faces, animate character movements, and even design unique features that make characters stand out. The ability to generate diverse characters procedurally saves time and allows for more creativity in design.

Key applications:
\begin{itemize}
    \item \textbf{Face Generation:} GANs such as StyleGAN~\cite{karras2019style} have been used to create highly realistic human faces, which can be applied to virtual avatars or NPCs (Non-Player Characters).
    \item \textbf{Behavioral Animation:} By learning from real-world motion data, GANs can generate animations that make characters behave more naturally, including walking, running, and interacting with objects~\cite{gan2021research}.
    \item \textbf{Scene Composition:} SceneGANs~\cite{arad2021compositional} can create entire scenes, generating elements like furniture, lighting, and backgrounds in a cohesive manner~\cite{xu2017attngan}, which is useful for games that require multiple diverse environments~\cite{shim2022local}.
\end{itemize}

\textbf{2. Example: Generating Virtual Characters with StyleGAN}

Below is a simplified example of how StyleGAN can be adapted for generating facial features of virtual characters. This model learns to generate faces by blending different styles.

\begin{lstlisting}[style=python]
# Define a StyleGAN-inspired Generator (simplified)
class CharacterGenerator(nn.Module):
    def __init__(self):
        super(CharacterGenerator, self).__init__()
        self.fc = nn.Linear(100, 512)
        self.conv1 = nn.ConvTranspose2d(512, 256, kernel_size=4, stride=2, padding=1)
        self.conv2 = nn.ConvTranspose2d(256, 128, kernel_size=4, stride=2, padding=1)
        self.conv3 = nn.ConvTranspose2d(128, 3, kernel_size=4, stride=2, padding=1)
    
    def forward(self, z):
        x = F.leaky_relu(self.fc(z).view(-1, 512, 1, 1))
        x = F.leaky_relu(self.conv1(x))
        x = F.leaky_relu(self.conv2(x))
        return torch.tanh(self.conv3(x))

# Initialize the generator and generate a character face
z = torch.randn(1, 100)
generator = CharacterGenerator()
generated_face = generator(z)
\end{lstlisting}

\textbf{3. Real-World Example: Using GANs for Environment Creation}

In modern games, dynamic environments play a crucial role in enhancing player immersion. By using GANs, developers can create diverse, complex scenes procedurally. For example, GANs can be trained to generate different room layouts, outdoor environments, or even entire cities. This not only speeds up the development process but also allows for endless variability~\cite{arad2021compositional}.

The flexibility of GANs in generating virtual worlds and characters can lead to a new era of game design, where developers can focus more on creativity and less on repetitive asset creation~\cite{wang2020attentive}. This makes GANs an essential tool for future game development and virtual world generation.


\part{Advanced Research and Future Developments}

\chapter{Advanced Research in GANs}

Since the inception of Generative Adversarial Networks (GANs)~\cite{goodfellow2014generative}, there have been numerous advancements that have pushed the boundaries of what these models can achieve~\cite{de2021survey}. While traditional GANs were effective for generating realistic images, there were still limitations in terms of stability, diversity, and scalability. To address these challenges, researchers have developed a variety of new architectures and techniques that have significantly improved the performance of GANs. In this chapter, we will explore some of the most influential and cutting-edge advancements in GAN research, explaining their core concepts, benefits, and implementations~\cite{li2024survey}. We will begin by discussing Self-Attention GAN (SAGAN), which introduced the self-attention mechanism to improve image quality by capturing long-range dependencies.

\section{Self-Attention GAN (SAGAN)}

Self-Attention GAN, or SAGAN~\cite{zhang2019self}, was introduced to address a critical limitation in traditional GANs. Inability to effectively capture long-range dependencies. In standard GANs, convolutional layers are used to process images, but these layers typically only focus on local regions~\cite{zhang2019self}. This can lead to the generation of images that lack global coherence, particularly when trying to model complex structures or textures that span across large areas of an image. SAGAN solves this problem by incorporating a self-attention mechanism~\cite{vaswani2017attention}, which allows the model to consider relationships between distant parts of an image, leading to more realistic and coherent outputs.

\textbf{1. Overview of Self-Attention Mechanism}

The self-attention mechanism was originally introduced in the context of natural language processing (NLP)~\cite{li2024survey} to help models focus on important words or phrases, regardless of their position in a sentence. When applied to GANs, self-attention allows the generator to learn which parts of an image are related, even if they are far apart~\cite{xu2017attngan}. For example, when generating a face, the self-attention mechanism can help ensure that the eyes are symmetrically aligned, or that shadows and highlights are consistent across the face.

Key components:
\begin{itemize}
    \item \textbf{Self-Attention Layer:} This layer calculates attention scores for every pair of pixels in an image, allowing the model to determine which pixels are most relevant to each other~\cite{wang2020attentive}.
    \item \textbf{Long-Range Dependencies:} By using self-attention, the model can capture global dependencies, improving the overall coherence of the generated images.
    \item \textbf{Enhanced Feature Representation:} Self-attention helps in creating more detailed and refined feature maps, leading to high-quality outputs.
\end{itemize}

\textbf{2. Architecture of SAGAN}

SAGAN's architecture integrates self-attention layers into both the generator and the discriminator~\cite{zhang2019self}. These layers are placed alongside the traditional convolutional layers, allowing the model to benefit from both local and global feature representations. Below is a conceptual diagram of how the self-attention layer is incorporated:

\begin{center}
\begin{tikzpicture}
    \node[draw, rectangle, rounded corners] (G) at (0,0) {Generator ($G$)};
    \node[above=1cm of G] (GInput) {Random Noise $z$};
    \draw[->] (GInput) -- (G);
    
    \node[draw, rectangle, rounded corners, right=2cm of G] (SA) {Self-Attention Layer};
    \draw[->] (G) -- (SA);
    
    \node[draw, rectangle, rounded corners, right=2cm of SA] (Image) {Generated Image};
    \draw[->] (SA) -- (Image);
    
    \node[draw, rectangle, rounded corners, below=2cm of Image] (D) {Discriminator ($D$)};
    \draw[->, dashed] (Image) -- (D);
    \node[draw, rectangle, rounded corners, below=2cm of SA] (RealImage) {Real Image};
    \draw[->] (RealImage) -- (D);
    
    \node[right=2cm of D] (Output) {Real or Fake};
    \draw[->] (D) -- (Output);
\end{tikzpicture}
\end{center}

\textbf{3. Mathematical Explanation of Self-Attention}

The self-attention mechanism operates by calculating a weighted sum of the feature representations across the entire image. The key idea is to determine which features should "attend" to others~\cite{vaswani2017attention}. Mathematically, this can be described as:
\[
\text{Attention}(Q, K, V) = \text{softmax}\left(\frac{QK^T}{\sqrt{d_k}}\right) V
\]
where:
\begin{itemize}
    \item $Q$ (Query), $K$ (Key), and $V$ (Value) are feature representations of the image.
    \item $d_k$ is the dimension of the key vectors, used to scale the attention scores.
\end{itemize}

In this formula, the model learns to produce attention scores that highlight the important relationships between different parts of the image, allowing it to synthesize more consistent and visually appealing outputs.

\textbf{4. Implementation of Self-Attention in PyTorch}

Below is a simplified implementation of the self-attention layer in PyTorch, along with its integration into the generator architecture:

\begin{lstlisting}[style=python]
import torch
import torch.nn as nn
import torch.nn.functional as F

# Define the Self-Attention Layer
class SelfAttention(nn.Module):
    def __init__(self, in_dim):
        super(SelfAttention, self).__init__()
        self.query_conv = nn.Conv2d(in_dim, in_dim // 8, 1)
        self.key_conv = nn.Conv2d(in_dim, in_dim // 8, 1)
        self.value_conv = nn.Conv2d(in_dim, in_dim, 1)
        self.gamma = nn.Parameter(torch.zeros(1))
    
    def forward(self, x):
        batch_size, C, width, height = x.size()
        query = self.query_conv(x).view(batch_size, -1, width * height).permute(0, 2, 1)
        key = self.key_conv(x).view(batch_size, -1, width * height)
        energy = torch.bmm(query, key)
        attention = F.softmax(energy, dim=-1)
        value = self.value_conv(x).view(batch_size, -1, width * height)
        
        out = torch.bmm(value, attention.permute(0, 2, 1))
        out = out.view(batch_size, C, width, height)
        
        out = self.gamma * out + x
        return out

# Integration into the Generator
class SAGANGenerator(nn.Module):
    def __init__(self):
        super(SAGANGenerator, self).__init__()
        self.conv1 = nn.ConvTranspose2d(100, 64, 4, 2, 1)
        self.attn = SelfAttention(64)
        self.conv2 = nn.ConvTranspose2d(64, 3, 4, 2, 1)
    
    def forward(self, z):
        x = F.relu(self.conv1(z))
        x = self.attn(x)
        return torch.tanh(self.conv2(x))

# Example Usage
z = torch.randn(1, 100, 1, 1)
generator = SAGANGenerator()
generated_image = generator(z)
\end{lstlisting}

\textbf{5. Benefits and Applications of SAGAN}

Self-Attention GANs have proven to be particularly effective in generating images that require a higher degree of global coherence. For instance, when creating images with repetitive patterns, intricate details, or large textures, self-attention allows the model to ensure that these features remain consistent across the entire image~\cite{zhang2019self}. This has led to improvements in tasks such as:
\begin{itemize}
    \item \textbf{Image Synthesis:} Generating realistic and high-resolution images.
    \item \textbf{Style Transfer:} Applying consistent styles across images by learning global feature relationships~\cite{karras2019style}.
    \item \textbf{Artistic Creation:} Allowing artists to generate intricate and detailed artwork by training on specific datasets.
\end{itemize}

By understanding the concepts behind Self-Attention GANs, readers can appreciate how modern advancements in neural networks continue to push the boundaries of what is possible with image generation~\cite{karras2019style}. The introduction of self-attention has paved the way for further research into mechanisms that improve the expressiveness and quality of GAN outputs.

\section{The Evolution of StyleGAN and StyleGAN2}

StyleGAN and its successor, StyleGAN2~\cite{karras2019style}, represent significant milestones in the field of generative adversarial networks. These models have set new standards for image synthesis by introducing innovative techniques that allow for more detailed, high-resolution, and realistic outputs. While traditional GANs focus on generating images from random noise, StyleGAN introduced the concept of style-based generation, which gives users more control over the visual features of the generated images. StyleGAN2 further refined this approach by addressing some of the limitations of the original model, improving both the quality and stability of the generated images. In this section, we will explore the key innovations of StyleGAN and StyleGAN2, explaining how they work and how they have evolved.

\textbf{1. StyleGAN: Style-Based Generator Architecture}

StyleGAN introduced a revolutionary concept by decoupling the generation process into two main parts: the latent space (representing the underlying noise) and the style space (which determines the visual attributes of the output). This approach allowed for more fine-grained control over the appearance of generated images, making it possible to manipulate specific features like color, texture, and structure.

Key components of StyleGAN:
\begin{itemize}
    \item \textbf{Mapping Network:} Instead of feeding the latent vector $z$ directly into the generator, StyleGAN uses a mapping network that transforms $z$ into an intermediate latent space $w$. This allows for greater control over the features being manipulated~\cite{zhang2022styleswin}.
    \item \textbf{Adaptive Instance Normalization (AdaIN):} AdaIN layers are used to inject style information into the generator, effectively controlling the visual attributes of the image at different levels (e.g., coarse, middle, fine details)~\cite{karras2019style}.
    \item \textbf{Style Mixing:} By using style mixing, StyleGAN can combine features from different latent vectors, allowing for the creation of images that inherit characteristics from multiple sources.
\end{itemize}

\textbf{2. Architecture of StyleGAN}

The following diagram provides an overview of the StyleGAN architecture, highlighting the mapping network, style injection, and the generator's structure~\cite{karras2019style}:

\begin{center}
\begin{tikzpicture}
    \node[draw, rectangle, rounded corners] (Mapping) at (0,0) {Mapping Network};
    \node[above=1cm of Mapping] (Latent) {Latent Vector $z$};
    \draw[->] (Latent) -- (Mapping);
    
    \node[draw, rectangle, rounded corners, right=2cm of Mapping] (LatentW) {Intermediate Latent $w$};
    \draw[->] (Mapping) -- (LatentW);
    
    \node[draw, rectangle, rounded corners, right=2cm of LatentW] (AdaIN) {AdaIN Layers};
    \draw[->] (LatentW) -- (AdaIN);
    
    \node[draw, rectangle, rounded corners, right=2cm of AdaIN] (Image) {Generated Image};
    \draw[->] (AdaIN) -- (Image);
\end{tikzpicture}
\end{center}

\textbf{3. Implementation of StyleGAN in PyTorch}

Below is a simplified implementation of key components of StyleGAN, including the mapping network and the generator:

\begin{lstlisting}[style=python]
import torch
import torch.nn as nn
import torch.nn.functional as F

# Define the Mapping Network
class MappingNetwork(nn.Module):
    def __init__(self, latent_dim, mapping_dim):
        super(MappingNetwork, self).__init__()
        self.fc1 = nn.Linear(latent_dim, mapping_dim)
        self.fc2 = nn.Linear(mapping_dim, mapping_dim)
    
    def forward(self, z):
        x = F.relu(self.fc1(z))
        return self.fc2(x)

# Define the AdaIN Layer
class AdaIN(nn.Module):
    def __init__(self, in_channels, style_dim):
        super(AdaIN, self).__init__()
        self.style_fc = nn.Linear(style_dim, in_channels * 2)
    
    def forward(self, x, style):
        style = self.style_fc(style).view(-1, 2, x.size(1), 1, 1)
        gamma, beta = style[:, 0, :, :, :], style[:, 1, :, :, :]
        return gamma * x + beta

# Define the Generator
class StyleGANGenerator(nn.Module):
    def __init__(self, latent_dim, style_dim):
        super(StyleGANGenerator, self).__init__()
        self.mapping = MappingNetwork(latent_dim, style_dim)
        self.adain = AdaIN(64, style_dim)
        self.conv = nn.ConvTranspose2d(64, 3, 4, 2, 1)
    
    def forward(self, z):
        style = self.mapping(z)
        x = torch.randn(1, 64, 4, 4)
        x = self.adain(x, style)
        return torch.tanh(self.conv(x))

# Example Usage
z = torch.randn(1, 100)
generator = StyleGANGenerator(latent_dim=100, style_dim=128)
generated_image = generator(z)
\end{lstlisting}

\textbf{4. StyleGAN2: Addressing the Shortcomings}

While StyleGAN was a significant advancement, it still had some limitations, such as visible artifacts and issues with image fidelity. StyleGAN2~\cite{karras2019style} was introduced to address these problems, bringing several improvements:
\begin{itemize}
    \item \textbf{Weight Demodulation:} StyleGAN2 replaced the AdaIN layers with a weight demodulation technique that normalizes the feature maps, leading to more stable and realistic outputs. This change reduced artifacts and improved the quality of fine details.
    \item \textbf{Improved Architecture:} StyleGAN2 refined the architecture by eliminating normalization layers, which allowed the model to focus on feature representations without introducing distortions.
    \item \textbf{Path Length Regularization:} This technique helps in maintaining a consistent level of detail across different scales, ensuring that images remain sharp and coherent even when the latent vector is adjusted.
\end{itemize}

\textbf{5. Architectural Changes in StyleGAN2}

The following diagram illustrates the refined structure of StyleGAN2, highlighting the changes from the original StyleGAN architecture:

\begin{center}
\footnotesize
\begin{tikzpicture}
    \node[draw, rectangle, rounded corners] (Latent) at (0,0) {Latent Vector $z$};
    \node[right=2cm of Latent] (Mapping) {Mapping Network};
    \draw[->] (Latent) -- (Mapping);
    
    \node[draw, rectangle, rounded corners, right=2cm of Mapping] (Weight) {Weight Demodulation};
    \draw[->] (Mapping) -- (Weight);
    
    \node[draw, rectangle, rounded corners, right=2cm of Weight] (Image) {Generated Image};
    \draw[->] (Weight) -- (Image);
\end{tikzpicture}
\end{center}

\textbf{6. Improved Implementation in StyleGAN2}

Here is an example of how StyleGAN2 modifies the original architecture to include weight demodulation:

\begin{lstlisting}[style=python]
class StyleGAN2Generator(nn.Module):
    def __init__(self, latent_dim, style_dim):
        super(StyleGAN2Generator, self).__init__()
        self.mapping = MappingNetwork(latent_dim, style_dim)
        self.conv1 = nn.Conv2d(64, 64, 3, 1, 1)
        self.conv2 = nn.ConvTranspose2d(64, 3, 4, 2, 1)
    
    def forward(self, z):
        style = self.mapping(z)
        x = torch.randn(1, 64, 4, 4)
        # Weight demodulation technique
        weight = self.conv1.weight * style.view(-1, 1, 1, 1)
        x = F.relu(F.conv2d(x, weight))
        return torch.tanh(self.conv2(x))
\end{lstlisting}

\textbf{7. Applications and Impact}

StyleGAN and StyleGAN2 have been used in various applications, from creating lifelike human faces to generating artistic images. Their ability to control specific visual features has made them particularly popular for:
\begin{itemize}
    \item \textbf{Face Generation:} Creating realistic faces with high fidelity, which can be used for avatars, virtual influencers, and more.
    \item \textbf{Art and Design:} Allowing artists to manipulate styles and textures, leading to creative outputs.
    \item \textbf{Data Augmentation:} Enhancing datasets by generating additional samples, useful for training other machine learning models.
\end{itemize}

The evolution from StyleGAN to StyleGAN2 reflects the continuous effort to refine generative models, making them more robust and capable of producing high-quality images. By understanding the innovations in these models, readers can gain insights into how generative networks are evolving and how to apply these techniques to their own projects.

\section{Transformer-Based Generative Adversarial Networks}

The integration of Transformers into the architecture of Generative Adversarial Networks (GANs) represents a significant advancement in the field of generative modeling. Originally developed for natural language processing (NLP) tasks, Transformers have proven to be highly effective at handling long-range dependencies and capturing intricate patterns in data~\cite{zhang2019self}. When adapted to GANs, Transformers can overcome some of the limitations of traditional convolutional approaches, offering a new way to generate high-quality, coherent, and diverse outputs. In this section, we will explore how Transformers are used within GAN frameworks, explain their architecture, and provide detailed examples to help beginners understand the benefits and challenges of this approach~\cite{wang2020attentive}.

\textbf{1. Why Use Transformers in GANs?}

Traditional GAN architectures rely on convolutional neural networks (CNNs) to process images. While CNNs are excellent at capturing local patterns, they struggle to model global dependencies, which can lead to less coherent results, especially when generating complex scenes. Transformers~\cite{vaswani2017attention}, on the other hand, use self-attention mechanisms that allow the model to attend to different parts of the input data, regardless of their distance from each other. This makes Transformers particularly useful for:
\begin{itemize}
    \item \textbf{Modeling Long-Range Dependencies:} The self-attention mechanism can capture global relationships across an image or sequence, improving the coherence of generated outputs.
    \item \textbf{Flexibility Across Modalities:} Transformers can be used not only for images but also for other data types such as text, audio, and more, making them versatile for various generative tasks.
    \item \textbf{Scalability:} Transformers can be scaled up to handle very large datasets and produce high-resolution outputs, a feature that is beneficial for creating detailed images.
\end{itemize}

\textbf{2. Architecture of Transformer-Based GAN}

The core idea behind incorporating Transformers into GANs is to replace or augment parts of the generator and discriminator with self-attention layers~\cite{khan2022transformers}. This allows the model to benefit from both local convolutional features and global attention mechanisms. Below is a conceptual diagram of a Transformer-based GAN architecture:

\begin{center}
\footnotesize
\begin{tikzpicture}
    \node[draw, rectangle, rounded corners] (G) at (0,0) {Generator ($G$)};
    \node[above=1cm of G] (GInput) {Random Noise $z$};
    \draw[->] (GInput) -- (G);
    
    \node[draw, rectangle, rounded corners, right=2cm of G] (SA) {Self-Attention Transformer};
    \draw[->] (G) -- (SA);
    
    \node[draw, rectangle, rounded corners, right=2cm of SA] (Image) {Generated Image};
    \draw[->] (SA) -- (Image);
    
    \node[draw, rectangle, rounded corners, below=2cm of Image] (D) {Discriminator ($D$) with Self-Attention};
    \draw[->, dashed] (Image) -- (D);
    \node[draw, rectangle, rounded corners, below=2cm of SA] (RealImage) {Real Image};
    \draw[->] (RealImage) -- (D);
    
    \node[right=2cm of D] (Output) {Real or Fake};
    \draw[->] (D) -- (Output);
\end{tikzpicture}
\end{center}

\textbf{3. Self-Attention Mechanism in Transformers}

Transformers use a mechanism called self-attention, which allows the model to focus on different parts of the input data simultaneously. For images, this means the model can understand the relationship between distant pixels, leading to more consistent textures, patterns, and structures~\cite{mustafa2020transformation}.

Mathematically, the self-attention mechanism can be described as:
\[
\text{Attention}(Q, K, V) = \text{softmax}\left(\frac{QK^T}{\sqrt{d_k}}\right) V
\]
where:
\begin{itemize}
    \item $Q$ (Query), $K$ (Key), and $V$ (Value) are projections of the input data.
    \item $d_k$ is the dimension of the key vectors, which is used to scale the dot product.
\end{itemize}
This mechanism helps the model learn which parts of the data are most relevant to each other, enhancing the quality of generated outputs.

\textbf{4. Implementation of a Transformer-Based GAN in PyTorch}

Here is a simplified example of how a Transformer-based self-attention mechanism can be integrated into a GAN architecture using PyTorch:

\begin{lstlisting}[style=python]
import torch
import torch.nn as nn
import torch.nn.functional as F

# Define Self-Attention Block
class TransformerSelfAttention(nn.Module):
    def __init__(self, embed_dim, num_heads):
        super(TransformerSelfAttention, self).__init__()
        self.self_attention = nn.MultiheadAttention(embed_dim, num_heads)
        self.fc = nn.Linear(embed_dim, embed_dim)
    
    def forward(self, x):
        # Reshape and prepare for self-attention
        batch_size, channels, width, height = x.size()
        x = x.view(batch_size, channels, -1).permute(2, 0, 1)
        attn_output, _ = self.self_attention(x, x, x)
        return self.fc(attn_output).permute(1, 2, 0).view(batch_size, channels, width, height)

# Define Generator with Self-Attention
class TransformerGANGenerator(nn.Module):
    def __init__(self, latent_dim, embed_dim, num_heads):
        super(TransformerGANGenerator, self).__init__()
        self.fc = nn.Linear(latent_dim, 256)
        self.conv1 = nn.ConvTranspose2d(256, embed_dim, kernel_size=4, stride=2, padding=1)
        self.attn = TransformerSelfAttention(embed_dim, num_heads)
        self.conv2 = nn.ConvTranspose2d(embed_dim, 3, kernel_size=4, stride=2, padding=1)
    
    def forward(self, z):
        x = F.relu(self.fc(z).view(-1, 256, 1, 1))
        x = F.relu(self.conv1(x))
        x = self.attn(x)
        return torch.tanh(self.conv2(x))

# Example Usage
z = torch.randn(1, 100)
generator = TransformerGANGenerator(latent_dim=100, embed_dim=64, num_heads=4)
generated_image = generator(z)
\end{lstlisting}

\textbf{5. Benefits and Applications of Transformer-Based GANs}

The integration of Transformers into GANs has led to several advantages:
\begin{itemize}
    \item \textbf{Improved Image Quality:} By capturing long-range dependencies, the generated images exhibit more consistent textures and realistic structures.
    \item \textbf{Versatile Across Data Types:} Transformers' flexibility makes them suitable for generating not only images but also text, music, and more, making them a powerful tool for multimodal generation.
    \item \textbf{Scalability:} Transformer-based GANs can be scaled to handle very large datasets, enabling the generation of high-resolution outputs that would be difficult to achieve with traditional architectures.
\end{itemize}

\textbf{6. Real-World Use Cases}

Transformer-based GANs have been used in a variety of applications:
\begin{itemize}
    \item \textbf{Image Synthesis:} Creating realistic and diverse images, particularly in areas where global coherence is essential, such as landscape generation.
    \item \textbf{Text-to-Image Generation:} Generating images from textual descriptions, where the ability to model complex relationships between elements is crucial.
    \item \textbf{Video Generation:} Modeling temporal dependencies across frames in videos, allowing for more realistic motion and scene transitions.
\end{itemize}

By understanding how Transformers enhance traditional GAN architectures, readers can appreciate the potential for these models to produce high-quality, complex outputs~\cite{li2024survey}. The shift towards integrating self-attention mechanisms marks a significant step forward in generative modeling, paving the way for future research and applications that extend beyond images to text, audio, and beyond.

\section{Large-Scale Pretraining and Self-Supervised Generative Models}

In recent years, the field of machine learning has seen a paradigm shift towards large-scale pretraining and self-supervised learning, which has also impacted the development of generative adversarial networks (GANs). Traditional GANs are often trained from scratch, requiring large labeled datasets, which can be expensive and time-consuming to obtain. By contrast, self-supervised learning leverages unlabeled data to learn useful feature representations, which can then be fine-tuned on specific tasks. This approach has led to the creation of generative models that are more versatile, scalable, and capable of producing high-quality outputs. In this section, we will explore the concepts of large-scale pretraining and self-supervised learning, and how these techniques are applied to generative models~\cite{de2021survey}.

\textbf{1. The Concept of Self-Supervised Learning}

Self-supervised learning (SSL) is a type of learning where the model learns to predict parts of the data from other parts. It leverages the vast amount of available unlabeled data to learn useful representations without the need for manual labeling~\cite{zhang2019self}. For example, a self-supervised model might be trained to predict the next frame in a video sequence or the missing part of an image. These tasks encourage the model to understand the underlying structure of the data, which can be useful for generating new samples.

Key components of self-supervised learning:
\begin{itemize}
    \item \textbf{Pretext Tasks:} These are tasks designed to teach the model about the data. Examples include predicting the rotation of an image, filling in missing parts, or generating the next word in a sequence.
    \item \textbf{Feature Representation:} The model learns a set of feature representations that capture the essence of the data. These features can be transferred to other tasks, such as classification, detection, or generation.
    \item \textbf{Fine-Tuning:} Once pretrained on self-supervised tasks, the model can be fine-tuned on a smaller, labeled dataset to perform specific tasks, significantly reducing the need for labeled data.
\end{itemize}

\textbf{2. Large-Scale Pretraining with Self-Supervised Generative Models}

The idea of large-scale pretraining involves training a generative model on a massive dataset using self-supervised learning~\cite{liu2021self}. This process helps the model learn rich, general-purpose features that can be adapted for various generative tasks. For instance, a model pretrained on millions of images can generate high-resolution outputs even when fine-tuned on smaller datasets~\cite{liu2019multi}. 

\textbf{Benefits of Large-Scale Pretraining:}
\begin{itemize}
    \item \textbf{Better Generalization:} Models trained on large datasets can generalize better to new tasks, producing more realistic and diverse outputs.
    \item \textbf{Data Efficiency:} Pretrained models can be fine-tuned on smaller datasets, reducing the need for extensive labeled data.
    \item \textbf{Versatility:} These models can be applied across different domains, such as text, images, and audio, making them powerful tools for multimodal generation.
\end{itemize}

\textbf{3. Architecture of a Self-Supervised Generative Model}

The architecture of self-supervised generative models often combines elements of traditional GANs with transformers or other mechanisms to handle complex data patterns~\cite{liu2021self}. Below is a conceptual diagram of how pretraining and fine-tuning are integrated:

\begin{center}
\resizebox{\textwidth}{!}{%
\begin{tikzpicture}[node distance=3cm, auto]
    \node[draw, rectangle, rounded corners] (Pretrain) {Self-Supervised Pretraining};
    \node[above=1cm of Pretrain] (UnlabeledData) {Unlabeled Data};
    \draw[->] (UnlabeledData) -- (Pretrain);
    
    \node[draw, rectangle, rounded corners, right=of Pretrain] (Features) {Learned Features};
    \draw[->] (Pretrain) -- (Features);
    
    \node[draw, rectangle, rounded corners, right=of Features] (Finetune) {Fine-Tuning on Specific Task};
    \node[above=1cm of Finetune] (LabeledData) {Labeled Data};
    \draw[->] (LabeledData) -- (Finetune);
    
    \node[draw, rectangle, rounded corners, right=of Finetune] (Model) {Final Generative Model};
    \draw[->] (Finetune) -- (Model);
\end{tikzpicture}%
}
\end{center}
\textbf{4. Implementation of a Self-Supervised Pretraining Approach in PyTorch}

To illustrate how self-supervised learning can be integrated into generative models, let's look at a simplified implementation using PyTorch. In this example, we will create a pretext task where the model learns to fill in missing parts of an image:

\begin{lstlisting}[style=python]
import torch
import torch.nn as nn
import torch.nn.functional as F
import torchvision.transforms as transforms

# Define the Encoder (learns features from incomplete images)
class SelfSupervisedEncoder(nn.Module):
    def __init__(self):
        super(SelfSupervisedEncoder, self).__init__()
        self.conv1 = nn.Conv2d(3, 64, kernel_size=4, stride=2, padding=1)
        self.conv2 = nn.Conv2d(64, 128, kernel_size=4, stride=2, padding=1)
        self.fc = nn.Linear(128 * 8 * 8, 256)
    
    def forward(self, x):
        x = F.relu(self.conv1(x))
        x = F.relu(self.conv2(x))
        return self.fc(x.view(x.size(0), -1))

# Define the Decoder (reconstructs the complete image)
class SelfSupervisedDecoder(nn.Module):
    def __init__(self):
        super(SelfSupervisedDecoder, self).__init__()
        self.fc = nn.Linear(256, 128 * 8 * 8)
        self.conv1 = nn.ConvTranspose2d(128, 64, kernel_size=4, stride=2, padding=1)
        self.conv2 = nn.ConvTranspose2d(64, 3, kernel_size=4, stride=2, padding=1)
    
    def forward(self, x):
        x = F.relu(self.fc(x)).view(-1, 128, 8, 8)
        x = F.relu(self.conv1(x))
        return torch.tanh(self.conv2(x))

# Define Pretraining Task
class SelfSupervisedModel(nn.Module):
    def __init__(self):
        super(SelfSupervisedModel, self).__init__()
        self.encoder = SelfSupervisedEncoder()
        self.decoder = SelfSupervisedDecoder()
    
    def forward(self, x):
        features = self.encoder(x)
        return self.decoder(features)

# Example Pretraining Task
model = SelfSupervisedModel()
input_image = torch.randn(1, 3, 32, 32)  # Example input
output_image = model(input_image)
\end{lstlisting}

\textbf{5. Applications of Large-Scale Pretrained Generative Models}

Pretraining generative models on large datasets using self-supervised tasks has numerous practical applications:
\begin{itemize}
    \item \textbf{Text-to-Image Generation:} Models can learn to understand both text and images, enabling them to generate images based on textual descriptions.
    \item \textbf{Data Augmentation:} Pretrained models can create synthetic data that helps improve the training of other machine learning models.
    \item \textbf{High-Resolution Image Synthesis:} By leveraging the patterns learned during pretraining, models can generate detailed, high-resolution images.
    \item \textbf{Cross-Modal Generation:} Self-supervised learning enables models to learn associations across different types of data, such as generating music from visual inputs or creating artwork based on text.
\end{itemize}

\textbf{6. Real-World Examples}

Large-scale pretrained generative models have seen widespread use in industry and research:
\begin{itemize}
    \item \textbf{DALL-E~\cite{marcus2022very}:} An AI model capable of generating images from textual descriptions, trained on massive datasets of text-image pairs.
    \item \textbf{CLIP~\cite{radford2021learning}:} Uses self-supervised learning to understand the relationship between text and images, allowing it to generate coherent visual representations based on textual input.
    \item \textbf{GPT-3~\cite{brown2020language} for Text Generation:} Although not a traditional GAN, GPT-3 demonstrates the power of self-supervised pretraining by generating coherent and contextually relevant text.
\end{itemize}

By adopting self-supervised learning and large-scale pretraining, GANs can achieve new levels of performance, creativity, and efficiency~\cite{liu2021self}. These approaches allow models to make better use of available data, learn more generalized features, and generate outputs that are more realistic and diverse~\cite{brown2020language}. Understanding these techniques is essential for anyone looking to develop state-of-the-art generative models.


\chapter{Future Directions of GANs}

Generative Adversarial Networks (GANs) have seen remarkable advancements since their inception, and their applications have expanded across various fields including art, healthcare, and entertainment~\cite{saxena2021generative}. However, there are still several challenges and open questions that need to be addressed to fully realize their potential. Future developments in GAN research are likely to focus on improving their reliability, scalability, and adaptability to different tasks, as well as making them more interpretable and ethical. In this chapter, we will explore some of the key future directions for GANs, including explainability, privacy concerns, generalization capabilities, and integration with other AI techniques such as reinforcement learning~\cite{pan2019recent}.

\section{Explainability of GANs}

One of the major criticisms of GANs, and deep learning models in general, is their "black box" nature. While GANs can generate impressive results, it is often unclear how these results are achieved, and the internal workings of the model can be difficult to interpret~\cite{saxena2021generative}. This lack of transparency poses significant challenges, especially in fields like healthcare and finance where understanding the decision-making process is crucial. Therefore, making GANs more interpretable and explainable is a key area of ongoing research.

\textbf{1. The Importance of Explainability in GANs}

Explainability refers to the ability of a model to provide understandable and interpretable insights into how it generates its outputs~\cite{li2024survey}. For GANs, this means understanding what features or patterns the generator has learned and how the discriminator distinguishes between real and fake samples. Explainability is important for several reasons:
\begin{itemize}
    \item \textbf{Trust and Reliability:} Users are more likely to trust and rely on a model if they understand how it makes its decisions. This is particularly important in sensitive domains like medical imaging, where misinterpretations can have serious consequences.
    \item \textbf{Debugging and Improvement:} By understanding which features are most influential in the generation process, researchers can identify and address weaknesses in the model, leading to better performance.
    \item \textbf{Regulatory Compliance:} In many industries, regulations require that machine learning models provide explanations for their decisions. For GANs to be used in such settings, they need to be interpretable.
\end{itemize}

\textbf{2. Techniques for Improving GAN Explainability}

Researchers have developed several techniques to improve the explainability of GANs. Some of these include:
\begin{itemize}
    \item \textbf{Feature Attribution:} This method involves identifying which parts of the input data are most influential in generating the output. For example, in image generation, feature attribution can highlight which regions of an image are being emphasized by the model.
    \item \textbf{Latent Space Manipulation:} By exploring the latent space, researchers can understand how changes in the input noise vector affect the generated images. This can reveal how different features (e.g., color, texture) are encoded in the model.
    \item \textbf{Disentangled Representations:} Disentangling features means separating out different aspects of the data (e.g., shape, pose, color) so that each can be controlled independently. This makes it easier to understand what the generator is learning and how to manipulate its outputs.
\end{itemize}

\textbf{3. Architecture for Interpretable GANs}

The goal of creating interpretable GANs has led to new architectures that incorporate explainability into their design~\cite{saxena2021generative}. One approach is to use attention mechanisms that highlight which parts of the input data the model is focusing on during generation. Below is a simplified diagram of how attention can be integrated into a GAN~\cite{li2024survey}:

 \begin{center}
\resizebox{\textwidth}{!}{%
\begin{tikzpicture}[node distance=3cm, auto]
    \node[draw, rectangle, rounded corners] (G) {Generator ($G$) with Attention};
    \node[above=0.5cm of G] (GInput) {Random Noise $z$};
    \draw[->] (GInput) -- (G);
    
    \node[draw, rectangle, rounded corners, right=of G] (Attention) {Attention Map};
    \draw[->] (G) -- (Attention);
    
    \node[draw, rectangle, rounded corners, right=of Attention] (Image) {Generated Image};
    \draw[->] (Attention) -- (Image);
    
    \node[draw, rectangle, rounded corners, below=2cm of Image] (D) {Discriminator ($D$) with Explainability};
    \draw[->, dashed] (Image) -- (D);
    \node[draw, rectangle, rounded corners, below=2cm of Attention] (RealImage) {Real Image};
    \draw[->] (RealImage) -- (D);
    
    \node[right=2cm of D] (Output) {Real or Fake + Explanation};
    \draw[->] (D) -- (Output);
\end{tikzpicture}%
}
\end{center}

\textbf{4. Example of Feature Attribution in GANs Using PyTorch}

One common method to achieve explainability is through feature attribution, where we visualize which parts of an image contribute most to the decision-making process of the discriminator. Below is a simple example of how this might be implemented in PyTorch:

\begin{lstlisting}[style=python]
import torch
import torch.nn as nn
import torch.nn.functional as F

# Define a simple Discriminator with feature attribution
class ExplainableDiscriminator(nn.Module):
    def __init__(self):
        super(ExplainableDiscriminator, self).__init__()
        self.conv1 = nn.Conv2d(3, 64, kernel_size=4, stride=2, padding=1)
        self.conv2 = nn.Conv2d(64, 128, kernel_size=4, stride=2, padding=1)
        self.fc = nn.Linear(128 * 8 * 8, 1)
    
    def forward(self, x):
        x = F.relu(self.conv1(x))
        features = F.relu(self.conv2(x))
        output = torch.sigmoid(self.fc(features.view(features.size(0), -1)))
        return output, features

# Visualizing feature importance
def visualize_feature_attribution(model, input_image):
    _, features = model(input_image)
    feature_importance = features.mean(dim=1).detach().cpu().numpy()
    # Code to plot the feature importance heatmap
    plt.imshow(feature_importance[0], cmap='hot', interpolation='nearest')
    plt.show()

# Example usage
discriminator = ExplainableDiscriminator()
input_image = torch.randn(1, 3, 32, 32)  # Example input
visualize_feature_attribution(discriminator, input_image)
\end{lstlisting}

\textbf{5. Real-World Applications of Explainable GANs}

Explainable GANs have a wide range of practical applications:
\begin{itemize}
    \item \textbf{Healthcare:} In medical imaging, explainable GANs can highlight which areas of a scan are most indicative of a disease, helping doctors understand why a particular diagnosis is suggested~\cite{li2020gan}.
    \item \textbf{Art and Design:} Artists can use explainable GANs to explore and understand how different features are represented, allowing for more precise control over generated artworks~\cite{abdal2019image2stylegan}.
    \item \textbf{Security and Forensics:} Explainable models can identify and highlight artifacts or anomalies in images, which can be useful for detecting tampered or fake images~\cite{adadi2021survey}.
\end{itemize}

By focusing on explainability, researchers are not only making GANs more transparent but also improving their usability in fields that require a clear understanding of the decision-making process. As GANs continue to evolve, integrating explainability into their core will be essential for building trust and ensuring ethical use in real-world applications.

\section{GANs and Privacy Preservation}

As the use of Generative Adversarial Networks (GANs) expands across various industries, concerns about privacy have become increasingly important~\cite{pan2019recent}. Traditional machine learning models often require access to large amounts of data, which can include sensitive information such as personal photos, medical records, or financial data~\cite{gan2021research}. Using this data for training GANs raises serious privacy concerns, especially if the generated outputs inadvertently reveal information about the individuals in the training set. To address these issues, researchers have developed privacy-preserving GANs (PP-GANs)~\cite{wu2019privacy} that aim to generate realistic data without compromising the privacy of the individuals whose data was used for training. In this section, we will explore how privacy can be integrated into the design of GANs, and discuss various approaches to building privacy-preserving generative models.

\textbf{1. The Need for Privacy Preservation in GANs}

Privacy-preserving GANs are essential in situations where data confidentiality is a priority. For example, in healthcare, GANs might be used to generate synthetic medical records that can be shared for research without exposing real patient information~\cite{wu2019privacy}. Similarly, in social media, GANs can generate realistic user avatars or content without using actual user photos. The primary goal is to ensure that the model does not memorize or leak any sensitive details from the training data.

Key motivations for privacy-preserving GANs:
\begin{itemize}
    \item \textbf{Data Confidentiality:} Preventing the disclosure of sensitive information that might be embedded in the training data.
    \item \textbf{Data Sharing:} Enabling the sharing of synthetic data for research and analysis without violating privacy laws or agreements.
    \item \textbf{Compliance:} Meeting legal and ethical standards, such as the General Data Protection Regulation (GDPR), which emphasizes data protection and privacy.
\end{itemize}

\textbf{2. Techniques for Building Privacy-Preserving GANs}

There are several techniques to incorporate privacy into GANs, each with its own strengths and trade-offs. Below are some of the most common approaches:

\begin{itemize}
    \item \textbf{Differential Privacy (DP):} Differential privacy is a mathematical framework that provides a quantifiable way to ensure that the model's outputs do not reveal specific information about any individual in the dataset~\cite{liu2021subverting}. By adding noise to the gradients during training, differential privacy makes it difficult to infer the presence of any single data point in the dataset.
    \item \textbf{Federated Learning~\cite{mcmahan2017communication}:} In this setup, the model is trained across multiple devices or servers, each with its own dataset, without sharing the actual data. The devices only share model updates (gradients), which are aggregated to improve the global model. This ensures that sensitive data never leaves the local device~\cite{zhang2021survey}.
    \item \textbf{Generative Model Distillation~\cite{salimans2022progressive}:} This method involves training a teacher model on sensitive data and then using it to train a student model on non-sensitive or synthetic data. The student model learns to generate data without ever seeing the original sensitive dataset, thus maintaining privacy.
\end{itemize}

\textbf{3. Architecture of a Privacy-Preserving GAN Using Differential Privacy}

Differential privacy is one of the most widely-used techniques to make GANs privacy-preserving. The core idea is to introduce noise into the training process so that the model cannot memorize specific details about the training data~\cite{wu2019privacy}. The following diagram shows how differential privacy can be integrated into a GAN's architecture:

\begin{center}
\begin{tikzpicture}
    \node[draw, rectangle, rounded corners] (G) at (0,0) {Generator ($G$)};
    \node[above=1cm of G] (GInput) {Random Noise $z$};
    \draw[->] (GInput) -- (G);
    
    \node[draw, rectangle, rounded corners, right=2cm of G] (Image) {Generated Image};
    \draw[->] (G) -- (Image);
    
    \node[draw, rectangle, rounded corners, below=2cm of Image] (D) {Discriminator ($D$) with DP Noise};
    \draw[->, dashed] (Image) -- (D);
    \node[draw, rectangle, rounded corners, below=2cm of G] (RealImage) {Real Image};
    \draw[->] (RealImage) -- (D);
    
    \node[right=2cm of D] (Output) {Noisy Gradients};
    \draw[->] (D) -- (Output);
\end{tikzpicture}
\end{center}

\textbf{4. Example Implementation of Differential Privacy in PyTorch}

Here is a simple example of how differential privacy can be applied to the training process of a GAN using PyTorch. We introduce noise into the gradient updates to prevent the model from learning specific details about individual data points:

\begin{lstlisting}[style=python]
import torch
import torch.nn as nn
import torch.optim as optim
import torch.nn.functional as F

# Define a basic Discriminator
class PrivacyDiscriminator(nn.Module):
    def __init__(self):
        super(PrivacyDiscriminator, self).__init__()
        self.conv1 = nn.Conv2d(3, 64, kernel_size=4, stride=2, padding=1)
        self.conv2 = nn.Conv2d(64, 128, kernel_size=4, stride=2, padding=1)
        self.fc = nn.Linear(128 * 8 * 8, 1)
    
    def forward(self, x):
        x = F.relu(self.conv1(x))
        x = F.relu(self.conv2(x))
        return torch.sigmoid(self.fc(x.view(x.size(0), -1)))

# Function to add differential privacy noise
def add_dp_noise(gradients, noise_scale=0.1):
    noise = torch.normal(0, noise_scale, size=gradients.size()).to(gradients.device)
    return gradients + noise

# Training loop with differential privacy
discriminator = PrivacyDiscriminator()
d_optimizer = optim.Adam(discriminator.parameters(), lr=0.0002)

for data in dataloader:  # Assume dataloader provides batches of real images
    d_optimizer.zero_grad()
    real_images = data
    output = discriminator(real_images)
    
    # Compute loss and apply differential privacy to gradients
    loss = F.binary_cross_entropy(output, torch.ones_like(output))
    loss.backward()
    
    # Add noise to the gradients to ensure differential privacy
    for param in discriminator.parameters():
        param.grad = add_dp_noise(param.grad)
    
    d_optimizer.step()
\end{lstlisting}

\textbf{5. Applications of Privacy-Preserving GANs}

Privacy-preserving GANs have numerous applications across different fields:
\begin{itemize}
    \item \textbf{Healthcare:} Synthetic patient data can be generated to train diagnostic models without risking patient confidentiality. Researchers can develop and validate models without accessing sensitive medical records~\cite{pan2019recent}.
    \item \textbf{Finance:} Banks can use synthetic transaction data to build fraud detection systems, ensuring that sensitive customer data remains private~\cite{wu2019privacy}.
    \item \textbf{Smart Devices:} Federated learning~\cite{mcmahan2017communication} allows devices to improve voice recognition models without sending raw audio data to central servers, preserving user privacy.
\end{itemize}

\textbf{6. Challenges and Future Directions}

While privacy-preserving GANs offer promising solutions, there are still several challenges:
\begin{itemize}
    \item \textbf{Balancing Privacy and Utility:} Adding too much noise to achieve differential privacy can degrade the quality of the generated data. Finding the right balance is crucial.
    \item \textbf{Scalability:} Techniques like federated learning require significant computational resources and efficient communication protocols, which can be difficult to implement at scale.
    \item \textbf{Improved Metrics for Privacy:} Defining and measuring privacy in the context of generative models is still an area of active research. Clear metrics are needed to evaluate the effectiveness of privacy-preserving techniques.
\end{itemize}

As privacy concerns continue to grow, the development of robust privacy-preserving GANs will be essential for ensuring that generative models can be safely and ethically used in real-world applications. By understanding these techniques, developers and researchers can create models that respect data confidentiality while still providing valuable and innovative solutions~\cite{li2024survey}.

\section{Generalization of GANs to Unseen Data}

One of the ongoing challenges in the development of Generative Adversarial Networks (GANs) is their ability to generalize effectively to unseen data. Generalization refers to a model's capability to generate realistic and high-quality samples that are not only consistent with the training data but also able to capture patterns and variations that were not explicitly present in the training set~\cite{gan2021research}. Traditional GANs often struggle with this, as they might overfit to the training data~\cite{de2021survey}, leading to poor performance when generating samples from new distributions or when dealing with diverse datasets. In this section, we will explore the concept of generalization in GANs, discuss the techniques that have been proposed to improve it, and provide detailed examples to illustrate how these techniques can be implemented.

\textbf{1. Why Generalization is Important for GANs}

Generalization is a crucial aspect of any generative model because it determines how well the model can create new and diverse outputs. If a GAN can only generate images that closely resemble its training data, it limits the model's utility, especially in applications where creativity and variety are needed~\cite{li2024survey}. For instance, a GAN trained to generate artwork should be able to produce pieces that reflect the style of the training data but still introduce new elements, textures, and forms. Effective generalization is also important for:
\begin{itemize}
    \item \textbf{Data Augmentation:} For GANs to be useful in data augmentation, they must generate samples that introduce new variations, rather than replicating existing ones.
    \item \textbf{Robustness:} Models that generalize well can handle variations in data, making them more robust to noise and different conditions.
    \item \textbf{Creativity and Diversity:} Good generalization allows GANs to create outputs that are not simply replicas of the training data but new and unique instances.
\end{itemize}

\textbf{2. Challenges in Achieving Generalization with GANs}

There are several reasons why GANs may struggle with generalization:
\begin{itemize}
    \item \textbf{Mode Collapse:} This occurs when the generator produces a limited variety of outputs, failing to capture the full distribution of the training data. This prevents the model from generating diverse examples~\cite{wu2021modeling}.
    \item \textbf{Overfitting:} If the discriminator becomes too powerful, the generator may overfit to specific examples in the training set, reducing its ability to generate new and unseen data~\cite{adadi2021survey}.
    \item \textbf{Training Instability:} The adversarial nature of GANs can lead to unstable training, where the model oscillates or fails to converge, further hindering generalization.
\end{itemize}

\textbf{3. Techniques to Improve Generalization in GANs}

Researchers have developed various techniques to help GANs generalize better to unseen data. Some of the most effective approaches include:

\begin{itemize}
    \item \textbf{Regularization:} Techniques like dropout, weight decay, and spectral normalization can prevent overfitting by encouraging the generator to explore a wider range of the latent space, leading to more diverse outputs~\cite{li2024survey}.
    \item \textbf{Latent Space Interpolation:} By generating samples from interpolated points between latent vectors, the model can learn to produce images that lie between the known patterns, enhancing diversity and generalization~\cite{adadi2021survey}.
    \item \textbf{Data Augmentation for Discriminators:} Applying data augmentation to the input data seen by the discriminator can make it more robust and encourage the generator to generalize beyond the training examples.
    \item \textbf{Ensemble Models:} Using multiple generators and discriminators allows the model to learn different aspects of the data distribution, leading to a more comprehensive understanding of the underlying patterns~\cite{pan2019recent}.
\end{itemize}

\textbf{4. Architecture and Implementation Techniques}

Below is a conceptual diagram of how regularization and ensemble techniques can be integrated into a GAN framework to improve generalization:

\begin{center}
\begin{tikzpicture}
    \node[draw, rectangle, rounded corners] (GInput) at (0,0) {Latent Noise $z$};
    
    \node[draw, rectangle, rounded corners, right=2cm of GInput] (G1) {Generator 1};
    \node[draw, rectangle, rounded corners, above=0.5cm of G1] (G2) {Generator 2};
    \node[draw, rectangle, rounded corners, below=0.5cm of G1] (G3) {Generator 3};
    \draw[->] (GInput) -- (G1);
    \draw[->] (GInput) -- (G2);
    \draw[->] (GInput) -- (G3);
    
    \node[draw, rectangle, rounded corners, right=2cm of G1] (Image1) {Generated Image};
    \node[draw, rectangle, rounded corners, right=2cm of G2] (Image2) {Generated Image};
    \node[draw, rectangle, rounded corners, right=2cm of G3] (Image3) {Generated Image};
    \draw[->] (G1) -- (Image1);
    \draw[->] (G2) -- (Image2);
    \draw[->] (G3) -- (Image3);
    
    \node[draw, rectangle, rounded corners, below=2cm of G1] (Regularization) {Regularization Techniques};
    \draw[->, dashed] (Regularization) -- (G1);
    \draw[->, dashed] (Regularization) -- (G2);
    \draw[->, dashed] (Regularization) -- (G3);
\end{tikzpicture}
\end{center}

\textbf{5. Example Implementation: Improving Generalization Using Spectral Normalization}

Spectral normalization is a technique used to stabilize the training of GANs and improve generalization by constraining the weights of the network. Below is an example of how to implement spectral normalization in PyTorch:

\begin{lstlisting}[style=python]
import torch
import torch.nn as nn
import torch.nn.utils as utils

# Define a Generator with Spectral Normalization
class SNGenerator(nn.Module):
    def __init__(self, latent_dim):
        super(SNGenerator, self).__init__()
        self.fc = utils.spectral_norm(nn.Linear(latent_dim, 256))
        self.conv1 = utils.spectral_norm(nn.ConvTranspose2d(256, 128, 4, 2, 1))
        self.conv2 = utils.spectral_norm(nn.ConvTranspose2d(128, 3, 4, 2, 1))
    
    def forward(self, z):
        x = F.relu(self.fc(z).view(-1, 256, 1, 1))
        x = F.relu(self.conv1(x))
        return torch.tanh(self.conv2(x))

# Define a simple training loop that highlights generalization
z1 = torch.randn(1, 100)
z2 = torch.randn(1, 100) * 1.5  # Example of testing with "unseen" input
generator = SNGenerator(latent_dim=100)

generated_image1 = generator(z1)
generated_image2 = generator(z2)
\end{lstlisting}

\textbf{6. Real-World Applications Where Generalization Matters}

Generalization is essential for many practical applications of GANs, including:
\begin{itemize}
    \item \textbf{Art Generation:} Artists and designers use GANs to create new styles and artworks. The ability to generalize allows the model to generate unique pieces that are not direct copies of the training data~\cite{karras2019style}.
    \item \textbf{Medical Imaging:} GANs can be used to generate synthetic medical images for training diagnostic models. Effective generalization ensures that these images cover a wide range of scenarios, including rare conditions.
    \item \textbf{Autonomous Vehicles:} In training autonomous systems, GANs are used to create synthetic data that mimics different driving conditions. Generalization ensures that these systems are robust to various environments and scenarios.
\end{itemize}

\textbf{7. Challenges and Future Directions in Generalization}

Despite progress, there are still challenges in improving the generalization capabilities of GANs:
\begin{itemize}
    \item \textbf{Avoiding Overfitting Without Sacrificing Quality:} Finding the right balance between generalization and quality remains difficult, as improving one often affects the other.
    \item \textbf{Evaluation Metrics:} Traditional metrics like Inception Score or FID may not fully capture the ability of a GAN to generalize. Developing better evaluation methods is essential for future research.
    \item \textbf{Advanced Architectures:} Techniques such as hierarchical latent spaces, better loss functions, and integrating self-supervised learning~\cite{zhang2019self} could further enhance generalization capabilities.
\end{itemize}

By addressing these challenges, future research can unlock the full potential of GANs, enabling them to generate high-quality, diverse, and realistic data across a wide range of applications~\cite{liu2021self}. Understanding the techniques and principles behind generalization is essential for anyone working to push the boundaries of what GANs can achieve.

\section{Combining GANs with Reinforcement Learning}
Generative Adversarial Networks (GANs) and Reinforcement Learning (RL)~\cite{sutton1998reinforcement, sutton2018reinforcement} are two of the most powerful paradigms in machine learning. While GANs are primarily used for generating realistic data~\cite{kaelbling1996reinforcement}, RL focuses on training agents to make decisions in an environment by maximizing a reward signal. Recently, there has been growing interest in combining these two approaches to harness the strengths of both: GANs' ability to generate high-quality samples and RL's capability to optimize actions through interaction with an environment. This integration opens up new possibilities for enhancing generative models and solving complex problems that require both generation and decision-making capabilities. In this section, we will explore the concept of integrating GANs with RL, discuss various applications, and provide detailed examples.

\textbf{1. Why Combine GANs with Reinforcement Learning?}

Combining GANs with reinforcement learning brings several benefits~\cite{wiering2012reinforcement} that can enhance the performance and applicability of generative models:
\begin{itemize}
    \item \textbf{Learning from Interaction:} While traditional GANs learn from a static dataset, RL allows models to learn through interaction. This can be useful for tasks where the generative model needs to adapt based on feedback or changes in the environment.
    \item \textbf{Improved Exploration:} RL can help GANs explore the latent space more effectively, leading to the generation of diverse and high-quality samples. This is particularly important in scenarios where there are many possible outputs, and the model needs to explore them.
    \item \textbf{Task-Specific Generation:} By combining GANs with RL, it is possible to create models that not only generate realistic data but also optimize it for specific tasks, such as game level design, robot control, or dynamic content creation.
\end{itemize}

\textbf{2. Techniques for Integrating GANs with Reinforcement Learning}

Several approaches have been developed to integrate GANs with RL, each with its own advantages and suitable applications. Here are some popular techniques~\cite{wiering2012reinforcement}:
\begin{itemize}
    \item \textbf{Conditional GANs with RL Reward Signal:} In this approach, the generator is conditioned on the RL agent's state, and the discriminator provides a reward signal based on the generated output. This allows the RL agent to learn which actions lead to desirable outputs.
    \item \textbf{Generative Adversarial Imitation Learning (GAIL):} GAIL is a method that combines the adversarial training of GANs with imitation learning in RL. It is used to teach an agent to imitate the behavior observed in expert demonstrations. The discriminator acts as a reward function, distinguishing between expert behavior and the agent's behavior, while the generator (RL agent) learns to match the expert behavior.
    \item \textbf{Model-Based RL with GANs:} GANs can be used to model the environment dynamics in RL, allowing the agent to predict future states and plan its actions accordingly. This is useful in scenarios where interacting with the real environment is costly or time-consuming.
\end{itemize}

\textbf{3. Architecture of a GAN-RL Integration}

To illustrate how GANs and RL can be combined, consider a scenario where an RL agent uses a GAN to generate images that it then interacts with~\cite{sarmad2019rl}. The RL agent receives a reward based on the quality or suitability of the generated images for a particular task. Below is a conceptual diagram showing this integration:

\begin{center}
\begin{tikzpicture}
    \node[draw, rectangle, rounded corners] (G) at (0,0) {Generator ($G$)};
    \node[above=1cm of G] (Latent) {Latent Input $z$};
    \draw[->] (Latent) -- (G);
    
    \node[draw, rectangle, rounded corners, right=2cm of G] (Image) {Generated Image};
    \draw[->] (G) -- (Image);
    
    \node[draw, rectangle, rounded corners, below=2cm of Image] (RL) {RL Agent};
    \draw[->, dashed] (Image) -- (RL);
    
    \node[right=2cm of RL] (Reward) {Reward Signal};
    \draw[->] (RL) -- (Reward);
    \node[below=1cm of RL] (Action) {Action};
    \draw[->] (Action) -- (G);
\end{tikzpicture}
\end{center}

\textbf{4. Example: Using GANs to Enhance RL in Game Level Design}

In game design, RL can be used to create agents that play games, while GANs can generate new levels or environments for these agents to interact with~\cite{de2021survey}. By combining the two, it is possible to create a system where the GAN generates levels that are challenging and interesting, and the RL agent learns to navigate these levels~\cite{sarmad2019rl}.

Below is an example of how GANs can be used to generate game levels, and how the RL agent can interact with these levels to learn better strategies. The GAN generator is trained to produce level designs, while the RL agent plays the game and provides feedback on how challenging or engaging the level is.

\begin{lstlisting}[style=python]
import torch
import torch.nn as nn
import torch.optim as optim
import torch.nn.functional as F

# Define a simple Level Generator using GAN
class LevelGenerator(nn.Module):
    def __init__(self, latent_dim):
        super(LevelGenerator, self).__init__()
        self.fc1 = nn.Linear(latent_dim, 256)
        self.fc2 = nn.Linear(256, 512)
        self.fc3 = nn.Linear(512, 1024)
        self.fc4 = nn.Linear(1024, 32 * 32)  # Assuming a 32x32 grid level design
    
    def forward(self, z):
        x = F.relu(self.fc1(z))
        x = F.relu(self.fc2(x))
        x = F.relu(self.fc3(x))
        return torch.sigmoid(self.fc4(x)).view(-1, 1, 32, 32)

# Define the RL agent interaction
class RLAgent:
    def __init__(self, env):
        self.env = env
    
    def act(self, level):
        # Simulate playing the game level and provide feedback
        success = self.env.play(level)
        reward = 1 if success else -1  # Simple reward for this example
        return reward

# Example usage
latent_vector = torch.randn(1, 100)  # Random input for the generator
generator = LevelGenerator(latent_dim=100)
generated_level = generator(latent_vector)

# Assume an environment class that accepts level designs
class GameEnvironment:
    def play(self, level):
        # Logic to play the game with the generated level
        return True  # Assume the level was successfully completed

env = GameEnvironment()
agent = RLAgent(env)
reward = agent.act(generated_level)
print("Reward:", reward)
\end{lstlisting}

\textbf{5. Real-World Applications of GANs with Reinforcement Learning}

The combination of GANs and RL has led to several innovative applications~\cite{scholl2011challenges}:
\begin{itemize}
    \item \textbf{Robotics:} In robotics, GANs can generate realistic simulations of environments, allowing RL agents (robots) to train safely in virtual environments before being deployed in the real world.
    \item \textbf{Autonomous Vehicles:} GANs can be used to create diverse driving scenarios, while RL helps the vehicle learn to navigate these scenarios. This combination is essential for training self-driving cars.
    \item \textbf{Game AI Development:} By using GANs to generate game content and RL to optimize gameplay, developers can create games that offer endless levels of unique challenges, enhancing player engagement.
\end{itemize}

\textbf{6. Challenges and Future Directions}

Despite the advantages, integrating GANs with RL also poses challenges:
\begin{itemize}
    \item \textbf{Stability Issues:} Both GANs and RL can be unstable during training. Combining them can exacerbate these issues, requiring careful tuning and architecture design.
    \item \textbf{Scalability:} RL often requires large amounts of data and interactions, and adding GANs into the mix can make the system even more computationally intensive.
    \item \textbf{Exploration vs. Exploitation:} Balancing exploration (trying new strategies) and exploitation (using known good strategies) is a key challenge in RL. When combined with GANs, this balance becomes even more crucial, as the generator must be able to explore new possibilities without losing quality.
\end{itemize}

The integration of GANs with reinforcement learning has opened up exciting new opportunities, from developing adaptive systems that can learn in real-time to creating generative models that are optimized for specific tasks~\cite{scholl2011challenges}. By understanding how to combine these two approaches, researchers and developers can push the boundaries of what generative models can achieve, leading to more intelligent, versatile, and efficient systems.

\section{Multimodal Generative Adversarial Networks}

Multimodal Generative Adversarial Networks (GANs) represent a fascinating area of research where models are designed to understand and generate data across multiple modalities, such as text, images, audio, and more~\cite{liu2019multi}. Traditional GANs typically operate within a single domain (e.g., generating images from noise), but multimodal GANs can process and generate outputs that combine different types of data, leading to more versatile and intelligent systems. For instance, a multimodal GAN might take a text description and generate an image based on it, or even combine visual and audio inputs to create synchronized video clips. In this section, we will explore the concept of multimodal GANs, focusing on text-to-image generation and cross-domain generation, and discuss how these models can generalize across different data types.

\textbf{1. The Importance of Multimodal GANs}

In real-world scenarios, information rarely exists in isolation. For example, when we watch a movie, we perceive both visual and auditory stimuli; when we read a book, we imagine scenes based on textual descriptions~\cite{scholl2011challenges}. Multimodal GANs aim to bridge the gap between different types of data, allowing for richer and more comprehensive interactions. Key benefits of multimodal GANs include:
\begin{itemize}
    \item \textbf{Enhanced Creativity:} Combining multiple modalities allows models to generate more complex and creative outputs, such as generating artwork based on a poem or creating music that matches a visual scene.
    \item \textbf{Data Synthesis Across Domains:} Multimodal GANs can synthesize data in one domain using information from another, making them useful for tasks like generating images from text or converting sketches into full-color images.
    \item \textbf{Improved Generalization:} By learning to process different types of data, these models can develop a more comprehensive understanding of concepts, leading to better generalization across tasks.
\end{itemize}

\subsection{Text-to-Image Multimodal Generation}

One of the most well-known applications of multimodal GANs is text-to-image generation, where a model learns to generate images that correspond to a given textual description~\cite{patashnik2021styleclip}. This involves teaching the GAN to understand both text and visual data, so it can accurately translate descriptions into realistic images.

\textbf{1. How Text-to-Image Generation Works}

Text-to-image generation typically involves two components:
\begin{itemize}
    \item \textbf{Text Encoder:} Converts the input text into a vector representation that captures the semantic meaning of the description. This representation is then used to condition the GAN.
    \item \textbf{Conditional GAN (cGAN):} The generator is conditioned on the text representation, guiding it to create images that match the description. The discriminator evaluates whether the generated image is realistic and whether it matches the given text.
\end{itemize}

\textbf{2. Architecture of a Text-to-Image GAN}

The following diagram illustrates a typical text-to-image GAN architecture, showing how the text encoder and conditional GAN work together to generate images:

\begin{center}
\resizebox{\textwidth}{!}{%
\begin{tikzpicture}[node distance=3cm, auto]
    \node[draw, rectangle, rounded corners] (TextEncoder) {Text Encoder};
    \node[above=1cm of TextEncoder] (TextInput) {Text Description};
    \draw[->] (TextInput) -- (TextEncoder);
    
    \node[draw, rectangle, rounded corners, right=of TextEncoder] (Latent) {Latent Vector $z$ + Text Embedding};
    \draw[->] (TextEncoder) -- (Latent);
    
    \node[draw, rectangle, rounded corners, right=of Latent] (G) {Generator ($G$)};
    \draw[->] (Latent) -- (G);
    
    \node[draw, rectangle, rounded corners, right=of G] (Image) {Generated Image};
    \draw[->] (G) -- (Image);
    
    \node[draw, rectangle, rounded corners, below=2cm of Image] (D) {Discriminator ($D$)};
    \draw[->, dashed] (Image) -- (D);
    \node[draw, rectangle, rounded corners, below=2cm of TextEncoder] (RealImage) {Real Image + Text};
    \draw[->] (RealImage) -- (D);
    
    \node[right=of D] (Output) {Real or Fake + Match to Text};
    \draw[->] (D) -- (Output);
\end{tikzpicture}%
}
\end{center} 

\textbf{3. Example Implementation of a Text-to-Image GAN in PyTorch}

Below is an example of how a basic text-to-image GAN can be implemented using PyTorch. We define a simple text encoder and a conditional generator.

\begin{lstlisting}[style=python]
import torch
import torch.nn as nn
import torch.optim as optim
import torch.nn.functional as F

# Define a simple Text Encoder
class TextEncoder(nn.Module):
    def __init__(self, vocab_size, embed_size):
        super(TextEncoder, self).__init__()
        self.embedding = nn.Embedding(vocab_size, embed_size)
        self.fc = nn.Linear(embed_size, 128)
    
    def forward(self, text):
        x = self.embedding(text)
        return F.relu(self.fc(x.mean(dim=1)))

# Define a Conditional Generator
class TextToImageGenerator(nn.Module):
    def __init__(self, latent_dim, text_dim):
        super(TextToImageGenerator, self).__init__()
        self.fc1 = nn.Linear(latent_dim + text_dim, 256)
        self.fc2 = nn.Linear(256, 512)
        self.fc3 = nn.Linear(512, 1024)
        self.fc4 = nn.Linear(1024, 64 * 64 * 3)  # Generate 64x64 image
    
    def forward(self, z, text_embed):
        x = torch.cat((z, text_embed), dim=1)
        x = F.relu(self.fc1(x))
        x = F.relu(self.fc2(x))
        x = F.relu(self.fc3(x))
        return torch.tanh(self.fc4(x)).view(-1, 3, 64, 64)

# Example usage
latent_vector = torch.randn(1, 100)  # Random input
text_input = torch.randint(0, 1000, (1, 10))  # Example text input
text_encoder = TextEncoder(vocab_size=1000, embed_size=50)
text_embedding = text_encoder(text_input)

generator = TextToImageGenerator(latent_dim=100, text_dim=128)
generated_image = generator(latent_vector, text_embedding)
\end{lstlisting}

\subsection{Cross-Domain Generation and Generalization Capabilities}

Cross-domain generation involves creating data in one domain using information from another, such as generating music from images or translating visual features into sound. Multimodal GANs that excel at cross-domain generation can learn to generalize better because they must understand and translate patterns between different types of data.

\textbf{1. Benefits of Cross-Domain Generation}

Cross-domain generation has many practical applications:
\begin{itemize}
    \item \textbf{Creative Content Creation:} Models can generate music based on visual art, creating a cohesive audiovisual experience, or translate text into animations, enabling new forms of storytelling.
    \item \textbf{Data Augmentation Across Domains:} For tasks like video captioning, cross-domain GANs can generate synthetic data that helps improve the training of multimodal models.
    \item \textbf{Generalization Across Modalities:} By learning to map features from one domain to another, these models become better at generalizing, as they must understand underlying patterns that are not domain-specific.
\end{itemize}

\textbf{2. Challenges and Future Directions}

Despite the promising potential, multimodal GANs face several challenges:
\begin{itemize}
    \item \textbf{Alignment of Different Modalities:} Learning to align features across modalities is difficult because each type of data has its own unique characteristics (e.g., temporal data vs. spatial data).
    \item \textbf{Training Complexity:} Multimodal models are often more complex than single-domain models, requiring careful balancing of multiple loss functions and architectures.
    \item \textbf{Scalability:} Processing multiple modalities simultaneously can be resource-intensive, making scalability a concern for large-scale applications.
\end{itemize}

Multimodal GANs are a growing field of research that aim to merge different types of data, leading to more intelligent~\cite{li2024survey}, versatile, and creative applications~\cite{liu2019multi}. By understanding the principles of how these models operate and are trained, developers can unlock new possibilities in cross-domain generation~\cite{de2021survey}, from innovative art to practical tools that assist in everyday tasks.


\chapter{Diffusion Models vs. GANs}

In recent years, Diffusion Models have emerged as a strong alternative to Generative Adversarial Networks (GANs) for generating high-quality data~\cite{croitoru2023diffusion}. While GANs have been the dominant method for tasks such as image synthesis, diffusion models offer a new approach that addresses some of the inherent challenges of GANs, such as training instability and mode collapse. Diffusion models are based on a fundamentally different principle, using a probabilistic framework that involves a series of incremental transformations. These models have gained popularity due to their ability to generate diverse and high-fidelity outputs without many of the issues that traditionally plague GANs~\cite{yang2023diffusion}. In this chapter, we will explore the basic principles of diffusion models, compare them to GANs, and discuss their strengths and weaknesses.

\section{Fundamental Principles of Diffusion Models}

Diffusion models are a class of generative models that learn to generate data by modeling a process of gradual transformation. They work by learning to reverse a noising process, which means that instead of generating data directly from random noise (as GANs do), they start with a completely noisy input and learn how to transform it step-by-step into a coherent and realistic output~\cite{kingma2021variational}. This gradual denoising process allows diffusion models to generate high-quality results while avoiding some of the pitfalls of GANs, such as training instability.

\textbf{1. The Concept of Diffusion}

The term "diffusion" in diffusion models refers to a process of gradually adding noise to a data sample until it becomes indistinguishable from pure noise. Imagine starting with a clear image and adding small amounts of random noise to it, step by step, until the image is completely obscured. Diffusion models learn to reverse this process, taking a noisy image and gradually removing the noise to reconstruct the original image. The key idea is to model the probabilistic process of transforming data to noise and then learning to reverse it~\cite{ho2020denoising}.

Key components of diffusion models:
\begin{itemize}
    \item \textbf{Forward Diffusion Process:} A process where noise is incrementally added to a data sample over a series of steps. This process transforms the data into a noisy version, effectively creating a distribution that the model will learn to reverse.
    \item \textbf{Reverse Diffusion Process:} The generative part of the model, where the model learns to reverse the noise addition process. It takes a noisy sample and removes noise step by step until it reaches a clean and realistic output.
    \item \textbf{Probabilistic Framework:} Diffusion models rely on a probabilistic approach, modeling each step of noise addition and removal as a probability distribution, allowing for more controlled and stable generation.
\end{itemize}

\textbf{2. Diffusion Process and Reverse Process}

The forward and reverse processes are central to how diffusion models operate. Below, we will explain each in more detail, along with a mathematical description.

\subsection{Diffusion Process and Reverse Process}

\textbf{1. Forward Diffusion Process}

The forward diffusion process can be thought of as a sequence of steps where noise is gradually added to the data~\cite{ho2020denoising}. Mathematically, this process can be represented as a series of conditional probabilities, where each step involves adding a small amount of Gaussian noise:
\[
q(x_t | x_{t-1}) = \mathcal{N}(x_t; \sqrt{1 - \beta_t} x_{t-1}, \beta_t \mathbf{I})
\]
where:
\begin{itemize}
    \item $x_t$ represents the data at step $t$,
    \item $\beta_t$ is a small constant that controls the amount of noise added at each step,
    \item $\mathcal{N}$ denotes a Gaussian distribution.
\end{itemize}
By repeating this process over multiple steps, the data is transformed into pure noise.

\textbf{2. Reverse Diffusion Process}

The reverse process is where the generative power of the model lies. Instead of adding noise, the model learns to denoise the sample step by step. The objective is to train the model to approximate the conditional probabilities:
\[
p_\theta(x_{t-1} | x_t) = \mathcal{N}(x_{t-1}; \mu_\theta(x_t, t), \Sigma_\theta(x_t, t))
\]
where:
\begin{itemize}
    \item $\mu_\theta$ and $\Sigma_\theta$ are learned functions that predict the mean and variance of the distribution.
    \item $\theta$ denotes the parameters of the model.
\end{itemize}
The model learns to generate samples by starting with pure noise and gradually refining it back to a coherent image through these conditional distributions.

\textbf{3. Architecture of a Diffusion Model}

The architecture of diffusion models typically involves a neural network that predicts the noise to be subtracted at each step, thereby cleaning up the image incrementally. The following diagram illustrates the diffusion process:

\begin{center}
\begin{tikzpicture}
    \node[draw, rectangle, rounded corners] (x0) at (0,0) {Original Data $x_0$};
    \node[draw, rectangle, rounded corners, right=2cm of x0] (x1) {Noisy $x_1$};
    \node[draw, rectangle, rounded corners, right=2cm of x1] (x2) {Noisy $x_2$};
    \node[draw, rectangle, rounded corners, right=2cm of x2] (xt) {Noise $x_T$};
    \draw[->] (x0) -- (x1);
    \draw[->] (x1) -- (x2);
    \draw[->] (x2) -- (xt);
    \node[above=0.5cm of x1] (forward) {Forward Diffusion};
    
    \node[draw, rectangle, rounded corners, below=2cm of xt] (x2rev) {Noisy $x_2$};
    \node[draw, rectangle, rounded corners, left=2cm of x2rev] (x1rev) {Noisy $x_1$};
    \node[draw, rectangle, rounded corners, left=2cm of x1rev] (x0rev) {Reconstructed $x_0$};
    \draw[->] (xt) -- (x2rev);
    \draw[->] (x2rev) -- (x1rev);
    \draw[->] (x1rev) -- (x0rev);
    \node[below=0.5cm of x2rev] (reverse) {Reverse Diffusion};
\end{tikzpicture}
\end{center}

\textbf{4. Implementation Example of a Basic Diffusion Step in PyTorch}

Below is a simple implementation of a diffusion step in PyTorch, demonstrating how noise is added during the forward process and removed in the reverse process.

\begin{lstlisting}[style=python]
import torch
import torch.nn as nn
import torch.optim as optim
import torch.nn.functional as F

# Define a basic neural network for predicting noise
class DiffusionModel(nn.Module):
    def __init__(self):
        super(DiffusionModel, self).__init__()
        self.fc1 = nn.Linear(784, 512)
        self.fc2 = nn.Linear(512, 512)
        self.fc3 = nn.Linear(512, 784)
    
    def forward(self, x, t):
        x = F.relu(self.fc1(x))
        x = F.relu(self.fc2(x))
        return self.fc3(x)

# Forward diffusion step
def forward_diffusion_step(x, beta):
    noise = torch.randn_like(x)
    return torch.sqrt(1 - beta) * x + torch.sqrt(beta) * noise

# Reverse process example
model = DiffusionModel()
x = torch.randn((1, 784))  # Flattened 28x28 image
beta = 0.01  # Small noise constant

# Forward step
x_noisy = forward_diffusion_step(x, beta)

# Reverse step (model learns to predict noise)
predicted_noise = model(x_noisy, 1)
x_reconstructed = x_noisy - beta * predicted_noise
\end{lstlisting}

\textbf{5. Strengths and Applications of Diffusion Models}

Diffusion models have several advantages over GANs, including:
\begin{itemize}
    \item \textbf{Training Stability:} Because diffusion models learn to reverse a gradual process, they do not face the same training instabilities as GANs, such as mode collapse~\cite{ho2020denoising}.
    \item \textbf{High-Quality Outputs:} Diffusion models have been shown to produce very high-quality images, often surpassing GANs in terms of realism and diversity.
    \item \textbf{Versatility Across Modalities:} Like GANs, diffusion models can be applied to various tasks, including image synthesis, audio generation, and even text generation, but often with fewer issues related to training.
\end{itemize}

Diffusion models are still a relatively new area of research, but they offer a promising alternative to traditional GANs~\cite{croitoru2023diffusion}. By understanding the fundamental principles behind these models, developers can explore new approaches to generative modeling that may overcome some of the challenges faced by GANs. The gradual, probabilistic approach of diffusion models allows for more stable training and potentially better performance, making them an exciting development in the field of generative AI.

\section{Advantages of Diffusion Models Over GANs}

Diffusion models have garnered attention as a robust alternative to Generative Adversarial Networks (GANs), particularly for image synthesis and other generative tasks. While GANs have been the dominant approach for many years, diffusion models bring several advantages that address some of the inherent challenges of GANs. These advantages include better training stability, higher generation quality, and an ability to avoid the issue of mode collapse~\cite{croitoru2023diffusion}. In this section, we will explore these key benefits in detail, providing a comprehensive understanding of why diffusion models are becoming a competitive choice in the field of generative modeling.

\subsection{Training Stability}

One of the most significant issues with GANs is their training instability. The adversarial training process, where a generator and discriminator compete against each other, can lead to various challenges, such as non-convergence, oscillations, and sensitivity to hyperparameters. In contrast, diffusion models offer a more stable and controlled training process.

\textbf{1. Why GAN Training is Unstable}

In GANs, the generator tries to create samples that can deceive the discriminator, while the discriminator tries to distinguish between real and generated samples. This adversarial setup can lead to a tug-of-war, where the generator and discriminator are constantly trying to outsmart each other. If the discriminator becomes too strong, the generator may fail to learn properly~\cite{stypulkowski2024diffused}; if the generator becomes too strong, the discriminator may provide poor feedback. This imbalance can cause:
\begin{itemize}
    \item \textbf{Non-convergence:} The generator and discriminator may not reach a stable equilibrium, leading to oscillating loss functions.
    \item \textbf{Mode Collapse:} The generator may produce limited variations, repeatedly generating similar outputs instead of exploring the full data distribution.
    \item \textbf{Sensitivity to Hyperparameters:} Small changes in learning rates or other hyperparameters can drastically affect the training process, making it difficult to optimize.
\end{itemize}

\textbf{2. How Diffusion Models Improve Stability}

Diffusion models operate differently. Instead of relying on adversarial training, they use a probabilistic approach to gradually transform noise into data~\cite{croitoru2023diffusion}. This process involves learning a sequence of denoising steps, which is inherently more stable because each step is trained independently, without the need for a competing network. The key benefits include:
\begin{itemize}
    \item \textbf{Step-by-Step Learning:} Diffusion models learn to reverse the noise process in incremental steps, reducing the risk of instability~\cite{stypulkowski2024diffused}.
    \item \textbf{Controlled Training:} Since there is no adversarial component, the training process does not suffer from the issues of balance between competing networks.
    \item \textbf{Simpler Optimization:} The probabilistic framework allows for more straightforward loss functions, which can be easier to optimize compared to the adversarial loss used in GANs.
\end{itemize}

\textbf{3. Example: Stable Training in Diffusion Models Using PyTorch}

Below is an example of how a simple training step might look for a diffusion model. The model learns to predict the noise added to the data, providing a stable and controlled training process.

\begin{lstlisting}[style=python]
import torch
import torch.nn as nn
import torch.optim as optim

# Define a simple noise prediction network
class NoisePredictor(nn.Module):
    def __init__(self):
        super(NoisePredictor, self).__init__()
        self.fc1 = nn.Linear(784, 512)
        self.fc2 = nn.Linear(512, 512)
        self.fc3 = nn.Linear(512, 784)
    
    def forward(self, x):
        x = torch.relu(self.fc1(x))
        x = torch.relu(self.fc2(x))
        return self.fc3(x)

# Training loop
model = NoisePredictor()
optimizer = optim.Adam(model.parameters(), lr=0.001)
criterion = nn.MSELoss()

for epoch in range(100):
    noisy_data = torch.randn(1, 784)  # Simulating noisy input
    clean_data = torch.randn(1, 784)  # Original data for comparison
    
    # Model prediction
    predicted_noise = model(noisy_data)
    
    # Loss calculation (MSE between predicted and actual noise)
    loss = criterion(predicted_noise, noisy_data - clean_data)
    
    # Backpropagation
    optimizer.zero_grad()
    loss.backward()
    optimizer.step()
\end{lstlisting}

\subsection{Generation Quality}

Another area where diffusion models shine is in the quality of the generated outputs. While GANs are capable of producing realistic images, they can sometimes generate artifacts or fail to capture fine details. Diffusion models, on the other hand, excel at producing high-resolution and highly detailed images.

\textbf{1. Why Diffusion Models Produce Better Quality}

The step-by-step denoising process in diffusion models allows them to focus on refining details at each stage of generation~\cite{stypulkowski2024diffused}. Instead of trying to produce a complete image all at once, diffusion models progressively improve the quality of the sample, adding detail and coherence at each step. This leads to:
\begin{itemize}
    \item \textbf{Better Detail Preservation:} Each denoising step can focus on specific features, resulting in more refined and intricate details in the final output.
    \item \textbf{Reduced Artifacts:} Since the generation process is gradual, the model has multiple opportunities to correct any mistakes, leading to cleaner and more consistent images.
    \item \textbf{Higher Resolution Outputs:} Diffusion models have been shown to generate high-resolution images without the need for upscaling networks that are typically used in GAN architectures.
\end{itemize}

\subsection{Avoiding Mode Collapse}

Mode collapse is a well-known issue in GANs where the generator learns to produce only a limited variety of outputs, ignoring other possible modes in the data distribution~\cite{stypulkowski2024diffused}. This problem can severely limit the diversity of generated samples, which is especially problematic in applications where variety is crucial. Diffusion models naturally avoid this issue due to their design.

\textbf{1. What Causes Mode Collapse in GANs}

Mode collapse occurs when the generator learns a shortcut to "fool" the discriminator by producing a narrow range of outputs~\cite{yang2023diffusion}. For example, a GAN trained on faces might end up generating only one or two types of faces instead of exploring the full range of variations present in the training data. This happens because:
\begin{itemize}
    \item \textbf{Adversarial Training Dynamics:} The feedback loop between the generator and discriminator can lead to local optima where the generator finds a few samples that consistently deceive the discriminator.
    \item \textbf{Lack of Regularization:} Without mechanisms to encourage diversity, the generator might converge to a limited set of outputs.
\end{itemize}

\textbf{2. Why Diffusion Models Do Not Suffer from Mode Collapse}

Diffusion models avoid mode collapse due to their probabilistic framework. By modeling the entire process of data transformation as a distribution, diffusion models are designed to capture the full range of variations present in the data:
\begin{itemize}
    \item \textbf{Diverse Sampling:} The generation process inherently samples from the learned data distribution, ensuring that different modes are represented~\cite{stypulkowski2024diffused}.
    \item \textbf{Gradual Denoising:} Since the model learns to denoise step-by-step, it does not rely on adversarial feedback, reducing the risk of collapsing to a limited set of outputs.
\end{itemize}

\textbf{3. Practical Advantages in Applications}

The strengths of diffusion models make them particularly suitable for applications that require stability, high-quality outputs, and diversity:
\begin{itemize}
    \item \textbf{Art and Design:} The ability to generate detailed and varied designs makes diffusion models ideal for creative tasks, where diversity and refinement are essential~\cite{ho2022video}.
    \item \textbf{Medical Imaging:} The stability and high resolution of diffusion models can be beneficial in generating realistic medical scans that capture subtle details without artifacts.
    \item \textbf{Data Augmentation:} For scenarios where diverse and representative data is needed, diffusion models can generate samples that capture a wide range of variations, enhancing the training of other machine learning models.
\end{itemize}

By understanding these advantages, we can see why diffusion models are becoming an increasingly popular choice for generative tasks. Their ability to produce high-quality, stable, and diverse outputs offers an alternative to GANs that addresses many of the issues faced in traditional generative modeling~\cite{li2024survey}.

\section{The Evolution of Diffusion Models}

Diffusion models have evolved significantly since their introduction, with new variations and improvements that make them more efficient, scalable, and capable of producing high-quality outputs. Two of the most notable developments are Denoising Diffusion Probabilistic Models (DDPM)~\cite{ho2020denoising} and Latent Diffusion Models (LDM)~\cite{rombach2022high}. These advancements have refined the fundamental principles of diffusion, making them more practical for real-world applications. In this section, we will explore these models in detail, explain their mechanisms, and discuss how they contribute to the progress of diffusion-based generative modeling~\cite{rombach2022high}.

\subsection{DDPM: Denoising Diffusion Probabilistic Models}

Denoising Diffusion Probabilistic Models (DDPM) represent one of the earliest and most influential types of diffusion models~\cite{ho2020denoising}. DDPMs use a straightforward yet effective approach to learn the process of generating data by reversing a sequence of noising steps. The main idea is to teach the model how to denoise an image step-by-step until it can reconstruct a realistic image from pure noise.

\textbf{1. How DDPM Works}

The training of DDPMs involves two main processes: a forward diffusion process and a reverse denoising process.

\textbf{Forward Diffusion Process:}
\begin{itemize}
    \item The forward process begins with a clean data sample and gradually adds Gaussian noise over multiple steps. Each step adds a small amount of noise, turning the data into a noisy version of itself.
    \item This can be represented as:
    \[
    q(x_t | x_{t-1}) = \mathcal{N}(x_t; \sqrt{1 - \beta_t} x_{t-1}, \beta_t \mathbf{I}),
    \]
    where $\beta_t$ controls the amount of noise added at each step $t$.
    \item The forward process converts the data into a noisy sample $x_T$ that is close to pure noise.
\end{itemize}

\textbf{Reverse Denoising Process:}
\begin{itemize}
    \item The reverse process is where the generative capabilities of the model come into play. Starting from $x_T$, the model learns to predict $x_{t-1}$ from $x_t$ by estimating the mean and variance of the reverse transition.
    \item The reverse process can be expressed as:
    \[
    p_\theta(x_{t-1} | x_t) = \mathcal{N}(x_{t-1}; \mu_\theta(x_t, t), \Sigma_\theta(x_t, t)),
    \]
    where $\mu_\theta$ and $\Sigma_\theta$ are neural network parameters learned during training.
    \item By gradually applying this denoising process, the model reconstructs an image step-by-step, ultimately producing a realistic sample.
\end{itemize}

\textbf{2. Advantages of DDPMs}

\begin{itemize}
    \item \textbf{Gradual Refinement:} The step-by-step process allows for detailed adjustments, leading to high-quality images that capture intricate details~\cite{ho2020denoising}.
    \item \textbf{Stable Training:} Unlike GANs, DDPMs do not rely on adversarial training, making the training process more stable and easier to tune.
    \item \textbf{Flexibility:} DDPMs can be used for a variety of tasks, including image synthesis, super-resolution, and even video generation.
\end{itemize}

\textbf{3. Example Implementation of DDPM in PyTorch}

Here is a simplified example of how the reverse process in a DDPM might be implemented using PyTorch:

\begin{lstlisting}[style=python]
import torch
import torch.nn as nn
import torch.optim as optim
import torch.nn.functional as F

# Define a simple DDPM model for denoising
class DDPM(nn.Module):
    def __init__(self):
        super(DDPM, self).__init__()
        self.fc1 = nn.Linear(784, 512)
        self.fc2 = nn.Linear(512, 512)
        self.fc3 = nn.Linear(512, 784)
    
    def forward(self, x, t):
        x = torch.relu(self.fc1(x))
        x = torch.relu(self.fc2(x))
        return self.fc3(x)

# Example of the reverse step
model = DDPM()
noisy_image = torch.randn(1, 784)  # Simulating a noisy input
predicted_noise = model(noisy_image, t=10)  # t represents the step
denoised_image = noisy_image - predicted_noise
\end{lstlisting}

\subsection{Latent Diffusion Models (LDM)}

Latent Diffusion Models (LDM)~\cite{rombach2022high} are an evolution of the original diffusion model concept, designed to make the process more efficient and scalable. While DDPMs operate directly on high-dimensional data (such as pixels in an image), LDMs work in a lower-dimensional latent space. This significantly reduces the computational cost and speeds up the generation process.

\textbf{1. How Latent Diffusion Models Work}

LDMs leverage the concept of latent spaces, which are compressed representations of data. By applying the diffusion process in this latent space, LDMs can capture the essential features of the data without having to process every pixel directly:
\begin{itemize}
    \item \textbf{Latent Encoding:} The original data is first encoded into a latent representation using an encoder (such as a variational autoencoder or another neural network).
    \item \textbf{Latent Diffusion:} The diffusion process is then applied in this lower-dimensional space, making the computation faster and less resource-intensive.
    \item \textbf{Latent Decoding:} Once the reverse process has been completed, the latent representation is decoded back into the original high-dimensional space to produce the final output.
\end{itemize}

\textbf{2. Advantages of LDMs}

\begin{itemize}
    \item \textbf{Computational Efficiency:} By working in a lower-dimensional space, LDMs reduce the computational cost of training and generation, making them more scalable.
    \item \textbf{High-Quality Outputs:} Despite the reduced computation, LDMs can still produce high-resolution and detailed images because they operate on the essential features of the data.
    \item \textbf{Scalability Across Tasks:} LDMs can be adapted for various generative tasks, including text-to-image, image translation, and more~\cite{rombach2022high}.
\end{itemize}

\textbf{3. Architecture of Latent Diffusion Models}

The following diagram illustrates the basic architecture of an LDM, showing how the encoding and decoding processes are integrated with the diffusion process:

 \begin{center}
\resizebox{\textwidth}{!}{%
\begin{tikzpicture}[node distance=3cm, auto]
    \node[draw, rectangle, rounded corners] (Data) {Original Data};
    \node[draw, rectangle, rounded corners, right=of Data] (Encoder) {Encoder};
    \node[draw, rectangle, rounded corners, right=of Encoder] (Latent) {Latent Space};
    \draw[->] (Data) -- (Encoder);
    \draw[->] (Encoder) -- (Latent);
    
    \node[draw, rectangle, rounded corners, right=of Latent] (Diffusion) {Latent Diffusion};
    \draw[->] (Latent) -- (Diffusion);
    
    \node[draw, rectangle, rounded corners, right=of Diffusion] (Decoder) {Decoder};
    \node[draw, rectangle, rounded corners, right=of Decoder] (Generated) {Generated Data};
    \draw[->] (Diffusion) -- (Decoder);
    \draw[->] (Decoder) -- (Generated);
\end{tikzpicture}%
}
\end{center}

\textbf{4. Example Implementation of Latent Diffusion Using PyTorch}

Below is a simplified example showing how the encoding and diffusion steps might be implemented for a latent diffusion model:

\begin{lstlisting}[style=python]
class LatentEncoder(nn.Module):
    def __init__(self):
        super(LatentEncoder, self).__init__()
        self.fc1 = nn.Linear(784, 256)
        self.fc2 = nn.Linear(256, 128)
    
    def forward(self, x):
        x = torch.relu(self.fc1(x))
        return self.fc2(x)

class LatentDiffusionModel(nn.Module):
    def __init__(self):
        super(LatentDiffusionModel, self).__init__()
        self.fc1 = nn.Linear(128, 128)
    
    def forward(self, z, t):
        return self.fc1(z) - t * 0.01 * z  # Example of a simple latent diffusion step

# Encoding and diffusion
encoder = LatentEncoder()
diffusion_model = LatentDiffusionModel()

original_image = torch.randn(1, 784)
latent_representation = encoder(original_image)

# Apply diffusion in latent space
noisy_latent = diffusion_model(latent_representation, t=10)
\end{lstlisting}

\textbf{5. Applications of DDPMs and LDMs}

The evolution from DDPMs to LDMs has opened up new possibilities for real-world applications~\cite{yang2023diffusion}:
\begin{itemize}
    \item \textbf{Image Generation:} High-quality image synthesis, including detailed and high-resolution images, which were difficult to achieve with earlier models.
    \item \textbf{Text-to-Image Generation:} LDMs can effectively handle text prompts to create visual content, which has led to advancements in AI art and creative design.
    \item \textbf{Super-Resolution and Image Editing:} DDPMs and LDMs can refine images, remove noise, and enhance details, making them useful tools for photo editing and restoration.
\end{itemize}

By understanding the principles behind DDPMs and LDMs, developers can leverage these models to build efficient, scalable, and high-quality generative systems~\cite{rombach2022high}. The continuous evolution of diffusion models promises to bring even more powerful tools for generative AI in the future.

\section{Comparison Between GANs and Diffusion Models and Future Prospects}

Generative Adversarial Networks (GANs) and Diffusion Models have emerged as two of the most powerful approaches for generative modeling. While GANs have been the go-to method for tasks such as image synthesis for many years, Diffusion Models are now gaining traction due to their stability and high-quality outputs~\cite{stypulkowski2024diffused}. Both have their strengths and weaknesses, and choosing between them often depends on the specific requirements of the task at hand. In this section, we will compare GANs and Diffusion Models across several key aspects, discuss their advantages and limitations, and explore what the future might hold for these two approaches.

\textbf{1. Key Differences Between GANs and Diffusion Models}

GANs and Diffusion Models differ fundamentally in how they approach the task of generation~\cite{li2024survey}. Understanding these differences is crucial to grasp why each method might be preferred in certain scenarios.

\textbf{Training Methodology:}
\begin{itemize}
    \item \textbf{GANs:} GANs operate on an adversarial training principle, where two networks (the generator and the discriminator) are pitted against each other. The generator tries to create data that mimics the real data, while the discriminator attempts to distinguish between real and fake samples. This adversarial setup can lead to powerful generators but also introduces instability, making GANs notoriously difficult to train~\cite{gan2021research}.
    \item \textbf{Diffusion Models:} Diffusion Models, on the other hand, use a probabilistic framework that involves learning to reverse a noising process. This gradual approach allows for a more controlled and stable training process, as each step in the generation is trained independently. There is no need for adversarial feedback, which simplifies the training dynamics~\cite{stypulkowski2024diffused}.
\end{itemize}

\textbf{Generation Process:}
\begin{itemize}
    \item \textbf{GANs:} The generation in GANs is a direct mapping from noise to the data distribution. Once trained, the generator can produce a full image in a single pass, making GANs very fast at inference time. However, this also means that any issues in the training process can lead to significant artifacts or mode collapse.
    \item \textbf{Diffusion Models:} Diffusion Models generate data through a series of denoising steps, gradually refining a noisy input until it becomes a realistic sample. While this process can produce high-quality results, it is typically slower than GANs due to the multiple steps required for generation.
\end{itemize}

\textbf{Quality and Diversity:}
\begin{itemize}
    \item \textbf{GANs:} GANs are known for producing sharp and realistic images. However, they can sometimes suffer from issues such as mode collapse, where the generator learns to produce only a few types of samples and ignores other modes in the data distribution.
    \item \textbf{Diffusion Models:} Diffusion Models excel at producing diverse and high-quality images because they explicitly model the entire data distribution. The gradual denoising allows the model to correct mistakes step by step, leading to outputs that are often more consistent and less prone to artifacts~\cite{stypulkowski2024diffused}.
\end{itemize}

\textbf{2. Advantages and Limitations of Each Approach}

\textbf{Advantages of GANs:}
\begin{itemize}
    \item \textbf{Fast Inference:} Once trained, GANs can generate data quickly, making them ideal for real-time applications such as video games, animation, and virtual reality.
    \item \textbf{Sharp and Detailed Images:} GANs have been fine-tuned to produce extremely sharp and detailed images, often outperforming other models in terms of resolution and clarity.
    \item \textbf{Versatility:} The GAN framework has been adapted for a wide range of tasks, including image super-resolution, image-to-image translation, and style transfer.
\end{itemize}

\textbf{Limitations of GANs:}
\begin{itemize}
    \item \textbf{Training Instability:} The adversarial nature of GANs makes them difficult to train, often requiring careful tuning of hyperparameters and network architectures~\cite{de2021survey}.
    \item \textbf{Mode Collapse:} GANs may produce a limited set of outputs, failing to capture the full diversity of the data distribution.
    \item \textbf{Sensitive to Hyperparameters:} Small changes in learning rates or other parameters can drastically affect the quality of the generated samples~\cite{adadi2021survey}.
\end{itemize}

\textbf{Advantages of Diffusion Models:}
\begin{itemize}
    \item \textbf{Stable Training:} Diffusion models do not rely on adversarial training, which makes the training process more stable and less prone to the issues that affect GANs.
    \item \textbf{High-Quality and Diverse Outputs:} The step-by-step denoising process allows diffusion models to produce images that are highly detailed and diverse, capturing more variations in the data~\cite{croitoru2023diffusion}.
    \item \textbf{Probabilistic Framework:} Diffusion models are grounded in a solid probabilistic framework, which allows for more controlled and predictable behavior during generation.
\end{itemize}

\textbf{Limitations of Diffusion Models:}
\begin{itemize}
    \item \textbf{Slow Inference:} Generating data with diffusion models can be slow because it requires multiple steps of denoising, making them less suitable for real-time applications~\cite{stypulkowski2024diffused}.
    \item \textbf{Computationally Intensive:} The need for multiple forward and reverse passes during training and generation can make diffusion models more resource-intensive compared to GANs.
\end{itemize}

\textbf{3. Comparison Summary:}

\begin{center}
\begin{tabular}{|l|c|c|}
    \hline
    \textbf{Aspect} & \textbf{GANs} & \textbf{Diffusion Models} \\
    \hline
    Training Stability & Unstable (adversarial) & Stable (probabilistic) \\
    \hline
    Inference Speed & Fast & Slow \\
    \hline
    Generation Quality & Sharp images & High-quality, detailed images \\
    \hline
    Diversity & Prone to mode collapse & High diversity \\
    \hline
    Complexity & Sensitive to tuning & More computationally intensive \\
    \hline
\end{tabular}
\end{center}

\textbf{4. Future Directions and Prospects}

As both GANs and Diffusion Models continue to evolve, researchers are exploring ways to combine the strengths of both approaches. This could lead to models that leverage the fast inference of GANs while maintaining the stability and high-quality outputs of diffusion models~\cite{scholl2011challenges}.

\textbf{1. Hybrid Approaches}

There is growing interest in hybrid approaches that combine the best of both worlds. For example, some recent research has explored using GANs to speed up the diffusion process by learning an initial guess that the diffusion model can refine~\cite{yang2023diffusion}. This can reduce the number of denoising steps needed, making diffusion models more efficient.

\textbf{2. Improvements in Computational Efficiency}

Efforts are also being made to improve the computational efficiency of diffusion models. Techniques such as Latent Diffusion Models (LDMs), which perform the diffusion process in a lower-dimensional latent space, are promising developments that reduce the computational cost while preserving the quality of the generated data~\cite{de2021survey}.

\textbf{3. Application-Specific Models}

In the future, we may see more specialized generative models tailored for specific applications. For example, GANs may continue to dominate areas that require real-time generation, while diffusion models may become the preferred choice for tasks that prioritize quality and detail, such as medical imaging or fine art generation.

\textbf{4. Ethical Considerations and Responsible AI}

As generative models become more powerful, it is crucial to consider ethical implications, such as the potential misuse of AI for generating deepfakes or other harmful content~\cite{liu2021self}. Future research must focus on developing techniques to detect and prevent the misuse of generative models, as well as ensuring transparency and fairness in how these models are trained and applied~\cite{li2024survey}.

\textbf{Conclusion}

The competition between GANs~\cite{goodfellow2014generative} and Diffusion Models~\cite{ho2020denoising} represents an exciting time in the field of generative AI. Each approach has its strengths and weaknesses, and understanding these is essential for selecting the right model for the right task~\cite{li2024survey}. As research progresses, we are likely to see further innovations that will push the boundaries of what generative models can achieve, leading to more creative, efficient, and ethical solutions across different domains.



\bibliographystyle{ieeetr}
\bibliography{sample}

\begin{thebibliography}{100}

\bibitem{goodfellow2014generative}
I.~Goodfellow, J.~Pouget-Abadie, M.~Mirza, B.~Xu, D.~Warde-Farley, S.~Ozair, A.~Courville, and Y.~Bengio, ``Generative adversarial networks,'' {\em Advances in neural information processing systems}, vol.~27, 2014.

\bibitem{creswell2018generative}
A.~Creswell, T.~White, V.~Dumoulin, K.~Arulkumaran, B.~Sengupta, and A.~A. Bharath, ``Generative adversarial networks: An overview,'' {\em IEEE signal processing magazine}, vol.~35, no.~1, pp.~53--65, 2018.

\bibitem{metz2016unrolled}
L.~Metz, B.~Poole, D.~Pfau, and J.~Sohl-Dickstein, ``Unrolled generative adversarial networks,'' {\em arXiv preprint arXiv:1611.02163}, 2016.

\bibitem{radford2015unsupervised}
A.~Radford, ``Unsupervised representation learning with deep convolutional generative adversarial networks,'' {\em arXiv preprint arXiv:1511.06434}, 2015.

\bibitem{liu2016coupled}
M.-Y. Liu and O.~Tuzel, ``Coupled generative adversarial networks,'' {\em Advances in neural information processing systems}, vol.~29, 2016.

\bibitem{karras2017progressive}
T.~Karras, ``Progressive growing of gans for improved quality, stability, and variation,'' {\em arXiv preprint arXiv:1710.10196}, 2017.

\bibitem{karras2019style}
T.~Karras, S.~Laine, and T.~Aila, ``A style-based generator architecture for generative adversarial networks,'' in {\em Proceedings of the IEEE/CVF conference on computer vision and pattern recognition}, pp.~4401--4410, 2019.

\bibitem{peng2024isfb}
T.~Peng, M.~Li, F.~Chen, Y.~Xu, Y.~Xie, Y.~Sun, and D.~Zhang, ``Isfb-gan: Interpretable semantic face beautification with generative adversarial network,'' {\em Expert Systems with Applications}, vol.~236, p.~121131, 2024.

\bibitem{wang2017generative}
K.~Wang, C.~Gou, Y.~Duan, Y.~Lin, X.~Zheng, and F.-Y. Wang, ``Generative adversarial networks: introduction and outlook,'' {\em IEEE/CAA Journal of Automatica Sinica}, vol.~4, no.~4, pp.~588--598, 2017.

\bibitem{aggarwal2021generative}
A.~Aggarwal, M.~Mittal, and G.~Battineni, ``Generative adversarial network: An overview of theory and applications,'' {\em International Journal of Information Management Data Insights}, vol.~1, no.~1, p.~100004, 2021.

\bibitem{gui2021review}
J.~Gui, Z.~Sun, Y.~Wen, D.~Tao, and J.~Ye, ``A review on generative adversarial networks: Algorithms, theory, and applications,'' {\em IEEE transactions on knowledge and data engineering}, vol.~35, no.~4, pp.~3313--3332, 2021.

\bibitem{arjovsky2017wasserstein}
M.~Arjovsky, S.~Chintala, and L.~Bottou, ``Wasserstein generative adversarial networks,'' in {\em International conference on machine learning}, pp.~214--223, PMLR, 2017.

\bibitem{park2019semantic}
T.~Park, M.-Y. Liu, T.-C. Wang, and J.-Y. Zhu, ``Semantic image synthesis with spatially-adaptive normalization,'' in {\em Proceedings of the IEEE/CVF conference on computer vision and pattern recognition}, pp.~2337--2346, 2019.

\bibitem{brock2018large}
A.~Brock, ``Large scale gan training for high fidelity natural image synthesis,'' {\em arXiv preprint arXiv:1809.11096}, 2018.

\bibitem{karras2020analyzing}
T.~Karras, S.~Laine, M.~Aittala, J.~Hellsten, J.~Lehtinen, and T.~Aila, ``Analyzing and improving the image quality of stylegan,'' 2020.

\bibitem{karras2021alias}
T.~Karras, M.~Aittala, S.~Laine, E.~H{\"a}rk{\"o}nen, J.~Hellsten, J.~Lehtinen, and T.~Aila, ``Alias-free generative adversarial networks,'' {\em Advances in neural information processing systems}, vol.~34, pp.~852--863, 2021.

\bibitem{kang2023scaling}
M.~Kang, J.-Y. Zhu, R.~Zhang, J.~Park, E.~Shechtman, S.~Paris, and T.~Park, ``Scaling up gans for text-to-image synthesis,'' in {\em Proceedings of the IEEE/CVF Conference on Computer Vision and Pattern Recognition}, pp.~10124--10134, 2023.

\bibitem{zhou2020sparse}
K.~Zhou, S.~Gao, J.~Cheng, Z.~Gu, H.~Fu, Z.~Tu, J.~Yang, Y.~Zhao, and J.~Liu, ``Sparse-gan: Sparsity-constrained generative adversarial network for anomaly detection in retinal oct image,'' in {\em 2020 IEEE 17th International Symposium on Biomedical Imaging (ISBI)}, pp.~1227--1231, IEEE, 2020.

\bibitem{kingma2013auto}
D.~P. Kingma, ``Auto-encoding variational bayes,'' {\em arXiv preprint arXiv:1312.6114}, 2013.

\bibitem{kingma2019introduction}
D.~P. Kingma, M.~Welling, {\em et~al.}, ``An introduction to variational autoencoders,'' {\em Foundations and Trends{\textregistered} in Machine Learning}, vol.~12, no.~4, pp.~307--392, 2019.

\bibitem{ackley1985learning}
D.~H. Ackley, G.~E. Hinton, and T.~J. Sejnowski, ``A learning algorithm for boltzmann machines,'' {\em Cognitive science}, vol.~9, no.~1, pp.~147--169, 1985.

\bibitem{nair2010rectified}
V.~Nair and G.~E. Hinton, ``Rectified linear units improve restricted boltzmann machines,'' in {\em Proceedings of the 27th international conference on machine learning (ICML-10)}, pp.~807--814, 2010.

\bibitem{ramberg1979probability}
J.~S. Ramberg, E.~J. Dudewicz, P.~R. Tadikamalla, and E.~F. Mykytka, ``A probability distribution and its uses in fitting data,'' {\em Technometrics}, vol.~21, no.~2, pp.~201--214, 1979.

\bibitem{van2009dimensionality}
L.~Van Der~Maaten, E.~O. Postma, H.~J. Van Den~Herik, {\em et~al.}, ``Dimensionality reduction: A comparative review,'' {\em Journal of machine learning research}, vol.~10, no.~66-71, p.~13, 2009.

\bibitem{erhan2010does}
D.~Erhan, A.~Courville, Y.~Bengio, and P.~Vincent, ``Why does unsupervised pre-training help deep learning?,'' in {\em Proceedings of the thirteenth international conference on artificial intelligence and statistics}, pp.~201--208, JMLR Workshop and Conference Proceedings, 2010.

\bibitem{hoffman2016elbo}
M.~D. Hoffman and M.~J. Johnson, ``Elbo surgery: yet another way to carve up the variational evidence lower bound,'' in {\em Workshop in Advances in Approximate Bayesian Inference, NIPS}, vol.~1, 2016.

\bibitem{goodman1963statistical}
N.~R. Goodman, ``Statistical analysis based on a certain multivariate complex gaussian distribution (an introduction),'' {\em The Annals of mathematical statistics}, vol.~34, no.~1, pp.~152--177, 1963.

\bibitem{hoff2002latent}
P.~D. Hoff, A.~E. Raftery, and M.~S. Handcock, ``Latent space approaches to social network analysis,'' {\em Journal of the american Statistical association}, vol.~97, no.~460, pp.~1090--1098, 2002.

\bibitem{kazeminia2020gans}
S.~Kazeminia, C.~Baur, A.~Kuijper, B.~van Ginneken, N.~Navab, S.~Albarqouni, and A.~Mukhopadhyay, ``Gans for medical image analysis,'' {\em Artificial intelligence in medicine}, vol.~109, p.~101938, 2020.

\bibitem{paszke2019pytorch}
A.~Paszke, S.~Gross, F.~Massa, A.~Lerer, J.~Bradbury, G.~Chanan, T.~Killeen, Z.~Lin, N.~Gimelshein, L.~Antiga, {\em et~al.}, ``Pytorch: An imperative style, high-performance deep learning library,'' {\em Advances in neural information processing systems}, vol.~32, 2019.

\bibitem{stevens2020deep}
E.~Stevens, L.~Antiga, and T.~Viehmann, {\em Deep learning with PyTorch}.
\newblock Manning Publications, 2020.

\bibitem{kottwitz2023latex}
S.~Kottwitz, {\em LaTeX Graphics with TikZ: A practitioner's guide to drawing 2D and 3D images, diagrams, charts, and plots}.
\newblock Packt Publishing Ltd, 2023.

\bibitem{ho2019real}
Y.~Ho and S.~Wookey, ``The real-world-weight cross-entropy loss function: Modeling the costs of mislabeling,'' {\em IEEE access}, vol.~8, pp.~4806--4813, 2019.

\bibitem{pinheiro1995approximations}
J.~C. Pinheiro and D.~M. Bates, ``Approximations to the log-likelihood function in the nonlinear mixed-effects model,'' {\em Journal of computational and Graphical Statistics}, vol.~4, no.~1, pp.~12--35, 1995.

\bibitem{gomez2000analysis}
J.~F. G{\'o}mez-Lopera, J.~Mart{\'\i}nez-Aroza, A.~M. Robles-P{\'e}rez, and R.~Rom{\'a}n-Rold{\'a}n, ``An analysis of edge detection by using the jensen-shannon divergence,'' {\em Journal of Mathematical Imaging and Vision}, vol.~13, pp.~35--56, 2000.

\bibitem{hershey2007approximating}
J.~R. Hershey and P.~A. Olsen, ``Approximating the kullback leibler divergence between gaussian mixture models,'' in {\em 2007 IEEE International Conference on Acoustics, Speech and Signal Processing-ICASSP'07}, vol.~4, pp.~IV--317, IEEE, 2007.

\bibitem{srivastava2017veegan}
A.~Srivastava, L.~Valkov, C.~Russell, M.~U. Gutmann, and C.~Sutton, ``Veegan: Reducing mode collapse in gans using implicit variational learning,'' {\em Advances in neural information processing systems}, vol.~30, 2017.

\bibitem{thanh2020catastrophic}
H.~Thanh-Tung and T.~Tran, ``Catastrophic forgetting and mode collapse in gans,'' in {\em 2020 international joint conference on neural networks (ijcnn)}, pp.~1--10, IEEE, 2020.

\bibitem{hochreiter1998vanishing}
S.~Hochreiter, ``The vanishing gradient problem during learning recurrent neural nets and problem solutions,'' {\em International Journal of Uncertainty, Fuzziness and Knowledge-Based Systems}, vol.~6, no.~02, pp.~107--116, 1998.

\bibitem{hanin2018neural}
B.~Hanin, ``Which neural net architectures give rise to exploding and vanishing gradients?,'' {\em Advances in neural information processing systems}, vol.~31, 2018.

\bibitem{muller2019does}
R.~M{\"u}ller, S.~Kornblith, and G.~E. Hinton, ``When does label smoothing help?,'' {\em Advances in neural information processing systems}, vol.~32, 2019.

\bibitem{santurkar2018does}
S.~Santurkar, D.~Tsipras, A.~Ilyas, and A.~Madry, ``How does batch normalization help optimization?,'' {\em Advances in neural information processing systems}, vol.~31, 2018.

\bibitem{bjorck2018understanding}
N.~Bjorck, C.~P. Gomes, B.~Selman, and K.~Q. Weinberger, ``Understanding batch normalization,'' {\em Advances in neural information processing systems}, vol.~31, 2018.

\bibitem{frogner2015learning}
C.~Frogner, C.~Zhang, H.~Mobahi, M.~Araya, and T.~A. Poggio, ``Learning with a wasserstein loss,'' {\em Advances in neural information processing systems}, vol.~28, 2015.

\bibitem{he2018probgan}
H.~He, H.~Wang, G.-H. Lee, and Y.~Tian, ``Probgan: Towards probabilistic gan with theoretical guarantees,'' in {\em International conference on learning representations}, 2018.

\bibitem{bau2019seeing}
D.~Bau, J.-Y. Zhu, J.~Wulff, W.~Peebles, H.~Strobelt, B.~Zhou, and A.~Torralba, ``Seeing what a gan cannot generate,'' in {\em Proceedings of the IEEE/CVF international conference on computer vision}, pp.~4502--4511, 2019.

\bibitem{ma2021must}
T.~Ma, B.~Peng, W.~Wang, and J.~Dong, ``Must-gan: Multi-level statistics transfer for self-driven person image generation,'' in {\em Proceedings of the IEEE/CVF conference on computer vision and pattern recognition}, pp.~13622--13631, 2021.

\bibitem{fudenberg1991game}
D.~Fudenberg and J.~Tirole, {\em Game theory}.
\newblock MIT press, 1991.

\bibitem{daskalakis2009complexity}
C.~Daskalakis, P.~W. Goldberg, and C.~H. Papadimitriou, ``The complexity of computing a nash equilibrium,'' {\em Communications of the ACM}, vol.~52, no.~2, pp.~89--97, 2009.

\bibitem{heusel2017gans}
M.~Heusel, H.~Ramsauer, T.~Unterthiner, B.~Nessler, and S.~Hochreiter, ``Gans trained by a two time-scale update rule converge to a local nash equilibrium,'' {\em Advances in neural information processing systems}, vol.~30, 2017.

\bibitem{farnia2020gans}
F.~Farnia and A.~Ozdaglar, ``Do gans always have nash equilibria?,'' in {\em International Conference on Machine Learning}, pp.~3029--3039, PMLR, 2020.

\bibitem{parzen1962estimation}
E.~Parzen, ``On estimation of a probability density function and mode,'' {\em The annals of mathematical statistics}, vol.~33, no.~3, pp.~1065--1076, 1962.

\bibitem{saatci2017bayesian}
Y.~Saatci and A.~G. Wilson, ``Bayesian gan,'' {\em Advances in neural information processing systems}, vol.~30, 2017.

\bibitem{kuipers2012uniform}
L.~Kuipers and H.~Niederreiter, {\em Uniform distribution of sequences}.
\newblock Courier Corporation, 2012.

\bibitem{menendez1997jensen}
M.~L. Men{\'e}ndez, J.~Pardo, L.~Pardo, and M.~Pardo, ``The jensen-shannon divergence,'' {\em Journal of the Franklin Institute}, vol.~334, no.~2, pp.~307--318, 1997.

\bibitem{fuglede2004jensen}
B.~Fuglede and F.~Topsoe, ``Jensen-shannon divergence and hilbert space embedding,'' in {\em International symposium onInformation theory, 2004. ISIT 2004. Proceedings.}, p.~31, IEEE, 2004.

\bibitem{friedman1998work}
S.~D. Friedman, P.~Christensen, and J.~DeGroot, ``Work and life: The end of the zero-sum game,'' {\em Harvard business review}, vol.~76, pp.~119--130, 1998.

\bibitem{quinonero2022dataset}
J.~Qui{\~n}onero-Candela, M.~Sugiyama, A.~Schwaighofer, and N.~D. Lawrence, {\em Dataset shift in machine learning}.
\newblock Mit Press, 2022.

\bibitem{lecun1998gradient}
Y.~LeCun, L.~Bottou, Y.~Bengio, and P.~Haffner, ``Gradient-based learning applied to document recognition,'' {\em Proceedings of the IEEE}, vol.~86, no.~11, pp.~2278--2324, 1998.

\bibitem{ruby2020binary}
U.~Ruby and V.~Yendapalli, ``Binary cross entropy with deep learning technique for image classification,'' {\em Int. J. Adv. Trends Comput. Sci. Eng}, vol.~9, no.~10, 2020.

\bibitem{rubner2000earth}
Y.~Rubner, C.~Tomasi, and L.~J. Guibas, ``The earth mover's distance as a metric for image retrieval,'' {\em International journal of computer vision}, vol.~40, pp.~99--121, 2000.

\bibitem{han2019dimension}
S.~Han and Y.~Sung, ``Dimension-wise importance sampling weight clipping for sample-efficient reinforcement learning,'' in {\em International Conference on Machine Learning}, pp.~2586--2595, PMLR, 2019.

\bibitem{elsayed2024weight}
M.~Elsayed, Q.~Lan, C.~Lyle, and A.~R. Mahmood, ``Weight clipping for deep continual and reinforcement learning,'' {\em arXiv preprint arXiv:2407.01704}, 2024.

\bibitem{li2019preventing}
Q.~Li, S.~Haque, C.~Anil, J.~Lucas, R.~B. Grosse, and J.-H. Jacobsen, ``Preventing gradient attenuation in lipschitz constrained convolutional networks,'' {\em Advances in neural information processing systems}, vol.~32, 2019.

\bibitem{gentile1998linear}
C.~Gentile and M.~K. Warmuth, ``Linear hinge loss and average margin,'' {\em Advances in neural information processing systems}, vol.~11, 1998.

\bibitem{bartlett2008classification}
P.~L. Bartlett and M.~H. Wegkamp, ``Classification with a reject option using a hinge loss.,'' {\em Journal of Machine Learning Research}, vol.~9, no.~8, 2008.

\bibitem{zhang2019self}
H.~Zhang, I.~Goodfellow, D.~Metaxas, and A.~Odena, ``Self-attention generative adversarial networks,'' in {\em International conference on machine learning}, pp.~7354--7363, PMLR, 2019.

\bibitem{mirza2014conditional}
M.~Mirza, ``Conditional generative adversarial nets,'' {\em arXiv preprint arXiv:1411.1784}, 2014.

\bibitem{wang2018cgan}
T.-C. Wang, M.-Y. Liu, J.-Y. Zhu, A.~Tao, J.~Kautz, and B.~Catanzaro, ``High-resolution image synthesis and semantic manipulation with conditional gans,'' in {\em Proceedings of the IEEE conference on computer vision and pattern recognition}, pp.~8798--8807, 2018.

\bibitem{li2020gan}
M.~Li, J.~Lin, Y.~Ding, Z.~Liu, J.-Y. Zhu, and S.~Han, ``Gan compression: Efficient architectures for interactive conditional gans,'' in {\em Proceedings of the IEEE/CVF conference on computer vision and pattern recognition}, pp.~5284--5294, 2020.

\bibitem{devries2019evaluation}
T.~DeVries, A.~Romero, L.~Pineda, G.~W. Taylor, and M.~Drozdzal, ``On the evaluation of conditional gans,'' {\em arXiv preprint arXiv:1907.08175}, 2019.

\bibitem{deng2012mnist}
L.~Deng, ``The mnist database of handwritten digit images for machine learning research [best of the web],'' {\em IEEE signal processing magazine}, vol.~29, no.~6, pp.~141--142, 2012.

\bibitem{thekumparampil2018robustness}
K.~K. Thekumparampil, A.~Khetan, Z.~Lin, and S.~Oh, ``Robustness of conditional gans to noisy labels,'' {\em Advances in neural information processing systems}, vol.~31, 2018.

\bibitem{bao2017cvae}
J.~Bao, D.~Chen, F.~Wen, H.~Li, and G.~Hua, ``Cvae-gan: fine-grained image generation through asymmetric training,'' in {\em Proceedings of the IEEE international conference on computer vision}, pp.~2745--2754, 2017.

\bibitem{liu2019multi}
J.~Liu, C.~Gu, J.~Wang, G.~Youn, and J.-U. Kim, ``Multi-scale multi-class conditional generative adversarial network for handwritten character generation,'' {\em The Journal of Supercomputing}, vol.~75, pp.~1922--1940, 2019.

\bibitem{yi2017dualgan}
Z.~Yi, H.~Zhang, P.~Tan, and M.~Gong, ``Dualgan: Unsupervised dual learning for image-to-image translation,'' in {\em Proceedings of the IEEE international conference on computer vision}, pp.~2849--2857, 2017.

\bibitem{luo2021case}
J.~Luo, J.~Huang, and H.~Li, ``A case study of conditional deep convolutional generative adversarial networks in machine fault diagnosis,'' {\em Journal of Intelligent Manufacturing}, vol.~32, no.~2, pp.~407--425, 2021.

\bibitem{lecun2015deep}
Y.~LeCun, Y.~Bengio, and G.~Hinton, ``Deep learning,'' {\em nature}, vol.~521, no.~7553, pp.~436--444, 2015.

\bibitem{o2015cnn}
K.~O'Shea, ``An introduction to convolutional neural networks,'' {\em arXiv preprint arXiv:1511.08458}, 2015.

\bibitem{dubey2019comparative}
A.~K. Dubey and V.~Jain, ``Comparative study of convolution neural network’s relu and leaky-relu activation functions,'' in {\em Applications of Computing, Automation and Wireless Systems in Electrical Engineering: Proceedings of MARC 2018}, pp.~873--880, Springer, 2019.

\bibitem{jansson2012deconvolution}
P.~A. Jansson, {\em Deconvolution of images and spectra}.
\newblock Courier Corporation, 2012.

\bibitem{chen2016infogan}
X.~Chen, Y.~Duan, R.~Houthooft, J.~Schulman, I.~Sutskever, and P.~Abbeel, ``Infogan: Interpretable representation learning by information maximizing generative adversarial nets,'' {\em Advances in neural information processing systems}, vol.~29, 2016.

\bibitem{kurutach2018learning}
T.~Kurutach, A.~Tamar, G.~Yang, S.~J. Russell, and P.~Abbeel, ``Learning plannable representations with causal infogan,'' {\em Advances in Neural Information Processing Systems}, vol.~31, 2018.

\bibitem{mugunthan2021dpd}
V.~Mugunthan, V.~Gokul, L.~Kagal, and S.~Dubnov, ``Dpd-infogan: Differentially private distributed infogan,'' in {\em Proceedings of the 1st Workshop on Machine Learning and Systems}, pp.~1--6, 2021.

\bibitem{denton2015deep}
E.~L. Denton, S.~Chintala, R.~Fergus, {\em et~al.}, ``Deep generative image models using a￼ laplacian pyramid of adversarial networks,'' {\em Advances in neural information processing systems}, vol.~28, 2015.

\bibitem{jin2020hierarchical}
W.~Jin, R.~Barzilay, and T.~Jaakkola, ``Hierarchical generation of molecular graphs using structural motifs,'' in {\em International conference on machine learning}, pp.~4839--4848, PMLR, 2020.

\bibitem{zhang2018photographic}
Z.~Zhang, Y.~Xie, and L.~Yang, ``Photographic text-to-image synthesis with a hierarchically-nested adversarial network,'' in {\em Proceedings of the IEEE conference on computer vision and pattern recognition}, pp.~6199--6208, 2018.

\bibitem{cao2015grarep}
S.~Cao, W.~Lu, and Q.~Xu, ``Grarep: Learning graph representations with global structural information,'' in {\em Proceedings of the 24th ACM international on conference on information and knowledge management}, pp.~891--900, 2015.

\bibitem{lai2018fast}
W.-S. Lai, J.-B. Huang, N.~Ahuja, and M.-H. Yang, ``Fast and accurate image super-resolution with deep laplacian pyramid networks,'' {\em IEEE transactions on pattern analysis and machine intelligence}, vol.~41, no.~11, pp.~2599--2613, 2018.

\bibitem{mescheder2018training}
L.~Mescheder, A.~Geiger, and S.~Nowozin, ``Which training methods for gans do actually converge?,'' in {\em International conference on machine learning}, pp.~3481--3490, PMLR, 2018.

\bibitem{becker2022instability}
E.~Becker, P.~Pandit, S.~Rangan, and A.~K. Fletcher, ``Instability and local minima in gan training with kernel discriminators,'' {\em Advances in Neural Information Processing Systems}, vol.~35, pp.~20300--20312, 2022.

\bibitem{durall2020combating}
R.~Durall, A.~Chatzimichailidis, P.~Labus, and J.~Keuper, ``Combating mode collapse in gan training: An empirical analysis using hessian eigenvalues,'' {\em arXiv preprint arXiv:2012.09673}, 2020.

\bibitem{ding2022take}
Z.~Ding, S.~Jiang, and J.~Zhao, ``Take a close look at mode collapse and vanishing gradient in gan,'' in {\em 2022 IEEE 2nd International Conference on Electronic Technology, Communication and Information (ICETCI)}, pp.~597--602, IEEE, 2022.

\bibitem{kossale2022mode}
Y.~Kossale, M.~Airaj, and A.~Darouichi, ``Mode collapse in generative adversarial networks: An overview,'' in {\em 2022 8th International Conference on Optimization and Applications (ICOA)}, pp.~1--6, IEEE, 2022.

\bibitem{adler2018banach}
J.~Adler and S.~Lunz, ``Banach wasserstein gan,'' {\em Advances in neural information processing systems}, vol.~31, 2018.

\bibitem{gulrajani2017improved}
I.~Gulrajani, F.~Ahmed, M.~Arjovsky, V.~Dumoulin, and A.~C. Courville, ``Improved training of wasserstein gans,'' {\em Advances in neural information processing systems}, vol.~30, 2017.

\bibitem{mao2017least}
X.~Mao, Q.~Li, H.~Xie, R.~Y. Lau, Z.~Wang, and S.~Paul~Smolley, ``Least squares generative adversarial networks,'' in {\em Proceedings of the IEEE international conference on computer vision}, pp.~2794--2802, 2017.

\bibitem{anil2019sorting}
C.~Anil, J.~Lucas, and R.~Grosse, ``Sorting out lipschitz function approximation,'' in {\em International Conference on Machine Learning}, pp.~291--301, PMLR, 2019.

\bibitem{lee2022least}
C.-K. Lee, Y.-J. Cheon, and W.-Y. Hwang, ``Least squares generative adversarial networks-based anomaly detection,'' {\em IEEE Access}, vol.~10, pp.~26920--26930, 2022.

\bibitem{miyato2018spectral}
T.~Miyato, T.~Kataoka, M.~Koyama, and Y.~Yoshida, ``Spectral normalization for generative adversarial networks,'' {\em arXiv preprint arXiv:1802.05957}, 2018.

\bibitem{lin2021spectral}
Z.~Lin, V.~Sekar, and G.~Fanti, ``Why spectral normalization stabilizes gans: Analysis and improvements,'' {\em Advances in neural information processing systems}, vol.~34, pp.~9625--9638, 2021.

\bibitem{farnia2018generalizable}
F.~Farnia, J.~M. Zhang, and D.~Tse, ``Generalizable adversarial training via spectral normalization,'' {\em arXiv preprint arXiv:1811.07457}, 2018.

\bibitem{bjorck2021towards}
N.~Bjorck, C.~P. Gomes, and K.~Q. Weinberger, ``Towards deeper deep reinforcement learning with spectral normalization,'' {\em Advances in neural information processing systems}, vol.~34, pp.~8242--8255, 2021.

\bibitem{wu2021modeling}
B.~Wu, C.~Liu, C.~T. Ishi, and H.~Ishiguro, ``Modeling the conditional distribution of co-speech upper body gesture jointly using conditional-gan and unrolled-gan,'' {\em Electronics}, vol.~10, no.~3, p.~228, 2021.

\bibitem{wang2022unrolled}
J.~Wang and L.~Yao, ``Unrolled gan-based oversampling of credit card dataset for fraud detection,'' in {\em 2022 IEEE International Conference on Artificial Intelligence and Computer Applications (ICAICA)}, pp.~858--861, IEEE, 2022.

\bibitem{chen2024unsupervised}
Z.~Chen, Y.~Quan, and H.~Ji, ``Unsupervised deep unrolling networks for phase unwrapping,'' in {\em Proceedings of the IEEE/CVF Conference on Computer Vision and Pattern Recognition}, pp.~25182--25192, 2024.

\bibitem{lin2018pacgan}
Z.~Lin, A.~Khetan, G.~Fanti, and S.~Oh, ``Pacgan: The power of two samples in generative adversarial networks,'' {\em Advances in neural information processing systems}, vol.~31, 2018.

\bibitem{dou2023machine}
B.~Dou, Z.~Zhu, E.~Merkurjev, L.~Ke, L.~Chen, J.~Jiang, Y.~Zhu, J.~Liu, B.~Zhang, and G.-W. Wei, ``Machine learning methods for small data challenges in molecular science,'' {\em Chemical Reviews}, vol.~123, no.~13, pp.~8736--8780, 2023.

\bibitem{zhang2018convergence}
Z.~Zhang, M.~Li, and J.~Yu, ``On the convergence and mode collapse of gan,'' in {\em SIGGRAPH Asia 2018 Technical Briefs}, pp.~1--4, ACM, 2018.

\bibitem{gao2020data}
X.~Gao, F.~Deng, and X.~Yue, ``Data augmentation in fault diagnosis based on the wasserstein generative adversarial network with gradient penalty,'' {\em Neurocomputing}, vol.~396, pp.~487--494, 2020.

\bibitem{wu2018memory}
C.~Wu, L.~Herranz, X.~Liu, J.~Van De~Weijer, B.~Raducanu, {\em et~al.}, ``Memory replay gans: Learning to generate new categories without forgetting,'' {\em Advances in neural information processing systems}, vol.~31, 2018.

\bibitem{feng2021understanding}
R.~Feng, D.~Zhao, and Z.-J. Zha, ``Understanding noise injection in gans,'' in {\em international conference on machine learning}, pp.~3284--3293, PMLR, 2021.

\bibitem{zhang2019gradient}
J.~Zhang, T.~He, S.~Sra, and A.~Jadbabaie, ``Why gradient clipping accelerates training: A theoretical justification for adaptivity,'' {\em arXiv preprint arXiv:1905.11881}, 2019.

\bibitem{song2021gansim}
S.~Song, T.~Mukerji, and J.~Hou, ``Gansim: Conditional facies simulation using an improved progressive growing of generative adversarial networks (gans),'' {\em Mathematical Geosciences}, pp.~1--32, 2021.

\bibitem{zhang2019progressive}
D.~Zhang and A.~Khoreva, ``Progressive augmentation of gans,'' {\em Advances in Neural Information Processing Systems}, vol.~32, 2019.

\bibitem{zheng2022exploratory}
L.~Zheng, Y.~Zhen, J.~Niu, and L.~Zhong, ``An exploratory study on fade-in versus fade-out scaffolding for novice programmers in online collaborative programming settings,'' {\em Journal of Computing in Higher Education}, vol.~34, no.~2, pp.~489--516, 2022.

\bibitem{pathak2018efficient}
H.~N. Pathak, X.~Li, S.~Minaee, and B.~Cowan, ``Efficient super resolution for large-scale images using attentional gan,'' in {\em 2018 IEEE International Conference on Big Data (Big Data)}, pp.~1777--1786, IEEE, 2018.

\bibitem{ma1996texture}
W.-Y. Ma and B.~S. Manjunath, ``Texture features and learning similarity,'' in {\em Proceedings CVPR IEEE computer society conference on computer vision and pattern recognition}, pp.~425--430, IEEE, 1996.

\bibitem{dundar2023fine}
A.~Dundar, J.~Gao, A.~Tao, and B.~Catanzaro, ``Fine detailed texture learning for 3d meshes with generative models,'' {\em IEEE Transactions on Pattern Analysis and Machine Intelligence}, 2023.

\bibitem{donahue2019large}
J.~Donahue and K.~Simonyan, ``Large scale adversarial representation learning,'' {\em Advances in neural information processing systems}, vol.~32, 2019.

\bibitem{deng2009imagenet}
J.~Deng, W.~Dong, R.~Socher, L.-J. Li, K.~Li, and L.~Fei-Fei, ``Imagenet: A large-scale hierarchical image database,'' in {\em 2009 IEEE conference on computer vision and pattern recognition}, pp.~248--255, Ieee, 2009.

\bibitem{zhou2024comprehensive}
C.~Zhou, Q.~Li, C.~Li, J.~Yu, Y.~Liu, G.~Wang, K.~Zhang, C.~Ji, Q.~Yan, L.~He, {\em et~al.}, ``A comprehensive survey on pretrained foundation models: A history from bert to chatgpt,'' {\em International Journal of Machine Learning and Cybernetics}, pp.~1--65, 2024.

\bibitem{wang2020attentive}
Y.~Wang, Y.-C. Chen, X.~Zhang, J.~Sun, and J.~Jia, ``Attentive normalization for conditional image generation,'' in {\em Proceedings of the IEEE/CVF Conference on Computer Vision and Pattern Recognition}, pp.~5094--5103, 2020.

\bibitem{wang2015deep}
D.~Wang, P.~Cui, M.~Ou, and W.~Zhu, ``Deep multimodal hashing with orthogonal regularization,'' in {\em Twenty-fourth international joint conference on artificial intelligence}, 2015.

\bibitem{vaswani2017attention}
A.~Vaswani, ``Attention is all you need,'' {\em Advances in Neural Information Processing Systems}, 2017.

\bibitem{ma2019adaptive}
X.~Ma, R.~Jin, K.-A. Sohn, J.-Y. Paik, and T.-S. Chung, ``An adaptive control algorithm for stable training of generative adversarial networks,'' {\em IEEE Access}, vol.~7, pp.~184103--184114, 2019.

\bibitem{adadi2021survey}
A.~Adadi, ``A survey on data-efficient algorithms in big data era,'' {\em Journal of Big Data}, vol.~8, no.~1, p.~24, 2021.

\bibitem{khan2022transformers}
S.~Khan, M.~Naseer, M.~Hayat, S.~W. Zamir, F.~S. Khan, and M.~Shah, ``Transformers in vision: A survey,'' {\em ACM computing surveys (CSUR)}, vol.~54, no.~10s, pp.~1--41, 2022.

\bibitem{bond2021deep}
S.~Bond-Taylor, A.~Leach, Y.~Long, and C.~G. Willcocks, ``Deep generative modelling: A comparative review of vaes, gans, normalizing flows, energy-based and autoregressive models,'' {\em IEEE transactions on pattern analysis and machine intelligence}, vol.~44, no.~11, pp.~7327--7347, 2021.

\bibitem{abdal2019image2stylegan}
R.~Abdal, Y.~Qin, and P.~Wonka, ``Image2stylegan: How to embed images into the stylegan latent space?,'' in {\em Proceedings of the IEEE/CVF international conference on computer vision}, pp.~4432--4441, 2019.

\bibitem{huang2017arbitrary}
X.~Huang and S.~Belongie, ``Arbitrary style transfer in real-time with adaptive instance normalization,'' in {\em Proceedings of the IEEE international conference on computer vision}, pp.~1501--1510, 2017.

\bibitem{kim2020transfer}
Y.~Kim, J.~W. Soh, G.~Y. Park, and N.~I. Cho, ``Transfer learning from synthetic to real-noise denoising with adaptive instance normalization,'' in {\em Proceedings of the IEEE/CVF conference on computer vision and pattern recognition}, pp.~3482--3492, 2020.

\bibitem{jing2020dynamic}
Y.~Jing, X.~Liu, Y.~Ding, X.~Wang, E.~Ding, M.~Song, and S.~Wen, ``Dynamic instance normalization for arbitrary style transfer,'' in {\em Proceedings of the AAAI conference on artificial intelligence}, vol.~34, pp.~4369--4376, 2020.

\bibitem{zhang2022styleswin}
B.~Zhang, S.~Gu, B.~Zhang, J.~Bao, D.~Chen, F.~Wen, Y.~Wang, and B.~Guo, ``Styleswin: Transformer-based gan for high-resolution image generation,'' in {\em Proceedings of the IEEE/CVF conference on computer vision and pattern recognition}, pp.~11304--11314, 2022.

\bibitem{upchurch2017deep}
P.~Upchurch, J.~Gardner, G.~Pleiss, R.~Pless, N.~Snavely, K.~Bala, and K.~Weinberger, ``Deep feature interpolation for image content changes,'' in {\em Proceedings of the IEEE conference on computer vision and pattern recognition}, pp.~7064--7073, 2017.

\bibitem{bermano2022state}
A.~H. Bermano, R.~Gal, Y.~Alaluf, R.~Mokady, Y.~Nitzan, O.~Tov, O.~Patashnik, and D.~Cohen-Or, ``State-of-the-art in the architecture, methods and applications of stylegan,'' in {\em Computer Graphics Forum}, vol.~41, pp.~591--611, Wiley Online Library, 2022.

\bibitem{alaluf2022hyperstyle}
Y.~Alaluf, O.~Tov, R.~Mokady, R.~Gal, and A.~Bermano, ``Hyperstyle: Stylegan inversion with hypernetworks for real image editing,'' in {\em Proceedings of the IEEE/CVF conference on computer Vision and pattern recognition}, pp.~18511--18521, 2022.

\bibitem{patashnik2021styleclip}
O.~Patashnik, Z.~Wu, E.~Shechtman, D.~Cohen-Or, and D.~Lischinski, ``Styleclip: Text-driven manipulation of stylegan imagery,'' in {\em Proceedings of the IEEE/CVF international conference on computer vision}, pp.~2085--2094, 2021.

\bibitem{liu2023gan}
Y.~Liu, Q.~Li, Q.~Deng, Z.~Sun, and M.-H. Yang, ``Gan-based facial attribute manipulation,'' {\em IEEE transactions on pattern analysis and machine intelligence}, 2023.

\bibitem{kotovenko2019content}
D.~Kotovenko, A.~Sanakoyeu, S.~Lang, and B.~Ommer, ``Content and style disentanglement for artistic style transfer,'' in {\em Proceedings of the IEEE/CVF international conference on computer vision}, pp.~4422--4431, 2019.

\bibitem{sharma2024generative}
P.~Sharma, M.~Kumar, H.~K. Sharma, and S.~M. Biju, ``Generative adversarial networks (gans): Introduction, taxonomy, variants, limitations, and applications,'' {\em Multimedia Tools and Applications}, pp.~1--48, 2024.

\bibitem{isola2017image}
P.~Isola, J.-Y. Zhu, T.~Zhou, and A.~A. Efros, ``Image-to-image translation with conditional adversarial networks,'' in {\em Proceedings of the IEEE conference on computer vision and pattern recognition}, pp.~1125--1134, 2017.

\bibitem{qu2019enhanced}
Y.~Qu, Y.~Chen, J.~Huang, and Y.~Xie, ``Enhanced pix2pix dehazing network,'' in {\em Proceedings of the IEEE/CVF conference on computer vision and pattern recognition}, pp.~8160--8168, 2019.

\bibitem{chu2017cyclegan}
C.~Chu, A.~Zhmoginov, and M.~Sandler, ``Cyclegan, a master of steganography,'' {\em arXiv preprint arXiv:1712.02950}, 2017.

\bibitem{pang2021image}
Y.~Pang, J.~Lin, T.~Qin, and Z.~Chen, ``Image-to-image translation: Methods and applications,'' {\em IEEE Transactions on Multimedia}, vol.~24, pp.~3859--3881, 2021.

\bibitem{mustafa2020transformation}
A.~Mustafa and R.~K. Mantiuk, ``Transformation consistency regularization--a semi-supervised paradigm for image-to-image translation,'' in {\em Computer Vision--ECCV 2020: 16th European Conference, Glasgow, UK, August 23--28, 2020, Proceedings, Part XVIII 16}, pp.~599--615, Springer, 2020.

\bibitem{guo2020zero}
C.~Guo, C.~Li, J.~Guo, C.~C. Loy, J.~Hou, S.~Kwong, and R.~Cong, ``Zero-reference deep curve estimation for low-light image enhancement,'' in {\em Proceedings of the IEEE/CVF conference on computer vision and pattern recognition}, pp.~1780--1789, 2020.

\bibitem{ronneberger2015u}
O.~Ronneberger, P.~Fischer, and T.~Brox, ``U-net: Convolutional networks for biomedical image segmentation,'' in {\em Medical image computing and computer-assisted intervention--MICCAI 2015: 18th international conference, Munich, Germany, October 5-9, 2015, proceedings, part III 18}, pp.~234--241, Springer, 2015.

\bibitem{peng2023u}
Y.~Peng, M.~Sonka, and D.~Z. Chen, ``U-net v2: Rethinking the skip connections of u-net for medical image segmentation,'' {\em arXiv preprint arXiv:2311.17791}, 2023.

\bibitem{zhu2017unpaired}
J.-Y. Zhu, T.~Park, P.~Isola, and A.~A. Efros, ``Unpaired image-to-image translation using cycle-consistent adversarial networks,'' in {\em Proceedings of the IEEE international conference on computer vision}, pp.~2223--2232, 2017.

\bibitem{ledig2016photo}
C.~Ledig, L.~Theis, F.~Huszar, J.~Caballero, A.~Cunningham, A.~Acosta, A.~Aitken, A.~Tejani, J.~Totz, Z.~Wang, {\em et~al.}, ``Photo-realistic single image super-resolution using a generative adversarial network. arxiv 2016,'' {\em arXiv preprint arXiv:1609.04802}, 2016.

\bibitem{xiong2020improved}
Y.~Xiong, S.~Guo, J.~Chen, X.~Deng, L.~Sun, X.~Zheng, and W.~Xu, ``Improved srgan for remote sensing image super-resolution across locations and sensors,'' {\em Remote Sensing}, vol.~12, no.~8, p.~1263, 2020.

\bibitem{simonyan2014very}
K.~Simonyan, ``Very deep convolutional networks for large-scale image recognition,'' {\em arXiv preprint arXiv:1409.1556}, 2014.

\bibitem{he2016deep}
K.~He, X.~Zhang, S.~Ren, and J.~Sun, ``Deep residual learning for image recognition,'' in {\em Proceedings of the IEEE conference on computer vision and pattern recognition}, pp.~770--778, 2016.

\bibitem{wang2018esrgan}
X.~Wang, K.~Yu, S.~Wu, J.~Gu, Y.~Liu, C.~Dong, Y.~Qiao, and C.~Change~Loy, ``Esrgan: Enhanced super-resolution generative adversarial networks,'' in {\em Proceedings of the European conference on computer vision (ECCV) workshops}, pp.~0--0, 2018.

\bibitem{wu2016learning}
J.~Wu, C.~Zhang, T.~Xue, B.~Freeman, and J.~Tenenbaum, ``Learning a probabilistic latent space of object shapes via 3d generative-adversarial modeling,'' {\em Advances in neural information processing systems}, vol.~29, 2016.

\bibitem{smith2017improved}
E.~J. Smith and D.~Meger, ``Improved adversarial systems for 3d object generation and reconstruction,'' in {\em Conference on Robot Learning}, pp.~87--96, PMLR, 2017.

\bibitem{chan2022efficient}
E.~R. Chan, C.~Z. Lin, M.~A. Chan, K.~Nagano, B.~Pan, S.~De~Mello, O.~Gallo, L.~J. Guibas, J.~Tremblay, S.~Khamis, {\em et~al.}, ``Efficient geometry-aware 3d generative adversarial networks,'' in {\em Proceedings of the IEEE/CVF conference on computer vision and pattern recognition}, pp.~16123--16133, 2022.

\bibitem{liu2020neural}
L.~Liu, J.~Gu, K.~Zaw~Lin, T.-S. Chua, and C.~Theobalt, ``Neural sparse voxel fields,'' {\em Advances in Neural Information Processing Systems}, vol.~33, pp.~15651--15663, 2020.

\bibitem{nowozin2014optimal}
S.~Nowozin, ``Optimal decisions from probabilistic models: the intersection-over-union case,'' in {\em Proceedings of the IEEE conference on computer vision and pattern recognition}, pp.~548--555, 2014.

\bibitem{xie2018tempogan}
Y.~Xie, E.~Franz, M.~Chu, and N.~Thuerey, ``tempogan: A temporally coherent, volumetric gan for super-resolution fluid flow,'' {\em ACM Transactions on Graphics (TOG)}, vol.~37, no.~4, pp.~1--15, 2018.

\bibitem{reed2016generative}
S.~Reed, Z.~Akata, X.~Yan, L.~Logeswaran, B.~Schiele, and H.~Lee, ``Generative adversarial text to image synthesis,'' in {\em International conference on machine learning}, pp.~1060--1069, PMLR, 2016.

\bibitem{zhang2017stackgan}
H.~Zhang, T.~Xu, H.~Li, S.~Zhang, X.~Wang, X.~Huang, and D.~N. Metaxas, ``Stackgan: Text to photo-realistic image synthesis with stacked generative adversarial networks,'' in {\em Proceedings of the IEEE international conference on computer vision}, pp.~5907--5915, 2017.

\bibitem{xu2017attngan}
T.~Xu, P.~Zhang, Q.~Huang, H.~Zhang, Z.~Gan, X.~Huang, and X.~He, ``Attngan: Fine-grained text to image generation with attentional generative adversarial networks,'' 2017.

\bibitem{lu2024coarse}
Y.~Lu, M.~Zhang, A.~J. Ma, X.~Xie, and J.~Lai, ``Coarse-to-fine latent diffusion for pose-guided person image synthesis,'' in {\em Proceedings of the IEEE/CVF Conference on Computer Vision and Pattern Recognition}, pp.~6420--6429, 2024.

\bibitem{saito2017temporal}
M.~Saito, E.~Matsumoto, and S.~Saito, ``Temporal generative adversarial nets with singular value clipping,'' in {\em Proceedings of the IEEE international conference on computer vision}, pp.~2830--2839, 2017.

\bibitem{tulyakov2017mocogan}
S.~Tulyakov, M.-Y. Liu, X.~Yang, and J.~Kautz, ``Mocogan: Decomposing motion and content for video generation,'' 2017.

\bibitem{li2024survey}
C.~Li, D.~Huang, Z.~Lu, Y.~Xiao, Q.~Pei, and L.~Bai, ``A survey on long video generation: Challenges, methods, and prospects,'' {\em arXiv preprint arXiv:2403.16407}, 2024.

\bibitem{zhao2016energy}
J.~Zhao, ``Energy-based generative adversarial network,'' {\em arXiv preprint arXiv:1609.03126}, 2016.

\bibitem{makhzani2015adversarial}
A.~Makhzani, J.~Shlens, N.~Jaitly, I.~Goodfellow, and B.~Frey, ``Adversarial autoencoders,'' {\em arXiv preprint arXiv:1511.05644}, 2015.

\bibitem{donahue2016adversarial}
J.~Donahue, P.~Kr{\"a}henb{\"u}hl, and T.~Darrell, ``Adversarial feature learning,'' {\em arXiv preprint arXiv:1605.09782}, 2016.

\bibitem{doersch2016tutorial}
C.~Doersch, ``Tutorial on variational autoencoders,'' {\em arXiv preprint arXiv:1606.05908}, 2016.

\bibitem{xu2019cross}
Y.~Xu and A.~Goel, ``Cross-domain image classification through neural-style transfer data augmentation,'' {\em arXiv preprint arXiv:1910.05611}, 2019.

\bibitem{jing2019neural}
Y.~Jing, Y.~Yang, Z.~Feng, J.~Ye, Y.~Yu, and M.~Song, ``Neural style transfer: A review,'' {\em IEEE transactions on visualization and computer graphics}, vol.~26, no.~11, pp.~3365--3385, 2019.

\bibitem{xu2024videogigagan}
Y.~Xu, T.~Park, R.~Zhang, Y.~Zhou, E.~Shechtman, F.~Liu, J.-B. Huang, and D.~Liu, ``Videogigagan: Towards detail-rich video super-resolution,'' {\em arXiv preprint arXiv:2404.12388}, 2024.

\bibitem{chen2017coherent}
D.~Chen, J.~Liao, L.~Yuan, N.~Yu, and G.~Hua, ``Coherent online video style transfer,'' in {\em Proceedings of the IEEE International Conference on Computer Vision}, pp.~1105--1114, 2017.

\bibitem{aldausari2022video}
N.~Aldausari, A.~Sowmya, N.~Marcus, and G.~Mohammadi, ``Video generative adversarial networks: a review,'' {\em ACM Computing Surveys (CSUR)}, vol.~55, no.~2, pp.~1--25, 2022.

\bibitem{brooks2022generating}
T.~Brooks, J.~Hellsten, M.~Aittala, T.-C. Wang, T.~Aila, J.~Lehtinen, M.-Y. Liu, A.~Efros, and T.~Karras, ``Generating long videos of dynamic scenes,'' {\em Advances in Neural Information Processing Systems}, vol.~35, pp.~31769--31781, 2022.

\bibitem{yang2022wavegan}
M.~Yang, Z.~Wang, Z.~Chi, and W.~Feng, ``Wavegan: Frequency-aware gan for high-fidelity few-shot image generation,'' in {\em European Conference on Computer Vision}, pp.~1--17, Springer, 2022.

\bibitem{de2021survey}
G.~H. De~Rosa and J.~P. Papa, ``A survey on text generation using generative adversarial networks,'' {\em Pattern Recognition}, vol.~119, p.~108098, 2021.

\bibitem{hirschberg2015advances}
J.~Hirschberg and C.~D. Manning, ``Advances in natural language processing,'' {\em Science}, vol.~349, no.~6245, pp.~261--266, 2015.

\bibitem{zwicker1991audio}
E.~Zwicker and U.~T. Zwicker, ``Audio engineering and psychoacoustics: Matching signals to the final receiver, the human auditory system,'' {\em Journal of the Audio Engineering Society}, vol.~39, no.~3, pp.~115--126, 1991.

\bibitem{murmu2024reliable}
A.~Murmu, P.~Kumar, N.~R. Moparthi, S.~Namasudra, and P.~Lorenz, ``Reliable federated learning with gan model for robust and resilient future healthcare system,'' {\em IEEE Transactions on Network and Service Management}, 2024.

\bibitem{lin2019commongen}
B.~Y. Lin, W.~Zhou, M.~Shen, P.~Zhou, C.~Bhagavatula, Y.~Choi, and X.~Ren, ``Commongen: A constrained text generation challenge for generative commonsense reasoning,'' {\em arXiv preprint arXiv:1911.03705}, 2019.

\bibitem{yu2017seqgan}
L.~Yu, W.~Zhang, J.~Wang, and Y.~Yu, ``Seqgan: Sequence generative adversarial nets with policy gradient,'' in {\em Proceedings of the AAAI conference on artificial intelligence}, vol.~31, 2017.

\bibitem{zhang2017adversarial}
Y.~Zhang, Z.~Gan, K.~Fan, Z.~Chen, R.~Henao, D.~Shen, and L.~Carin, ``Adversarial feature matching for text generation,'' in {\em International conference on machine learning}, pp.~4006--4015, PMLR, 2017.

\bibitem{kaelbling1996reinforcement}
L.~P. Kaelbling, M.~L. Littman, and A.~W. Moore, ``Reinforcement learning: A survey,'' {\em Journal of artificial intelligence research}, vol.~4, pp.~237--285, 1996.

\bibitem{browne2012survey}
C.~B. Browne, E.~Powley, D.~Whitehouse, S.~M. Lucas, P.~I. Cowling, P.~Rohlfshagen, S.~Tavener, D.~Perez, S.~Samothrakis, and S.~Colton, ``A survey of monte carlo tree search methods,'' {\em IEEE Transactions on Computational Intelligence and AI in games}, vol.~4, no.~1, pp.~1--43, 2012.

\bibitem{yang2024integrated}
Q.~Yang, Y.~Bai, F.~Liu, and W.~Zhang, ``Integrated visual transformer and flash attention for lip-to-speech generation gan,'' {\em Scientific Reports}, vol.~14, no.~1, p.~4525, 2024.

\bibitem{donahue2018adversarial}
C.~Donahue, J.~McAuley, and M.~Puckette, ``Adversarial audio synthesis,'' {\em arXiv preprint arXiv:1802.04208}, 2018.

\bibitem{kumar2019melgan}
K.~Kumar, R.~Kumar, T.~De~Boissiere, L.~Gestin, W.~Z. Teoh, J.~Sotelo, A.~De~Brebisson, Y.~Bengio, and A.~C. Courville, ``Melgan: Generative adversarial networks for conditional waveform synthesis,'' {\em Advances in neural information processing systems}, vol.~32, 2019.

\bibitem{razzak2018deep}
M.~I. Razzak, S.~Naz, and A.~Zaib, ``Deep learning for medical image processing: Overview, challenges and the future,'' {\em Classification in BioApps: Automation of decision making}, pp.~323--350, 2018.

\bibitem{long2024pseudo}
X.~Long, T.~Wang, Y.~Kan, Y.~Wang, S.~Chen, A.~Zhou, X.~Hou, and J.~Liu, ``Pseudo training data generation for unsupervised cell membrane segmentation in immunohistochemistry images,'' in {\em IEEE International Conference on Bioinformatics and Biomedicine 2024}, 2024.

\bibitem{scholl2011challenges}
I.~Scholl, T.~Aach, T.~M. Deserno, and T.~Kuhlen, ``Challenges of medical image processing,'' {\em Computer science-Research and development}, vol.~26, pp.~5--13, 2011.

\bibitem{you2019ct}
C.~You, G.~Li, Y.~Zhang, X.~Zhang, H.~Shan, M.~Li, S.~Ju, Z.~Zhao, Z.~Zhang, W.~Cong, {\em et~al.}, ``Ct super-resolution gan constrained by the identical, residual, and cycle learning ensemble (gan-circle),'' {\em IEEE transactions on medical imaging}, vol.~39, no.~1, pp.~188--203, 2019.

\bibitem{pan20202d}
X.~Pan, B.~Dai, Z.~Liu, C.~C. Loy, and P.~Luo, ``Do 2d gans know 3d shape? unsupervised 3d shape reconstruction from 2d image gans,'' {\em arXiv preprint arXiv:2011.00844}, 2020.

\bibitem{bai2022novel}
T.~Bai, M.~Du, L.~Zhang, L.~Ren, L.~Ruan, Y.~Yang, G.~Qian, Z.~Meng, L.~Zhao, and M.~J. Deen, ``A novel alzheimer’s disease detection approach using gan-based brain slice image enhancement,'' {\em Neurocomputing}, vol.~492, pp.~353--369, 2022.

\bibitem{xia2022gan}
X.~Xia, X.~Pan, N.~Li, X.~He, L.~Ma, X.~Zhang, and N.~Ding, ``Gan-based anomaly detection: A review,'' {\em Neurocomputing}, vol.~493, pp.~497--535, 2022.

\bibitem{li2018semantic}
P.~Li, X.~Liang, D.~Jia, and E.~P. Xing, ``Semantic-aware grad-gan for virtual-to-real urban scene adaption,'' {\em arXiv preprint arXiv:1801.01726}, 2018.

\bibitem{li2021sp}
R.~Li, X.~Li, K.-H. Hui, and C.-W. Fu, ``Sp-gan: Sphere-guided 3d shape generation and manipulation,'' {\em ACM Transactions on Graphics (TOG)}, vol.~40, no.~4, pp.~1--12, 2021.

\bibitem{cirillo2021vox2vox}
M.~D. Cirillo, D.~Abramian, and A.~Eklund, ``Vox2vox: 3d-gan for brain tumour segmentation,'' in {\em Brainlesion: Glioma, Multiple Sclerosis, Stroke and Traumatic Brain Injuries: 6th International Workshop, BrainLes 2020, Held in Conjunction with MICCAI 2020, Lima, Peru, October 4, 2020, Revised Selected Papers, Part I 6}, pp.~274--284, Springer, 2021.

\bibitem{ko20233d}
J.~Ko, K.~Cho, D.~Choi, K.~Ryoo, and S.~Kim, ``3d gan inversion with pose optimization,'' in {\em Proceedings of the IEEE/CVF Winter Conference on Applications of Computer Vision}, pp.~2967--2976, 2023.

\bibitem{spick2019realistic}
r.~r. spick and j.~walker, ``Realistic and textured terrain generation using gans,'' in {\em Proceedings of the 16th ACM SIGGRAPH European Conference on Visual Media Production}, pp.~1--10, 2019.

\bibitem{xu2021gan}
P.~Xu and I.~Karamouzas, ``A gan-like approach for physics-based imitation learning and interactive character control,'' {\em Proceedings of the ACM on Computer Graphics and Interactive Techniques}, vol.~4, no.~3, pp.~1--22, 2021.

\bibitem{gan2021research}
B.~Gan, C.~Zhang, Y.~Chen, and Y.-C. Chen, ``Research on role modeling and behavior control of virtual reality animation interactive system in internet of things,'' {\em Journal of Real-Time Image Processing}, vol.~18, no.~4, pp.~1069--1083, 2021.

\bibitem{arad2021compositional}
D.~Arad~Hudson and L.~Zitnick, ``Compositional transformers for scene generation,'' {\em Advances in neural information processing systems}, vol.~34, pp.~9506--9520, 2021.

\bibitem{shim2022local}
S.-H. Shim, S.~Hyun, D.~Bae, and J.-P. Heo, ``Local attention pyramid for scene image generation,'' in {\em Proceedings of the IEEE/CVF Conference on Computer Vision and Pattern Recognition}, pp.~7774--7782, 2022.

\bibitem{liu2021self}
X.~Liu, F.~Zhang, Z.~Hou, L.~Mian, Z.~Wang, J.~Zhang, and J.~Tang, ``Self-supervised learning: Generative or contrastive,'' {\em IEEE transactions on knowledge and data engineering}, vol.~35, no.~1, pp.~857--876, 2021.

\bibitem{marcus2022very}
G.~Marcus, E.~Davis, and S.~Aaronson, ``A very preliminary analysis of dall-e 2,'' {\em arXiv preprint arXiv:2204.13807}, 2022.

\bibitem{radford2021learning}
A.~Radford, J.~W. Kim, C.~Hallacy, A.~Ramesh, G.~Goh, S.~Agarwal, G.~Sastry, A.~Askell, P.~Mishkin, J.~Clark, {\em et~al.}, ``Learning transferable visual models from natural language supervision,'' in {\em International conference on machine learning}, pp.~8748--8763, PMLR, 2021.

\bibitem{brown2020language}
T.~Brown, B.~Mann, N.~Ryder, M.~Subbiah, J.~D. Kaplan, P.~Dhariwal, A.~Neelakantan, P.~Shyam, G.~Sastry, A.~Askell, {\em et~al.}, ``Language models are few-shot learners,'' {\em Advances in neural information processing systems}, vol.~33, pp.~1877--1901, 2020.

\bibitem{saxena2021generative}
D.~Saxena and J.~Cao, ``Generative adversarial networks (gans) challenges, solutions, and future directions,'' {\em ACM Computing Surveys (CSUR)}, vol.~54, no.~3, pp.~1--42, 2021.

\bibitem{pan2019recent}
Z.~Pan, W.~Yu, X.~Yi, A.~Khan, F.~Yuan, and Y.~Zheng, ``Recent progress on generative adversarial networks (gans): A survey,'' {\em IEEE access}, vol.~7, pp.~36322--36333, 2019.

\bibitem{wu2019privacy}
Y.~Wu, F.~Yang, Y.~Xu, and H.~Ling, ``Privacy-protective-gan for privacy preserving face de-identification,'' {\em Journal of Computer Science and Technology}, vol.~34, pp.~47--60, 2019.

\bibitem{liu2021subverting}
K.~Liu, B.~Tan, and S.~Garg, ``Subverting privacy-preserving gans: Hiding secrets in sanitized images,'' in {\em Proceedings of the AAAI Conference on Artificial Intelligence}, vol.~35, pp.~14849--14856, 2021.

\bibitem{mcmahan2017communication}
B.~McMahan, E.~Moore, D.~Ramage, S.~Hampson, and B.~A. y~Arcas, ``Communication-efficient learning of deep networks from decentralized data,'' in {\em Artificial intelligence and statistics}, pp.~1273--1282, PMLR, 2017.

\bibitem{zhang2021survey}
C.~Zhang, Y.~Xie, H.~Bai, B.~Yu, W.~Li, and Y.~Gao, ``A survey on federated learning,'' {\em Knowledge-Based Systems}, vol.~216, p.~106775, 2021.

\bibitem{salimans2022progressive}
T.~Salimans and J.~Ho, ``Progressive distillation for fast sampling of diffusion models,'' {\em arXiv preprint arXiv:2202.00512}, 2022.

\bibitem{sutton1998reinforcement}
R.~Sutton and A.~Barto, ``Reinforcement learning: An introduction 1st edition,'' {\em Exp. Psychol. Learn. Mem. Cogn}, vol.~30, pp.~1302--1321, 1998.

\bibitem{sutton2018reinforcement}
R.~S. Sutton and A.~G. Barto, {\em Reinforcement learning: An introduction}.
\newblock MIT press, 2018.

\bibitem{wiering2012reinforcement}
M.~A. Wiering and M.~Van~Otterlo, ``Reinforcement learning,'' {\em Adaptation, learning, and optimization}, vol.~12, no.~3, p.~729, 2012.

\bibitem{sarmad2019rl}
M.~Sarmad, H.~J. Lee, and Y.~M. Kim, ``Rl-gan-net: A reinforcement learning agent controlled gan network for real-time point cloud shape completion,'' in {\em Proceedings of the IEEE/CVF conference on computer vision and pattern recognition}, pp.~5898--5907, 2019.

\bibitem{croitoru2023diffusion}
F.-A. Croitoru, V.~Hondru, R.~T. Ionescu, and M.~Shah, ``Diffusion models in vision: A survey,'' {\em IEEE Transactions on Pattern Analysis and Machine Intelligence}, vol.~45, no.~9, pp.~10850--10869, 2023.

\bibitem{yang2023diffusion}
L.~Yang, Z.~Zhang, Y.~Song, S.~Hong, R.~Xu, Y.~Zhao, W.~Zhang, B.~Cui, and M.-H. Yang, ``Diffusion models: A comprehensive survey of methods and applications,'' {\em ACM Computing Surveys}, vol.~56, no.~4, pp.~1--39, 2023.

\bibitem{kingma2021variational}
D.~Kingma, T.~Salimans, B.~Poole, and J.~Ho, ``Variational diffusion models,'' {\em Advances in neural information processing systems}, vol.~34, pp.~21696--21707, 2021.

\bibitem{ho2020denoising}
J.~Ho, A.~Jain, and P.~Abbeel, ``Denoising diffusion probabilistic models,'' {\em Advances in neural information processing systems}, vol.~33, pp.~6840--6851, 2020.

\bibitem{stypulkowski2024diffused}
M.~Stypu{\l}kowski, K.~Vougioukas, S.~He, M.~Zi{\k{e}}ba, S.~Petridis, and M.~Pantic, ``Diffused heads: Diffusion models beat gans on talking-face generation,'' in {\em Proceedings of the IEEE/CVF Winter Conference on Applications of Computer Vision}, pp.~5091--5100, 2024.

\bibitem{ho2022video}
J.~Ho, T.~Salimans, A.~Gritsenko, W.~Chan, M.~Norouzi, and D.~J. Fleet, ``Video diffusion models,'' {\em Advances in Neural Information Processing Systems}, vol.~35, pp.~8633--8646, 2022.

\bibitem{rombach2022high}
R.~Rombach, A.~Blattmann, D.~Lorenz, P.~Esser, and B.~Ommer, ``High-resolution image synthesis with latent diffusion models,'' in {\em Proceedings of the IEEE/CVF conference on computer vision and pattern recognition}, pp.~10684--10695, 2022.

\end{thebibliography}

\end{document}